\documentclass{article}

\PassOptionsToPackage{numbers, compress}{natbib}

\usepackage[final]{neurips_2024}

\usepackage[utf8]{inputenc} 
\usepackage[T1]{fontenc}    
\usepackage{url}            
\usepackage{booktabs}       
\usepackage{amsfonts}       
\usepackage{nicefrac}       
\usepackage{microtype}      
\usepackage[table,dvipsnames]{xcolor}         

\usepackage{minitoc}
\usepackage[toc,page,header]{appendix}

\definecolor{citecolor}{HTML}{0071BC}
\definecolor{linkcolor}{HTML}{ED1C24}

\definecolor{myyellow}{RGB}{244,164,96}
\definecolor{myblue}{RGB}{65,105,225}
\definecolor{mygreen}{RGB}{47,139,87}

\usepackage[pdftex]{graphicx}
\usepackage{bbding}
\usepackage{soul}
\usepackage{colortbl}
\usepackage{bibunits}
\usepackage{bm}
\usepackage{array}
\usepackage{subfigure}
\usepackage{pdflscape}
\usepackage{makecell}
\usepackage{framed,multirow}
\usepackage{threeparttable}
\usepackage{amsmath}
\usepackage{amssymb}
\usepackage{pifont}
\usepackage{algorithm}
\usepackage{layouts}
\usepackage{listings}
\usepackage{pythonhighlight}
\usepackage{tabularray} 
\usepackage[colorlinks,
            anchorcolor=red,
            citecolor=citecolor, 
            linkcolor=linkcolor,
            ]{hyperref}
\makeatletter
\newcommand\footnoteref[1]{\protected@xdef\@thefnmark{\ref{#1}}\@footnotemark}
\makeatother

\newcolumntype{P}[1]{>{\centering\arraybackslash}p{#1}}

\newlength\savewidth
\newcommand{\etal}{\mbox{et al.}}

\def\arrvline{\hfil\kern\arraycolsep\vline\kern-\arraycolsep\hfilneg}

\definecolor{iblue}{rgb}{0.06, 0.75, 1.0}
\definecolor{igray}{rgb}{0.00, 0.00, 0.00}
\definecolor{Highlight}{HTML}{39b54a}  

\newcolumntype{R}[1]{>{\raggedleft\arraybackslash}p{#1}}
\newcolumntype{L}[1]{>{\raggedright\arraybackslash}p{#1}}

\newcommand{\ourdataset}{AbdomenAtlas 1.0}
\newcommand{\ourproject}{Touchstone}

\newcommand{\numofct}{11,098}
\newcommand{\numoftrct}{5,195}
\newcommand{\numoftect}{5,903}
\newcommand{\numoftetotalsegct}{743}
\newcommand{\numofofficialtetotalsegct}{59}
\newcommand{\numoftejhhct}{5,160}

\newcommand{\numoftrcountry}{8}
\newcommand{\numoftrhospital}{76}
\newcommand{\numoftrpubdataset}{16}
\newcommand{\numoftehospital}{11}
\newcommand{\numoftetotalhospital}{10}

\newcommand{\numofinventor}{14}
\newcommand{\numofalgorithm}{19}
\newcommand{\numofframework}{3}

\newcommand{\numofclass}{9}

\title{A Large-Scale AI Benchmark for 3D Multi-Organ Segmentation}
\title{\ourproject\ Benchmark: Are We on the Right Way for Evaluating AI Algorithms for Medical Segmentation?}

\author{
Pedro R. A. S. Bassi\textsuperscript{1,2,3}\thanks{Equal contribution. Authors are permitted to list their name first in their CVs.} \quad
Wenxuan Li\textsuperscript{1}\footnotemark[1] \quad 
Yucheng Tang\textsuperscript{4} \quad
\bf Fabian Isensee\textsuperscript{5,6} \\
\bf Zifu Wang\textsuperscript{7} \quad 
\bf Jieneng Chen\textsuperscript{1} \quad 
\bf Yu-Cheng Chou\textsuperscript{1} \quad
\bf Saikat Roy\textsuperscript{5,8} \quad
\bf Yannick Kirchhoff\textsuperscript{5,8,9} \\
\bf Maximilian Rokuss\textsuperscript{5,8} \quad
\bf Ziyan Huang\textsuperscript{10} \quad
\bf Jin Ye\textsuperscript{11} \quad
\bf Junjun He\textsuperscript{11} \quad
\bf Tassilo Wald\textsuperscript{5,6} \\
\bf Constantin Ulrich\textsuperscript{5} \quad
\bf Michael Baumgartner\textsuperscript{5,6} \quad
\bf Klaus H. Maier-Hein\textsuperscript{5,12} \quad
\bf Paul Jaeger\textsuperscript{6,13} \\
\bf Yiwen Ye\textsuperscript{14} \quad
\bf Yutong Xie\textsuperscript{15} \quad
\bf Jianpeng Zhang\textsuperscript{16} \quad
\bf Ziyang Chen\textsuperscript{14} \quad
\bf Yong Xia\textsuperscript{14} \\
\bf Zhaohu Xing\textsuperscript{17} \quad
\bf Lei Zhu\textsuperscript{17, 18} \quad
\bf Yousef Sadegheih\textsuperscript{19} \quad
\bf Afshin Bozorgpour\textsuperscript{19} \\
\bf Pratibha Kumari\textsuperscript{19} \quad
\bf Reza Azad\textsuperscript{20} \quad
\bf Dorit Merhof\textsuperscript{19,21} \quad
\bf Pengcheng Shi\textsuperscript{22} \\
\bf Ting Ma\textsuperscript{22} \quad
\bf Yuxin Du\textsuperscript{10,23} \quad
\bf Fan Bai\textsuperscript{10,23,24} \quad
\bf Tiejun Huang\textsuperscript{23,25} \quad
\bf Bo Zhao\textsuperscript{10,23} \\
\bf Haonan Wang\textsuperscript{18} \quad
\bf Xiaomeng Li\textsuperscript{18} \quad
\bf Hanxue Gu\textsuperscript{26} \quad 
\bf Haoyu Dong\textsuperscript{26} \\ 
\bf Jichen Yang\textsuperscript{26} \quad 
\bf Maciej A. Mazurowski\textsuperscript{26} \quad 
\bf Saumya Gupta\textsuperscript{27} \quad
\bf Linshan Wu\textsuperscript{18} \\
\bf Jiaxin Zhuang\textsuperscript{18} \quad
\bf Hao Chen\textsuperscript{28} \quad
\bf Holger Roth\textsuperscript{4} \quad
\bf Daguang Xu\textsuperscript{4} \\
\bf Matthew B. Blaschko\textsuperscript{7} \quad 
\bf Sergio Decherchi\textsuperscript{29} \quad
\bf Andrea Cavalli\textsuperscript{2,29,30} \\
\bf Alan L. Yuille\textsuperscript{1}\thanks{Correspondence to: Alan L. Yuille (\href{mailto:ayuille1@jhu.edu}{\textsc{ayuille1@jhu.edu}}) and Zongwei Zhou (\href{mailto:zzhou82@jh.edu}{\textsc{zzhou82@jh.edu}})} \quad
\bf Zongwei Zhou\textsuperscript{1}\footnotemark[2] \\[2.5mm]
\textsuperscript{1}Department of Computer Science, Johns Hopkins University \\
\textsuperscript{2}Department of Pharmacy and Biotechnology, University of Bologna \\
\textsuperscript{3}Center for Biomolecular Nanotechnologies, Istituto Italiano di Tecnologia \\
\textsuperscript{4}NVIDIA \\
\textsuperscript{5}Division of Medical Image Computing, German Cancer Research Center (DKFZ) \\
\textsuperscript{6}Helmholtz Imaging, German Cancer Research Center (DKFZ) \\
Full affiliations are given in Appendix~\ref{sec:full_affiliation}. \\[2mm]{\small \texttt{Code, Models \& Data:} \href{https://github.com/MrGiovanni/\ourproject}{\texttt{https://github.com/MrGiovanni/Touchstone}}} \\
{\small \texttt{Leaderboard:} \href{https://mrgiovanni.github.io/Leaderboard/}{\texttt{https://mrgiovanni.github.io/Leaderboard}}}
}

\begin{document}

\maketitle

\doparttoc 
\faketableofcontents 

\begin{abstract}

\textit{How can we test AI performance?} This question seems trivial, but it isn't. Standard benchmarks often have problems such as in-distribution and small-size test sets, oversimplified metrics, unfair comparisons, and short-term outcome pressure. As a consequence, good performance on standard benchmarks does not guarantee success in real-world scenarios. To address these problems, we present \ourproject, a large-scale collaborative segmentation benchmark of 9 types of abdominal organs. This benchmark is based on \numoftrct\ training CT scans from \numoftrhospital\ hospitals around the world and \numoftect\ testing CT scans from \numoftehospital\ additional hospitals. This diverse test set enhances the statistical significance of benchmark results and rigorously evaluates AI algorithms across out-of-distribution scenarios. We invited \numofinventor\ inventors of \numofalgorithm\ AI algorithms to train their algorithms, while our team, as a third party, independently evaluated these algorithms. In addition, we also evaluated pre-existing AI frameworks---which, differing from algorithms, are more flexible and can support different algorithms---including MONAI from NVIDIA, nnU-Net from DKFZ, and numerous other open-source frameworks. We are committed to expanding this benchmark to encourage more innovation of AI algorithms for the medical domain.


\end{abstract}

\setcounter{footnote}{0} 
\section{Introduction}\label{sec:introduction}

The development of AI algorithms has led to enormous progress in medical segmentation, but few algorithms are reliable enough for clinical use~\cite{ardila2019end,isensee2021nnu,cao2023large}. 
Most AI algorithms fall short of expert radiologists, who are much more reliable and consistent when dealing with medical images from multiple hospitals, varied in different scanners, clinical protocols, patient demographics, or disease prevalences \cite{svanera2024fighting,lin2024shortcut,huang2023eval,zhou2022interpreting}. 
Therefore, the question remains:
\textit{How can we test medical AI in the diverse scenarios that are encountered by radiologists?} Establishing a trustworthy AI benchmark is important but exceptionally challenging, and seldom achieved in the medical domain. Tougher tests, like out-of-distribution evaluation on large, varied datasets, are needed.

Standard benchmarks have underlying problems that cause confusion in algorithm comparisons and delay progress. 
\textit{\textbf{First}, in-distribution test sets.} In the medical domain, CT scans in the test set often share sources, scanners, and populations with the training set. As a result, AI algorithms may perform well on the test set but generalize poorly to out-of-distribution (OOD) scenarios \cite{geirhos2020shortcut,banerjee2023shortcuts,bassi2024improving,lin2024shortcut,huang2023eval}. For example, Xia~\etal~\cite{xia2022felix} found that AI algorithms trained on data from Johns Hopkins Hospital (Baltimore, USA) lose accuracy in pancreatic tumor detection when evaluated on CT scans from Heidelberg Medical School (Heidelberg, Germany).
\textit{\textbf{Second}, small-size test sets.} Annotating medical data is expensive and time-consuming, but training AI requires substantial annotated data \cite{park2020annotated,qu2023annotating}. Therefore, most annotated data is used for training, leaving very little assigned for testing. Recent CT datasets such as TotalSegmentator \cite{wasserthal2022totalsegmentator}, WORD \cite{luo2021word}, and MSD \cite{antonelli2021medical}, offered fewer than 100 CT scans for testing. Even a single success or failure can skew results, reducing the statistical power and potentially misleading conclusions.
\textit{\textbf{Third}, over-simplified metrics.} Most standard benchmarks only compare average performance, failing to identify each AI algorithm's strengths and weaknesses in different scenarios. For instance, one algorithm might excel at segmenting small, circular structures (like the gall bladder) while another performs better on long, tubular ones (such as the aorta). Average performance across many classes can hide these nuances. 
\textit{\textbf{Fourth}, unfair comparisons.} Almost every paper reports that the newly `proposed AI' outperforms existing `alternative AIs.' The improvement becomes more significant if alternative AIs are reproduced and evaluated on an unknown training/test split. There are biases in comparison due to asymmetric efforts made in optimizing the proposed and alternative AIs. Many independent studies have reported these comparison biases over the years \cite{isensee2021nnu,isensee2024nnu} but remain unresolved. There is a need to have more widely adopted benchmarks (e.g., challenges) where all AI algorithms are trained by their inventors and evaluated by third parties. 
\textit{\textbf{Fifth}, short-term outcome pressure.} Standard benchmarks are often in short-term and non-recurring, requiring a final solution within several months. For example, RSNA 2024 Abdominal Trauma Detection \cite{rsna-2023-abdominal-trauma-detection} only opened for three months for data access and AI development \& evaluation. The short-term outcome pressure can discourage new classes of AI algorithms that need considerable time and computational resources for a thorough investigation, as their vanilla versions (e.g., Mamba \cite{gu2023mamba,yu2024mambaout} in early 2024 and Transformers \cite{dosovitskiy2020image} in early 2021) might not outperform all the alternatives judged. The benchmark must have long-term commitment and allowance.

To address this AI mismeasurement issue, we present the \ourproject\ benchmark, an effort towards the objective of creating a fair, large-scale, and widely-adopted medical AI benchmark. Its scale is large, featuring a training set of \numoftrct\ publicly available CT scans from \numoftrhospital\ hospitals and a test set of \numoftect\ CT scans from additional \numoftehospital\ hospitals. Test sets were unknown to the participants of the benchmark. All \numofct\ scans are annotated per voxel for \numofclass\ anatomical structures. The training set annotations were created by collaboration between AI specialists and radiologists followed by manual revision \cite{qu2023annotating}, \numoftejhhct\ out of \numoftect\ test scans are proprietary and manually annotated, and the remaining test datasets are publicly available, annotated by AI-radiologist collaboration. As of May 2024, \numofinventor\ global teams from eight countries have contributed to our benchmark. These teams are known for inventing novel AI algorithms for medical segmentation. In summary, the \ourproject\ benchmark explores an evaluation philosophy defined by the following \textbf{five contributions}:

\begin{enumerate}

    \item \textit{Evaluating on out-of-distribution data:} The JHH test set (Sec. \ref{sec:dataset}) presents 5,160 CT scans from an hospital never seen during training, introducing a new scale of external validation for abdominal CT benchmarks. The test data distribution varies in contrast enhancement (pre, venous, arterial, post-phases), disease condition (30\% containing abdominal tumors at varied stages), demographics (age, gender, race), image quality (e.g., slice thickness of 0.5--1.5 mm), and scanner types. We have collected metadata information for 72\% of the test set ($N$=\numoftejhhct) and reported AI performance in each sub-group.
    
    \item \textit{Providing a large test set:} Our test set ($N$=\numoftect) is much larger than the test sets of all current public CT benchmarks combined. It can enhance the statistical significance of the benchmark results: a 1\% average accuracy increment across 5,000 CT scans is more indicative of a genuine algorithmic improvement than a 1\% variation across 50 CT scans. 
    
    \item \textit{Analyzing pros/cons from multiple perspectives:} We evaluated segmentation performance of \numofclass\ anatomical structures, comparing the average results and analyzing them by metadata groups. We also reported per-class algorithm rankings and visualized worst-case performance. Moreover, we assessed inference time and computational cost, key factors for the clinical deployment of AI algorithms. 
    
    \item \textit{Inviting inventors to train their own algorithms:} 
    Each AI algorithm is configured by its own inventors, who know it best and have the most interest in its success.
    In our benchmark, each inventor trained their AI algorithm on \numoftrct\ annotated CT scans in AbdomenAtlas \cite{qu2023annotating}, and we, as a third party, independently evaluated these algorithms on \numoftect\ CT scans that are unknown and inaccessible to the AI inventors. This setting protects the integrity of our results (i.e., precluding the use of test data for hyperparameter tuning).
    
    \item \textit{Evaluating new algorithms with long-term commitment:} Our \ourproject\ benchmark not only invited established AI algorithms that are already published in major conferences/journals, but also invited newly developed algorithms appearing in recent pre-prints. We have a long-term commitment to this benchmark by organizing recurring challenges for at least five years, curating larger datasets, and improving label quality and task diversity. The first edition was featured as an invitation-only challenge at ISBI-2024.
    
\end{enumerate}

\textbf{Related benchmarks/challenges \& our innovations.} 
In a general sense, we define a \textit{benchmark} as an algorithmic comparison. Accordingly, the most common type of benchmark are the standard comparisons found in thousands of research papers \cite{oktay2018attention,zhou2018unet++,zhou2019unet++,chen2021transunet,he2021dints,hatamizadeh2022unetformer,liu2024universal,wu2024medsegdiff} where authors present new algorithms and compare baselines. As previously explained, this type of benchmark incurs the risk of unfairness, due to possible asymmetric efforts made in optimizing the proposed and alternative algorithms. However, open \textit{challenges} are a different type of benchmark, where developers train their own algorithms and submit them for third-party evaluation, mitigating the risk of unfair comparisons. For this reason, Table \ref{tab:related} contrasts our \ourproject\ benchmark to a non-exhaustive list of the most influential abdominal CT segmentation challenges. Notably, our training dataset is considerably larger and comes from more hospitals than any CT dataset ever used in a challenge. Furthermore, the only challenge training datasets on a scale similar to \ourdataset\ have partial labels and/or unlabeled portions \cite{antonelli2021medical,ma2022fast}. Our dataset is 17.3\(\times\) larger than the second-largest fully-annotated CT dataset \cite{heller2023kits21} in Table \ref{tab:related}. Boosting our results' statistical significance, our evaluation dataset is 8.6\(\times\) larger than any CT segmentation challenge test dataset. Moreover, \ourproject\ is the only benchmark in Table \ref{tab:related} to, simultaneously, explicitly analyze the performance of AI algorithms controlled by age, sex, race, and other metadata information. Lastly, this work is the starting point of a long-term benchmark, which we commit to maintain and improve over the years. Considering the importance of long-term commitment, we must acclaim KiTS, an abdominal segmentation challenge that had 3 editions since 2019 \cite{heller2019kits19,heller2020international,heller2021state,heller2023kits21} and FLARE, a challenge being consistently held yearly since 2021 \cite{ma2021abdomenct,ma2022fast,ma2024unleashing,ma2024automatic}.

\begin{table}[t]
\scriptsize
\centering
\caption{\textbf{Related benchmarks \& our innovations.} We compare \ourproject\ with influential CT segmentation benchmarks in light of the five contributions presented in the introduction.}
\label{tab:related}
\begin{tabular}{llllllll}
\toprule
contribution                                        & \multicolumn{3}{l}{\begin{tabular}[c]{@{}l@{}}promoting superior OOD performance\\ with a large and diverse training dataset \\ (\#1)\end{tabular}}                                                                                                  & \begin{tabular}[c]{@{}l@{}}boosting results' significance \\ \& large-scale OOD test \\ (\#1, \#2)\end{tabular} & \begin{tabular}[c]{@{}l@{}}multi-faceted\\ evaluation \\ (\#3)\end{tabular} & \begin{tabular}[c]{@{}l@{}}encouraging \\ innovative AI \\ (\#4, \#5)\end{tabular} \\ \midrule
benchmark                                           & \begin{tabular}[c]{@{}l@{}}\# CT scans\\ train\end{tabular} & \begin{tabular}[c]{@{}l@{}}\# hospitals\\ train\end{tabular} & \begin{tabular}[c]{@{}l@{}}\# countries\\ train\end{tabular} & \begin{tabular}[c]{@{}l@{}}\# CT scans\\ test\end{tabular}                   & \begin{tabular}[c]{@{}l@{}}AI consistency \\ analysis\end{tabular}       & \begin{tabular}[c]{@{}l@{}}targeted\\ invitation\end{tabular}                  \\ \midrule
MSD-CT \cite{antonelli2021medical} & 947$^{\dagger}$                                               & 1                                                               & 1                                                            & 465 IID                                                                        & none                                                                     & no                                                                             \\
FLARE'22 \cite{ma2023unleashing}         & 2,050$^{\dagger}$                                              & 22                                                              & 5+                                                           & 200 IID, 600 OOD                                                               & sex, age                                                                 & no                                                                             \\
FLARE'23 \cite{ma2024automatic}         & 4,000$^{\dagger}$                                              & 30                                                              & n/a                                                          & n/a                                                                            & n/a                                                                      & no                                                                             \\
KiTS21 \cite{heller2023kits21}     & 300                                                           & 50+                                                             & 1                                                            & 100 OOD                                                                        & sex, race                                                                & no                                                                             \\
AMOS22-CT \cite{ji2022amos}        & 200                                                           & 3                                                               & 1                                                            & 78 IID, 122 OOD                                                                & none                                                                     & no                                                                             \\
LiTS \cite{bilic2019liver}         & 130                                                           & 7                                                               & 5                                                            & 70 IID                                                                         & none                                                                     & no                                                                             \\
BTCV \cite{landman2015miccai}      & 30                                                            & 1                                                               & 1                                                            & 20 IID                                                                         & none                                                                     & no                                                                             \\
CHAOS-CT \cite{valindria2018multi} & 20                                                            & 1                                                               & 1                                                            & 20 IID                                                                         & none                                                                     & no                                                                             \\ 
\textbf{\ourproject\ (ours)}                          & \textbf{\numoftrct}                                                & \textbf{\numoftrhospital}                                                    & \textbf{\numoftrcountry}                                                   & \textbf{\numoftect\ OOD}                                                             & \textbf{sex, age, race}                                                  & \textbf{yes}                                                                            \\ \bottomrule
\end{tabular}
\begin{tablenotes}
    \item $^\dagger$Partially labeled: annotations for each organ do not cover the entire dataset, and/or may contain unlabeled samples.
\end{tablenotes}
\end{table}

\begin{figure}[t]
	\centering
	\includegraphics[width=\columnwidth]{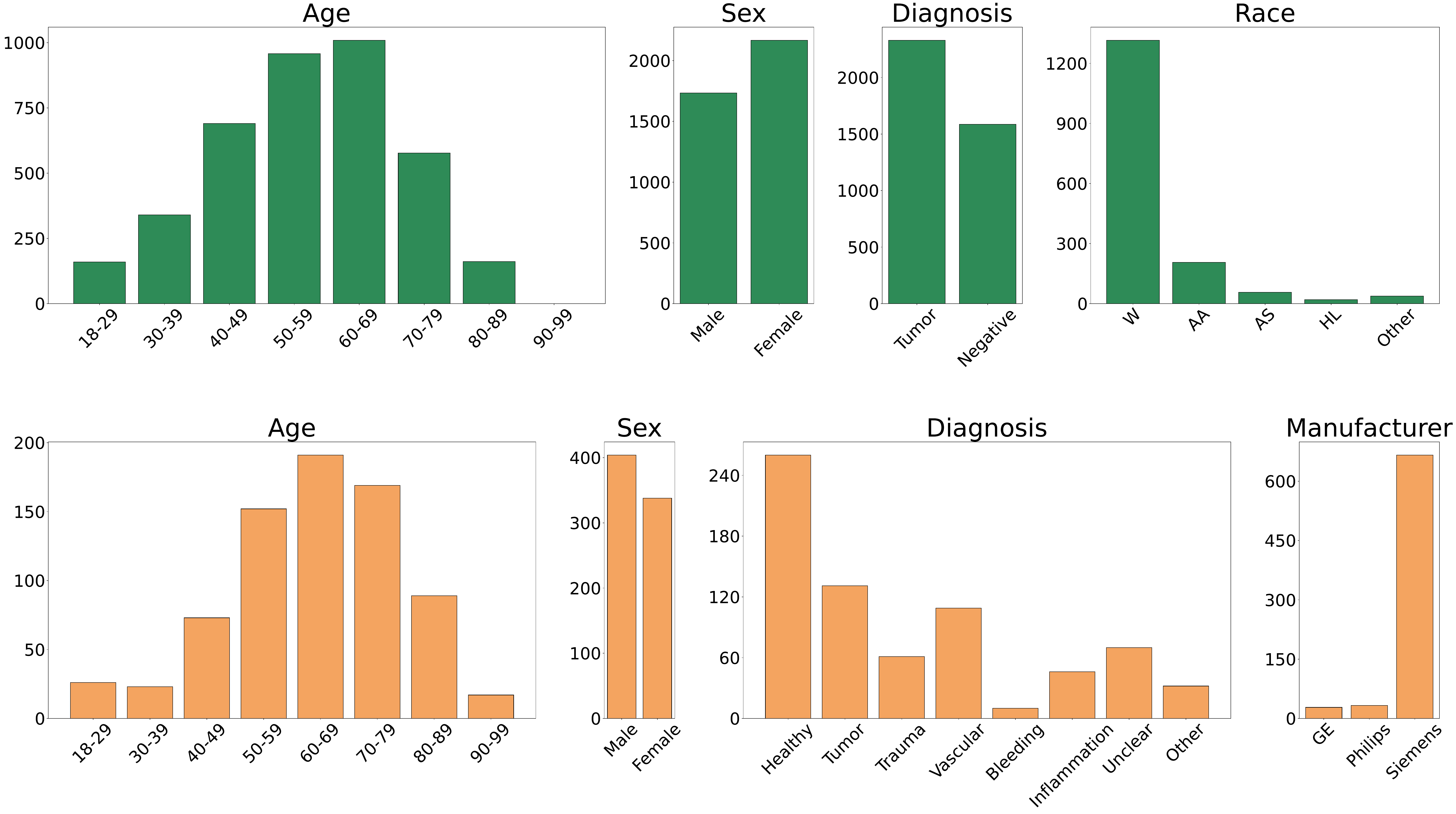}
    \caption{Summary of \textcolor{mygreen}{JHH} and \textcolor{myyellow}{TotalSegmentator} metadata.
    The diversity of data distribution includes more than just the number of centers; it also includes age, sex, manufacturer, diagnosis, and many other factors. JHH is the only dataset that provides race information, allowing us to compare the results; the race information is unknown in TotalSegmentator and most publicly available datasets. Therefore, the inclusion of JHH is value-added because it enabled the analysis on race. Races HL, W, AS, AA, O, and U indicate Hispanic \& Latino, White, Asian, African American, other and unknown, respectively. 
    }
	\label{fig:general_meta}
\end{figure}

\section{\ourproject\ Benchmark}\label{sec:method}

\subsection{Datasets -- Annotations, Statistics, Distribution, \& Characteristics}\label{sec:dataset}

We used one training dataset and two test datasets to perform a comprehensive out-of-distribution benchmark. The training and test datasets were collected from many hospitals worldwide. Figure \ref{fig:general_meta} shows the demographics of the two test datasets, JHH and TotalSegmentator; Appendix Figures \ref{fig:dataset_visualization_jhh}--\ref{fig:dataset_visualization_totalsegmentator} provide examples of CT scans and per-voxel annotations for various demographic groups across all datasets. The JHH dataset is proprietary and used for third-party evaluation; participants do not have access to the CT scans or their annotations. TotalSegmentator is a publicly available dataset; we did not inform the inventors beforehand of its use in our evaluation and confirmed that their AI algorithms had not been trained on this dataset. We included this public dataset to enable future participants to easily compare their algorithms with our benchmark. 

\textbf{\ourdataset}---\textit{$N$=\numoftrct; publicly available for training purposes}---is the largest multi-organ fully-annotated CT dataset to date, encompassing \numoftrhospital\ hospitals in \numoftrcountry\ countries \cite{qu2023annotating}. It leveraged a human-in-the-loop active learning strategy to empower radiologists to feasibly annotate \numoftrct\ CT scans from \numoftrpubdataset\ public datasets (listed in Appendix Table~\ref{tab:public_dataset_list}) and is fully annotated for 9 anatomical structures, i.e., spleen, liver, L\&R kidneys, stomach, gallbladder, pancreas, aorta, and postcava. \ourdataset, under \href{https://archive.data.jhu.edu/dataset.xhtml?persistentId=doi:10.7281/T1/7ELIJW}{CC BY-NC 4.0 License}, is derived from publicly available datasets, so detailed metadata information is unfortunately not available. 

\textbf{JHH}---\textit{$N$=\numoftejhhct; reserved for out-of-distribution test purposes\footnote{Out-of-distribution (OOD) test data (both images and annotations) must remain private, as public release can lead to overfitting and compromise OOD evaluation integrity \cite{geirhos2020shortcut,recht2019imagenet}. If any OOD data is released, a new, privately preserved test set will be required to ensure reliable evaluation.}}---provides contrast-enhanced CT scans in venous and arterial phases. Collected from Johns Hopkins Hospital using two Siemens scanners, this dataset includes metadata on age, race, gender, and diagnosis. Notably, all per-voxel annotations in JHH were manually created by radiologists \cite{park2020annotated,xia2022felix}. Annotation time for a single structure ranges from minutes to hours, depending on the size and complexity of the regions of interest to annotate and the local surrounding anatomical structures. Each CT scan was annotated by a team of radiologists, and confirmed by one of three additional experienced radiologists to ensure the quality of the annotation. All personally identifiable information was removed and the use of this dataset has received IRB approval from Johns Hopkins Medicine under IRB00403268. JHH is considered here an OOD test set because no CT scan from the Johns Hopkins hospital is present in the training dataset.

\textbf{TotalSegmentatorV2}---\textit{$N$=\numoftetotalsegct; publicly available for out-of-distribution test purposes}---is from \numoftetotalhospital\ institutes within the University Hospital Basel (Switzerland) picture archiving and communication system (PACS) \cite{wasserthal2022totalsegmentator}.
Being one of the largest public CT datasets, TotalSegmentator, under \href{https://github.com/wasserth/TotalSegmentator/blob/master/LICENSE#}{Apache License 2.0}, was annotated by AI-assisted radiologists. It comprises both contrast-enhanced and non-contrast images, with per-sample metadata including age, sex, scanner details, diagnosis, and institution. We report AI performance on a subset of TotalSegmentator dataset\footnote{TotalSegmentator offers 1,228 CT scans, but 485 scans were included into FLARE and subsequently inherited by \ourdataset. As a result, we used only the remaining \numoftetotalsegct\ scans for evaluation. Unlike JHH, this evaluation set does not come from completely unseen hospitals. However, there is a significant distribution shift between the TotalSegmentator data within AbdomenAtlas and the data in our test set (see Appendix \ref{sec:TS}).} in Table~\ref{tab:totalseg_train_plus_test_dsc} and its official test set in Appendix Tables~\ref{tab:totalseg_test_dsc}--\ref{tab:totalseg_test_nsd}.

\subsection{Evaluation Protocols -- Architectures, Frameworks, Metrics, \& Statistical Analysis}\label{sec:evaluation_protocols}

In this study, we define an \textit{architecture} as the overall design and structure of the entire neural network model; and define a \textit{framework} as a set of tools or protocols that can accommodate multiple AI architectures. We evaluated \numofalgorithm\ architectures and \numofframework\ frameworks trained by their inventors on our \ourdataset\footnote{Appendix \ref{sec:backbone_description}--\ref{sec:implementation_details} describe in-depth the description and configuration of each architecture/framework.}.
We used Dice Similarity Coefficient (DSC) and Normalized Surface Distance (NSD) to evaluate segmentation performance. 
We enforced that the inference speed must be faster than $1e^6$ mm$^3$ per second. The inference speed for each algorithm is summarized in Appendix Table~\ref{tab:algorithm_details}. We employed the same computer to evaluate all submitted algorithms. Its specifications are CPU: AMD EPYC 7713 @ 2,0Ghz$\times$64; GPU: NVIDIA Ampere A100 (80GB); RAM: 2TB. We applied statistical hypothesis testing to each possible pair of algorithms to ensure their performance differences are significant. Following Wiesenfarth~\etal~\cite{wiesenfarth2021methods}, we used the one-sided Wilcoxon signed rank test with Holm's adjustment for multiplicity at 5\% significance level and summarized results in significance maps. Per-group metadata analysis in Appendix \ref{sec:per_group_analysis} considers Kruskal–Wallis tests, followed by post-hoc Mann-Whitney U Tests with Bonferroni correction. More statistical analyses, such as ranking stability \cite{wiesenfarth2021methods}, are presented in Appendix \ref{sec:supplementary_statistics}.

\begin{table*}[h]
\caption{
\textbf{External validation on proprietary JHH dataset ($N$=\numoftejhhct).} Performance is given as DSC score (mean$\pm$s.d.). For each class, we bold the best-performing results and highlight the runners-up, which show no significant difference from the best results at $p=0.05$ level, in red. Architectures are grouped by their frameworks and sorted in ascending order based on the number of parameters. CNNs based on the nnU-Net framework have the best performance on most classes, but other models excel at specific structures (e.g., the graph neural network-based NeXToU for aorta, and the diffusion-based Diff-UNet for kidneys). The NSD results are reported in Appendix Table~\ref{tab:jhh_nsd}. We measured inference speed in cm$^3$/s (see Table~\ref{tab:algorithm_details} for details).
}\vspace{2px}
\centering
\scriptsize
\begin{tabular}{p{0.12\linewidth}p{0.19\linewidth}p{0.07\linewidth}P{0.075\linewidth}P{0.075\linewidth}P{0.075\linewidth}P{0.075\linewidth}P{0.075\linewidth}}
\toprule
framework & architecture & param & spleen & kidneyR & kidneyL & gallbladder & liver \\
\midrule
\multirow{9}{*}{nnU-Net}
& UniSeg$^\dagger$~\cite{ye2023uniseg} & 31.0M & 94.9$\pm$6.0 & 92.2$\pm$7.2 & 91.5$\pm$7.0 & 84.7$\pm$12.6 & 96.1$\pm$4.4 \\ 
& MedNeXt~\cite{roy2023mednext} & 61.8M & 95.2$\pm$6.3 & 92.6$\pm$7.4 & 91.8$\pm$7.3 & 85.3$\pm$12.9 & 96.3$\pm$4.5 \\ 
& NexToU~\cite{shi2023nextou} & 81.9M & 94.7$\pm$8.1 & 90.1$\pm$9.5 & 89.6$\pm$9.3 & 82.3$\pm$17.0 & 95.7$\pm$5.5 \\ 
& STU-Net-B~\cite{huang2023stu} & 58.3M & 95.1$\pm$6.4 & 92.5$\pm$7.3 & 91.9$\pm$7.2 & 85.5$\pm$12.3 & 96.2$\pm$4.8 \\ 
& STU-Net-L~\cite{huang2023stu} & 440.3M & 95.2$\pm$6.1 & 92.5$\pm$7.1 & 91.8$\pm$7.1 & 85.7$\pm$11.8 & 96.3$\pm$4.4 \\ 
& STU-Net-H~\cite{huang2023stu} & 1457.3M & 95.2$\pm$5.9 & 92.6$\pm$6.9 & 91.9$\pm$7.1 & \cellcolor{red!20}{\textbf{86.0$\pm$11.6}} & 96.3$\pm$4.4 \\ 
& U-Net~\cite{ronneberger2015u} & 31.1M & 95.1$\pm$6.3 & 92.7$\pm$6.9 & 91.9$\pm$7.2 & 84.7$\pm$13.1 & 96.2$\pm$4.5 \\ 
& ResEncL~\cite{isensee2021nnu,isensee2024nnu} & 102.0M & 95.2$\pm$6.3 & 92.6$\pm$7.0 & 91.9$\pm$6.9 & 84.9$\pm$13.0 & 96.3$\pm$4.5 \\ 
& \textcolor{lightgray}{ResEncL$^\bigstar$} & \textcolor{lightgray}{102.0M} & \textcolor{lightgray}{95.1$\pm$6.2} & \textcolor{lightgray}{92.7$\pm$6.9} & \textcolor{lightgray}{91.9$\pm$7.1} & \textcolor{lightgray}{84.9$\pm$12.8} & \textcolor{lightgray}{96.3$\pm$4.5} \\ 
\arrayrulecolor{gray}\midrule
\multirow{2}{*}{Vision-Language}
& U-Net \& CLIP~\cite{liu2023clip} & 19.1M & 94.3$\pm$6.9 & 91.9$\pm$7.8 & 91.1$\pm$8.8 & 82.1$\pm$15.4 & 96.0$\pm$4.3 \\ 
& Swin UNETR \& CLIP~\cite{liu2023clip} & 62.2M & 94.1$\pm$7.7 & 91.7$\pm$9.1 & 91.0$\pm$9.1 & 80.2$\pm$18.3 & 95.8$\pm$5.6 \\ 
\midrule
\multirow{6}{*}{MONAI}
& LHU-Net~\cite{sadegheih2024lhu} & 8.6M & 94.9$\pm$6.3 & 92.5$\pm$7.0 & 91.8$\pm$7.4 & 83.9$\pm$14.5 & 96.2$\pm$4.3 \\ 
& UCTransNet~\cite{wang2022uctransnet} & 68.0M & 90.2$\pm$11.9 & 86.5$\pm$14.6 & 86.9$\pm$12.8 & 77.8$\pm$19.5 & 93.6$\pm$6.4 \\ 
& Swin UNETR~\cite{tang2022self} & 72.8M & 92.7$\pm$8.8 & 89.8$\pm$11.1 & 89.7$\pm$10.2 & 76.9$\pm$20.7 & 95.2$\pm$5.3 \\ 
& UNesT~\cite{yu2023unest} & 87.2M & 93.2$\pm$7.1 & 90.9$\pm$8.1 & 90.1$\pm$8.2 & 75.1$\pm$21.2 & 95.3$\pm$5.0 \\ 
& UNETR~\cite{hatamizadeh2022unetr} & 101.8M & 91.7$\pm$10.1 & 90.1$\pm$9.4 & 89.2$\pm$9.6 & 74.7$\pm$20.4 & 95.0$\pm$5.3 \\ 
& SegVol$^\dagger$~\cite{du2023segvol} & 181.0M & 94.5$\pm$6.9 & 92.5$\pm$7.1 & 91.8$\pm$7.3 & 79.3$\pm$18.8 & 96.0$\pm$4.7 \\ 
\midrule
\multirow{3}{*}{n/a}
& SAM-Adapter$^\dagger$~\cite{gu2024build} & 11.6M & 90.5$\pm$8.8 & 90.4$\pm$7.9 & 87.3$\pm$9.6 & 49.4$\pm$22.9 & 94.1$\pm$5.3 \\ 
& MedFormer~\cite{gao2022data} & 38.5M & \cellcolor{red!20}{\textbf{95.5$\pm$6.1}} & 92.8$\pm$7.3 & 91.9$\pm$7.4 & \cellcolor{red!20}{85.3$\pm$13.6} & \cellcolor{red!20}{\textbf{96.4$\pm$4.4}} \\ 
& Diff-UNet~\cite{xing2023diff} & 434.0M & 95.0$\pm$6.9 & \cellcolor{red!20}{\textbf{92.8$\pm$7.4}} & \cellcolor{red!20}{\textbf{91.9$\pm$7.5}} & 83.8$\pm$14.8 & 96.2$\pm$4.7 \\ 
\arrayrulecolor{black}\midrule

framework & architecture & speed & stomach & aorta & postcava & pancreas & average \\
\midrule
\multirow{9}{*}{nnU-Net}
& UniSeg$^\dagger$~\cite{ye2023uniseg} & 198 & 93.3$\pm$6.0 & 82.3$\pm$10.3 & 81.2$\pm$8.1 & 82.7$\pm$10.4 & 88.8$\pm$8.0 \\ 
& MedNeXt~\cite{roy2023mednext} & 308 & 93.5$\pm$6.0 & 83.1$\pm$10.2 & \cellcolor{red!20}{81.3$\pm$8.3} & \cellcolor{red!20}{83.3$\pm$11.0} & \cellcolor{red!20}{\textbf{89.2$\pm$8.2}} \\ 
& NexToU~\cite{shi2023nextou} & 654 & 92.7$\pm$7.5 & \cellcolor{red!20}{86.4$\pm$8.7} & 78.1$\pm$9.1 & 80.2$\pm$13.5 & 87.8$\pm$9.8 \\ 
& STU-Net-B~\cite{huang2023stu} & 418 & 93.5$\pm$6.0 & 82.1$\pm$10.5 & \cellcolor{red!20}{\textbf{81.3$\pm$8.2}} & 83.2$\pm$10.7 & 89.0$\pm$8.1 \\ 
& STU-Net-L~\cite{huang2023stu} & 179 & 93.7$\pm$5.6 & 81.0$\pm$10.9 & \cellcolor{red!20}{81.3$\pm$8.2} & 83.4$\pm$10.7 & 89.0$\pm$8.0 \\ 
& STU-Net-H~\cite{huang2023stu} & 73 & \cellcolor{red!20}{\textbf{93.7$\pm$5.7}} & 81.1$\pm$10.9 & 81.1$\pm$8.2 & \cellcolor{red!20}{\textbf{83.4$\pm$10.7}} & 89.0$\pm$7.9 \\ 
& U-Net~\cite{ronneberger2015u} & 1064 & 93.3$\pm$6.0 & 82.8$\pm$10.2 & 81.0$\pm$8.2 & 82.3$\pm$11.4 & 88.9$\pm$8.2 \\ 
& ResEncL~\cite{isensee2021nnu,isensee2024nnu} & 794 & 93.4$\pm$6.0 & 81.4$\pm$11.1 & 80.5$\pm$8.8 & 82.9$\pm$10.8 & 88.8$\pm$8.3 \\ 
& \textcolor{lightgray}{ResEncL$^\bigstar$} & \textcolor{lightgray}{794} & \textcolor{lightgray}{93.5$\pm$5.9} & \textcolor{lightgray}{88.0$\pm$7.3} & \textcolor{lightgray}{80.5$\pm$8.7} & \textcolor{lightgray}{82.8$\pm$11.1} & \textcolor{lightgray}{89.5$\pm$7.8} \\ 
\arrayrulecolor{gray}\midrule
\multirow{2}{*}{Vision-Language}
& U-Net \& CLIP~\cite{liu2023clip} & 543 & 92.4$\pm$6.8 & 77.1$\pm$12.7 & 78.5$\pm$9.6 & 80.8$\pm$11.5 & 87.1$\pm$9.3 \\ 
& Swin UNETR \& CLIP~\cite{liu2023clip} & 606 & 92.2$\pm$8.3 & 78.1$\pm$12.6 & 76.8$\pm$11.0 & 80.2$\pm$12.5 & 86.7$\pm$10.5 \\ 
\midrule
\multirow{6}{*}{MONAI}
& LHU-Net~\cite{sadegheih2024lhu} & 2273 & 93.0$\pm$6.1 & 79.5$\pm$11.2 & 79.4$\pm$9.3 & 81.0$\pm$11.3 & 88.0$\pm$8.6 \\ 
& UCTransNet~\cite{wang2022uctransnet} & 1163 & 81.9$\pm$12.9 & \cellcolor{red!20}{\textbf{86.5$\pm$8.0}} & 68.1$\pm$15.8 & 59.0$\pm$21.6 & 81.1$\pm$13.7 \\ 
& Swin UNETR~\cite{tang2022self} & 2222 & 90.5$\pm$8.6 & 77.2$\pm$15.1 & 75.4$\pm$11.8 & 75.6$\pm$14.5 & 84.8$\pm$11.8 \\ 
& UNesT~\cite{yu2023unest} & 2703 & 90.9$\pm$7.3 & 77.7$\pm$16.1 & 74.4$\pm$11.8 & 76.2$\pm$12.1 & 84.9$\pm$10.8 \\ 
& UNETR~\cite{hatamizadeh2022unetr} & 2703 & 88.8$\pm$8.4 & 76.5$\pm$16.4 & 71.5$\pm$12.8 & 72.3$\pm$14.5 & 83.3$\pm$11.9 \\ 
& SegVol$^\dagger$~\cite{du2023segvol} & 1923 & 92.5$\pm$7.0 & 80.2$\pm$11.3 & 77.8$\pm$9.7 & 79.1$\pm$12.4 & 87.1$\pm$9.5 \\ 
\midrule
\multirow{3}{*}{n/a}
& SAM-Adapter$^\dagger$~\cite{gu2024build} & 1639 & 88.0$\pm$9.3 & 62.8$\pm$12.2 & 48.0$\pm$14.2 & 50.2$\pm$12.6 & 73.4$\pm$11.4 \\ 
& MedFormer~\cite{gao2022data} & 535 & 93.4$\pm$6.4 & 82.1$\pm$11.7 & \cellcolor{red!20}{80.7$\pm$10.1} & 83.1$\pm$11.2 & \cellcolor{red!20}{89.0$\pm$8.7} \\ 
& Diff-UNet~\cite{xing2023diff} & 442 & 93.1$\pm$6.5 & 81.2$\pm$11.3 & 80.8$\pm$8.9 & 81.9$\pm$11.4 & 88.5$\pm$8.8 \\ 
\arrayrulecolor{black}\bottomrule
\end{tabular}
\begin{tablenotes}
    \item $^\dagger$These architectures were pre-trained (Appendix \ref{sec:implementation_details}).
    \item $^\bigstar$These architectures were trained on \ourdataset\ with enhanced label quality for the aorta and kidney classes (discussed in \S\ref{sec:conclusion}).
\end{tablenotes}
\label{tab:jhh_dsc}
\end{table*}

\begin{table*}[t]
\caption{\textbf{Validation on TotalSegmentator ($N$=\numoftetotalsegct).} Performances given as DSC score (mean$\pm$s.d.). For each class, we bold the best-performing results and highlight the runners-up, which show no significant difference from the best results at $p=0.05$ level, in red. To ease the direct comparison with other literature, we also reported the \textit{official} test set performance in Appendix Tables~\ref{tab:totalseg_test_dsc}--\ref{tab:totalseg_test_nsd}.  We measured inference speed in cm$^3$/s (see Table~\ref{tab:algorithm_details} for details).
}\vspace{2px}
\centering
\scriptsize

\begin{tabular}{p{0.12\linewidth}p{0.19\linewidth}p{0.07\linewidth}P{0.075\linewidth}P{0.075\linewidth}P{0.075\linewidth}P{0.075\linewidth}P{0.075\linewidth}}
\toprule
framework & architecture & param & spleen & kidneyR & kidneyL & gallbladder & liver \\
\midrule
\multirow{9}{*}{nnU-Net}
& UniSeg$^\dagger$~\cite{ye2023uniseg} & 31.0M & 89.4$\pm$19.4 & 84.5$\pm$23.8 & 81.9$\pm$27.9 & 74.6$\pm$27.4 & 91.7$\pm$16.5 \\ 
& MedNeXt~\cite{roy2023mednext} & 61.8M & \cellcolor{red!20}{91.6$\pm$18.3} & 85.5$\pm$24.8 & 86.0$\pm$23.8 & \cellcolor{red!20}{75.8$\pm$28.5} & \cellcolor{red!20}{93.0$\pm$15.8} \\ 
& NexToU~\cite{shi2023nextou} & 81.9M & 83.0$\pm$29.5 & 78.2$\pm$32.7 & 78.7$\pm$30.8 & 72.0$\pm$31.2 & 87.6$\pm$23.0 \\ 
& STU-Net-B~\cite{huang2023stu} & 58.3M & \cellcolor{red!20}{92.3$\pm$15.4} & 87.1$\pm$20.3 & 86.8$\pm$22.1 & \cellcolor{red!20}{\textbf{78.5$\pm$25.0}} & 93.0$\pm$13.9 \\ 
& STU-Net-L~\cite{huang2023stu} & 440.3M & \cellcolor{red!20}{91.6$\pm$17.8} & 88.2$\pm$18.6 & 86.3$\pm$22.9 & \cellcolor{red!20}{78.1$\pm$24.7} & \cellcolor{red!20}{\textbf{94.2$\pm$11.2}} \\ 
& STU-Net-H~\cite{huang2023stu} & 1457.3M & \cellcolor{red!20}{\textbf{92.4$\pm$14.6}} & 88.9$\pm$16.3 & 86.5$\pm$23.4 & \cellcolor{red!20}{77.7$\pm$25.4} & 94.0$\pm$11.4 \\ 
& U-Net~\cite{ronneberger2015u} & 31.1M & \cellcolor{red!20}{91.2$\pm$17.8} & 88.4$\pm$18.3 & 87.7$\pm$20.8 & \cellcolor{red!20}{78.3$\pm$25.5} & 93.4$\pm$13.8 \\ 
& ResEncL~\cite{isensee2021nnu,isensee2024nnu} & 102.0M & \cellcolor{red!20}{91.8$\pm$17.5} & \cellcolor{red!20}{\textbf{88.9$\pm$18.0}} & \cellcolor{red!20}{\textbf{88.2$\pm$20.5}} & \cellcolor{red!20}{78.0$\pm$25.2} & 91.7$\pm$18.4 \\ 
& \textcolor{lightgray}{ResEncL$^\bigstar$} & \textcolor{lightgray}{102.0M} & \textcolor{lightgray}{92.0$\pm$16.7} & \textcolor{lightgray}{89.9$\pm$15.3} & \textcolor{lightgray}{89.5$\pm$18.3} & \textcolor{lightgray}{78.0$\pm$24.7} & \textcolor{lightgray}{92.4$\pm$17.4} \\ 
\arrayrulecolor{gray}\midrule
\multirow{2}{*}{Vision-Language}
& U-Net \& CLIP~\cite{liu2023clip} & 19.1M & 87.4$\pm$23.8 & 83.6$\pm$25.6 & 82.7$\pm$26.6 & 73.1$\pm$29.1 & 91.6$\pm$14.8 \\ 
& Swin UNETR \& CLIP~\cite{liu2023clip} & 62.2M & 87.1$\pm$22.4 & 81.1$\pm$29.0 & 77.0$\pm$32.3 & 70.3$\pm$31.0 & 91.6$\pm$16.0 \\ 
\midrule
\multirow{6}{*}{MONAI}
& LHU-Net~\cite{sadegheih2024lhu} & 8.6M & 86.0$\pm$25.7 & 81.8$\pm$29.3 & 82.4$\pm$27.0 & 71.3$\pm$32.2 & 87.7$\pm$22.9 \\ 
& UCTransNet~\cite{wang2022uctransnet} & 68.0M & 76.4$\pm$34.5 & 74.3$\pm$35.2 & 62.0$\pm$41.5 & 69.6$\pm$31.9 & 82.6$\pm$28.2 \\ 
& Swin UNETR~\cite{tang2022self} & 72.8M & 66.3$\pm$36.4 & 59.7$\pm$39.4 & 58.5$\pm$40.2 & 50.6$\pm$40.6 & 80.2$\pm$28.7 \\ 
& UNesT~\cite{yu2023unest} & 87.2M & 79.5$\pm$26.7 & 73.8$\pm$32.4 & 72.0$\pm$33.8 & 50.3$\pm$40.0 & 87.6$\pm$20.9 \\ 
& UNETR~\cite{hatamizadeh2022unetr} & 101.8M & 60.4$\pm$37.9 & 47.9$\pm$39.6 & 41.9$\pm$39.8 & 40.0$\pm$36.9 & 78.1$\pm$29.9 \\
& SegVol$^\dagger$~\cite{du2023segvol} & 181.0M & 87.1$\pm$23.0 & 82.8$\pm$23.5 & 82.6$\pm$24.8 & 68.1$\pm$29.3 & 89.4$\pm$20.5 \\ 
\midrule
\multirow{3}{*}{n/a}
& SAM-Adapter$^\dagger$~\cite{gu2024build} & 11.6M & 53.5$\pm$33.4 & 8.5$\pm$11.1 & 19.9$\pm$22.1 & 11.5$\pm$17.6 & 66.4$\pm$35.5 \\ 
& MedFormer~\cite{gao2022data} & 38.5M & 90.7$\pm$15.0 & 85.5$\pm$18.5 & 84.0$\pm$21.5 & 74.1$\pm$26.8 & 92.8$\pm$12.4 \\ 
& Diff-UNet~\cite{xing2023diff} & 434.0M & 88.3$\pm$23.6 & 81.3$\pm$27.9 & 81.0$\pm$28.4 & 71.8$\pm$30.0 & 92.4$\pm$14.9 \\ 
\arrayrulecolor{black}\midrule
framework & architecture & speed & stomach & aorta & IVC$^\ddagger$ & pancreas & average \\
\midrule
\multirow{9}{*}{nnU-Net}
& UniSeg$^\dagger$~\cite{ye2023uniseg} & 198 & 74.0$\pm$29.5 & 69.2$\pm$31.5 & \cellcolor{red!20}{72.8$\pm$25.9} & 70.3$\pm$30.9 & 78.7$\pm$25.9 \\ 
& MedNeXt~\cite{roy2023mednext} & 308 & \cellcolor{red!20}{77.2$\pm$28.7} & 71.9$\pm$30.1 & \cellcolor{red!20}{75.2$\pm$23.5} & 71.6$\pm$31.4 & 80.9$\pm$25.0 \\ 
& NexToU~\cite{shi2023nextou} & 654 & 69.0$\pm$34.7 & 61.5$\pm$33.0 & 59.4$\pm$32.7 & 66.8$\pm$32.0 & 72.9$\pm$31.1 \\ 
& STU-Net-B~\cite{huang2023stu} & 418 & \cellcolor{red!20}{78.6$\pm$26.5} & \cellcolor{red!20}{74.2$\pm$28.9} & \cellcolor{red!20}{77.3$\pm$19.6} & \cellcolor{red!20}{74.9$\pm$27.5} & \cellcolor{red!20}{82.5$\pm$22.1} \\ 
& STU-Net-L~\cite{huang2023stu} & 179 & \cellcolor{red!20}{79.7$\pm$24.6} & \cellcolor{red!20}{\textbf{75.7$\pm$27.0}} & \cellcolor{red!20}{\textbf{77.6$\pm$18.7}} & \cellcolor{red!20}{75.2$\pm$27.0} & \cellcolor{red!20}{\textbf{83.0$\pm$21.4}} \\ 
& STU-Net-H~\cite{huang2023stu} & 73 & \cellcolor{red!20}{78.5$\pm$25.5} & \cellcolor{red!20}{74.7$\pm$28.1} & \cellcolor{red!20}{76.9$\pm$19.0} & \cellcolor{red!20}{74.5$\pm$27.6} & \cellcolor{red!20}{82.7$\pm$21.3} \\ 
& U-Net~\cite{ronneberger2015u} & 1064 & \cellcolor{red!20}{78.9$\pm$26.3} & 71.0$\pm$28.4 & \cellcolor{red!20}{76.4$\pm$21.8} & \cellcolor{red!20}{75.2$\pm$27.0} & 82.3$\pm$22.2 \\ 
& ResEncL~\cite{isensee2021nnu,isensee2024nnu} & 794 & \cellcolor{red!20}{78.9$\pm$25.3} & 73.8$\pm$25.9 & \cellcolor{red!20}{76.4$\pm$20.2} & \cellcolor{red!20}{\textbf{76.3$\pm$25.9}} & \cellcolor{red!20}{82.7$\pm$21.9} \\ 
& \textcolor{lightgray}{ResEncL$^\bigstar$} & \textcolor{lightgray}{794} & \textcolor{lightgray}{80.9$\pm$23.0} & \textcolor{lightgray}{84.2$\pm$20.5} & \textcolor{lightgray}{76.3$\pm$20.0} & \textcolor{lightgray}{77.3$\pm$24.9} & \textcolor{lightgray}{84.5$\pm$20.1} \\ 
\arrayrulecolor{gray}\midrule
\multirow{2}{*}{Vision-Language}
& U-Net \& CLIP~\cite{liu2023clip} & 543 & \cellcolor{red!20}{77.7$\pm$26.8} & 59.0$\pm$32.8 & 65.8$\pm$27.2 & 74.6$\pm$25.7 & 77.3$\pm$25.8 \\ 
& Swin UNETR \& CLIP~\cite{liu2023clip} & 606 & 71.2$\pm$30.7 & 58.6$\pm$34.5 & 63.6$\pm$27.4 & 70.3$\pm$28.9 & 74.5$\pm$28.0 \\ 
\midrule
\multirow{6}{*}{MONAI}
& LHU-Net~\cite{sadegheih2024lhu} & 2273 & 71.3$\pm$31.8 & 63.0$\pm$34.1 & 67.5$\pm$28.5 & 68.6$\pm$32.6 & 75.5$\pm$29.3 \\ 
& UCTransNet~\cite{wang2022uctransnet} & 1163 & 61.6$\pm$36.1 & 49.7$\pm$34.8 & 49.3$\pm$36.4 & 59.0$\pm$35.1 & 64.9$\pm$34.9 \\ 
& Swin UNETR~\cite{tang2022self} & 2222 & 52.2$\pm$35.2 & 54.5$\pm$37.0 & 38.1$\pm$34.7 & 42.3$\pm$34.5 & 55.8$\pm$36.3 \\ 
& UNesT~\cite{yu2023unest} & 2703 & 63.9$\pm$31.5 & 54.7$\pm$37.0 & 38.9$\pm$36.2 & 50.0$\pm$33.0 & 63.4$\pm$32.4 \\ 
& UNETR~\cite{hatamizadeh2022unetr} & 2703 & 42.1$\pm$32.1 & 41.0$\pm$31.4 & 41.3$\pm$32.3 & 28.2$\pm$29.2 & 46.8$\pm$34.3 \\
& SegVol$^\dagger$~\cite{du2023segvol} & 1923 & 71.6$\pm$29.9 & 60.8$\pm$29.8 & 63.0$\pm$24.3 & 66.3$\pm$28.1 & 74.6$\pm$25.9 \\ 
\midrule
\multirow{3}{*}{n/a}
& SAM-Adapter$^\dagger$~\cite{gu2024build} & 1639 & 48.4$\pm$30.9 & 15.2$\pm$18.6 & 4.8$\pm$8.1 & 30.9$\pm$21.7 & 28.8$\pm$22.1 \\ 
& MedFormer~\cite{gao2022data} & 535 & \cellcolor{red!20}{\textbf{80.4$\pm$23.6}} & 70.3$\pm$28.0 & 70.0$\pm$24.5 & 72.5$\pm$27.9 & 80.0$\pm$22.0 \\ 
& Diff-UNet~\cite{xing2023diff} & 442 & 73.4$\pm$29.8 & 61.0$\pm$34.5 & 60.7$\pm$33.3 & 69.7$\pm$29.8 & 75.5$\pm$28.0 \\ 
\arrayrulecolor{black}\bottomrule
\end{tabular}
\begin{tablenotes}
    \item $^\dagger$These architectures were pre-trained (Appendix \ref{sec:implementation_details}).
    \item $^\ddagger$The class IVC (inferior vena cava) shares the same meaning as the class postcava in other datasets (e.g., \ourdataset\ and JHH).
    \item $^\bigstar$These architectures were trained on \ourdataset\ with enhanced label quality for the aorta and kidney classes (discussed in \S\ref{sec:conclusion}).
\end{tablenotes}
\label{tab:totalseg_train_plus_test_dsc}
\end{table*}

\begin{figure}[t]
	\centering
	\includegraphics[width=\columnwidth]{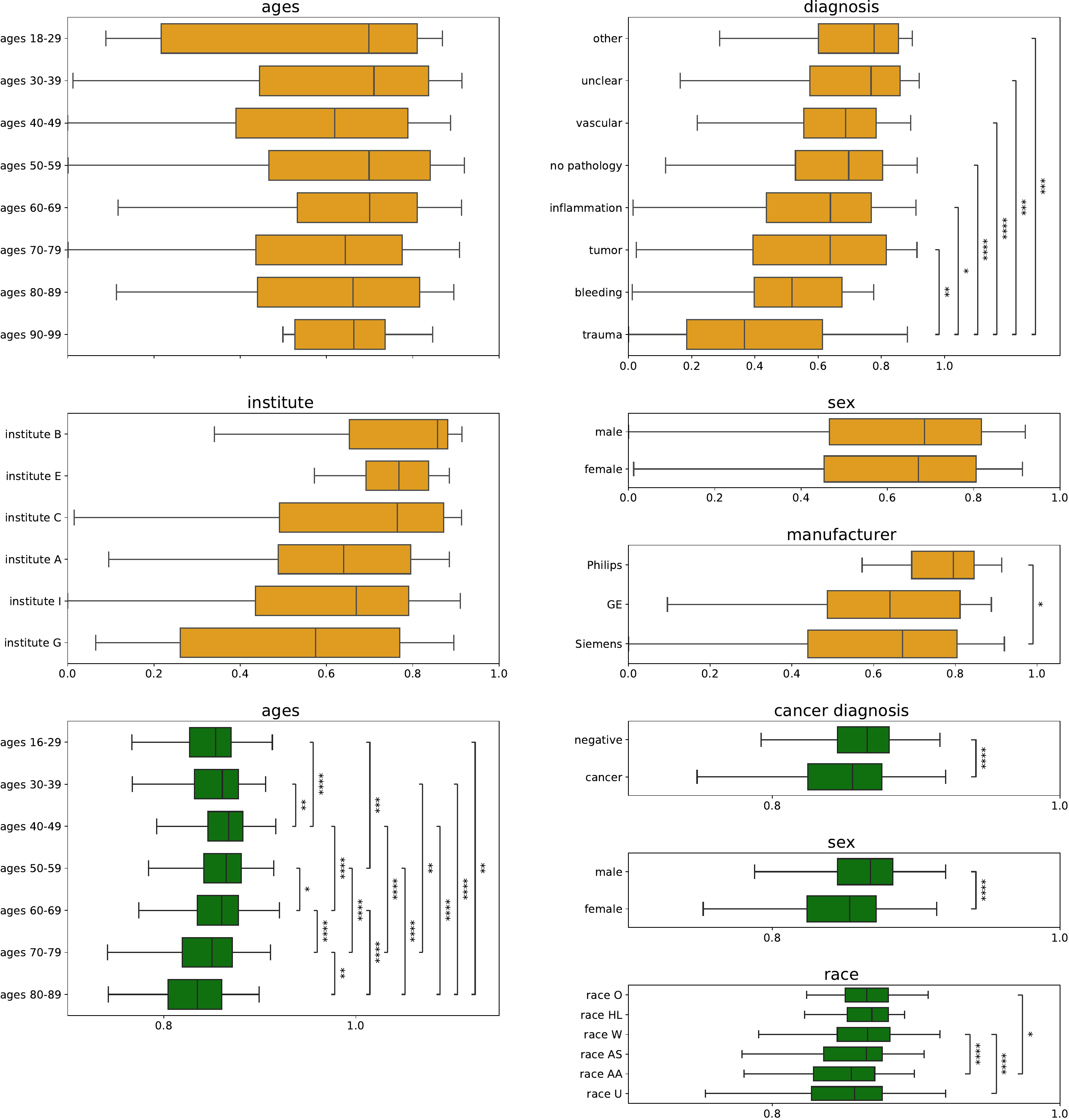}
    \caption{\textbf{Potential confounders significantly impact AI performance.} Boxplots showing the average DSC score of nine classes and \numofalgorithm\ algorithms for diverse demographic groups in two OOD test sets: \textcolor{myyellow}{TotalSegmentator} and \textcolor{mygreen}{JHH}. 
    Whiskers indicate 1.5\(\times\)IQR (interquartile range). Statistical significance is indicated by stars: $*~p<0.05$, $**~p<0.01$, $***~p<0.001$, $****~p<0.0001$. We perform Kruskal–Wallis tests followed by post-hoc Mann-Whitney U Tests with Bonferroni correction. Greater performance differences are observed in the JHH dataset compared to TotalSegmentator, likely due to the larger number of CT scans. Differences are apparent across demographic groups such as age, diagnoses, scanner manufacturer, sex, and medical institutions. Races HL, W, AS, AA, O, and U indicate Hispanic\&Latino, White, Asian, African American, other and unknown, respectively.}
	\label{fig:plot_groups}
\end{figure}

\section{Benchmark Results}\label{sec:result}

\subsection{Performances According to Out-of-distribution Evaluation on Large Datasets}
\label{sec:overall}


We started by comparing the average DSC score over the \numofclass\ classes. MedNeXt and MedFormer are the winners of the JHH dataset; STU-Net and ResEncL are the winners of the TotalSegmentator dataset. Among these winners, three are CNNs (STU-Net, ResEncL and MedNeXt) and one is a CNN Transformer hybrid (MedFormer). There is no significant difference among these winners at $p=0.05$ level, evidenced by the statistical analysis in Tables \ref{tab:jhh_dsc}--\ref{tab:totalseg_train_plus_test_dsc}. Regarding frameworks, nnU-Net \cite{isensee2021nnu} is the winner since 3 out of 4 of the aforementioned winners were developed on the self-configuring nnU-Net framework.

In addition to reporting the average performance ranking, we examined the \ul{per-class} performance and made the following findings.
\textit{\textbf{First}, diversified OOD evaluation is necessary.} For multiple algorithms, the DSC score for a given organ varied 15\% or more across diverse test sets. E.g., the SAM-Adapter, a transformer-based 2D model, generalizes much better to JHH than to TotalSegmentator: in kidney segmentation, its DSC score differs by more than 80\% across the datasets (see Appendix \ref{sec:sam} for explanations). Such stark performance variations reveal the importance of evaluating models on diverse OOD test sets. 
\textit{\textbf{Second}, test dataset size matters.}
More test samples increase statistical power, enabling benchmarks to more reliably detect differences between algorithms and produce stable, trustworthy rankings. Higher statistical power allows us to better differentiate the best performing model from the others: for JHH ($N$=5,160), there is at most four winners for any class, but for TotalSegmentator, there is up to eight (Tables \ref{tab:jhh_dsc}--\ref{tab:totalseg_train_plus_test_dsc}). Appendix \ref{sec:appendix_boxplots} uses box-plots and significance heatmaps \cite{wiesenfarth2021methods} to confirm these findings, and Appendix \ref{sec:supplementary_statistics} shows ranking order is much more stable for JHH than for smaller test sets. This finding emphasizes the importance of test dataset size for accurate and reliable algorithm comparisons. 
\textit{\textbf{Third}, average-based rankings are not enough.}
Tables \ref{tab:jhh_dsc}--\ref{tab:totalseg_train_plus_test_dsc} show that, for the same AI algorithm, DSC scores on difficult-to-segment structures, like the gallbladder and the pancreas, are usually 10--20\% lower than performance on easily identifiable structures, like the liver and the spleen. Usually, the best models for average DSC are also the best at individual structures, but per-class results reveal notable exceptions. E.g., in JHH, NexToU, a graph neural network-based hybrid architecture, excels at aorta segmentation, and Diff-UNet, a diffusion-based model, excels at kidney segmentation. Accordingly, per-class results reveal hidden strengths of AI algorithms. For a more comprehensive evaluation, Appendix \ref{sec:full_results} analyzes performance measured by NSD scores.
\textit{\textbf{Fourth}, inviting innovation is important.}
As in past 3D medical segmentation challenges \cite{antonelli2021medical}, CNNs with the nnU-Net framework \cite{isensee2021nnu} showed strong performance in our benchmark. However, by searching for innovative algorithms, sending target invitations to their inventors, and performing comprehensive evaluations, we could reveal strengths of new and less well known models, such as vision-language algorithms and Diff-UNet, the first 3D medical image segmentation method based on diffusion models, and MedFormer, a hybrid architecture that combines convolutional inductive bias with efficient, scalable bidirectional multi-head attention. Meanwhile, the LHU-Net, a hybrid architecture combining CNN and transformer attention mechanisms, excels in computational efficiency: it is 2 to 4 times faster than models with similar accuracy.

\subsection{Potential Confounders Significantly Impact AI Performance}
\label{sec:per_group_analysis}

We leveraged the metadata available in test datasets to assess AI' performance consistency across diverse demographic groups. We studied correlation between AI performance and the five types of metadata: age, sex, and diagnosis are analyzed on all two datasets, race is only analyzed on one dataset, JHH, since most public test sets lack this information, and manufacturer is only analyzed in one dataset.

Figure \ref{fig:plot_groups} displays per-group DSC for an average AI model, i.e., the average performance across our \numofalgorithm\ evaluated algorithms. The statistical analysis further highlights the need for large test datasets: JHH’s large sample size ($N$=5,160) allows detection of statistically significant DSC differences across all metadata, but some of these differences (for age and sex) are noticeable but not significant in the smaller TotalSegmentator dataset.
\textit{Notably, AI performance reduces for advanced \textbf{age}.} 
Median DSC starts dropping around the fifties. JHH shows multiple statistically significant performance drops after this age. The creators of the TotalSegmentator observed that aging caused attenuation in CT scans \cite{wasserthal2022totalsegmentator}, which may explain the common descending DSC trend after age 50, despite the fact that the 60-69 age group is the most populous in most datasets (Figure \ref{fig:general_meta}). This trend exists for all tested AI algorithms (Appendix \ref{sec:plots_per_group} displays per-group performances for each algorithm and organ). \textit{\textbf{Sex} only significantly confounds some AI algorithms.} The median DSC is significantly smaller for women in JHH. However, multiple top-performing models show no significant performance difference across sexes in any dataset (e.g., nnU-Net, STU-Net, and Diff-UNet), showcasing current AI can be robust to this confounder. \textit{We found significant performance differences for diverse \textbf{races}}. AI performance for white patients was significantly superior to the performance for African Americans, showing the need to increase the presence of this demographic group in public CT scan datasets. Again, many of the best performing algorithms did not present statistically significant differences for these two races (Appendix \ref{sec:plots_per_group}). \textit{In all datasets, diagnosis significantly impacted  AI performance.} Cancer patients have significantly smaller DSC scores in JHH ($p<0.0001$), and trauma patients have median DSC scores below other groups in TotalSegmentator. \textit{\textbf{Scanner manufacturer} changes cause significant DSC differences ($p<0.05$) in TotalSegmentator}. 

\section{Conclusion \& Discussion}\label{sec:conclusion}

\textbf{Conclusion.} \textit{Are we on the right way for evaluating AI algorithms for medical segmentation?} This paper outlines five properties of an ideal benchmark: (I) diverse data distribution in both training and test datasets, (II) a large number of test samples, (III) varied evaluation perspectives, (IV) equitably optimized AI algorithms, and (V) a long-term commitment. \ourproject\ sets itself apart from previous benchmarks in these criteria, enabling us to share unique insights that often missing in standard benchmarks. Our findings indicate: (1) AI performance can vary significantly across different datasets, with per-class differences of 10--20\% common, and up to 80\% observed (SAM-Adapter in kidney); thus, out-of-distribution evaluation across multiple datasets is crucial for ensuring AI's reliability and clinical adoption. (2) Larger test datasets reveal more significant differences between AI algorithms, allowing for meaningful rankings and nuanced analyses. (3) Average rankings can obscure AI's specific strengths; per-organ and metadata analysis is crucial in highlighting the benefits of innovative vision-language algorithms and the first diffusion-based 3D medical segmentation model. (4) By evaluating diverse AI architectures trained by their inventors, we establish a fair reference point for future development, which \ourproject\ will continually support with a long-term commitment.

\textbf{Label Noise in Training Set.} There is no perfect ground truth in segmentation datasets (except for synthetic data \cite{hu2023label,li2023early,chen2024towards,du2024boosting,chen2024analyzing,lai2024pixel,li2024text}), especially in the abdominal region where anatomical boundaries can be blurry due to disease or age (examples in Appendix~\ref{sec:dataset_visualization}). Identifying these boundaries is challenging for both human annotators and AI algorithms. Many recent datasets, including TotalSegmentator \cite{wasserthal2022totalsegmentator} and \ourdataset\ \cite{qu2023annotating}, use human-in-the-loop strategies, combining AI-predicted annotations and manual annotations by radiologists, which inevitably contain label errors. The errors in \ourdataset\ arise from poor CT image quality (e.g., BDMAP\_00000339, BDMAP\_00001044, BDMAP\_00003725), mistakes in AI predictions but not revised by humans, and inconsistency in label standards across the public datasets incorporated into \ourdataset\ \cite{li2024abdomenatlas}. With the feedback from our benchmark participants, we can \textit{partially} detect these label errors, primarily in the aorta (32.4\%), a structure with high annotation standard inconsistency in public data (e.g., in BTCV and FLARE) \cite{liu2023clip,liu2024universal}, and in the L\&R kidneys (2.6\%). We revised \ourdataset\ by reducing label errors in the aorta to 5.4\% and in the L\&R kidneys to 0.6\%. A ResEncL trained on the revised \ourdataset\ showed statistically significant performance gains in the aorta, but gains for kidneys were small and not always statistically significant (see Tables~\ref{tab:jhh_dsc}--\ref{tab:totalseg_train_plus_test_dsc}). These results highlight that current AI may be resistant to moderate levels of label noise (2.6\%), but not to high levels (32.4\%), as we detail in Appendix~\ref{sec:label_noise}. As future work, an improved label error detector will be a valuable tool for automatically assessing the quality of publicly available datasets and quickly improving quality through human annotation based on detected errors.

\textbf{High-Quality, Proprietary Test Set.} Having JHH ($N$=\numoftejhhct) available for third-party evaluation is a big plus for OOD benchmarks. It was completely annotated by radiologists, manually and following a well-defined annotation standard, for several years \cite{park2020annotated}. Thus, it can serve as a gold standard for our benchmark. The fact that JHH is a private dataset has both problems and benefits. It can significantly increases feedback time for AI performance evaluation, as it requires additional procedures to submit the AI to a third party, set it up, and run it on over 5,000 CT scans. If a benchmark takes too much work to run, it will not gain wide traction. But making test set (either images or annotations) publicly available can cause more problems—including completely destroying the OOD benchmark. For example, Medical Segmentation Decathlon (MSD) \cite{antonelli2021medical} was a benchmark with publicly accessible test images and its test annotations were private. Similarly, BTCV \cite{landman2015miccai} released both testing images and annotations. However, due to the growing need for more annotated data in the medical domain, even MSD/BTCV test sets have been annotated and integrated into recent public datasets, like FLARE \cite{ma2022fast,ma2023unleashing,ma2024automatic} and AbdomenAtlas \cite{qu2023annotating,li2024well,li2024abdomenatlas}. Therefore, any AI models trained or pre-trained on these public datasets are problematic in the MSD/BTCV leaderboard. With widespread access to test data, it becomes challenging to fairly compare models, as some may be overly optimized for the benchmark rather than for real-world performance. As a result, researchers must continue to seek or develop new datasets---preferably with images and annotations that have never been disclosed. This is critical in many fields as well. Yann Lecun---\textit{beware of testing on the training set}---in response to the incredible results achieved by GPT. Therefore, our proprietary JHH dataset is a valuable resource that other researchers can exploit to reduce data leakage risks and improve the reliability of OOD benchmark results. Our \ourproject\ Benchmark is still in the initial stage, so we are very careful with the decision of releasing JHH images/annotations. It must be managed carefully to ensure its benefits outweigh the risks.

\textbf{Per-Group Metadata Analysis.} Our study underscores the need for detailed metadata for algorithmic benchmark, which is currently a big limitation in the medical domain. Evidenced by Table \ref{tab:related}, only KiTS \& FLARE provided metadata analysis on sex, age, and/or race. Our \ourproject\ not only provides more extensive metadata analyses, including diagnosis, but also offers an order of magnitude more test data ($N$=\numoftect) for benchmarking. We have analyzed AI performance by metadata such as sex, age, and race but realized that a more rigorous analysis could be based on combined criteria (e.g., white females aged 30--40). Therefore, in the next round of benchmarking, instead of only providing average performance per class, we will also offer participants per-case performance along with each case's metadata information. This approach will provide a richer understanding of the pros/cons of AI algorithms and potentially stimulate AI innovation.

\textbf{Architectural Insights.} In Appendix \ref{sec:per_algorithm_analysis}, we have provided architectural comparison of both the top-ranking and bottom-ranking algorithms. But we find it difficult to extract trustworthy architectural insights directly from our current benchmark results. For example, Tables~\ref{tab:jhh_dsc}--\ref{tab:totalseg_train_plus_test_dsc} show that top performing models in our benchmark are usually CNNs within the nnU-Net framework. However, it is unclear if this is due to an intrinsic advantage of CNNs over Transformers or just an indication of nnU-Net’s superior pipeline configuration. Given that Transformers are newer, future frameworks, designed for them, could potentially enhance their performance. I.e., mature frameworks that extract the best from both CNNs and transformers should allow fairer architectural comparisons in the future. Beyond medical imaging, the architectural debate between CNNs and Transformers in computer vision has been ongoing and remains unresolved \cite{bai2021transformers,wang2022can}. Our benchmark provides `predictions-only' results, which can be heavily influenced by many factors such as preprocessing, data augmentation, post-processing, and training hyper-parameters \cite{isensee2021nnu}. To draw convincing architectural insights, extensive ablation studies under controlled settings are required. However, conducting ablation studies for all \numofalgorithm\ AI algorithms would be extremely costly for us.
We anticipate further insights and details from the AI inventors' upcoming technical reports, including extensive ablation studies. We are also happy to assist the inventors in their ablation studies by providing feedback on the OOD evaluation results of their algorithm variants.

With the success of the first edition of \ourproject\ Benchmark, we are actively pursuing multi-center, OOD datasets, to further enhance the benchmark. This is difficult for many well-known reasons---patient privacy, ethical compliance, data annotation, intellectual property, etc. \textit{Rome wasn't built in a day.} A multi-center, OOD dataset can never be made without accumulating the contribution of every single-center dataset. We hope this benchmark initiative at Johns Hopkins University, a highly regarded institution, could also inspire more institutes to contribute their private datasets for third-party OOD evaluation.

\clearpage
\section*{Acknowledgements and Disclosure of Funding}\label{sec:acknowledgement}

This work was supported by the Lustgarten Foundation for Pancreatic Cancer Research and the Patrick J. McGovern Foundation Award. We gratefully acknowledge the Data Science and Computation Facility and its HPC Support Team at Fondazione Istituto Italiano di Tecnologia. P.R.A.S.B. thanks the funding from the Center for Biomolecular Nanotechnologies, Istituto Italiano di Tecnologia (73010, Arnesano, LE, Italy). A.C. and S.D. thank the funding from the Istituto Italiano di Tecnologia (16163, Genova, GE, Italy). Z.W. and M.B.B. acknowledge support from the Research Foundation - Flanders (FWO) through project numbers G0A1319N and S001421N, and funding from the Flemish Government under the Onderzoeksprogramma Artifici\"{e}le Intelligentie (AI) Vlaanderen programme. Z.W. and M.B.B. acknowledge LUMI-BE for awarding this project access to the LUMI supercomputer, owned by the EuroHPC JU, hosted by CSC (Finland) and the LUMI consortium, and EuroHPC JU for awarding this project access to the Leonardo supercomputer, hosted by CINECA. Y.S., A.B., P.K., R.A. and D.M. acknowledge the scientific support and HPC resources provided by the Erlangen National High-Performance Computing Center (NHR@FAU) of the Friedrich-Alexander-Universit\"{a}t Erlangen-N\"{u}rnberg (FAU) under the NHR project "DeepNeuro - Exploring novel deep learning approaches for the analysis of diffusion imaging data." NHR funding is provided by federal and Bavarian state authorities. NHR@FAU hardware is partially funded by the German Research Foundation (DFG) – 440719683. Part of this work was funded by Helmholtz Imaging (HI), a platform of the Helmholtz Incubator on Information and Data Science. This work is partially funded by NSFC-62306046. We thank Thomas Brox for supporting the benchmark of the U-Net architecture.

We thank Di Liang for providing consultant of the statistical analysis in this benchmark; thank Zeyu Zhao for creating an live leaderboard of Touchstone; thank Xiaoxi Chen for analyzing AI predictions; thank Seth Zonies and Andrew Wichmann for providing legal advice on the release of AbdomenAtlas 1.0. The content of this paper covered by patents pending.

\clearpage
\bibliographystyle{plain}
{\small
  \bibliography{refs,zzhou}
}

\clearpage
\section*{Checklist}

\begin{enumerate}

\item For all authors...
\begin{enumerate}
  \item Do the main claims made in the abstract and introduction accurately reflect the paper's contributions and scope?
    \answerYes{}
  \item Did you describe the limitations of your work?
    \answerYes{See Section~\ref{sec:conclusion}.}
  \item Did you discuss any potential negative societal impacts of your work?
    \answerYes{See Section~\ref{sec:introduction} and Appendix~\ref{sec:negative_societal_impacts}.}
  \item Have you read the ethics review guidelines and ensured that your paper conforms to them?
    \answerYes{}
\end{enumerate}

\item If you are including theoretical results...
\begin{enumerate}
  \item Did you state the full set of assumptions of all theoretical results?
    \answerNA{}
	\item Did you include complete proofs of all theoretical results?
    \answerNA{}
\end{enumerate}

\item If you ran experiments (e.g. for benchmarks)...
\begin{enumerate}
  \item Did you include the code, data, and instructions needed to reproduce the main experimental results (either in the supplemental material or as a URL)?
    \answerYes{See the last sentence of Abstract.}
  \item Did you specify all the training details (e.g., data splits, hyperparameters, how they were chosen)?
    \answerYes{See Section~\ref{sec:dataset} and Appendix~\ref{sec:backbone_description}--\ref{sec:implementation_details}.}
	\item Did you report error bars (e.g., with respect to the random seed after running experiments multiple times)?
    \answerYes{See Table~\ref{tab:jhh_dsc} and Appendix~\ref{sec:full_results}.}
	\item Did you include the total amount of compute and the type of resources used (e.g., type of GPUs, internal cluster, or cloud provider)?
    \answerYes{See Section~\ref{sec:evaluation_protocols} and Appendix~\ref{sec:implementation_details}.}
\end{enumerate}

\item If you are using existing assets (e.g., code, data, models) or curating/releasing new assets...
\begin{enumerate}
  \item If your work uses existing assets, did you cite the creators?
    \answerYes{See Section~\ref{sec:dataset} for the use of existing datasets; Tables~\ref{tab:jhh_dsc}--\ref{tab:totalseg_train_plus_test_dsc} for the use of existing code and models. A more detailed description is given in Appendix~\ref{sec:backbone_description}--\ref{sec:implementation_details}.}
  \item Did you mention the license of the assets?
    \answerYes{See Section~\ref{sec:dataset}.}
  \item Did you include any new assets either in the supplemental material or as a URL?
    \answerYes{We have publicly released the \href{https://github.com/MrGiovanni/\ourproject}{evaluation code} used in our benchmark (given in the abstract) and provided the download link of our datasets, i.e., \href{https://huggingface.co/datasets/AbdomenAtlas/AbdomenAtlas1.0Mini}{\ourdataset} and \href{https://huggingface.co/datasets/MrGiovanni/AbdomenAtlas1.0C}{\ourdataset C} (given in Section~\ref{sec:dataset} and Appendix \ref{sec:label_noise}).}
  \item Did you discuss whether and how consent was obtained from people whose data you're using/curating?
    \answerYes{See Section~\ref{sec:dataset}.}
  \item Did you discuss whether the data you are using/curating contains personally identifiable information or offensive content?
    \answerYes{See Section~\ref{sec:dataset}.}
\end{enumerate}

\item If you used crowdsourcing or conducted research with human subjects...
\begin{enumerate}
  \item Did you include the full text of instructions given to participants and screenshots, if applicable?
    \answerNA{}
  \item Did you describe any potential participant risks, with links to Institutional Review Board (IRB) approvals, if applicable?
    \answerNA{}
  \item Did you include the estimated hourly wage paid to participants and the total amount spent on participant compensation?
    \answerNA{}
\end{enumerate}

\end{enumerate}

\clearpage

\appendix

\renewcommand \thepart{}
\renewcommand \partname{}

\part{Appendix} 
\setcounter{secnumdepth}{4}
\setcounter{tocdepth}{4}
\parttoc 

\clearpage
\section{Extensive Datasets in \ourproject}

\subsection{Construction of \ourdataset}\label{sec:abdomenatlas_construction}

\begin{table*}[h]
\caption{
\textbf{Public datasets composing \ourdataset\ and their details \cite{qu2023annotating}.} The naive aggregation of these public datasets results in a database with partial and incomplete labels, e.g., LiTS only had labels for the liver and its tumors, and KiTS only had labels for the kidneys and its tumors. Conversely, our \ourdataset\ is fully-annotated, offering detailed per-voxel labels for all \numofclass\ organs within each CT scan. We detected and removed duplicated CT scans across public datasets like LiTS and FLARE'23. Duplicate scans were identified by generating a 3D perceptual hash \cite{xu2015robust} for each image in the dataset. By comparing the similarity of these hashes, duplicates were reliably detected, a finding that was further confirmed through manual inspection of CT scans with high perceptual hash similarities. After aggregating all datasets and removing duplicates, we obtained a total of 5,195 fully-annotated CT scans.
}\vspace{2px}
\centering
\scriptsize
\begin{tabular}{p{0.22\linewidth}P{0.05\linewidth}P{0.06\linewidth}P{0.05\linewidth}P{0.12\linewidth}P{0.14\linewidth}P{0.16\linewidth}}
    \toprule
     Dataset & \makecell{\# of\\organs} & \makecell{\# of\\scans} & \makecell{\# of\\centers} & \makecell{source\\countries} & license & \makecell{\# of \textit{unique} scans\\in \ourdataset} \\
    \midrule
    \cellcolor{igray!5}1. Pancreas-CT \cite{roth2015deeporgan} &\cellcolor{igray!5} 1 & \cellcolor{igray!5}82 &\cellcolor{igray!5} 1 &\cellcolor{igray!5}US &\cellcolor{igray!5} CC BY 3.0 &\cellcolor{igray!5} 42 \\
    2. LiTS \cite{bilic2019liver} & 1 & 201 & 7 & \makecell{DE, NL, CA,\\FR, IL} &CC BY-SA 4.0 & 131 \\
    \cellcolor{igray!5}3. KiTS \cite{heller2020international} &\cellcolor{igray!5} 1 & \cellcolor{igray!5}300 &\cellcolor{igray!5} 50+ &\cellcolor{igray!5} US& \cellcolor{igray!5}CC BY-NC-SA 4.0 &\cellcolor{igray!5} 300 \\
    4. AbdomenCT-1K \cite{ma2021abdomenct} & 4 & 1,112 & 12 & \makecell{DE, NL, CA,\\FR, IL, US, CN} &CC BY-NC-SA & 1000 \\
    \cellcolor{igray!5}5. CT-ORG \cite{rister2020ct} &\cellcolor{igray!5} 5 &\cellcolor{igray!5} 140 &\cellcolor{igray!5} 8 &\cellcolor{igray!5}\makecell{DE, NL, CA,\\FR, IL, US} &\cellcolor{igray!5}CC BY 3.0 &\cellcolor{igray!5} 140 \\
    6. CHAOS \cite{valindria2018multi} & 4 & 40 & 1 &TR &CC BY-SA 4.0 & 20 \\
    \cellcolor{igray!5}7-11. MSD CT Tasks \cite{antonelli2021medical} & \cellcolor{igray!5}9 &\cellcolor{igray!5} 947 & \cellcolor{igray!5}1 &\cellcolor{igray!5}US &\cellcolor{igray!5}CC BY-SA 4.0 &\cellcolor{igray!5} 945 \\
    12. BTCV \cite{landman2015miccai} & 12 & 50 & 1 &US &CC BY 4.0 & 47 \\
    \cellcolor{igray!5}13. AMOS22 \cite{ji2022amos} & \cellcolor{igray!5}15 &\cellcolor{igray!5} 500 &\cellcolor{igray!5} 2 &\cellcolor{igray!5}CN &\cellcolor{igray!5}CC BY-NC-SA &\cellcolor{igray!5} 200 \\
    14. WORD \cite{luo2021word} & 16 & 150 & 1 &CN &GNU GPL 3.0 & 120 \\
    \cellcolor{igray!5}15. FLARE'23 &\cellcolor{igray!5} 13 &\cellcolor{igray!5} 4,000 &\cellcolor{igray!5} 30 &\cellcolor{igray!5}- &\cellcolor{igray!5}CC BY-NC-ND 4.0 &\cellcolor{igray!5} 2200\\
    16. AbdomenCT-12organ \cite{ma2023unleashing} & 12 & 50 & - & - &CC BY-NC-SA & 50 \\
    \bottomrule
\end{tabular}
\begin{tablenotes}
    \item US: United States \quad DE: Germany \quad NL: Netherlands \quad CA: Canada \quad FR: France \quad IL: Israel 
    \item CN: China \quad TR: Turkey \quad CH: Switzerland
\end{tablenotes}
\label{tab:public_dataset_list}
\end{table*}

\subsection{Domain Shift in TotalSegmentator}
\label{sec:TS}
\begin{table*}[h]
\caption{
\textbf{Percentage of Missing Classes in the two Partitions of TotalSegmentator.} Part of TotalSegmentator is included in the AbdomenAtlas dataset ($N$=485), because it is contained in FLARE, one of the AbdomenAtlas constituents. We leveraged the remaining sample of TotalSegmentator ($N$=743) for testing, providing a public test set anyone can easily use to compare segmentation models to Touchstone results. Unlike for the JHH test set, the hospitals in TotalSegmentator are present in AbdomenAtlas. However, the part of TotalSementator inside AbdomenAtlas ($N$=485) and the 743 test samples are not identically distributed. Table \ref{tab:percentage_empty_cell_counts} analyzes these two subsets, and shows that the one inside AbdomenAtlas was carefully selected to focus on the abdominal region, with a regular Region of Interest: almost all of these 485 images contain the 9 abdominal organs considered in this Touchstone. Conversely, the 743 TotalSegmentator images in our test set are more challenging, presenting varying regions of interest, which can extend outside of the abdomen and usually crop out some of the 9 classes in this benchmark. Therefore, Table \ref{tab:percentage_empty_cell_counts} demonstrates a substantial distribution shift between the two TotalSegmentator partitions, making our TotalSegmentator test images ($N$=743) out-of-distribution and a challenging test scenario. Interestingly, our results show this scenario was even more challenging to the AI algorithms than the JHH test set, which contains only images from an unseen hospital (see Sec. \ref{sec:result}).
}\vspace{2px}
\centering
\scriptsize
\begin{tabular}{p{0.15\linewidth}P{0.06\linewidth}P{0.06\linewidth}P{0.06\linewidth}P{0.06\linewidth}P{0.06\linewidth}P{0.06\linewidth}P{0.06\linewidth}P{0.06\linewidth}P{0.06\linewidth}}
\toprule
dataset & aorta & gallbladder & kidneyL & kidneyR & liver & spleen & stomach & pancreas & postcava \\
\midrule
AbdomenAtlas1.0 & 0\% & 3.9\% & 0.4\% & 0.4\% & 0\% & 0\% & 0\% & 0\% & 0\% \\
TotalSegmentator & 17.4\% & 81.8\% & 60.3\% & 63.0\% & 40.4\% & 60.3\% & 35.3\% & 47.2\% & 45.1\% \\
\bottomrule
\end{tabular}
\label{tab:percentage_empty_cell_counts}
\end{table*}

\clearpage
\subsection{Dataset Visualization by Metadata Information}\label{sec:dataset_visualization}

\begin{figure}[h]
	\centering
	\includegraphics[width=0.94\columnwidth]{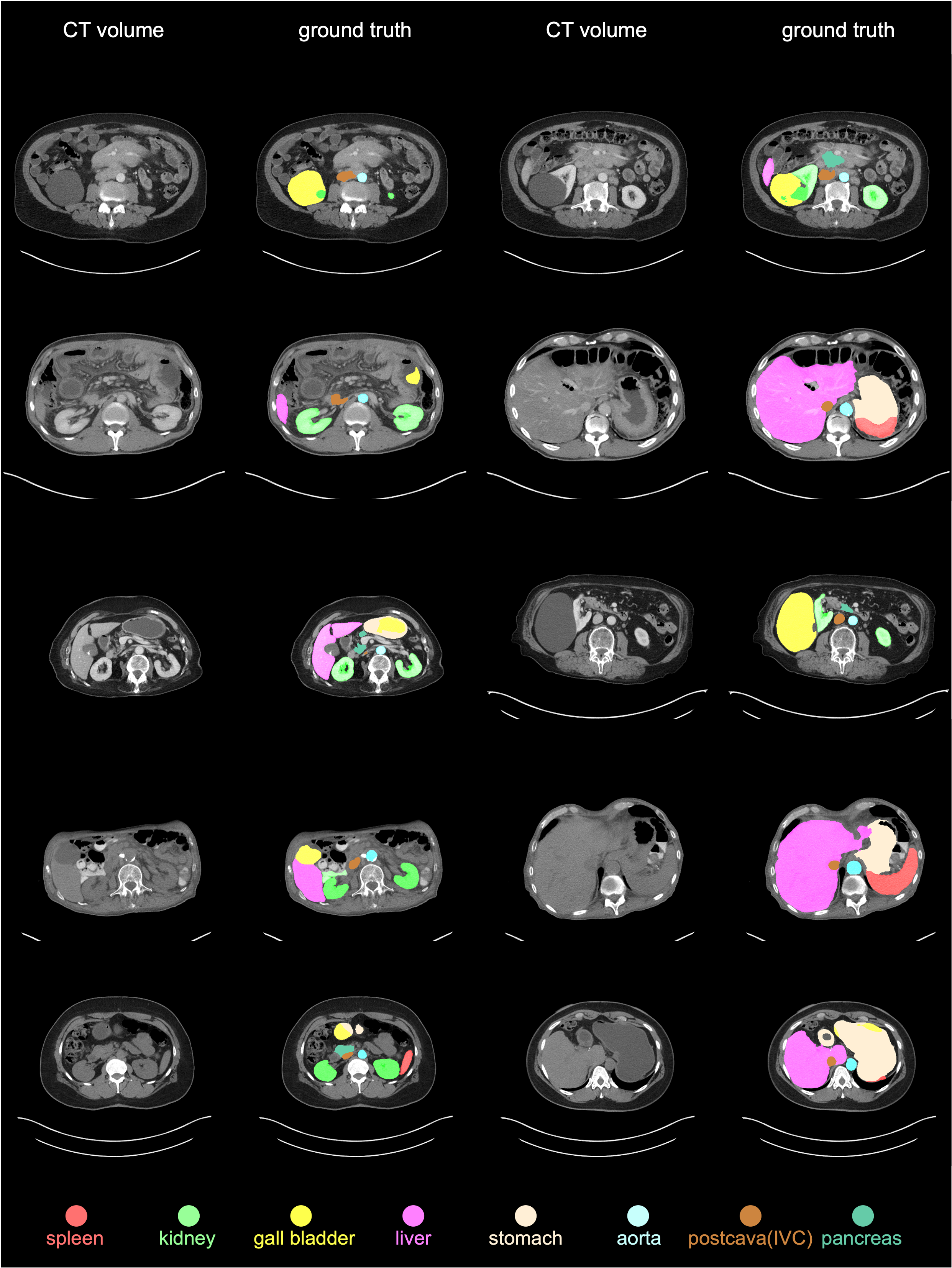}
    \caption{Anatomical boundaries and structures can be indistinct due to disease, as seen in the JHH dataset. We display CT volumes with patients depicted under unhealthy conditions that are challenging for most AI algorithms to identify. The CT volumes are from patients in unhealthy conditions. As shown in the first row on the left side, a kidney cyst is mistakenly annotated as the gall bladder. This example highlights that in the abdominal region, diseases can obscure anatomical boundaries and even lead to misidentification of structures.}
	\label{fig:dataset_visualization_jhh}
\end{figure}

\begin{figure}[h]
	\centering
	\includegraphics[width=0.94\columnwidth]{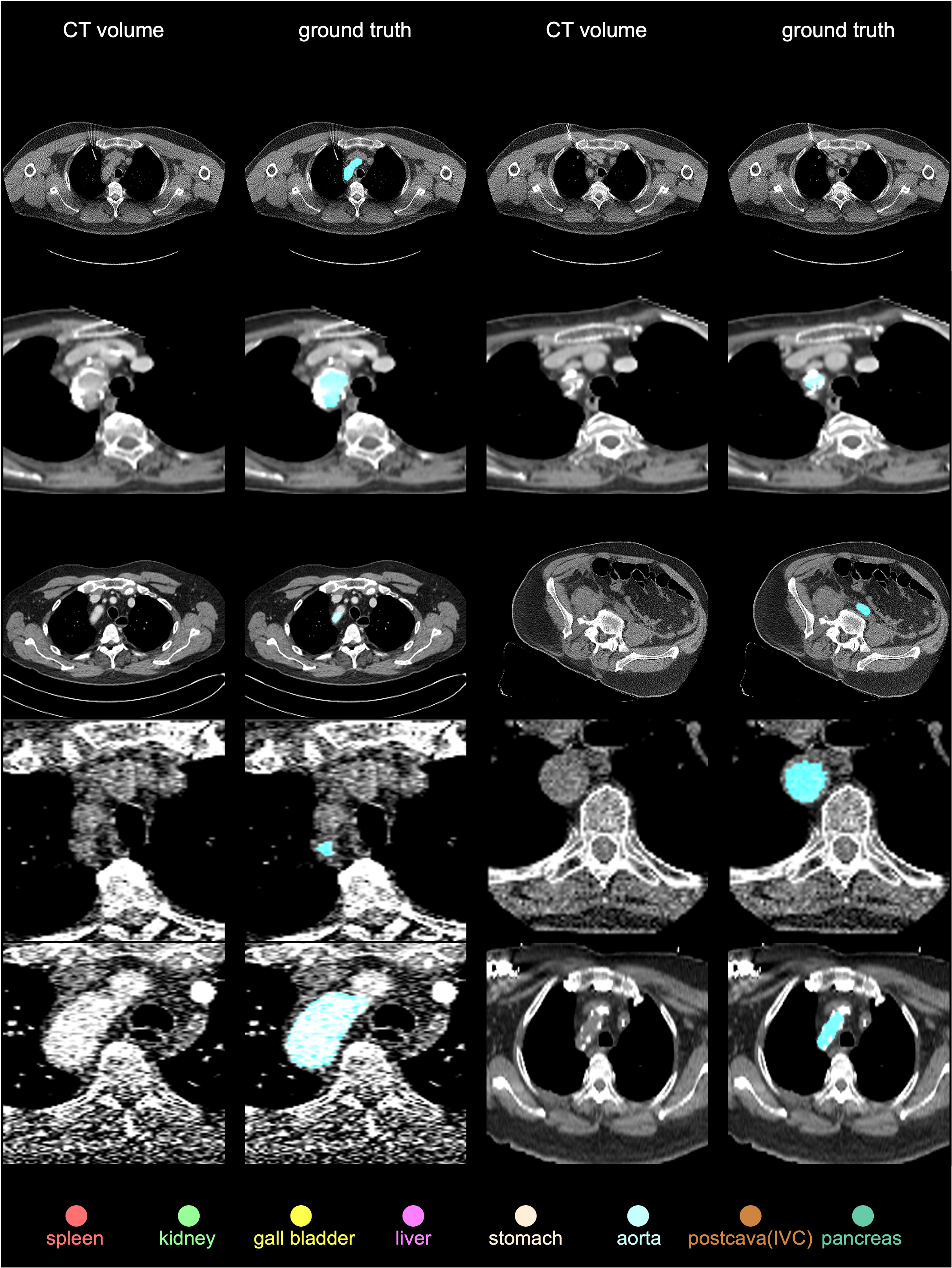}
    \caption{\textbf{Anatomical boundaries can be blurry due to factors such as patient disease, age, and CT scan quality in TotalSegmentator.} We display CT scans that are challenging for most AI algorithms to identify. The CT scans in the top three rows are from patients diagnosed with the tumors specified in the pathology metadata. The remaining images feature patients in their 70s and 80s. As shown in the fourth row on the right side, the boundary of the aorta in a 78-year-old patient is challenging to identify, not only for AI algorithms but also for human annotators in determining the ground truth.}
	\label{fig:dataset_visualization_totalsegmentator}
\end{figure}

\clearpage

\clearpage
\section{Extensive Number of AI Algorithms in \ourproject}
\subsection{Description of AI Architectures}\label{sec:backbone_description}

\subsubsection{Category CNN}

\textbf{U-Net.} The U-Net \cite{ronneberger2015u} is a fully-convolutional neural network, based on an encoder-decoder structure joint by multiple skip-connections. The encoder performs down-sampling operations, and it is designed to capture high-level semantics and context information. The decoder conducts up-sampling, and the long-range skip connections allow it to fuse the high-level semantics available at deep encoder layers, with the precise spatial information extracted from earlier encoder layers. The U-Net is the most influential architecture in biomedical segmentation; almost one decade after its release, the model is still the base of multiple novel architectures in this Benchmark.

\textbf{ResEncL.} nnU-Net \cite{isensee2021nnu,isensee2024nnu} is a self-configuring segmentation framework. It automatically configures pre-processing, network architecture, training and post-processing. Auto-configuration is guided by fixed parameters, interdependent rules that consider dataset properties and computational limitations, and empirical parameters. nnU-Net's default model has recently been updated to ResEncL default, which is based on a U-Net architecture with residual connections in its encoder \cite{isensee2024nnu}. The encoder is computationally expensive while the decoder is as lightweight as possible. For hyper-parameter configuration the nnU-Net default values are used except for the modality which was declared as ``nonCT'', resulting in z-score intensity normalization. ResEncL serves as a modernized nnU-Net baseline to compare new methodological innovations against.

\textbf{MedNeXt.} 
MedNeXt \cite{roy2023mednext} is a fully ConvNeXt-based 3D Encoder-Decoder Network designed to benefit from the scalability of Transformer-based networks while leveraging the inductive bias inherent to convolutions. This enables effective training on large datasets while still being beneficial on small data-scarce settings common to 3D medical image segmentation in the last decade. In the 3-layer residual structure of a MedNeXt block, the first layer computes features using a depthwise convolutional kernel, and it is followed by an expansion and compression layer, akin to a Swin Transformer. The architecture primarily benefits from using its MedNeXt blocks in all layers of the architecture, including up and downsampling blocks. The MedNeXt block enables effective representation learning in standard layers while allowing the network to maintain semantic richness in all resampling operations.

\textbf{STU-Net.} 
STU-Net \cite{huang2023stu} is a family of scalable and transferable medical image segmentation models based on the nnU-Net framework and the U-Net architecture. The STU-Net models introduce innovations such as refined convolutional blocks with residual connections for better scalability and weight-free interpolation for enhanced transferability. The models are available in different sizes: STU-Net-S with 14 million parameters, STU-Net-B (with 58.3M), STU-Net-L (440.3M), and STU-Net-H with 1.4 billion parameters. Improvements in segmentation accuracy stem from the empirical scaling of network depth and width. The primary goal of STU-Net is to enhance the scalability and transferability of medical image segmentation algorithms, facilitating their application across a variety of downstream tasks in transfer learning.

\textbf{UniSeg.} 
UniSeg \cite{ye2023uniseg} is a prompt-driven universal segmentation framework designed for multi-task medical image segmentation, offering transfer capabilities across various modalities and domains. Based on the nnU-Net framework, UniSeg has a vision encoder and a fusion module, which together enable a prompt-driven decoder. A key innovation of UniSeg is its universal learnable prompt that models complex inter-task relationships. UniSeg integrates task-specific prompts early in the training process, enhancing the training effectiveness of the entire decoder. The primary goal of UniSeg is not only to excel in multi-task learning but also to serve as a pre-trained model that improves the accuracy of downstream segmentation tasks. UniSeg was pre-trained (supervised) on 5 datasets before fine-tuning on \ourdataset: MOTS \cite{zhang2021dodnet,xie2023learning}, VerSe20 \cite{loffler2020vertebral}, Prostate\cite{liu2020ms}, BraTS21\cite{baid2021rsna}, and AutoPET2022 \cite{gatidis2022whole}.

\subsubsection{Category Transformer}

\textbf{UNETR.} 
UNETR \cite{hatamizadeh2022unetr} was proposed as a 3D transformer-based segmentation backbone network. The method leverages the Transformer model and CNN as a hybrid architecture, to capture long-range dependencies within volumetric medical data. The architecture integrates a Vision Transformer (ViT) as the encoder to handle the 3D input patches and extract rich feature representations. These features are then progressively merged with a convolutional neural network (CNN)-based decoder in a UNet-like structure.

\textbf{Swin UNETR.} 
SwinUNETR \cite{tang2022self} adapted Swin Transformers to enhance volumetric medical image segmentation by capturing both local and global features through a hierarchical, window-based self-attention mechanism, outperforming the original UNETR, and using Swin-transformers for global context. Additionally, self-supervised pre-training of Swin Transformers on large-scale unlabeled 3D medical images datasets, using techniques like masked autoencoding, can significantly boost the model's robustness and performance on downstream tasks. These features enabled leading results in various 3D medical image analysis applications, especially in CT segmentation tasks.

\textbf{UNEST.}
UNEST~\cite{yu2023unest} is an advanced 3D segmentation model designed to leverage the strengths of the hierarchical vision transformer architecture for handling 3D medical image data. UNEST also employed a U-shape encoder-decoder structure, where the encoder is based on the 2-stage nested ViT. This transformer-based encoder extracts hierarchical features from the input CT scan using self-attention mechanisms, which capture long-range dependencies and spatial relationships efficiently. The decoder consists of 4-levels of CNN-based blocks that reconstruct the segmentation map by upsampling the features and incorporating skip connections from the encoder to retain spatial information. The model's architecture and training protocol are optimized to provide a robust and efficient solution for 3D segmentation tasks such as whole body, regional, and whole brain segmentation.

\textbf{SegVol.}
SegVol \cite{du2023segvol} is based on the SAM architecture \cite{kirillov2023segment} and 3D transformers, enabling universal and interactive volumetric medical image segmentation on over 200 anatomical categories. SegVol supports spatial-prompt, semantic-prompt, and combined-prompt segmentation, aiming for high-precision segmentation and semantic disambiguation. SegVol introduced a zoom-out-zoom-in mechanism to provide users with an easy SAM-like interface on volumetric images, while significantly reducing computational cost and preserving the segmentation precision. Pseudo labels are used to relieve the problem of spurious correlation between predictions and data distributions. Prior to training on AbdomenAtlas, SegVol was pre-trained on 90K unlabeled CT scans from M3D-Cap, and 5,772 labeled CT scans from M3D-Seg \cite{bai2024m3d}.

\textbf{SAM-Adapter.} The SAM-Adapter \cite{gu2024build} is a 2D segmentation model, unlike the other networks in this study. Thus, it individually analyzes the 2D slices that compose a CT scan. The model is based on fine-tuning the MobileSAM \cite{zhang2023faster} encoder and decoder using Adapter layers. The SAM-Adapter follows the philosophy that model size has limited effect over the accuracy of medical segmentation algorithms \cite{gu2024build}.

\subsubsection{Hybrid Architectures}

\textbf{LHU-Net.} 
LHU-Net \cite{sadegheih2024lhu} is a compact and efficient U-Net-based architecture created for 3D medical image segmentation. It utilizes a hierarchical encoder-decoder structure with convolutional layers followed by hybrid attention mechanisms to capture both local and global features. Key innovations include the integration of CNN-based spatial attention and Vision Transformer (ViT) attention mechanisms, such as the OmniFocus attention and self-adaptive contextual fusion modules, which enhance discriminative feature extraction while keeping the model lightweight. These attention mechanisms' objective is to ensure high precision and detail in the segmentation results. The main aim of LHU-Net is to achieve high segmentation accuracy with minimal computational resources and parameters, making it a practical and accessible tool for medical imaging tasks.

\textbf{UCTransNet.} 
UCTransNet \cite{wang2022uctransnet} is a hybrid architecture, based on U-Net with transformer blocks as skip connections. It introduces the Channel-wise Cross Fusion Transformer (CCT) to fuse multi-scale context with cross attention from a channel-wise perspective. CCT captures local cross-channel interaction for adaptive fusing of multi-scale features with possible scale semantic gap. Additionally, a channel-wise cross attention (CCA) module is proposed for fusing features from decoder stages and fused multi-scale features to solve inconsistent semantic levels. Both cross attention modules are called CTrans and replace the original skip connections in the U-Net. Here, the UCTransNet 2D components were substituted by their 3D versions, including convolution layers, patch embedding layers, and patch merging layers. The main goal is to discover an efficient approach for integrating CNNs and Transformers for medical image segmentation.

\textbf{Diff-UNet.} 
Diff-UNet \cite{xing2023diff} is the first generic 3D medical image segmentation model based on a denoising diffusion model. It mainly consists of two branches: the boundary prediction branch and the diffusion denoising branch. The boundary prediction branch is based on the U-Net structure, while the diffusion denoising branch is based on a denoising U-Net structure with noise input. To aggregate the low-level and high-level features from both branches for better boundary perception, Diff-UNet also includes a Multi-granularity Boundary Aggregation (MBA) module. Next, Diff-U-Net proposes a Monte Carlo Diffusion (MC Diffusion) module to obtain uncertainty maps and guide segmentation loss to focus on hard-to-segment regions during training. Finally, Diff-UNet devises a Progressive Uncertainty-driven REfinement (PURE) strategy to obtain a more robust prediction result during inference, based on the inference steps and uncertainty maps estimated by the MC Diffusion module.

\textbf{NexToU.} 
NexToU \cite{shi2023nextou} is a hybrid architecture that follows a hierarchical 3D U-shaped encoder-decoder structure, based on CNNs and graph neural networks (GNNs). It incorporates a hierarchical, topology-aware strategy inspired by human cognitive processes, progressively decomposing anatomical semantics from simpler to more complex structures. Concurrently, it also learns containment, connection, and exclusion relationships among various anatomical classes. To facilitate learning and speed up training, NexToU employs a semantic tree and a novel hierarchical topological interaction (HTI) module. Additionally, it enhances spatial topology perception by incorporating Vision GNN \cite{han2022vision} and Swin GNN modules, which adeptly represent topology on both global and local scales. The primary goal of NexToU's innovations is to improve segmentation accuracy for homogeneous multi-class anatomical structures, such as vasculature and skeletons. The HTI module is designed to be more effective when dealing with a large number of classes.

\textbf{MedFormer.}
MedFormer~\cite{gao2022data} is a hybrid architecture that combines the inductive bias of convolution with the global modeling capabilities of Transformers. A key innovation in the design is the bidirectional multi-head attention (B-MHA) mechanism, which addresses the quadratic complexity typically associated with self-attention on long sequences. B-MHA employs a low-rank projection mechanism to achieve linear complexity attention, making it computationally efficient for both low- and high-resolution feature maps. Furthermore, B-MHA's architecture captures the most salient features in its hidden state, enhancing model robustness by reducing focus on irrelevant details. Through this design, MedFormer demonstrates good scalability, efficiency, and generalizability, performing effectively on both small and large datasets without requiring pre-trained weights.

\clearpage
\subsection{Description of AI Frameworks}
\label{sec:framework_description}

\subsubsection{nnU-Net}

nnU-Net \cite{isensee2021nnu} is a framework for automatically configuring AI-based semantic segmentation pipelines. Given a new segmentation dataset, it will extract relevant metadata from the training cases to automatically determine its hyperparameters. Despite its first release dating back to 2019 and despite its use of a standard U-Net \cite{ronneberger2015u}, it stood the test of time and continues to produce state-of-the-art results. nnU-Net powerfully demonstrates that carefully configuring and validating segmentation pipelines across a wide range of segmentation tasks yields a surprisingly potent algorithm. As a framework for method development, it is widely used and extended by the community to push the boundaries of semantic segmentation \cite{ye2023uniseg,huang2023stu,roy2023mednext,shi2023nextou,ulrich2023multitalent,isensee2023extending}. A recent update to the nnU-Net presets \cite{isensee2024nnu} includes reference implementations for a U-Net with residual connections in the encoder, optimized for different VRAM budgets.

\subsubsection{MONAI}
MONAI (Medical Open Network for AI) \cite{cardoso2022monai} is an open-source framework designed to support artificial intelligence in healthcare data. Built on top of PyTorch, MONAI facilitates a comprehensive suite of tools for configuring, training, inference, and deploying AI models tailored to medical applications. It includes components for data loading, preprocessing, and augmentation, as well as prebuilt architectures for common tasks such as segmentation, registration, detection, and classification. MONAI is designed to be flexible, extensible, and performance-optimized, enabling researchers and practitioners to accelerate their AI development cycle in the medical domain.

\subsubsection{Vision-Language Models}

\textbf{CLIP-Driven Universal Model.} 
The CLIP-Driven Universal Model~\cite{liu2023clip} framework, which is designed for organ and tumor segmentation, integrates a label taxonomy from various public datasets. The architecture consists of a text branch and a vision branch. In the text branch, the model generates CLIP embeddings for each organ and tumor using label prompts, enhancing the anatomical structure of the feature embedding. These embeddings are concatenated with global image features, termed the text-based controller, to produce prompt features for segmentation. The vision branch pre-calculates CT scans to mitigate domain gaps across different datasets. These extracted features are processed by three sequential convolutional layers, referred to as the text-driven segmentor, which utilize the parameters generated by the text branch to predict segmentation masks for each class. The decoder also includes a "one vs. all" approach, using Sigmoid activation for each class to generate individual predictions, ensuring robust and dynamic segmentation across diverse medical imaging datasets.

\clearpage
\subsection{Implementation and Configuration Details}\label{sec:implementation_details}

Tables~\ref{tab:algorithm_details}-\ref{tab:training_details} present details on the algorithms we benchmarked, and on their training configurations, respectively.

\begin{table*}[h]
\caption{
\textbf{Details on the AI algorithms and speed.}
}\vspace{2px}
\centering
\label{tab:algorithm_details}
\scriptsize
\begin{tabular}{p{0.12\linewidth}p{0.17\linewidth}p{0.08\linewidth}p{0.10\linewidth}P{0.10\linewidth}P{0.13\linewidth}}
\toprule
framework & architecture & parameters & category & inference time ($\mu s/mm^3$)$^\dagger$ & inference memory (average)$^\dagger$ \\ 
\midrule
\multirow{6}{*}{nnU-Net}
& UniSeg & 31.0M & CNN & 5.04 & 3.9 GB \\
& MedNeXt & 61.8M & CNN & 3.25 & 4.3 GB \\
& NexToU & 81.9M & Hybrid & 1.53 & 1.9 GB \\
& STU-Net-B & 58.3M & CNN & 2.39 & 2.0 GB \\
& STU-Net-L & 440.3M & CNN & 5.6 & 5.4 GB \\
& STU-Net-H & 1457.3M & CNN & 13.66 & 12.5 GB \\
& U-Net & 31.1M & CNN & 0.94 & 1.9 GB \\
& ResEncL & 102.0M & CNN & 1.26 & 3.7 GB \\
\midrule
\multirow{2}{*}{Vision-Language}
& U-Net \& CLIP & 19.1M & Hybrid & 1.84 & 8.0 GB \\
& SwinUNETR \& CLIP &62.2M & Hybrid & 1.65 & 7.5 GB \\
\midrule
\multirow{6}{*}{MONAI}
& LHU-Net & 8.6M & Hybrid & 0.44 & 0.6 GB \\
& UCTransNet & 68.0M & Hybrid & 0.86 & 2.8 GB \\
& SwinUNETR &72.8M & Hybrid & 0.45 & 4.2 GB \\
& UNesT &87.2M & Hybrid & 0.37 & 2.4 GB \\
& UNETR &101.8M & Hybrid & 0.37 & 2.4 GB \\
& SegVol & 181.0M & Transformer & 0.52 & 0.8 GB \\
\midrule
\multirow{3}{*}{n/a}
& SAM-Adapter & 11.6M & Transformer & 0.61 & 0.5 GB \\
& MedFormer & 38.5M & Hybrid & 1.87 & 2.8 GB \\
& Diff-UNet & 434.0M & Hybrid & 2.26 & 3.9 GB \\
\bottomrule
\end{tabular}
\begin{tablenotes}
    \item $^\dagger$ The time and average GPU memory for inference were measured with an NVIDIA V100 GPU and an Intel Xeon Silver 4210 CPU, evaluating a CT scan with 259$\times$259$\times$283 voxels and spacing of 1.5 mm/voxel. Measurements consider the entire segmentation pipeline, from loading the CT scan and the AI algorithm, to saving the inference. We observed that the way each AI algorithm deals with spacing and re-shapes its input scan plays a major role in their inference speed.
\end{tablenotes}
\end{table*}

\begin{table*}[h]

\caption{
\textbf{Training configuration on \ourdataset.}
}\vspace{2px}
\centering
\label{tab:training_details}
\scriptsize
\begin{tabular}{p{0.16\linewidth}p{0.08\linewidth}p{0.06\linewidth}p{0.03\linewidth}p{0.11\linewidth}p{0.1\linewidth}p{0.245\linewidth}}
\toprule
architecture & pre-trained & iterations$^\dagger$ & hours & GPU$^\ddagger$ & GPU memory & hyper-parameter \\ 
\midrule
UniSeg & Yes & 2M & 186 & 1$\times$RTX 3090 & 8.2 GB & Self-configuring \\
MedNeXt & No & 250K & 67 & 4$\times$A100 & 17.6 GB & Manual trial-and-error \\
NexToU & No & 500K & 186 & 1$\times$RTX 3090 & 17.2 GB & Self-configuring \\
STU-Net-B & No & 500K & 30 & 1$\times$A100 & 8.8 GB & Self-configuring \\
U-Net & No & 250k & 7.5 & 1$\times$A100 & 7 GB & Self-configuring \\
ResEncL & No & 250K & 28 & 1$\times$A100 & 24 GB & Self-configuring \\
U-Net \& CLIP & No & 200K & 120 & 8$\times$RTX 8000 & 12GB & Self-configuring \\
SwinUNETR \& CLIP & No & 200K & 120 & 4$\times$A100 & 48 GB  & Self-configuring \\
LHU-Net & No & 250K & 40 & 1$\times$A100 & 8 GB & Pre-defined, from \cite{sadegheih2024lhu,li2024well} \\
UCTransNet & No & 200K & 20 & 2$\times$A100 & 16 GB & Self-configuring \\
SwinUNETR & Yes & 250k & 24 & 8$\times$V100 & 32 GB & Self-configuring \\
UNesT & No & 250k & 24 & 8$\times$V100 & 16 GB & Self-configuring \\
UNETR & No & 250k & 24 & 8$\times$V100 & 12 GB & Self-configuring \\
SegVol & Yes & 18.75K & 60 & 8$\times$A800 & 50 GB & Manual trial-and-error \\
SAM-Adapter & Yes & 32.5K & 170 & 1$\times$RTX A6000 & 37 GB & Pre-defined, from \cite{gu2024build} \\
MedFormer & No & 300K & 72 & 16$\times$V100 & 27.5 GB & Pre-defined, Manual trial-and-error \\
Diff-UNet & No & 500K & 48 & 1$\times$RTX 4090 & 16 GB & Self-configuring \\
\bottomrule
\end{tabular}
\begin{tablenotes}
    \item $^\dagger$1 iteration is 1 batch, not a full iteration over all dataset.
    \item $^\ddagger$GPU: number of GPUs used for training $\times$ specific (NVIDIA) GPU.
\end{tablenotes}
\end{table*}

\begin{table}[]
\caption{
\textbf{Additional Training Hyper-parameters.}
}\vspace{2px}
\centering
\label{tab:training_details}
\scriptsize
\begin{tabular}{lllllll}
\toprule
architecture & patch size & batch size & optimizer & learning rate & loss function & WD \\ \hline

UniSeg & {[}48, 160, 224{]} & 2 & SGD & 0.01, PolyLRScheduler & Dice, CE & 3.00E-05 \\
MedNeXt & {[}128, 128, 128{]} & 8 & AdamW & 1.00E-03 & Dice, CE & 3.00E-05 \\
NexToU & {[}96, 160, 160{]} & 2 & SGD & 0.01, PolyLRScheduler & Dice, CE, HTI & 3.00E-05 \\
STU-Net-B & {[}80, 128, 192{]} & 2 & SGD & 0.01, PolyLRScheduler & Dice, CE & 3.00E-05 \\
U-Net & {[}64, 160, 192{]} & 2 & SGD & 0.01, PolyLRScheduler & Dice, CE & 3.00E-05 \\
ResEncL & {[}96, 192, 288{]} & 2 & SGD & 0.01, PolyLRScheduler & Dice, CE & 3.00E-05 \\
U-Net \& CLIP & {[}96,96,96{]} & 2 & AdamW & 1,00E-4, cosineScheduler & Dice, BCE & 1.00E-05\\
SwinUNETR \& CLIP & {[}96,96,96{]} & 2 & AdamW & 1,00E-4, cosineScheduler & Dice, BCE & 1.00E-05\\
LHU-Net & {[}96,96,96{]} & 2 & SGD & 0.01 & Dice, CE & 1,00E-05 \\
UCTransNet & [128,128,128] & 4 & AdamW & 1.00E-04 & Dice, CE & 1.00E-04 \\
SwinUNETR & {[}96,96,96{]} & 2 & AdamW & 1.00E-3, cosineScheduler & Dice, CE & 1.00E-05 \\
UNesT & {[}96,96,96{]} & 2 & AdamW & 1,00E-3, cosineScheduler & Dice, CE & 1.00E-05 \\
UNETR & {[}96,96,96{]} & 2 & AdamW & 1,00E-3, cosineScheduler & Dice, CE & 1.00E-05 \\
SegVol & {[}4, 16, 16{]} & 64 & AdamW & 1,00E-04 & Dice, BCE & 1.00E-05 \\

SAM-Adapter & {[}1, 1024, 1024{]} & 32 & AdamW & 1.00E-03, warmup      & Dice, CE & 0.1 \\
MedFormer & {[}128,128,128{]} & 32 & AdamW & 6.00E-4 & Dice, CE & 5.00E-2 \\
Diff-UNet & {[}128,128,128{]} & 2 & SGD & 0.01, PolyLRScheduler & CE & 1.00E-03 \\
\bottomrule
\end{tabular}
\end{table}

\clearpage

\clearpage
\section{Extensive Results on Four Test Datasets}\label{sec:full_results}
\subsection{NSD scores on the entire JHH dataset}\label{sec:entire_JHH}

\begin{table*}[h]
\caption{
\textbf{External validation on proprietary JHH dataset ($N$=\numoftejhhct) - NSD.} For each class, we bold the best-performing results and highlight the runners-up, which show no significant difference ($P>0.05$) from the best results, in red. Architectures are grouped by their frameworks and sorted in ascending order based on the number of parameters. NSD considers a tolerance of 1.5mm.
}\vspace{2px}
\centering
\scriptsize
\begin{tabular}{p{0.12\linewidth}p{0.20\linewidth}p{0.05\linewidth}P{0.075\linewidth}P{0.075\linewidth}P{0.075\linewidth}P{0.075\linewidth}P{0.075\linewidth}}
\toprule
framework & architecture & param & spleen & kidneyR & kidneyL & gallbladder & liver \\
\midrule
\multirow{9}{*}{nnU-Net}
& UniSeg$^\dagger$~\cite{ye2023uniseg} & 31.0M & 88.8$\pm$9.7 & 79.8$\pm$10.5 & 78.7$\pm$9.8 & 75.6$\pm$16.8 & 79.5$\pm$8.9 \\ 
& MedNeXt~\cite{roy2023mednext} & 61.8M & 88.9$\pm$10.3 & 80.0$\pm$11.2 & 78.8$\pm$10.3 & 75.2$\pm$17.5 & 79.0$\pm$9.3 \\ 
& NexToU~\cite{shi2023nextou} & 81.9M & 88.2$\pm$11.6 & 75.7$\pm$13.0 & 75.1$\pm$11.7 & 72.2$\pm$20.6 & 76.2$\pm$10.3 \\ 
& STU-Net-B~\cite{huang2023stu} & 58.3M & 88.7$\pm$10.4 & 80.2$\pm$11.1 & 79.3$\pm$10.2 & 75.6$\pm$16.8 & 78.6$\pm$9.4 \\ 
& STU-Net-L~\cite{huang2023stu} & 440.3M & 89.1$\pm$10.1 & 79.7$\pm$11.2 & 79.0$\pm$10.2 & 76.1$\pm$16.9 & 79.0$\pm$9.3 \\ 
& STU-Net-H~\cite{huang2023stu} & 1457.3M & 89.1$\pm$10.0 & 80.1$\pm$10.9 & 79.2$\pm$10.2 & 76.8$\pm$16.6 & 79.4$\pm$9.3 \\ 
& U-Net~\cite{ronneberger2015u} & 31.1M & 88.6$\pm$10.5 & 79.9$\pm$11.1 & 79.1$\pm$10.4 & 73.6$\pm$17.9 & 78.1$\pm$9.5 \\ 
& ResEncL~\cite{isensee2021nnu,isensee2024nnu} & 102.0M & 89.0$\pm$10.3 & 80.3$\pm$11.0 & 79.1$\pm$10.2 & 74.1$\pm$18.1 & 78.9$\pm$9.5 \\ 
& \textcolor{lightgray}{ResEncL$^\bigstar$} & \textcolor{lightgray}{102.0M} & \textcolor{lightgray}{88.6$\pm$10.4} & \textcolor{lightgray}{80.0$\pm$11.1} & \textcolor{lightgray}{78.8$\pm$10.3} & \textcolor{lightgray}{74.0$\pm$17.9} & \textcolor{lightgray}{78.9$\pm$9.5} \\ 
\hline
\multirow{2}{*}{Vision-Language}
& U-Net \& CLIP~\cite{liu2023clip} & 19.1M & 86.5$\pm$10.8 & 78.7$\pm$10.2 & 78.7$\pm$10.4 & 71.4$\pm$18.5 & 77.8$\pm$8.9 \\ 
& Swin UNETR \& CLIP~\cite{liu2023clip} & 62.2M & 86.0$\pm$11.4 & 79.0$\pm$11.1 & 78.1$\pm$10.7 & 70.2$\pm$20.4 & 78.1$\pm$9.8 \\ 
\hline
\multirow{6}{*}{MONAI}
& LHU-Net~\cite{sadegheih2024lhu} & 8.6M & 87.1$\pm$10.9 & 79.1$\pm$10.8 & 78.7$\pm$10.1 & 73.0$\pm$18.1 & 77.8$\pm$9.1 \\ 
& UCTransNet~\cite{wang2022uctransnet} & 68.0M & 78.7$\pm$16.0 & 73.3$\pm$15.5 & 73.3$\pm$13.5 & 66.0$\pm$21.8 & 71.4$\pm$11.6 \\ 
& Swin UNETR~\cite{tang2022self} & 72.8M & 80.5$\pm$13.4 & 73.7$\pm$13.1 & 74.6$\pm$12.2 & 62.5$\pm$20.6 & 73.7$\pm$9.6 \\
& UNesT~\cite{yu2023unest} & 87.2M & 80.7$\pm$12.4 & 72.6$\pm$12.2 & 72.2$\pm$12.1 & 57.8$\pm$20.1 & 73.3$\pm$9.1 \\ 
& UNETR~\cite{hatamizadeh2022unetr} & 101.8M & 78.4$\pm$15.0 & 73.2$\pm$12.3 & 72.8$\pm$12.5 & 59.2$\pm$21.4 & 73.1$\pm$9.6 \\ 
& SegVol$^\dagger$~\cite{du2023segvol} & 181.0M & 86.7$\pm$11.1 & 80.2$\pm$10.5 & 79.2$\pm$9.9 & 68.5$\pm$20.7 & 77.9$\pm$9.7 \\ 
\hline
\multirow{3}{*}{n/a}
& SAM-Adapter$^\dagger$~\cite{gu2024build} & 11.6M & 70.9$\pm$15.2 & 70.0$\pm$11.6 & 66.2$\pm$11.8 & 19.8$\pm$11.7 & 62.3$\pm$9.7 \\ 
& MedFormer~\cite{gao2022data} & 38.5M & \cellcolor{red!20}{\textbf{91.3$\pm$9.5}} & \cellcolor{red!20}{\textbf{83.0$\pm$10.3}} & \cellcolor{red!20}{\textbf{80.7$\pm$9.7}} & \cellcolor{red!20}{\textbf{77.3$\pm$17.0}} & \cellcolor{red!20}{\textbf{81.2$\pm$9.1}} \\ 
& Diff-UNet~\cite{xing2023diff} & 434.0M & 88.7$\pm$10.7 & 81.0$\pm$11.0 & 79.5$\pm$10.4 & 72.1$\pm$18.9 & 78.2$\pm$9.5 \\ 
\bottomrule
framework & architecture & param & stomach & aorta & postcava & pancreas & average \\
\midrule
\multirow{9}{*}{nnU-Net}
& UniSeg$^\dagger$~\cite{ye2023uniseg} & 31.0M & 72.4$\pm$11.2 & 78.3$\pm$13.2 & 70.2$\pm$10.8 & 69.9$\pm$11.1 & 77.0$\pm$11.4 \\ 
& MedNeXt~\cite{roy2023mednext} & 61.8M & 71.5$\pm$11.8 & 80.2$\pm$12.9 & 70.8$\pm$11.0 & 69.3$\pm$11.7 & 77.1$\pm$11.8 \\ 
& NexToU~\cite{shi2023nextou} & 81.9M & 70.0$\pm$12.7 & \cellcolor{red!20}{\textbf{83.8$\pm$11.9}} & 66.2$\pm$11.2 & 68.6$\pm$14.1 & 75.1$\pm$13.0 \\ 
& STU-Net-B~\cite{huang2023stu} & 58.3M & 70.5$\pm$12.1 & 78.3$\pm$13.4 & 70.5$\pm$10.9 & 69.0$\pm$11.5 & 76.8$\pm$11.8 \\ 
& STU-Net-L~\cite{huang2023stu} & 440.3M & 71.7$\pm$12.0 & 77.4$\pm$13.8 & 70.7$\pm$10.9 & 69.7$\pm$11.5 & 76.9$\pm$11.8 \\ 
& STU-Net-H~\cite{huang2023stu} & 1457.3M & 72.4$\pm$11.9 & 78.0$\pm$13.6 & 70.7$\pm$10.9 & 69.7$\pm$11.5 & 77.3$\pm$11.7 \\ 
& U-Net~\cite{ronneberger2015u} & 31.1M & 70.1$\pm$11.8 & 79.4$\pm$13.4 & 70.2$\pm$11.0 & 67.4$\pm$11.9 & 76.3$\pm$11.9 \\ 
& ResEncL~\cite{isensee2021nnu,isensee2024nnu} & 102.0M & 70.7$\pm$11.8 & 78.2$\pm$14.1 & 69.8$\pm$11.2 & 68.2$\pm$11.5 & 76.5$\pm$12.0 \\ 
& \textcolor{lightgray}{ResEncL$^\bigstar$} & \textcolor{lightgray}{102.0M} & \textcolor{lightgray}{71.0$\pm$11.8} & \textcolor{lightgray}{81.1$\pm$12.5} & \textcolor{lightgray}{69.7$\pm$11.1} & \textcolor{lightgray}{68.3$\pm$11.8} & \textcolor{lightgray}{76.7$\pm$11.8} \\ 
\hline
\multirow{2}{*}{Vision-Language}
& U-Net \& CLIP~\cite{liu2023clip} & 19.1M & 69.9$\pm$11.5 & 74.4$\pm$13.9 & 68.0$\pm$11.0 & 67.4$\pm$12.0 & 74.7$\pm$11.9 \\ 
& Swin UNETR \& CLIP~\cite{liu2023clip} & 62.2M & 70.1$\pm$12.0 & 75.0$\pm$13.6 & 66.2$\pm$11.7 & 66.9$\pm$12.7 & 74.4$\pm$12.6 \\ 
\hline
\multirow{6}{*}{MONAI}
& LHU-Net~\cite{sadegheih2024lhu} & 8.6M & 69.3$\pm$11.9 & 75.5$\pm$13.3 & 68.1$\pm$11.3 & 65.1$\pm$11.9 & 74.9$\pm$11.9 \\ 
& UCTransNet~\cite{wang2022uctransnet} & 68.0M & 51.4$\pm$13.4 & 82.0$\pm$11.5 & 56.3$\pm$16.1 & 44.9$\pm$18.1 & 66.4$\pm$15.3 \\ 
& Swin UNETR~\cite{tang2022self} & 72.8M & 61.6$\pm$11.8 & 72.1$\pm$15.6 & 60.8$\pm$12.6 & 59.2$\pm$13.5 & 68.7$\pm$13.6 \\
& UNesT~\cite{yu2023unest} & 87.2M & 61.6$\pm$11.2 & 71.3$\pm$16.0 & 60.4$\pm$12.1 & 58.0$\pm$11.3 & 67.6$\pm$12.9 \\ 
& UNETR~\cite{hatamizadeh2022unetr} & 101.8M & 53.8$\pm$11.8 & 69.2$\pm$15.3 & 54.7$\pm$12.4 & 54.5$\pm$12.8 & 65.4$\pm$13.7 \\
& SegVol$^\dagger$~\cite{du2023segvol} & 181.0M & 68.2$\pm$11.9 & 78.0$\pm$13.9 & 66.7$\pm$11.4 & 65.9$\pm$12.3 & 74.6$\pm$12.4 \\ 
\hline
\multirow{3}{*}{n/a}
& SAM-Adapter$^\dagger$~\cite{gu2024build} & 11.6M & 48.0$\pm$10.5 & 48.8$\pm$8.1 & 38.2$\pm$9.7 & 22.4$\pm$6.2 & 49.6$\pm$10.5 \\ 
& MedFormer~\cite{gao2022data} & 38.5M & \cellcolor{red!20}{\textbf{72.9$\pm$12.2}} & \cellcolor{red!20}{82.8$\pm$13.4} & \cellcolor{red!20}{\textbf{71.8$\pm$11.8}} & \cellcolor{red!20}{\textbf{71.4$\pm$12.2}} & \cellcolor{red!20}{\textbf{79.2$\pm$11.7}} \\ 
& Diff-UNet~\cite{xing2023diff} & 434.0M & 68.9$\pm$12.0 & 79.3$\pm$13.4 & 70.2$\pm$11.5 & 66.9$\pm$12.3 & 76.1$\pm$12.2 \\ 
\bottomrule
\end{tabular}
\begin{tablenotes}
    \item $^\dagger$These architectures were pre-trained (Appendix \ref{sec:implementation_details}).
    \item $^\bigstar$These architectures were trained on \ourdataset\ with enhanced label quality for the aorta class (discussed in \S\ref{sec:conclusion}).
\end{tablenotes}
\label{tab:jhh_nsd}
\end{table*}

\clearpage
\subsection{NSD scores on the TotalSegmentator dataset}

\begin{table*}[h]
\caption{
\textbf{Validation on the TotalSegmentator dataset ($N$=\numoftetotalsegct) - NSD.}  For each class, we bold the best-performing results and highlight the runners-up, which show no significant difference ($P>0.05$) from the best results, in red. Architectures are grouped by their frameworks and sorted in ascending order based on the number of parameters. NSD considers a tolerance of 1.5mm.
}\vspace{2px}
\centering
\scriptsize
\begin{tabular}{p{0.12\linewidth}p{0.20\linewidth}p{0.05\linewidth}P{0.075\linewidth}P{0.075\linewidth}P{0.075\linewidth}P{0.075\linewidth}P{0.075\linewidth}}
\toprule
framework & architecture & param & spleen & kidneyR & kidneyL & gallbladder & liver \\
\midrule
\multirow{9}{*}{nnU-Net}
& UniSeg$^\dagger$~\cite{ye2023uniseg} & 31.0M & 87.1$\pm$21.6 & 81.1$\pm$24.7 & 78.9$\pm$27.3 & \cellcolor{red!20}{73.2$\pm$29.4} & 83.5$\pm$19.9 \\ 
& MedNeXt~\cite{roy2023mednext} & 61.8M & \cellcolor{red!20}{90.1$\pm$20.1} & 82.4$\pm$24.8 & \cellcolor{red!20}{82.8$\pm$24.2} & \cellcolor{red!20}{74.9$\pm$29.1} & 86.7$\pm$18.3 \\ 
& NexToU~\cite{shi2023nextou} & 81.9M & 79.7$\pm$30.6 & 74.1$\pm$31.5 & 74.6$\pm$29.9 & 70.4$\pm$31.5 & 78.5$\pm$24.9 \\ 
& STU-Net-B~\cite{huang2023stu} & 58.3M & \cellcolor{red!20}{90.6$\pm$17.8} & 83.4$\pm$21.5 & \cellcolor{red!20}{83.3$\pm$23.0} & \cellcolor{red!20}{\textbf{77.5$\pm$25.3}} & 85.4$\pm$18.8 \\ 
& STU-Net-L~\cite{huang2023stu} & 440.3M & \cellcolor{red!20}{90.0$\pm$20.0} & 84.4$\pm$20.4 & \cellcolor{red!20}{83.0$\pm$23.6} & \cellcolor{red!20}{76.7$\pm$25.3} & \cellcolor{red!20}{\textbf{87.9$\pm$15.3}} \\ 
& STU-Net-H~\cite{huang2023stu} & 1457.3M & \cellcolor{red!20}{\textbf{90.6$\pm$17.0}} & 85.1$\pm$18.5 & \cellcolor{red!20}{82.9$\pm$24.3} & \cellcolor{red!20}{76.5$\pm$25.6} & 87.2$\pm$16.4 \\ 
& U-Net~\cite{ronneberger2015u} & 31.1M & \cellcolor{red!20}{89.6$\pm$19.4} & 84.4$\pm$19.3 & 83.9$\pm$21.7 & \cellcolor{red!20}{77.5$\pm$26.0} & 86.6$\pm$15.9 \\ 
& ResEncL~\cite{isensee2021nnu,isensee2024nnu} & 102.0M & \cellcolor{red!20}{90.4$\pm$19.1} & \cellcolor{red!20}{\textbf{85.6$\pm$19.2}} & \cellcolor{red!20}{\textbf{85.2$\pm$21.1}} & \cellcolor{red!20}{76.6$\pm$26.6} & 85.1$\pm$20.0 \\ 
& \textcolor{lightgray}{ResEncL$^\bigstar$} & \textcolor{lightgray}{102.0M} & \textcolor{lightgray}{90.4$\pm$18.7} & \textcolor{lightgray}{86.6$\pm$17.2} & \textcolor{lightgray}{86.6$\pm$19.2} & \textcolor{lightgray}{76.6$\pm$25.9} & \textcolor{lightgray}{86.0$\pm$18.6} \\ 
\hline
\multirow{2}{*}{Vision-Language}
& U-Net \& CLIP~\cite{liu2023clip} & 19.1M & 84.3$\pm$26.0 & 79.8$\pm$25.4 & 78.9$\pm$25.9 & 71.5$\pm$29.0 & 81.9$\pm$18.6 \\ 
& Swin UNETR \& CLIP~\cite{liu2023clip} & 62.2M & 83.2$\pm$25.6 & 78.0$\pm$28.3 & 74.2$\pm$31.3 & 68.8$\pm$31.4 & 82.0$\pm$19.7 \\ 
\hline
\multirow{6}{*}{MONAI}
& LHU-Net~\cite{sadegheih2024lhu} & 8.6M & 82.2$\pm$28.3 & 77.4$\pm$28.9 & 78.0$\pm$27.3 & 69.8$\pm$32.2 & 77.9$\pm$27.0 \\ 
& UCTransNet~\cite{wang2022uctransnet} & 68.0M & 72.2$\pm$35.3 & 71.1$\pm$34.3 & 59.6$\pm$39.7 & 67.3$\pm$32.1 & 71.3$\pm$29.5 \\ 
& Swin UNETR~\cite{tang2022self} & 72.8M & 58.9$\pm$35.3 & 53.2$\pm$36.9 & 53.1$\pm$37.4 & 46.3$\pm$38.6 & 65.1$\pm$27.1 \\ 
& UNesT~\cite{yu2023unest} & 87.2M & 71.7$\pm$27.5 & 69.5$\pm$30.8 & 66.7$\pm$32.6 & 45.7$\pm$38.4 & 75.8$\pm$20.8 \\ 
& UNETR~\cite{hatamizadeh2022unetr} & 101.8M & 48.8$\pm$34.9 & 40.1$\pm$35.0 & 35.5$\pm$35.3 & 32.9$\pm$32.1 & 58.0$\pm$25.3 \\ 
& SegVol$^\dagger$~\cite{du2023segvol} & 181.0M & 83.2$\pm$24.1 & 77.2$\pm$23.1 & 76.6$\pm$25.1 & 63.3$\pm$28.8 & 79.0$\pm$21.5 \\ 
\hline
\multirow{3}{*}{n/a}
& SAM-Adapter$^\dagger$~\cite{gu2024build} & 11.6M & 36.7$\pm$25.2 & 8.8$\pm$9.8 & 24.3$\pm$19.7 & 6.4$\pm$10.5 & 40.8$\pm$26.1 \\ 
& MedFormer~\cite{gao2022data} & 38.5M & 86.5$\pm$19.0 & 79.7$\pm$20.6 & 79.2$\pm$23.0 & 71.0$\pm$27.4 & 83.0$\pm$17.2 \\ 
& Diff-UNet~\cite{xing2023diff} & 434.0M & 85.4$\pm$25.9 & 76.5$\pm$27.5 & 76.2$\pm$28.3 & 68.9$\pm$31.5 & 84.7$\pm$17.5 \\ 
\bottomrule
framework & architecture & param & stomach & aorta & IVC$^\ddagger$ & pancreas & average \\
\midrule
\multirow{9}{*}{nnU-Net}
& UniSeg$^\dagger$~\cite{ye2023uniseg} & 31.0M & 64.0$\pm$31.1 & 67.3$\pm$31.9 & \cellcolor{red!20}{68.3$\pm$26.7} & 67.8$\pm$30.8 & 74.6$\pm$27.1 \\ 
& MedNeXt~\cite{roy2023mednext} & 61.8M & \cellcolor{red!20}{67.1$\pm$30.9} & 69.5$\pm$31.0 & \cellcolor{red!20}{70.0$\pm$24.8} & 68.6$\pm$31.2 & 76.9$\pm$26.1 \\ 
& NexToU~\cite{shi2023nextou} & 81.9M & 58.6$\pm$34.2 & 59.5$\pm$32.9 & 54.0$\pm$31.3 & 62.1$\pm$31.6 & 68.0$\pm$31.0 \\ 
& STU-Net-B~\cite{huang2023stu} & 58.3M & \cellcolor{red!20}{68.1$\pm$30.2} & 71.8$\pm$30.0 & \cellcolor{red!20}{71.8$\pm$22.1} & \cellcolor{red!20}{72.0$\pm$27.3} & 78.2$\pm$24.0 \\ 
& STU-Net-L~\cite{huang2023stu} & 440.3M & \cellcolor{red!20}{69.2$\pm$28.6} & \cellcolor{red!20}{\textbf{74.0$\pm$27.5}} & \cellcolor{red!20}{\textbf{72.0$\pm$21.2}} & \cellcolor{red!20}{72.9$\pm$26.9} & \cellcolor{red!20}{\textbf{78.9$\pm$23.2}} \\ 
& STU-Net-H~\cite{huang2023stu} & 1457.3M & \cellcolor{red!20}{68.4$\pm$28.8} & 72.7$\pm$28.6 & \cellcolor{red!20}{71.5$\pm$21.2} & \cellcolor{red!20}{71.9$\pm$27.8} & 78.5$\pm$23.2 \\ 
& U-Net~\cite{ronneberger2015u} & 31.1M & \cellcolor{red!20}{68.6$\pm$28.6} & 68.4$\pm$28.5 & \cellcolor{red!20}{71.0$\pm$24.0} & \cellcolor{red!20}{72.1$\pm$27.4} & 78.0$\pm$23.4 \\ 
& ResEncL~\cite{isensee2021nnu,isensee2024nnu} & 102.0M & \cellcolor{red!20}{68.7$\pm$28.8} & 71.3$\pm$26.6 & \cellcolor{red!20}{70.9$\pm$22.2} & \cellcolor{red!20}{\textbf{73.5$\pm$26.6}} & 78.6$\pm$23.3 \\ 
& \textcolor{lightgray}{ResEncL$^\bigstar$} & \textcolor{lightgray}{102.0M} & \textcolor{lightgray}{70.1$\pm$27.1} & \textcolor{lightgray}{81.9$\pm$22.0} & \textcolor{lightgray}{71.0$\pm$21.9} & \textcolor{lightgray}{74.5$\pm$25.6} & \textcolor{lightgray}{80.4$\pm$21.8} \\ 
\hline
\multirow{2}{*}{Vision-Language}
& U-Net \& CLIP~\cite{liu2023clip} & 19.1M & 66.7$\pm$28.2 & 57.5$\pm$32.2 & 61.6$\pm$26.8 & 70.6$\pm$26.1 & 72.5$\pm$26.5 \\ 
& Swin UNETR \& CLIP~\cite{liu2023clip} & 62.2M & 58.9$\pm$31.3 & 56.6$\pm$34.0 & 58.9$\pm$26.9 & 66.2$\pm$28.8 & 69.6$\pm$28.6 \\ 
\hline
\multirow{6}{*}{MONAI}
& LHU-Net~\cite{sadegheih2024lhu} & 8.6M & 60.5$\pm$32.5 & 59.2$\pm$33.5 & 62.6$\pm$27.9 & 65.4$\pm$32.0 & 70.3$\pm$30.0 \\ 
& UCTransNet~\cite{wang2022uctransnet} & 68.0M & 48.1$\pm$34.2 & 48.1$\pm$33.8 & 45.2$\pm$34.8 & 54.4$\pm$33.7 & 59.7$\pm$34.1 \\ 
& Swin UNETR~\cite{tang2022self} & 72.8M & 37.1$\pm$29.4 & 51.2$\pm$36.1 & 31.6$\pm$30.8 & 35.1$\pm$30.4 & 48.0$\pm$33.6 \\ 
& UNesT~\cite{yu2023unest} & 87.2M & 48.8$\pm$28.7 & 51.6$\pm$35.5 & 34.0$\pm$32.9 & 42.8$\pm$28.8 & 56.3$\pm$30.7 \\ 
& UNETR~\cite{hatamizadeh2022unetr} & 101.8M & 25.3$\pm$22.7 & 36.8$\pm$28.8 & 32.4$\pm$27.3 & 21.2$\pm$22.8 & 36.8$\pm$29.3 \\ 
& SegVol$^\dagger$~\cite{du2023segvol} & 181.0M & 58.7$\pm$28.7 & 57.6$\pm$28.8 & 56.1$\pm$23.5 & 59.9$\pm$26.5 & 68.0$\pm$25.6 \\ 
\hline
\multirow{3}{*}{n/a}
& SAM-Adapter$^\dagger$~\cite{gu2024build} & 11.6M & 27.0$\pm$19.6 & 17.1$\pm$17.2 & 5.4$\pm$8.1 & 21.7$\pm$14.4 & 20.9$\pm$16.7 \\ 
& MedFormer~\cite{gao2022data} & 38.5M & \cellcolor{red!20}{\textbf{69.3$\pm$26.7}} & 67.9$\pm$29.2 & 65.5$\pm$25.5 & 69.0$\pm$27.8 & 74.6$\pm$24.0 \\ 
& Diff-UNet~\cite{xing2023diff} & 434.0M & 59.8$\pm$31.2 & 57.7$\pm$34.6 & 55.4$\pm$33.4 & 65.5$\pm$29.4 & 70.0$\pm$28.8 \\ 
\bottomrule
\end{tabular}
\begin{tablenotes}
    \item $^\dagger$These architectures were pre-trained (Appendix \ref{sec:implementation_details}).
    \item $^\ddagger$The class IVC (inferior vena cava) shares the same meaning as the class postcava in other datasets (e.g., \ourdataset\ and JHH).
    \item $^\bigstar$These architectures were trained on \ourdataset\ with enhanced label quality for the aorta class (discussed in \S\ref{sec:conclusion}).
\end{tablenotes}
\label{tab:totalseg_train_plus_test_nsd}
\end{table*}

\clearpage
\subsection{DSC/NSD scores on the official test set of TotalSegmentator}

\begin{table*}[h]
\caption{
\textbf{Validation on the official test set of TotalSegmentator ($N$=\numofofficialtetotalsegct) - DSC.} TotalSegmentator provides an official split of training and testing sets. To align with other papers, we hereby also provide the benchmark results on the test set of TotalSegmentator ($N$=\numofofficialtetotalsegct). Notably, the average scores in the official test set are usually higher than the ones in the entire TotalSegmentator dataset.
}\vspace{2px}
\centering
\scriptsize
\begin{tabular}{p{0.12\linewidth}p{0.20\linewidth}p{0.05\linewidth}P{0.075\linewidth}P{0.075\linewidth}P{0.075\linewidth}P{0.075\linewidth}P{0.075\linewidth}}
\toprule
framework & architecture & param & spleen & kidneyR & kidneyL & gallbladder & liver \\
\midrule
\multirow{9}{*}{nnU-Net}
& UniSeg$^\dagger$~\cite{ye2023uniseg} & 31.0M & 94.7$\pm$6.8 & 86.5$\pm$17.8 & 88.2$\pm$13.3 & \cellcolor{red!20}{78.0$\pm$27.8} & 96.2$\pm$2.4 \\ 
& MedNeXt~\cite{roy2023mednext} & 61.8M & \cellcolor{red!20}{93.5$\pm$12.0} & \cellcolor{red!20}{83.6$\pm$24.8} & \cellcolor{red!20}{89.7$\pm$14.8} & \cellcolor{red!20}{73.1$\pm$34.7} & \cellcolor{red!20}{96.8$\pm$2.3} \\ 
& NexToU~\cite{shi2023nextou} & 81.9M & \cellcolor{red!20}{90.0$\pm$22.8} & 82.1$\pm$26.2 & 79.4$\pm$26.4 & \cellcolor{red!20}{76.2$\pm$32.8} & 90.8$\pm$18.5 \\ 
& STU-Net-B~\cite{huang2023stu} & 58.3M & \cellcolor{red!20}{\textbf{96.5$\pm$2.6}} & \cellcolor{red!20}{86.8$\pm$18.3} & \cellcolor{red!20}{90.2$\pm$9.7} & \cellcolor{red!20}{78.4$\pm$30.9} & \cellcolor{red!20}{96.4$\pm$4.9} \\ 
& STU-Net-L~\cite{huang2023stu} & 440.3M & \cellcolor{red!20}{96.1$\pm$3.4} & \cellcolor{red!20}{85.2$\pm$22.0} & \cellcolor{red!20}{89.4$\pm$14.5} & \cellcolor{red!20}{82.0$\pm$24.6} & \cellcolor{red!20}{96.8$\pm$2.6} \\ 
& STU-Net-H~\cite{huang2023stu} & 1457.3M & \cellcolor{red!20}{96.3$\pm$3.2} & \cellcolor{red!20}{85.7$\pm$19.9} & \cellcolor{red!20}{\textbf{92.5$\pm$5.6}} & \cellcolor{red!20}{\textbf{84.4$\pm$22.2}} & \cellcolor{red!20}{\textbf{97.2$\pm$1.6}} \\ 
& U-Net~\cite{ronneberger2015u} & 31.1M & \cellcolor{red!20}{94.9$\pm$12.3} & \cellcolor{red!20}{\textbf{88.3$\pm$18.1}} & 88.6$\pm$12.3 & \cellcolor{red!20}{78.3$\pm$29.7} & 95.7$\pm$5.8 \\ 
& ResEncL~\cite{isensee2021nnu,isensee2024nnu} & 102.0M & \cellcolor{red!20}{94.7$\pm$12.3} & \cellcolor{red!20}{84.9$\pm$23.5} & \cellcolor{red!20}{90.7$\pm$11.0} & \cellcolor{red!20}{78.4$\pm$29.7} & \cellcolor{red!20}{95.7$\pm$8.2} \\ 
& \textcolor{lightgray}{ResEncL$^\bigstar$} & \textcolor{lightgray}{102.0M} & \textcolor{lightgray}{95.6$\pm$8.8} & \textcolor{lightgray}{87.0$\pm$20.7} & \textcolor{lightgray}{91.6$\pm$10.3} & \textcolor{lightgray}{78.0$\pm$29.0} & \textcolor{lightgray}{96.7$\pm$2.7} \\ 
\hline
\multirow{2}{*}{Vision-Language}
& U-Net \& CLIP~\cite{liu2023clip} & 19.1M & 94.6$\pm$7.0 & \cellcolor{red!20}{85.2$\pm$22.5} & 83.1$\pm$24.0 & 70.1$\pm$33.9 & 95.3$\pm$4.6 \\ 
& Swin UNETR \& CLIP~\cite{liu2023clip} & 62.2M & 92.5$\pm$10.1 & \cellcolor{red!20}{76.7$\pm$34.6} & 73.4$\pm$34.8 & \cellcolor{red!20}{72.2$\pm$34.2} & 96.2$\pm$2.8 \\ 
\hline
\multirow{6}{*}{MONAI}
& LHU-Net~\cite{sadegheih2024lhu} & 8.6M & \cellcolor{red!20}{92.3$\pm$15.5} & \cellcolor{red!20}{84.9$\pm$21.4} & \cellcolor{red!20}{89.5$\pm$10.6} & 74.8$\pm$33.3 & 94.2$\pm$10.0 \\ 
& UCTransNet~\cite{wang2022uctransnet} & 68.0M & 89.3$\pm$19.4 & \cellcolor{red!20}{82.7$\pm$27.6} & 59.3$\pm$41.7 & 70.3$\pm$32.5 & 92.8$\pm$15.9 \\ 
& Swin UNETR~\cite{tang2022self} & 72.8M & 80.8$\pm$28.9 & 69.9$\pm$35.7 & 57.7$\pm$40.2 & 47.4$\pm$44.1 & 89.8$\pm$16.5 \\ 
& UNesT~\cite{yu2023unest} & 87.2M & 90.2$\pm$11.3 & 79.0$\pm$26.7 & 70.4$\pm$34.6 & 49.7$\pm$40.2 & 95.0$\pm$3.3 \\ 
& UNETR~\cite{hatamizadeh2022unetr} & 101.8M & 74.4$\pm$31.3 & 60.0$\pm$37.1 & 47.5$\pm$39.7 & 40.1$\pm$40.2 & 84.6$\pm$23.9 \\ 
& SegVol$^\dagger$~\cite{du2023segvol} & 181.0M & 91.2$\pm$16.7 & 82.1$\pm$21.2 & 82.5$\pm$21.9 & 69.9$\pm$30.8 & 94.8$\pm$5.6 \\ 
\hline
\multirow{3}{*}{n/a}
& SAM-Adapter$^\dagger$~\cite{gu2024build} & 11.6M & 50.4$\pm$34.1 & 9.2$\pm$10.5 & 18.0$\pm$21.2 & 7.2$\pm$12.3 & 77.5$\pm$21.3 \\ 
& MedFormer~\cite{gao2022data} & 38.5M & 95.4$\pm$1.7 & 84.0$\pm$22.5 & 89.2$\pm$9.3 & 76.5$\pm$28.5 & 96.2$\pm$2.7 \\ 
& Diff-UNet~\cite{xing2023diff} & 434.0M & \cellcolor{red!20}{95.3$\pm$6.3} & \cellcolor{red!20}{85.0$\pm$22.9} & \cellcolor{red!20}{86.7$\pm$16.9} & \cellcolor{red!20}{72.3$\pm$34.5} & 93.6$\pm$15.9 \\ 
\bottomrule
framework & architecture & param & stomach & aorta & IVC$^\ddagger$ & pancreas & average \\
\midrule
\multirow{9}{*}{nnU-Net}
& UniSeg$^\dagger$~\cite{ye2023uniseg} & 31.0M & 80.8$\pm$27.3 & \cellcolor{red!20}{82.6$\pm$19.7} & \cellcolor{red!20}{79.5$\pm$20.1} & 82.1$\pm$17.2 & 85.4$\pm$16.9 \\ 
& MedNeXt~\cite{roy2023mednext} & 61.8M & \cellcolor{red!20}{\textbf{87.8$\pm$13.3}} & \cellcolor{red!20}{84.9$\pm$17.2} & \cellcolor{red!20}{82.2$\pm$16.0} & \cellcolor{red!20}{83.9$\pm$16.8} & \cellcolor{red!20}{86.2$\pm$16.9} \\ 
& NexToU~\cite{shi2023nextou} & 81.9M & \cellcolor{red!20}{82.4$\pm$25.8} & 72.5$\pm$27.1 & 66.4$\pm$30.2 & 78.9$\pm$19.2 & 79.9$\pm$25.4 \\ 
& STU-Net-B~\cite{huang2023stu} & 58.3M & \cellcolor{red!20}{86.1$\pm$20.1} & \cellcolor{red!20}{85.5$\pm$16.3} & \cellcolor{red!20}{82.1$\pm$17.3} & \cellcolor{red!20}{\textbf{84.1$\pm$15.9}} & \cellcolor{red!20}{87.3$\pm$15.1} \\ 
& STU-Net-L~\cite{huang2023stu} & 440.3M & \cellcolor{red!20}{88.7$\pm$14.2} & \cellcolor{red!20}{\textbf{87.0$\pm$11.2}} & \cellcolor{red!20}{\textbf{84.5$\pm$8.9}} & \cellcolor{red!20}{83.4$\pm$17.2} & \cellcolor{red!20}{88.1$\pm$13.2} \\ 
& STU-Net-H~\cite{huang2023stu} & 1457.3M & \cellcolor{red!20}{88.4$\pm$14.2} & \cellcolor{red!20}{86.7$\pm$11.1} & \cellcolor{red!20}{84.0$\pm$9.7} & \cellcolor{red!20}{82.9$\pm$17.5} & \cellcolor{red!20}{\textbf{88.7$\pm$11.7}} \\ 
& U-Net~\cite{ronneberger2015u} & 31.1M & \cellcolor{red!20}{85.7$\pm$21.1} & 82.6$\pm$18.8 & \cellcolor{red!20}{79.7$\pm$20.4} & \cellcolor{red!20}{83.1$\pm$16.0} & 86.3$\pm$17.2 \\ 
& ResEncL~\cite{isensee2021nnu,isensee2024nnu} & 102.0M & \cellcolor{red!20}{85.4$\pm$21.1} & \cellcolor{red!20}{83.7$\pm$17.6} & \cellcolor{red!20}{79.0$\pm$20.4} & \cellcolor{red!20}{83.4$\pm$16.7} & \cellcolor{red!20}{86.2$\pm$17.8} \\ 
& \textcolor{lightgray}{ResEncL$^\bigstar$} & \textcolor{lightgray}{102.0M} & \textcolor{lightgray}{86.9$\pm$17.8} & \textcolor{lightgray}{91.1$\pm$8.8} & \textcolor{lightgray}{80.6$\pm$16.2} & \textcolor{lightgray}{83.8$\pm$16.3} & \textcolor{lightgray}{87.9$\pm$14.5} \\ 
\hline
\multirow{2}{*}{Vision-Language}
& U-Net \& CLIP~\cite{liu2023clip} & 19.1M & \cellcolor{red!20}{84.0$\pm$19.1} & 70.7$\pm$28.7 & 77.0$\pm$20.4 & 79.8$\pm$21.7 & 82.2$\pm$20.2 \\ 
& Swin UNETR \& CLIP~\cite{liu2023clip} & 62.2M & 79.9$\pm$25.7 & 72.3$\pm$27.7 & 72.9$\pm$21.9 & 77.6$\pm$21.8 & 79.3$\pm$23.7 \\ 
\hline
\multirow{6}{*}{MONAI}
& LHU-Net~\cite{sadegheih2024lhu} & 8.6M & \cellcolor{red!20}{80.5$\pm$26.4} & 72.2$\pm$29.9 & 73.6$\pm$24.5 & \cellcolor{red!20}{80.0$\pm$21.9} & 82.5$\pm$21.5 \\ 
& UCTransNet~\cite{wang2022uctransnet} & 68.0M & 74.4$\pm$31.8 & 61.7$\pm$32.8 & 63.7$\pm$32.6 & 76.0$\pm$18.1 & 74.5$\pm$28.0 \\ 
& Swin UNETR~\cite{tang2022self} & 72.8M & 55.1$\pm$36.8 & 69.9$\pm$27.5 & 52.7$\pm$32.0 & 57.2$\pm$32.8 & 64.5$\pm$32.7 \\ 
& UNesT~\cite{yu2023unest} & 87.2M & 70.4$\pm$30.0 & 65.0$\pm$33.9 & 53.2$\pm$33.8 & 65.2$\pm$25.7 & 70.9$\pm$26.6 \\ 
& UNETR~\cite{hatamizadeh2022unetr} & 101.8M & 52.6$\pm$31.0 & 50.4$\pm$30.2 & 52.7$\pm$28.8 & 45.1$\pm$30.8 & 56.4$\pm$32.6 \\ 
& SegVol$^\dagger$~\cite{du2023segvol} & 181.0M & 78.5$\pm$26.5 & 74.4$\pm$21.8 & 69.9$\pm$19.5 & 76.0$\pm$16.9 & 79.9$\pm$20.1 \\ 
\hline
\multirow{3}{*}{n/a}
& SAM-Adapter$^\dagger$~\cite{gu2024build} & 11.6M & 48.7$\pm$32.9 & 25.1$\pm$23.3 & 7.0$\pm$8.6 & 37.7$\pm$20.0 & 31.2$\pm$20.5 \\ 
& MedFormer~\cite{gao2022data} & 38.5M & \cellcolor{red!20}{87.8$\pm$13.9} & \cellcolor{red!20}{83.9$\pm$15.8} & 79.6$\pm$10.5 & 81.2$\pm$18.5 & 86.0$\pm$13.7 \\ 
& Diff-UNet~\cite{xing2023diff} & 434.0M & \cellcolor{red!20}{82.0$\pm$25.0} & 74.4$\pm$26.8 & 73.6$\pm$27.4 & 79.0$\pm$21.4 & 82.4$\pm$21.9 \\ 
\bottomrule
\end{tabular}
\begin{tablenotes}
    \item $^\dagger$These architectures were pre-trained (Appendix \ref{sec:implementation_details}).
    \item $^\ddagger$The class IVC (inferior vena cava) shares the same meaning as the class postcava in other datasets (e.g., \ourdataset\ and JHH).
    \item $^\bigstar$These architectures were trained on \ourdataset\ with enhanced label quality for the aorta class (discussed in \S\ref{sec:conclusion}).
\end{tablenotes}
\label{tab:totalseg_test_dsc}
\end{table*}

\clearpage
\begin{table*}[t]
\caption{
\textbf{Validation on the official test set of TotalSegmentator ($N$=\numofofficialtetotalsegct) - NSD.} TotalSegmentator provides an official split of training and testing sets. To align with other papers, we hereby also provide the benchmark results on the test set of TotalSegmentator ($N$=\numofofficialtetotalsegct). Notably, the average scores in the official test set are usually higher than the ones in the entire TotalSegmentator dataset. NSD considers a tolerance of 1.5mm.
}\vspace{2px}
\centering
\scriptsize
\begin{tabular}{p{0.12\linewidth}p{0.20\linewidth}p{0.05\linewidth}P{0.075\linewidth}P{0.075\linewidth}P{0.075\linewidth}P{0.075\linewidth}P{0.075\linewidth}}
\toprule
framework & architecture & param & spleen & kidneyR & kidneyL & gallbladder & liver \\
\midrule
\multirow{9}{*}{nnU-Net}
& UniSeg$^\dagger$~\cite{ye2023uniseg} & 31.0M & 89.3$\pm$13.2 & \cellcolor{red!20}{81.7$\pm$19.1} & \cellcolor{red!20}{83.5$\pm$13.1} & \cellcolor{red!20}{75.2$\pm$29.8} & 87.4$\pm$10.7 \\ 
& MedNeXt~\cite{roy2023mednext} & 61.8M & \cellcolor{red!20}{89.5$\pm$15.9} & \cellcolor{red!20}{79.2$\pm$25.0} & \cellcolor{red!20}{84.5$\pm$17.1} & \cellcolor{red!20}{71.2$\pm$35.9} & \cellcolor{red!20}{90.4$\pm$9.0} \\ 
& NexToU~\cite{shi2023nextou} & 81.9M & 84.3$\pm$25.1 & 75.9$\pm$25.2 & 74.3$\pm$24.7 & \cellcolor{red!20}{73.3$\pm$33.5} & 80.2$\pm$23.9 \\ 
& STU-Net-B~\cite{huang2023stu} & 58.3M & \cellcolor{red!20}{91.7$\pm$11.4} & \cellcolor{red!20}{81.9$\pm$19.2} & \cellcolor{red!20}{85.5$\pm$12.6} & \cellcolor{red!20}{76.3$\pm$30.3} & \cellcolor{red!20}{89.1$\pm$12.7} \\ 
& STU-Net-L~\cite{huang2023stu} & 440.3M & \cellcolor{red!20}{\textbf{91.1$\pm$12.3}} & \cellcolor{red!20}{80.9$\pm$22.6} & \cellcolor{red!20}{84.9$\pm$15.0} & \cellcolor{red!20}{78.4$\pm$28.6} & \cellcolor{red!20}{90.8$\pm$7.2} \\ 
& STU-Net-H~\cite{huang2023stu} & 1457.3M & 90.8$\pm$13.0 & \cellcolor{red!20}{81.1$\pm$22.0} & \cellcolor{red!20}{\textbf{87.2$\pm$12.0}} & \cellcolor{red!20}{\textbf{81.0$\pm$24.1}} & \cellcolor{red!20}{\textbf{91.6$\pm$6.4}} \\ 
& U-Net~\cite{ronneberger2015u} & 31.1M & \cellcolor{red!20}{91.5$\pm$14.5} & \cellcolor{red!20}{\textbf{82.3$\pm$20.7}} & 82.1$\pm$17.0 & \cellcolor{red!20}{75.8$\pm$30.4} & 89.0$\pm$9.4 \\ 
& ResEncL~\cite{isensee2021nnu,isensee2024nnu} & 102.0M & \cellcolor{red!20}{90.8$\pm$15.0} & \cellcolor{red!20}{81.2$\pm$23.1} & \cellcolor{red!20}{84.9$\pm$15.3} & \cellcolor{red!20}{76.7$\pm$31.8} & \cellcolor{red!20}{89.9$\pm$10.6} \\ 
& \textcolor{lightgray}{ResEncL$^\bigstar$} & \textcolor{lightgray}{102.0M} & \textcolor{lightgray}{91.4$\pm$13.4} & \textcolor{lightgray}{82.7$\pm$21.6} & \textcolor{lightgray}{86.5$\pm$12.8} & \textcolor{lightgray}{75.4$\pm$31.5} & \textcolor{lightgray}{90.1$\pm$8.6} \\ 
\hline
\multirow{2}{*}{Vision-Language}
& U-Net \& CLIP~\cite{liu2023clip} & 19.1M & 87.4$\pm$16.3 & \cellcolor{red!20}{78.6$\pm$23.9} & 77.1$\pm$24.6 & 68.9$\pm$31.9 & 85.5$\pm$12.5 \\ 
& Swin UNETR \& CLIP~\cite{liu2023clip} & 62.2M & 84.1$\pm$20.2 & \cellcolor{red!20}{72.5$\pm$34.2} & 68.1$\pm$32.9 & \cellcolor{red!20}{69.5$\pm$35.8} & 87.2$\pm$10.7 \\ 
\hline
\multirow{6}{*}{MONAI}
& LHU-Net~\cite{sadegheih2024lhu} & 8.6M & 85.1$\pm$22.9 & \cellcolor{red!20}{79.1$\pm$21.9} & \cellcolor{red!20}{83.1$\pm$16.3} & \cellcolor{red!20}{72.9$\pm$32.3} & 83.9$\pm$19.4 \\ 
& UCTransNet~\cite{wang2022uctransnet} & 68.0M & 82.2$\pm$25.1 & \cellcolor{red!20}{77.7$\pm$27.6} & 55.8$\pm$39.1 & 65.4$\pm$32.8 & 83.3$\pm$18.2 \\ 
& Swin UNETR~\cite{tang2022self} & 72.8M & 71.8$\pm$29.2 & 62.8$\pm$34.7 & 51.2$\pm$36.4 & 44.9$\pm$41.5 & 73.3$\pm$21.0 \\ 
& UNesT~\cite{yu2023unest} & 87.2M & 79.2$\pm$18.3 & 72.1$\pm$26.8 & 62.8$\pm$33.0 & 43.7$\pm$39.4 & 82.5$\pm$9.2 \\ 
& UNETR~\cite{hatamizadeh2022unetr} & 101.8M & 61.0$\pm$33.3 & 49.0$\pm$34.0 & 39.4$\pm$34.8 & 33.2$\pm$33.2 & 62.9$\pm$25.1 \\ 
& SegVol$^\dagger$~\cite{du2023segvol} & 181.0M & 83.5$\pm$19.4 & 74.2$\pm$21.0 & 73.6$\pm$22.8 & 62.4$\pm$29.6 & 82.1$\pm$12.8 \\ 
\hline
\multirow{3}{*}{n/a}
& SAM-Adapter$^\dagger$~\cite{gu2024build} & 11.6M & 34.6$\pm$27.8 & 9.1$\pm$9.8 & 21.5$\pm$19.5 & 4.2$\pm$8.0 & 44.6$\pm$23.0 \\ 
& MedFormer~\cite{gao2022data} & 38.5M & 90.0$\pm$10.0 & 78.2$\pm$22.5 & 82.6$\pm$15.3 & 70.3$\pm$30.3 & 87.6$\pm$6.5 \\ 
& Diff-UNet~\cite{xing2023diff} & 434.0M & \cellcolor{red!20}{89.8$\pm$14.8} & \cellcolor{red!20}{79.8$\pm$22.8} & \cellcolor{red!20}{79.9$\pm$17.6} & \cellcolor{red!20}{69.0$\pm$36.6} & 86.0$\pm$16.8 \\ 
\bottomrule
framework & architecture & param & stomach & aorta & IVC$^\dagger$ & pancreas & average \\
\midrule
\multirow{9}{*}{nnU-Net}
& UniSeg$^\dagger$~\cite{ye2023uniseg} & 31.0M & 72.1$\pm$29.9 & \cellcolor{red!20}{81.8$\pm$21.5} & \cellcolor{red!20}{76.5$\pm$21.0} & \cellcolor{red!20}{78.3$\pm$17.8} & 80.6$\pm$19.6 \\ 
& MedNeXt~\cite{roy2023mednext} & 61.8M & \cellcolor{red!20}{\textbf{77.9$\pm$21.9}} & \cellcolor{red!20}{84.0$\pm$19.0} & \cellcolor{red!20}{78.4$\pm$17.6} & \cellcolor{red!20}{79.2$\pm$17.6} & \cellcolor{red!20}{81.6$\pm$19.9} \\ 
& NexToU~\cite{shi2023nextou} & 81.9M & \cellcolor{red!20}{72.9$\pm$29.4} & 71.2$\pm$28.3 & 62.2$\pm$30.2 & 71.9$\pm$21.8 & 74.0$\pm$26.9 \\ 
& STU-Net-B~\cite{huang2023stu} & 58.3M & \cellcolor{red!20}{76.6$\pm$27.4} & \cellcolor{red!20}{83.7$\pm$18.5} & \cellcolor{red!20}{78.5$\pm$19.0} & \cellcolor{red!20}{79.3$\pm$16.5} & \cellcolor{red!20}{82.5$\pm$18.6} \\ 
& STU-Net-L~\cite{huang2023stu} & 440.3M & \cellcolor{red!20}{80.0$\pm$21.7} & \cellcolor{red!20}{\textbf{86.2$\pm$13.6}} & \cellcolor{red!20}{80.9$\pm$12.3} & \cellcolor{red!20}{\textbf{78.7$\pm$18.0}} & \cellcolor{red!20}{83.5$\pm$16.8} \\ 
& STU-Net-H~\cite{huang2023stu} & 1457.3M & \cellcolor{red!20}{79.4$\pm$22.2} & \cellcolor{red!20}{86.1$\pm$12.9} & \cellcolor{red!20}{\textbf{81.3$\pm$12.4}} & \cellcolor{red!20}{78.0$\pm$18.2} & \cellcolor{red!20}{\textbf{84.0$\pm$15.9}} \\ 
& U-Net~\cite{ronneberger2015u} & 31.1M & \cellcolor{red!20}{76.3$\pm$26.7} & \cellcolor{red!20}{81.1$\pm$20.6} & \cellcolor{red!20}{76.7$\pm$21.1} & \cellcolor{red!20}{78.3$\pm$16.6} & 81.4$\pm$19.7 \\ 
& ResEncL~\cite{isensee2021nnu,isensee2024nnu} & 102.0M & \cellcolor{red!20}{76.7$\pm$26.2} & \cellcolor{red!20}{83.6$\pm$18.2} & 75.6$\pm$21.8 & \cellcolor{red!20}{78.6$\pm$18.1} & \cellcolor{red!20}{82.0$\pm$20.0} \\ 
& \textcolor{lightgray}{ResEncL$^\bigstar$} & \textcolor{lightgray}{102.0M} & \textcolor{lightgray}{77.7$\pm$23.8} & \textcolor{lightgray}{91.0$\pm$9.8} & \textcolor{lightgray}{77.4$\pm$18.4} & \textcolor{lightgray}{79.5$\pm$16.8} & \textcolor{lightgray}{83.5$\pm$17.4} \\ 
\hline
\multirow{2}{*}{Vision-Language}
& U-Net \& CLIP~\cite{liu2023clip} & 19.1M & 73.0$\pm$25.4 & 70.5$\pm$28.6 & 73.6$\pm$21.5 & 74.5$\pm$21.8 & 76.6$\pm$22.9 \\ 
& Swin UNETR \& CLIP~\cite{liu2023clip} & 62.2M & 69.7$\pm$27.8 & 71.4$\pm$28.1 & 69.1$\pm$22.9 & 71.9$\pm$21.6 & 73.7$\pm$26.0 \\ 
\hline
\multirow{6}{*}{MONAI}
& LHU-Net~\cite{sadegheih2024lhu} & 8.6M & 71.2$\pm$28.3 & 68.7$\pm$31.5 & 69.7$\pm$24.9 & \cellcolor{red!20}{74.7$\pm$22.7} & 76.5$\pm$24.5 \\ 
& UCTransNet~\cite{wang2022uctransnet} & 68.0M & 62.3$\pm$34.0 & 60.0$\pm$32.7 & 60.7$\pm$31.8 & 68.8$\pm$18.8 & 68.5$\pm$28.9 \\ 
& Swin UNETR~\cite{tang2022self} & 72.8M & 41.1$\pm$33.5 & 65.9$\pm$28.1 & 44.6$\pm$29.6 & 48.7$\pm$31.2 & 56.0$\pm$31.7 \\ 
& UNesT~\cite{yu2023unest} & 87.2M & 56.2$\pm$28.8 & 61.0$\pm$32.9 & 47.9$\pm$31.5 & 55.2$\pm$24.4 & 62.3$\pm$27.1 \\ 
& UNETR~\cite{hatamizadeh2022unetr} & 101.8M & 34.8$\pm$26.4 & 44.8$\pm$27.2 & 43.5$\pm$25.4 & 34.6$\pm$26.2 & 44.8$\pm$29.5 \\ 
& SegVol$^\dagger$~\cite{du2023segvol} & 181.0M & 65.7$\pm$24.4 & 71.9$\pm$21.8 & 64.8$\pm$19.8 & 66.9$\pm$16.0 & 71.7$\pm$20.9 \\ 
\hline
\multirow{3}{*}{n/a}
& SAM-Adapter$^\dagger$~\cite{gu2024build} & 11.6M & 25.4$\pm$19.1 & 24.2$\pm$17.3 & 8.6$\pm$9.9 & 24.9$\pm$14.1 & 21.9$\pm$16.5 \\ 
& MedFormer~\cite{gao2022data} & 38.5M & \cellcolor{red!20}{77.3$\pm$21.7} & \cellcolor{red!20}{83.8$\pm$17.9} & 77.4$\pm$13.7 & 76.5$\pm$18.8 & 80.4$\pm$17.4 \\ 
& Diff-UNet~\cite{xing2023diff} & 434.0M & \cellcolor{red!20}{71.2$\pm$29.9} & 71.4$\pm$28.6 & 69.9$\pm$28.8 & \cellcolor{red!20}{72.7$\pm$22.1} & 76.6$\pm$24.2 \\ 
\bottomrule
\end{tabular}
\begin{tablenotes}
    \item $^\dagger$These architectures were pre-trained (Appendix \ref{sec:implementation_details}).
    \item $^\ddagger$The class IVC (inferior vena cava) shares the same meaning as the class postcava in other datasets (e.g., \ourdataset\ and JHH).
    \item $^\bigstar$These architectures were trained on \ourdataset\ with enhanced label quality for the aorta class (discussed in \S\ref{sec:conclusion}).
\end{tablenotes}
\label{tab:totalseg_test_nsd}
\end{table*}

\clearpage
\section{Additional Analysis of Benchmark Results}\label{sec:analysis}

\subsection{Worst-case Analysis}\label{sec:wrost_case_analysis}

\begin{figure}[h]
	\centering
	\includegraphics[width=\columnwidth]{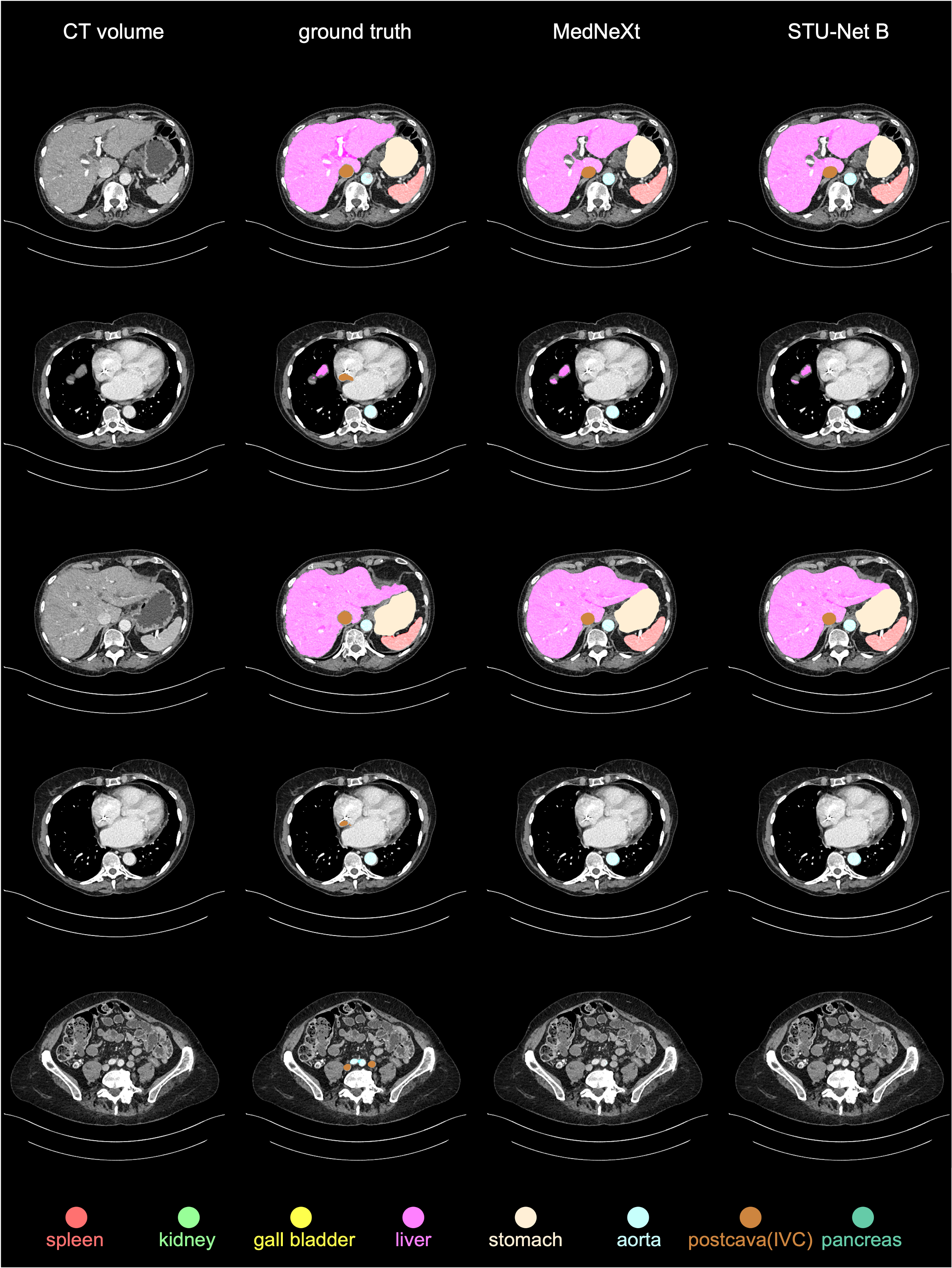}
    \caption{\textbf{Worst case analysis for JHH.} This figure displays CT scans that are particularly challenging for most AI algorithms to identify. To illustrate these difficult cases, we also include visualizations from the top-performing algorithm, MedNext, and the first runner-up, STU-Net Base.}
	\label{fig:worst_case_analysis_jhh}
\end{figure}

\clearpage
\begin{figure}[t]
	\centering
	\includegraphics[width=\columnwidth]{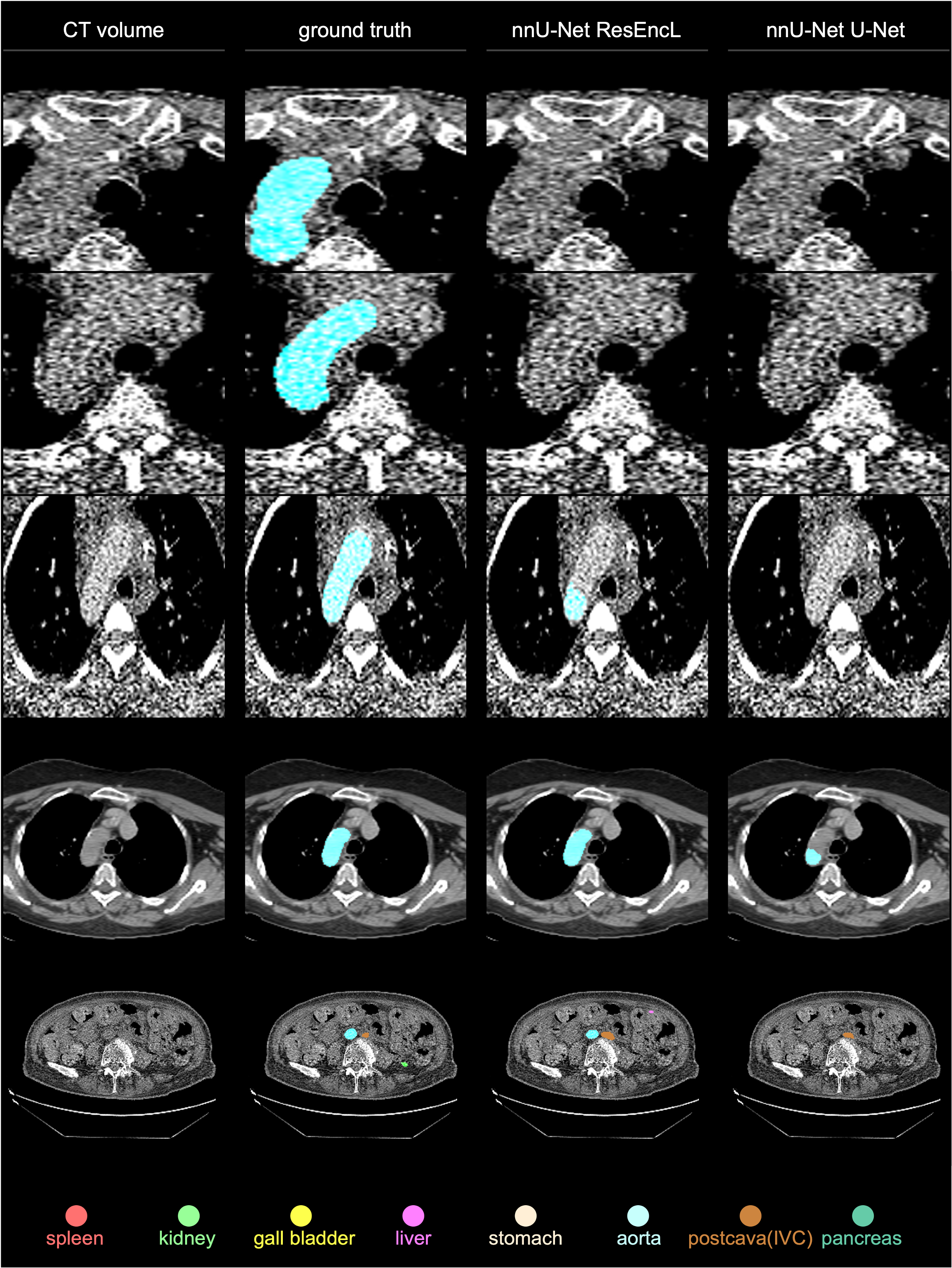}
    \caption{\textbf{Worst case analysis for TotalSegmentator.} 
    This figure displays CT scans that are particularly challenging for most AI algorithms to identify. To illustrate these difficult cases, we also include visualizations from the top-performing algorithm, ResEncL, and the first runner-up, U-Net. The results show ResEncL does perform better than U-Net in these worst cases.}
    \label{fig:worst_case_analysis_totalseg_train}
\end{figure}

\clearpage
\subsection{Ranking Stability Analyses}\label{sec:supplementary_statistics}

\subsubsection{Evaluation Metrics}
Every evaluation metric reflects a certain aspect of the results and choosing the right one is important to emphasize those properties that we care about. In this section, we assess the ranking stability with respect to different evaluation metrics.

The Dice Similarity Coefficient (DSC) is a widely used metric in medical imaging to measure the overlap between the prediction and the ground truth. Additionally, Normalized Surface Distance (NSD) focuses on the segmentation quality between two boundaries. 

Due to the existence of NaNs (which represent some organs that are missing in some CT scans), averaging per-case-per-class values by case first and then by class differs from averaging them by class first and then by case~\cite{wang2023revisiting}. Let's focus on DSC (note that this also applies to other metrics such as NSD) and denote the first version as DSC\textsuperscript{C} and the second as DSC\textsuperscript{I}. DSC\textsuperscript{C} allows us to evaluate model performance on a class-wise scale, emphasizing difficult classes, and it alleviates the limitation of DSC\textsuperscript{I}, which can be biased towards classes with less NaNs. On the other hand, DSC\textsuperscript{I} facilitates statistical tests across different cases. Due to these considerations, we use DSC\textsuperscript{C} for reporting per-class performance and utilize DSC\textsuperscript{I} to conduct statistical tests. In the rest of the paper, we drop the superscripts for simplicity unless stated otherwise. 

Besides the standard DSC, in this section, we also consider a worst-case metric to emphasize difficult cases~\cite{wang2023revisiting}. In particular, it only averages over cases whose scores fall below the 10\% quantile.

Except accuracy metrics such DSC and NSD, we also study bias metrics. Specifically, we choose Demographic Parity Difference (DPD)~\cite{agarwal2018reductions,tian2023fairseg}, which captures bias across diverse demographic groups. Originally proposed for classification problems, we extend it to medical segmentation and define it as the maximum differences in DSC among different sensitive demographic groups.

\begin{figure}[h]
    \centering
    \includegraphics[width=\linewidth]{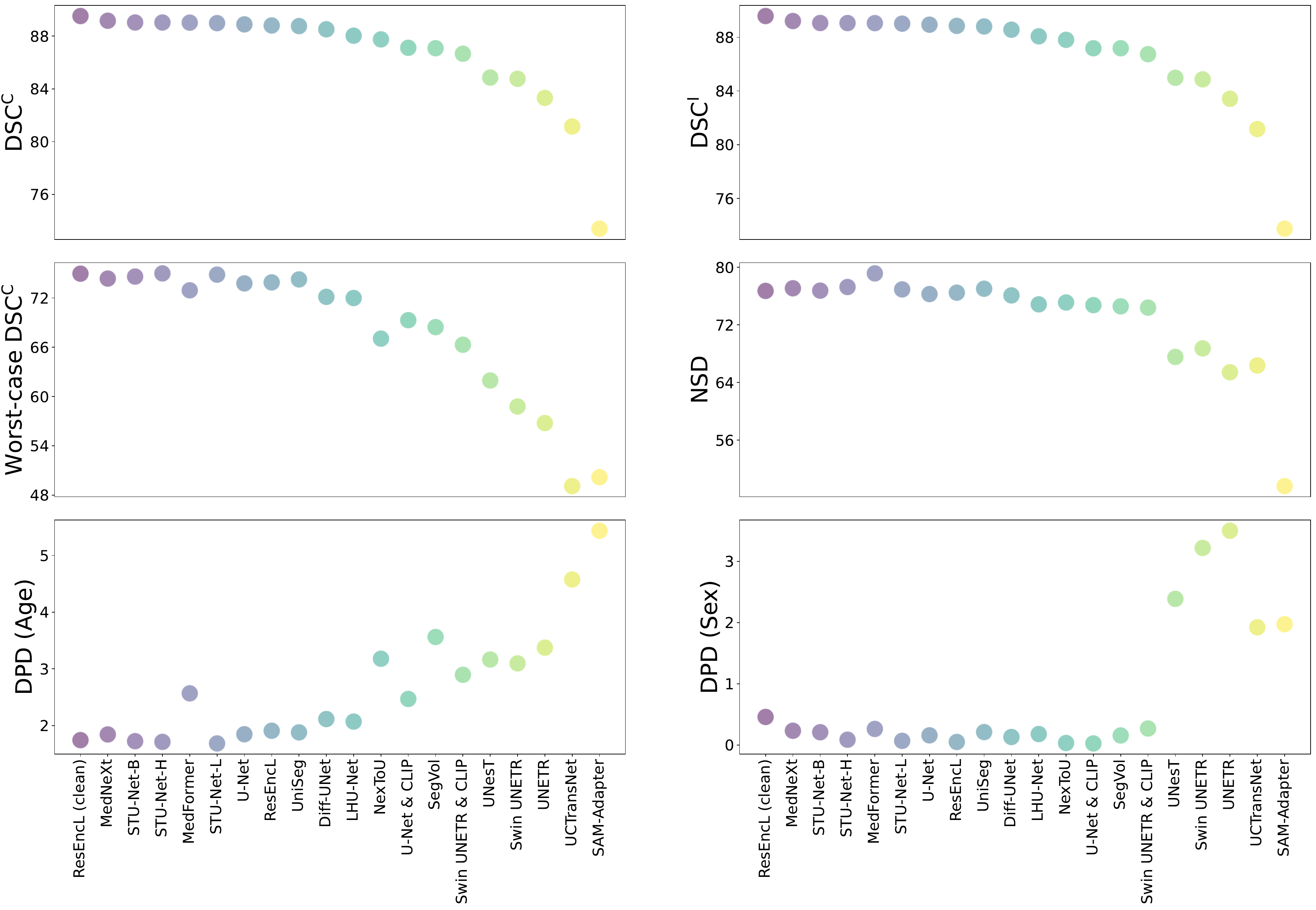}
    \caption{\textbf{Comparison of different evaluation metrics on proprietary JHH dataset.} Models tend to retain a similar rank across different metrics.}
    \label{fig:metrics_jhh}
\end{figure}

\begin{figure}[t]
    \centering
    \includegraphics[width=\linewidth]{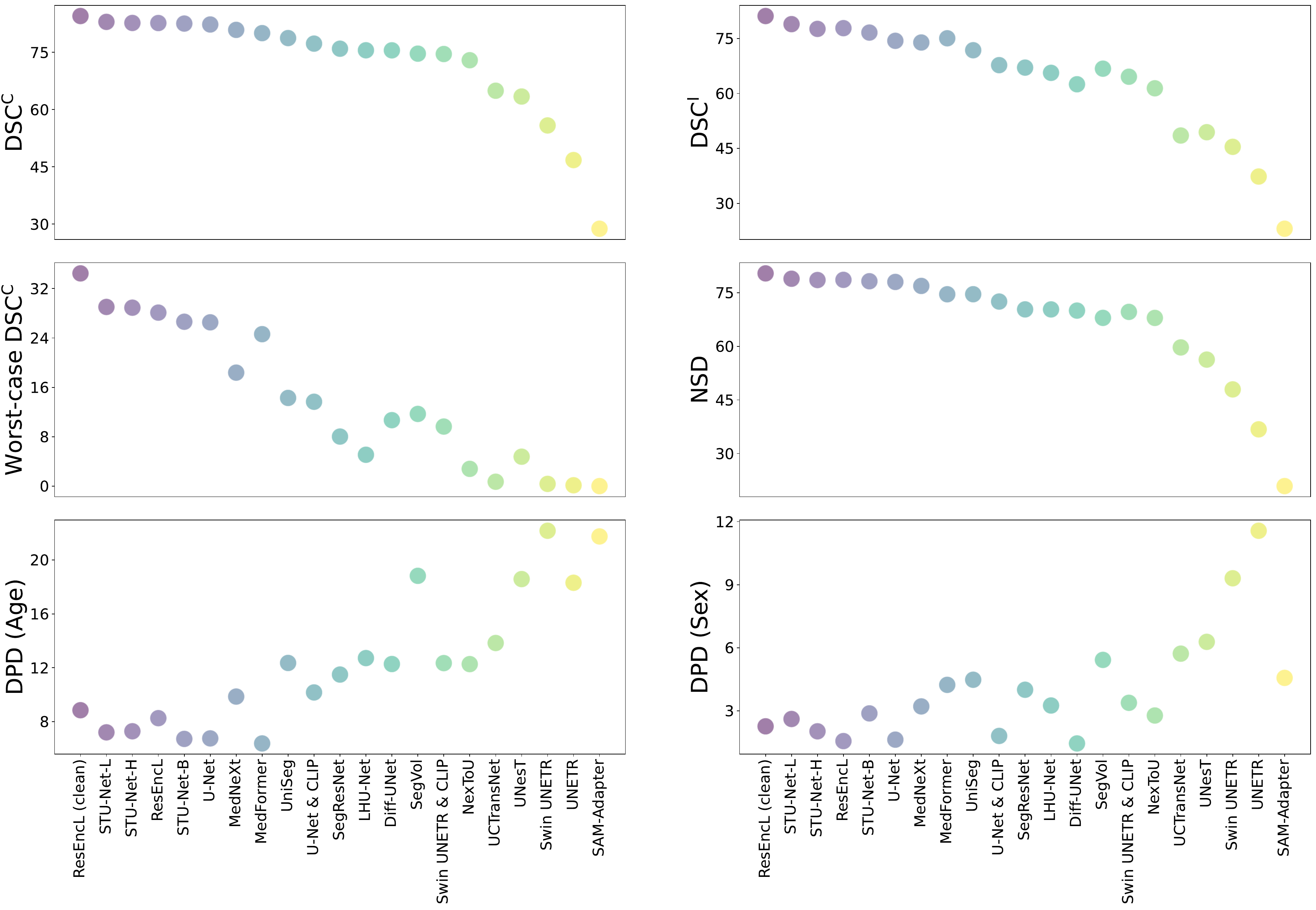}
    \caption{\textbf{Comparison of different evaluation metrics on TotalSegmentator.} Models tend to retain a similar rank across different metrics.}
    \label{fig:metrics_ts}
\end{figure}

\begin{figure}[h]
    \centering
    \includegraphics[width=\linewidth]{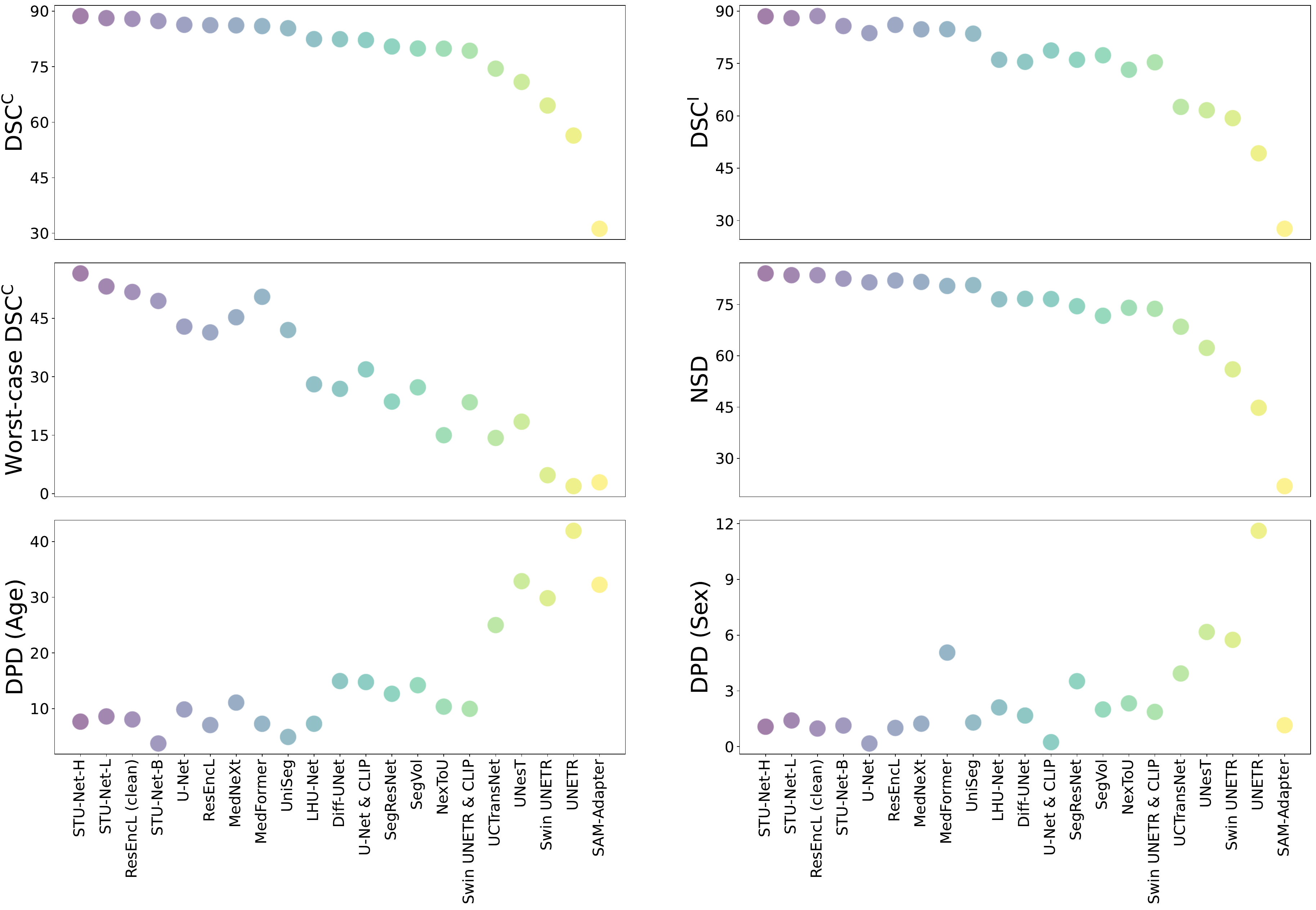}
    \caption{\textbf{Comparison of different evaluation metrics on the official test set of TotalSegmentator.} Models tend to retain a similar rank across different metrics.}
    \label{fig:metrics_ts_test}
\end{figure}

The results for different metrics are shown in Figures~\ref{fig:metrics_jhh}-\ref{fig:metrics_ts_test}. We find that models tend to retain a similar rank across different accuracy metrics, indicating that these models do not overfit to a specific metric. However, performance on the worst-case DSC\textsuperscript{C} is significantly lower than the DSC\textsuperscript{C} itself, showing that a need for improvements in model performance on these hard cases, or indicating the existence of some label noise in test sets. We visualize some worse-case examples in Appendix~\ref{sec:wrost_case_analysis}. Regarding the bias metrics, although there are some variations in rankings, we find models with high accuracy usually have low bias.

\clearpage

\clearpage

\subsubsection{Bootstrap Sampling}
\begin{figure}[h]
    \centering
    \includegraphics[width=0.5\linewidth]{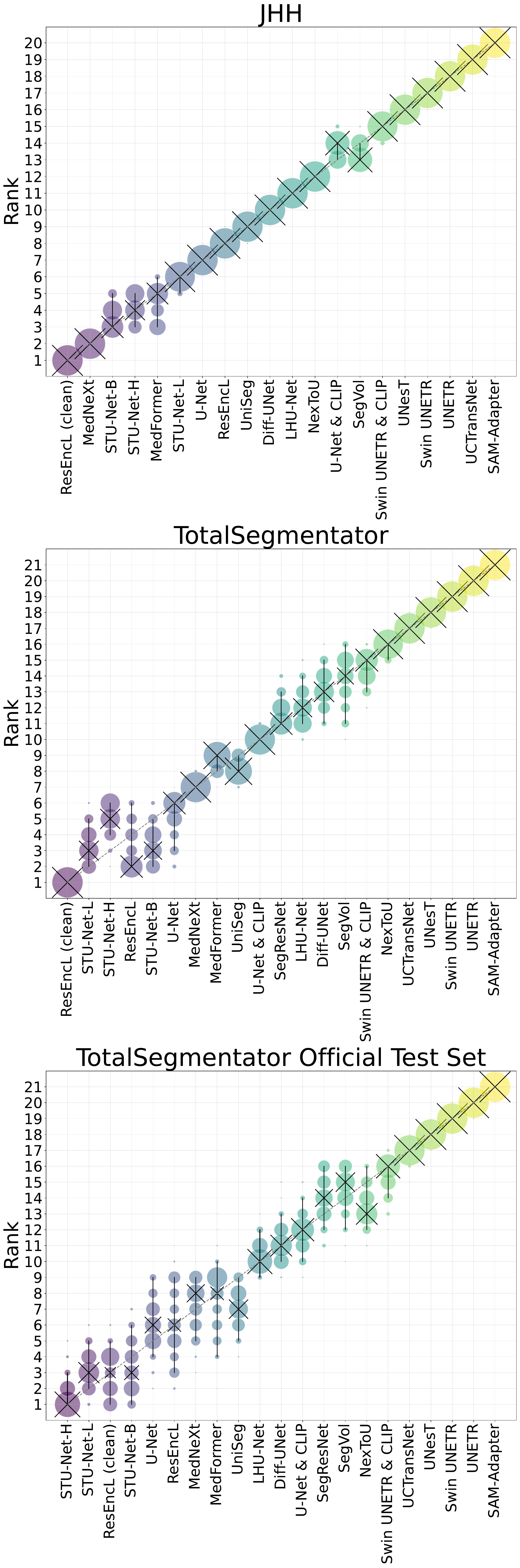}
    \caption{\textbf{Blob plots for visualizing ranking stability based on 1,000 bootstrap samples.} The area of each blob is proportional to the relative frequency. The median rank for each model is marked by a black cross. 95\% bootstrap intervals (ranging from the 2.5th to the 97.5th percentile of the bootstrap distribution) are connected by black lines. We observe more stable rankings for larger tests sets.}
    \label{fig:bootstrap}
\end{figure}

To evaluate ranking stability, we perform bootstrap sampling as described in \cite{wiesenfarth2021methods}. A bootstrap sample of a dataset with $n$ test cases consists of $n$ test cases randomly drawn from the dataset with replacement. A total of 1,000 bootstrap samples are drawn, and the results are visualized as blob plots in Figure~\ref{fig:bootstrap}.

Our findings indicate that datasets with more test cases tend to present fewer variations in ranks. For example, we find fewer variations in ranks on the entire TotalSegmentator ($N$ = \numoftetotalsegct) compared with the ranks on the TotalSegmentator official test set ($N$ = \numofofficialtetotalsegct). On the proprietary JHH dataset ($N$ = \numoftejhhct), we observe minimum ranking variations due to its large number of test cases. Additionally, the ranks are relatively robust for the highest- and lowest-performing models, but they can be more unstable for models in the middle range.

\clearpage

\clearpage

\subsubsection{Significance Maps}

To further investigate ranking stability, we performed pair-wise comparisons between each possible pair of algorithms. Comparisons use statistical tests to understand if an algorithm's scores are significantly better than the other model's results. We employed one-sided Wilcoxon signed rank tests with Holm's adjustment and 5\% significance level.

\begin{figure}[h]
    \centering
    \includegraphics[width=0.9\linewidth]{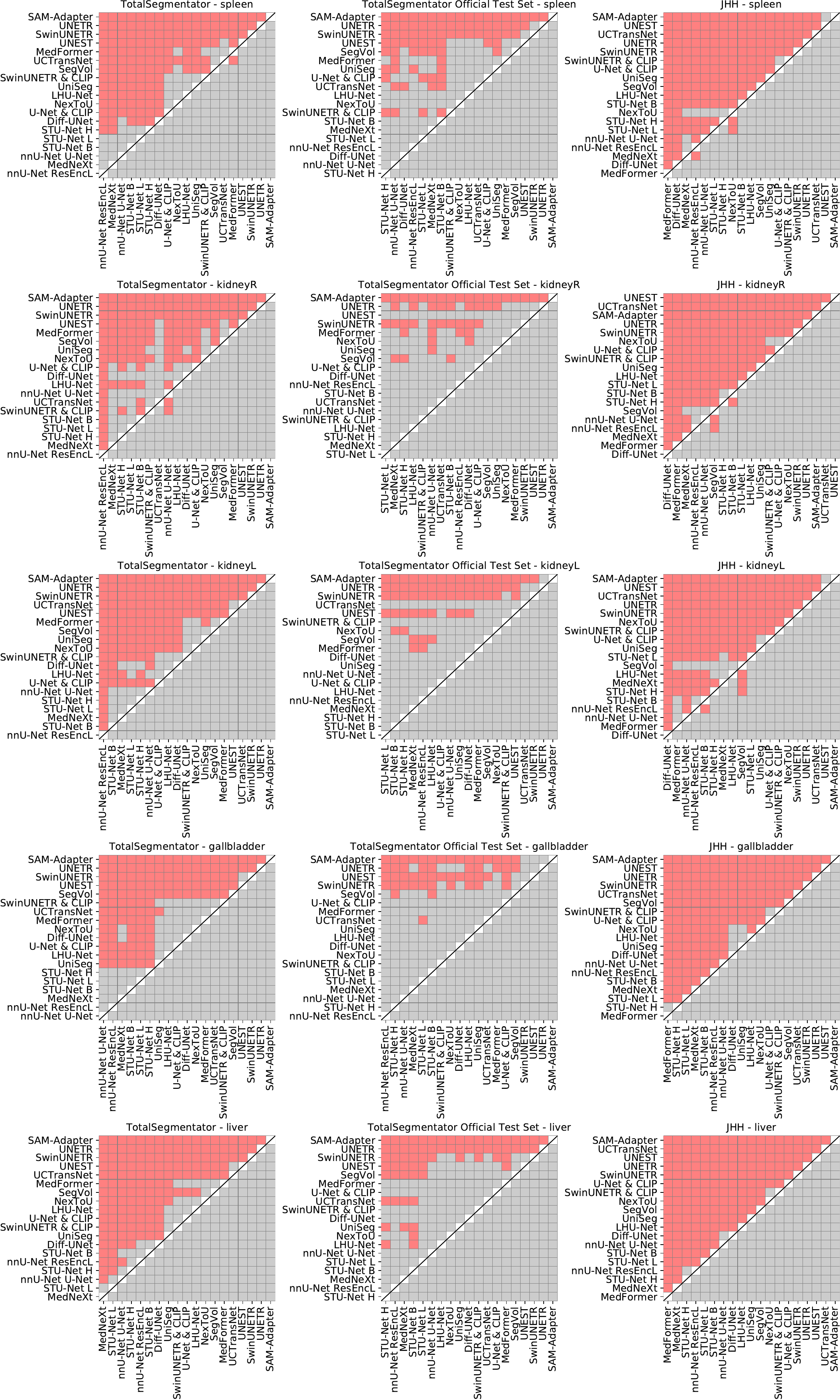}
    \caption{\textbf{DSC significance maps.} Each cell represents a pair-wise comparison between two algorithms, according to DSC score. Yellow colors indicate that the x-axis AI algorithm is significantly superior to the y-axis algorithm in terms of DSC score (considering all organs). Blue represents no significant superiority. Comparisons employed one-sided Wilcoxon signed rank tests with Holm's adjustment and 5\% significance level.}
    \label{fig:heatmap}
\end{figure}

\begin{figure}[h]
    \centering
    \includegraphics[width=0.9\linewidth]{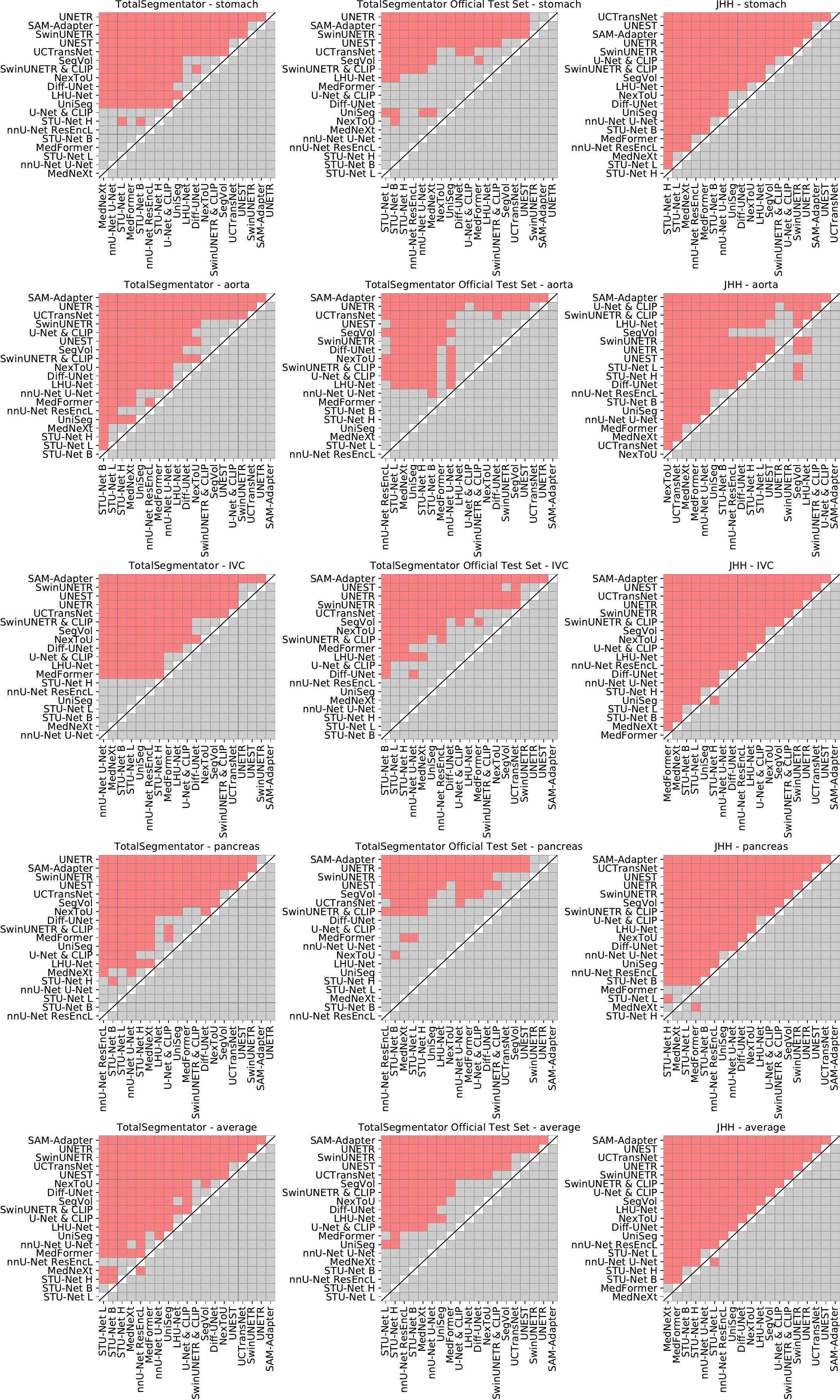}
    \caption{\textbf{Continuation of DSC significance maps.} Each cell represents a pair-wise comparison between two algorithms, according to DSC score. Yellow colors indicate that the x-axis AI algorithm is significantly superior to the y-axis algorithm in terms of DSC score (considering all organs). Blue represents no significant superiority. Comparisons employed one-sided Wilcoxon signed rank tests with Holm's adjustment and 5\% significance level.}
    \label{fig:heatmap2}
\end{figure}

\begin{figure}[h]
    \centering
    \includegraphics[width=0.9\linewidth]{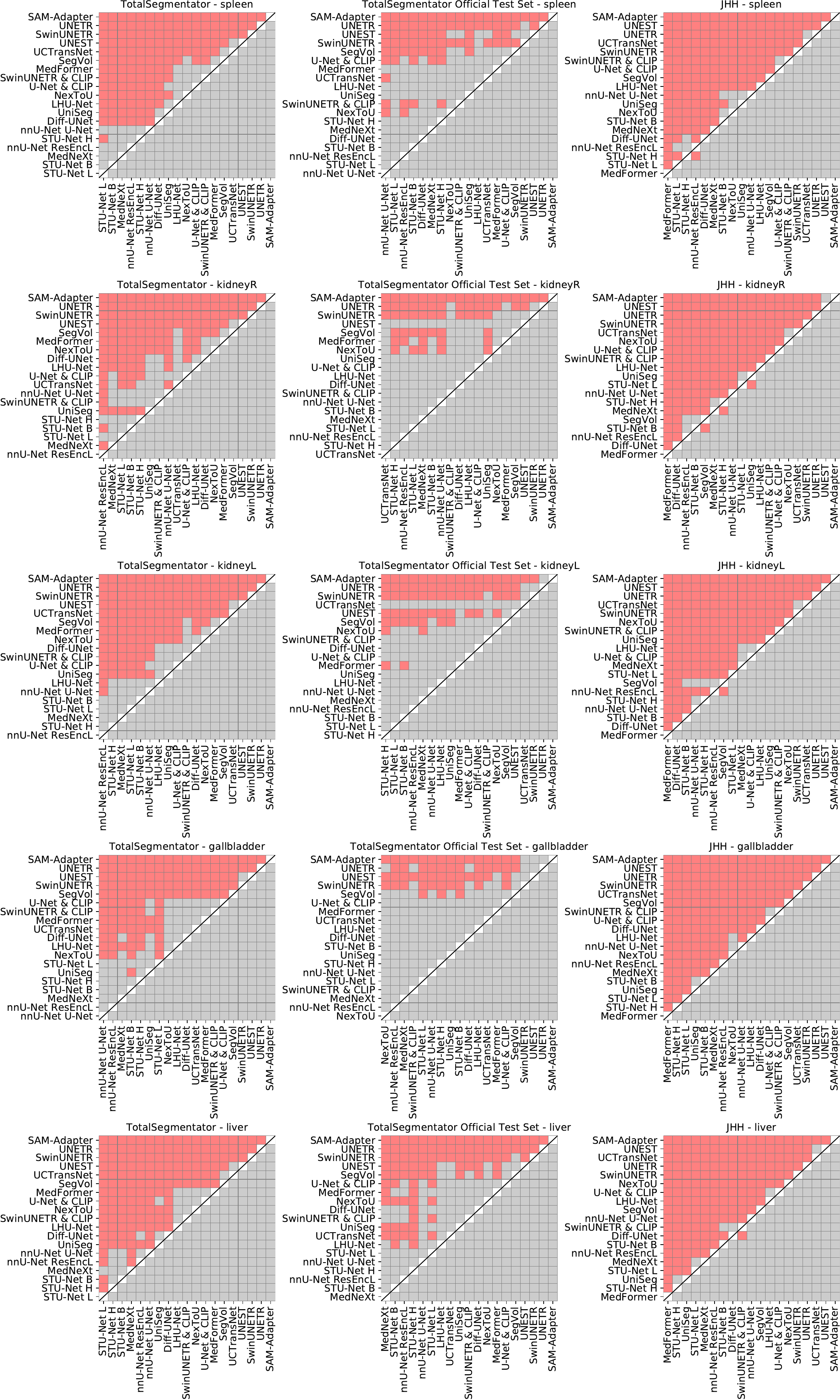}
    \caption{\textbf{NSD significance maps.} Each cell represents a pair-wise comparison between two algorithms, according to NSD. Yellow colors indicate that the x-axis AI algorithm is significantly superior to the y-axis algorithm in terms of NSD score (considering all organs). Blue represents no significant superiority. Comparisons employed one-sided Wilcoxon signed rank tests with Holm's adjustment and 5\% significance level. NSD considers a threshold of 1.5mm.}
    \label{fig:heatmap}
\end{figure}

\begin{figure}[h]
    \centering
    \includegraphics[width=0.9\linewidth]{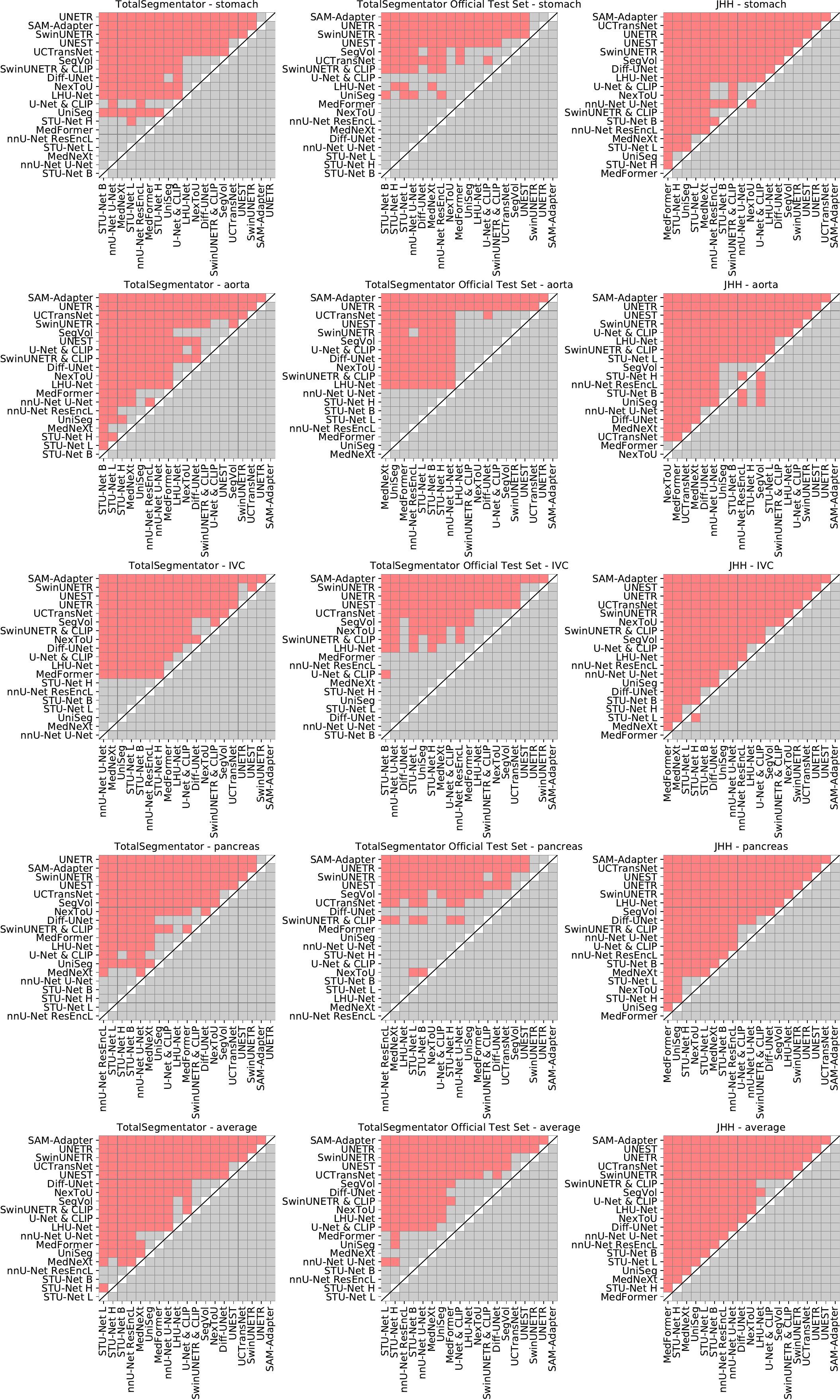}
    \caption{\textbf{Continuation of NSD significance maps.} Each cell represents a pair-wise comparison between two algorithms, according to NSD. Yellow colors indicate that the x-axis AI algorithm is significantly superior to the y-axis algorithm in terms of NSD score (considering all organs). Blue represents no significant superiority. Comparisons employed one-sided Wilcoxon signed rank tests with Holm's adjustment and 5\% significance level. NSD considers a threshold of 1.5mm.}
    \label{fig:heatmap2}
\end{figure}

\clearpage
\subsection{Per-Algorithm Analysis}\label{sec:per_algorithm_analysis}

\subsubsection{MedNeXt}

The central theme of the ConvNeXt \cite{liu2022convnet} architecture was decoupling the scalability of the Transformer architecture and using it in a convolutional fashion, without self-attention. Scalability becomes relevant for medical images when creating large 3D networks while not overfitting. MedNeXt \cite{roy2023mednext} builds upon this principle by using these blocks across the network, leading to the performance seen in this work.

\subsubsection{STU-Net}

The STU-Net \cite{huang2023stu} is built upon the nnU-Net framework, which was proven effective in our experiments. Additionally, STU-Net is based on scaling the AI model size, which may be exceptionally useful for dealing with large-scale datasets like \ourdataset. The combination of a high-performance framework and an appropriately scaled model may be the key for STU-Net's high segmentation accuracy in this study.

\subsubsection{NexToU}

NexToU \cite{shi2023nextou} is a hybrid architecture that combines a hierarchical 3D U-shaped encoder-decoder structure with both Convolutional Neural Networks (CNNs) and Graph Neural Networks (GNNs). This innovative approach employs a hierarchical, topology-aware strategy inspired by human cognitive processes, allowing the model to progressively decompose anatomical semantics from simpler to more complex structures. On the JHH dataset, NexToU's results were relatively close to the best-performing models. However, we observed a significant performance difference on the TotalSegmentator dataset. This discrepancy is likely due to our model not utilizing a resampling step to the average spatial resolution during inference for data with fewer slices along the z-axis. While this approach saves inference time, it compromises performance on data with low z-axis resolution. Additionally, to further reduce inference time, Test Time Augmentation (TTA) was minimized, leading to a decline in performance for bilaterally symmetric classes like kidneyR and kidneyL, as well as for some small sample classes.

\subsubsection{DiffU-Net}

We hypothesize that two main factors contributed to Diff-UNet's \cite{xing2023diff} high segmentation accuracy: its nnU-Net-inspired hyper-parameter selection procedure and the use of stable diffusion. The diffusion model excels in handling details, generating high-resolution images when used as a generative model. During inference, the model predicts multiple times using the DDIM sampling strategy, further enhancing Diff-UNet's outputs. Moreover, considering that the diffusion model includes noised information, DiffU-Net has a boundary branch, which takes the 3D medical image as input. This branch supplies clear image information to complement the diffusion branch, further improving segmentation accuracy.

\subsubsection{SAM-Adapter}\label{sec:sam}

We observed a lower performance for the fine-tuned Segment Anything model, which we hypothesize may be due to the following reasons:
\begin{itemize}
    \item  The SAM-based model is a 2D-based model that performs multi-class segmentation solely on 2D slices. This approach relies mainly on 2D information, such as location relations, rather than 3D organ shape information. When tested on out-of-distribution (OOD) sets, images from different hospitals may introduce spatial variations and voxel spacing, leading to varying spatial distributions of abdomen regions compared to the training images. These spatial changes can cause the 2D-based model to lose its segmentation accuracy.
    
   \item During the training of this fine-tuned model, no spatial transformations for augmentation were used, which might have been used in other comparison methods. This lack of augmentation could lead to poorer generalization on spatial changes in OOD data.
\end{itemize}

Possibly, the use of spatial transformations during training and inference could improve the SAM-Adapter results.

\clearpage
\subsection{Per-Class Analysis}\label{sec:appendix_boxplots}

\subsubsection{JHH}

\begin{figure}[h]
	\centering
	\includegraphics[width=\columnwidth]{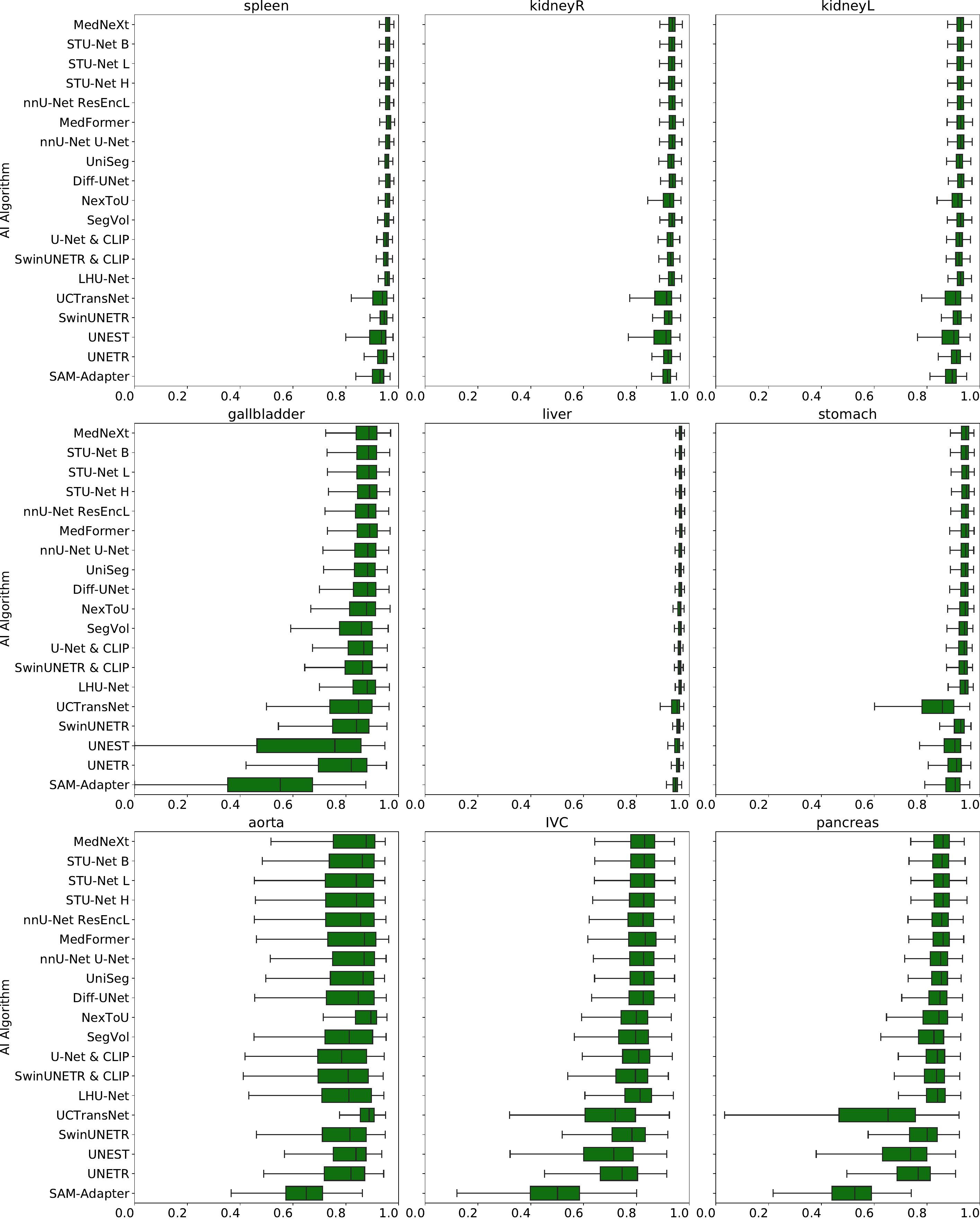}
    \caption{\textbf{Boxplots showing DSC score in JHH, per class.} Performances are not homogeneous across classes: structures like the liver, which are easier to segment, show higher median scores and smaller score variation, when compared to more difficult structures, like the gallbladder.}
	\label{fig:plot_TS_liver}
\end{figure}

\begin{figure}[h]
	\centering
	\includegraphics[width=\columnwidth]{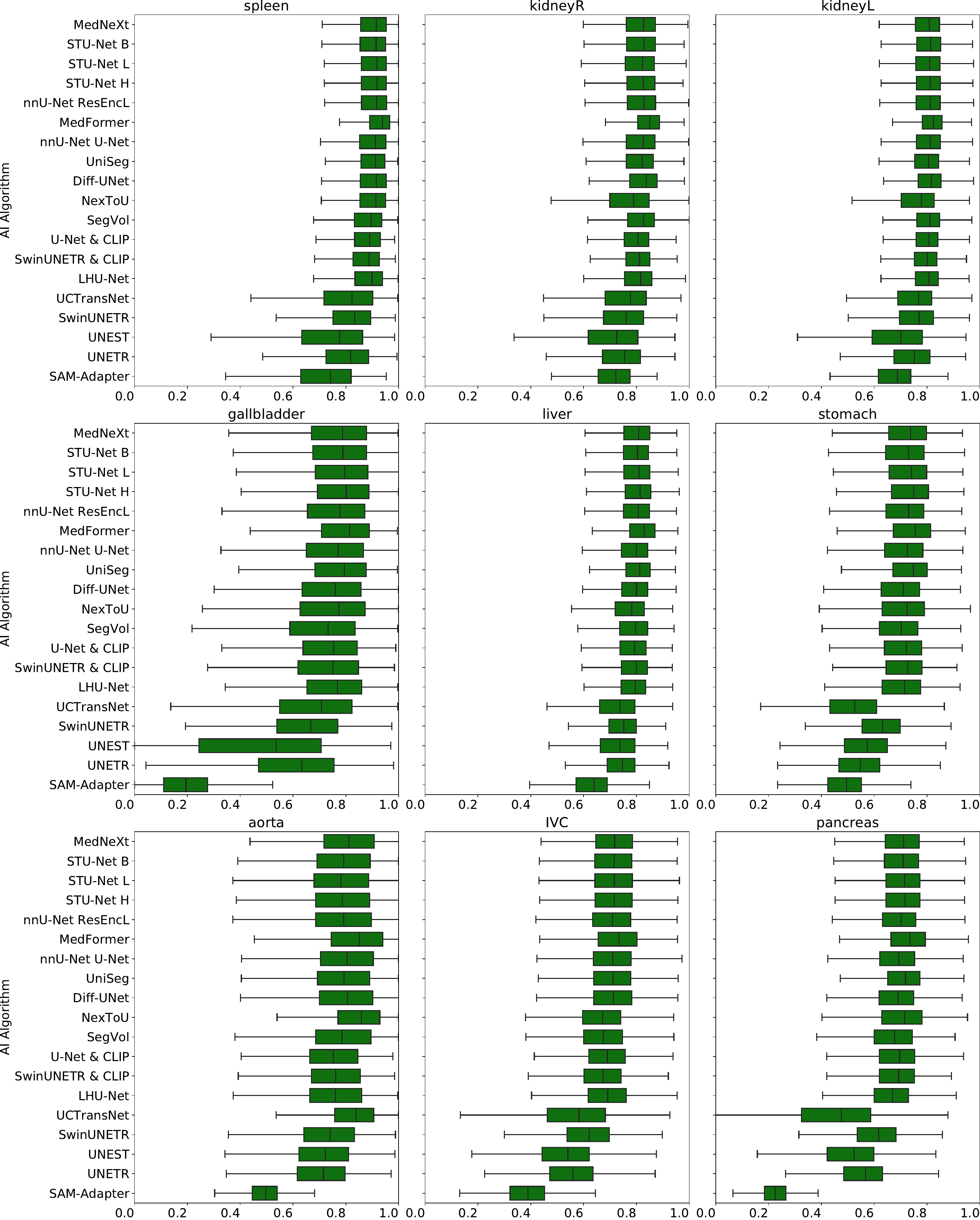}
    \caption{\textbf{Boxplots showing NSD score in JHH, per class.} Performances are not homogeneous across classes: structures like the liver, which are easier to segment, show higher median scores and smaller score variation, when compared to more difficult structures, like the gallbladder. NSD considers a threshold of 1.5mm.}
	\label{fig:plot_TS_liver}
\end{figure}

\clearpage
\subsubsection{TotalSegmentator}

\begin{figure}[h]
	\centering
	\includegraphics[width=\columnwidth]{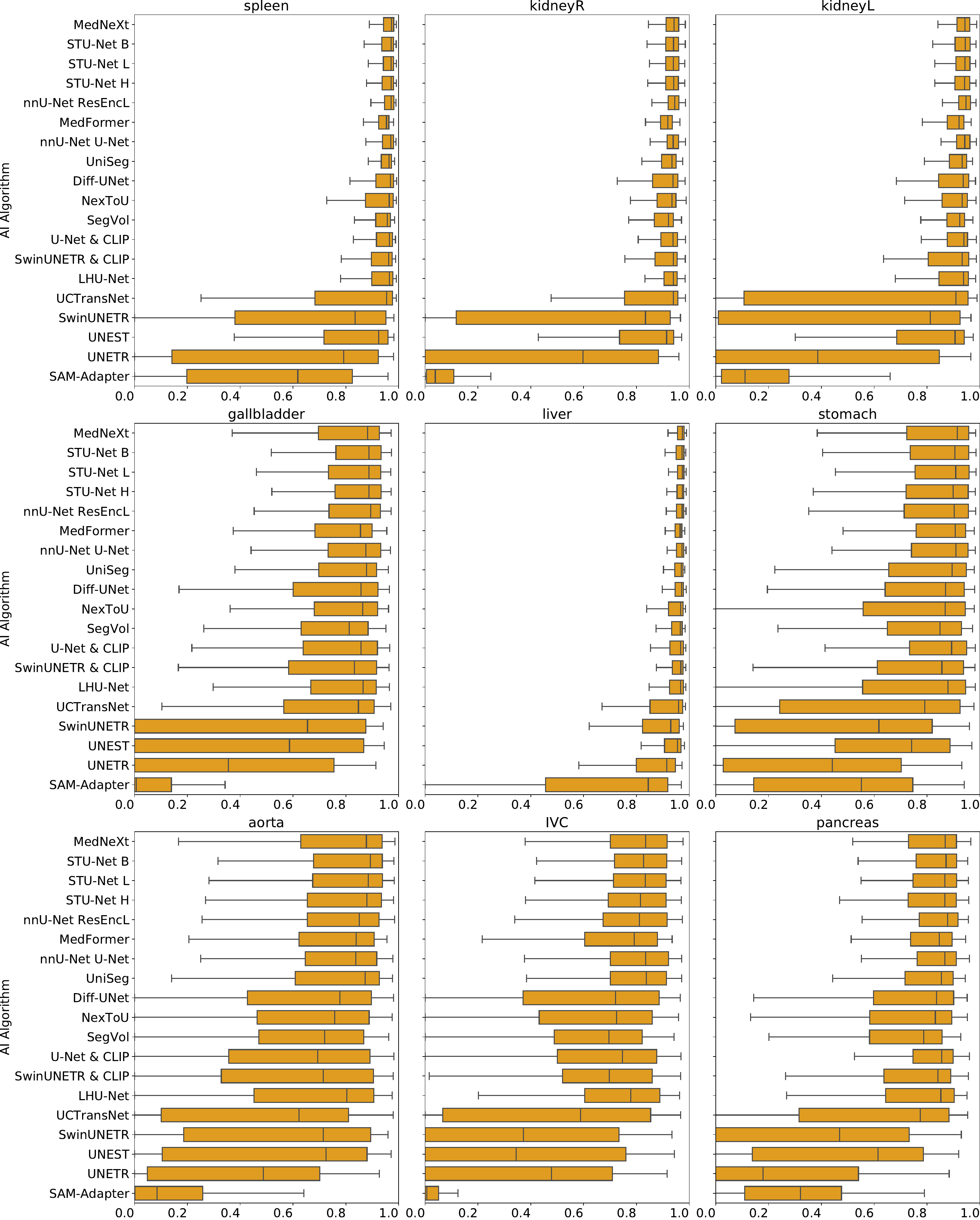}
    \caption{\textbf{Boxplots showing DSC score in the entire TotalSegmentator dataset, per class.} Performances are not homogeneous across classes: structures like the liver, which are easier to segment, show higher median scores and smaller score variation, when compared to more difficult structures, like the gallbladder.}
	\label{fig:plot_TS_liver}
\end{figure}

\begin{figure}[h]
	\centering
	\includegraphics[width=\columnwidth]{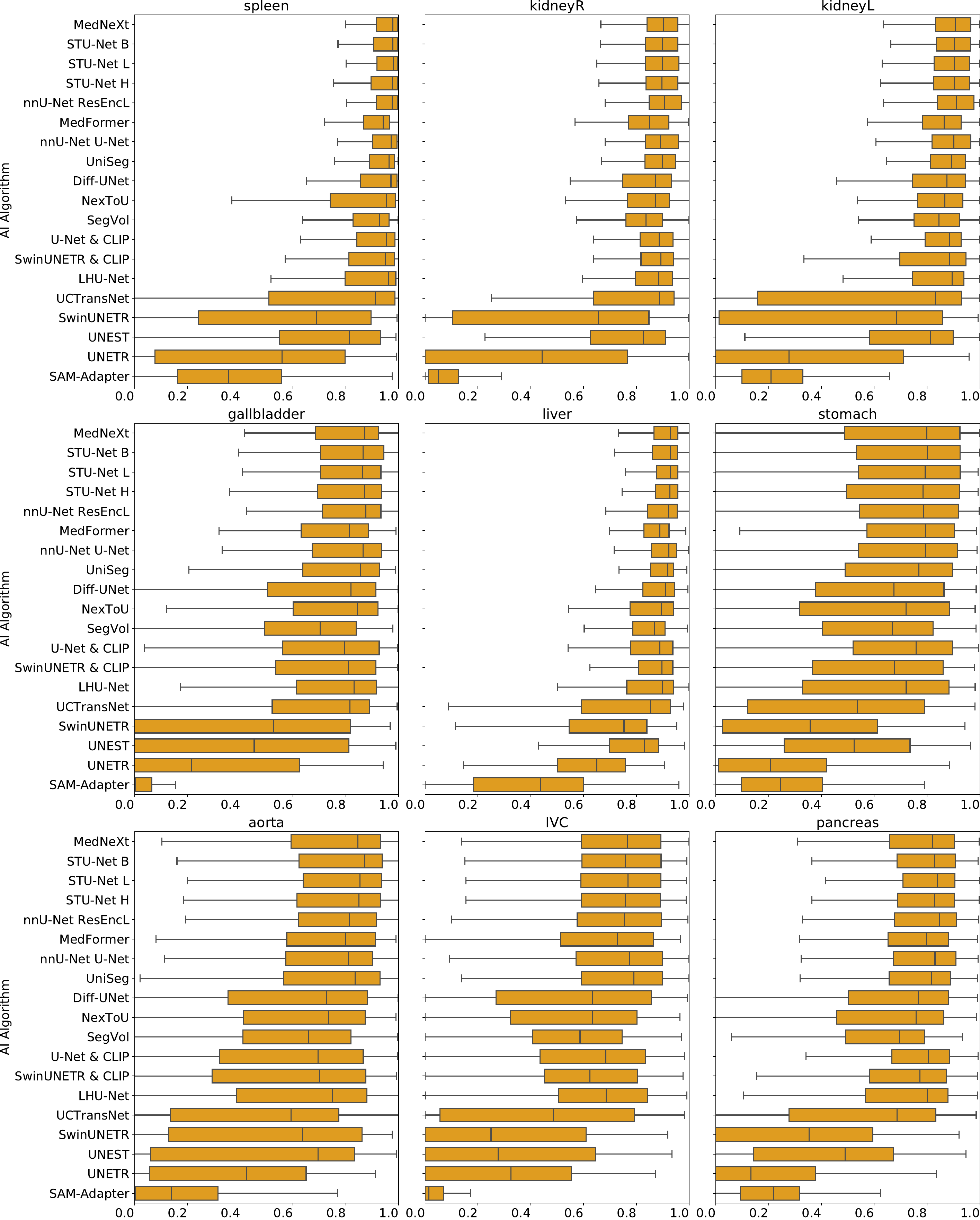}
    \caption{\textbf{Boxplots showing NSD score in the entire TotalSegmentator dataset, per class.} Performances are not homogeneous across classes: structures like the liver, which are easier to segment, show higher median scores and smaller score variation, when compared to more difficult structures, like the gallbladder. NSD considers a threshold of 1.5mm.}
	\label{fig:plot_TS_liver}
\end{figure}

\clearpage
\subsubsection{TotalSegmentator Official Test Set}

\begin{figure}[h]
	\centering
	\includegraphics[width=\columnwidth]{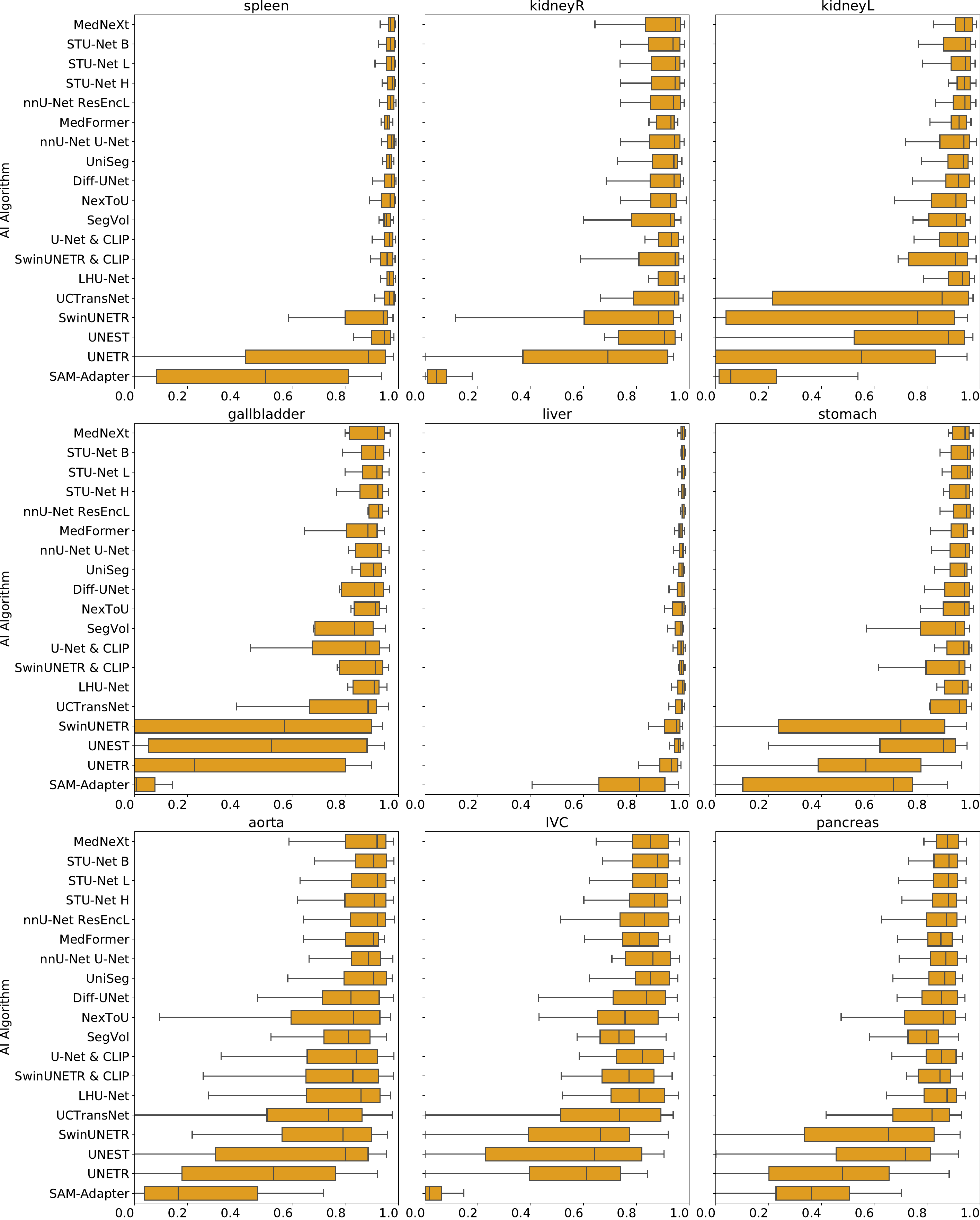}
    \caption{\textbf{Boxplots showing DSC score in the TotalSegmentator official test dataset, per class.} Performances are not homogeneous across classes: structures like the liver, which are easier to segment, show higher median scores and smaller score variation, when compared to more difficult structures, like the gallbladder.}
	\label{fig:plot_TS_liver}
\end{figure}

\begin{figure}[h]
	\centering
	\includegraphics[width=\columnwidth]{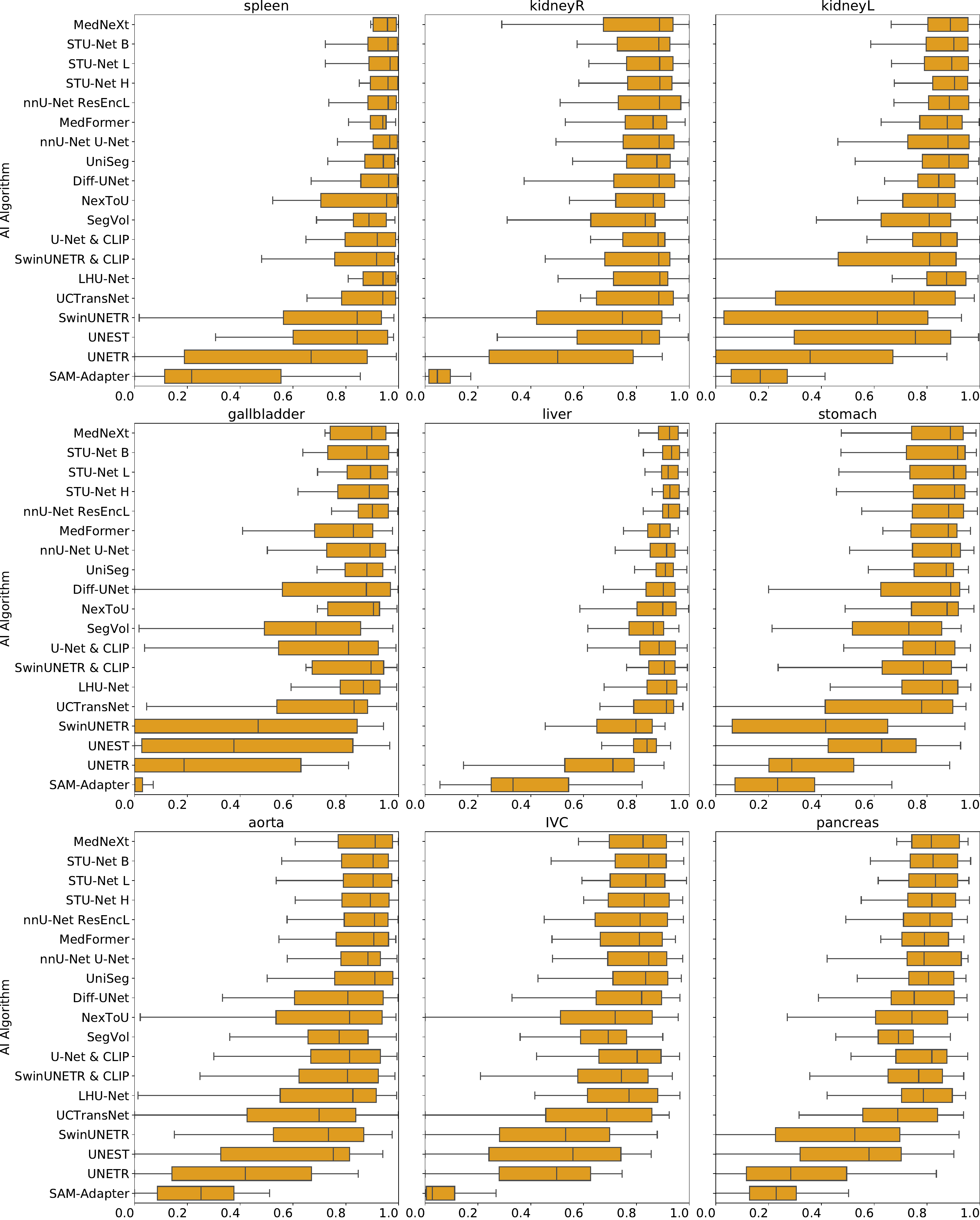}
    \caption{\textbf{Boxplots showing NSD score in the TotalSegmentator official test dataset, per class.} Performances are not homogeneous across classes: structures like the liver, which are easier to segment, show higher median scores and smaller score variation, when compared to more difficult structures, like the gallbladder. NSD considers a threshold of 1.5mm.}
	\label{fig:plot_TS_liver}
\end{figure}

\clearpage
\subsection{Per-Group Metadata Analysis}\label{sec:per_group_analysis}

\label{sec:plots_per_group}

\subsubsection{Age}

\begin{figure}[!h]
	\centering
	\includegraphics[height=0.8\textheight]{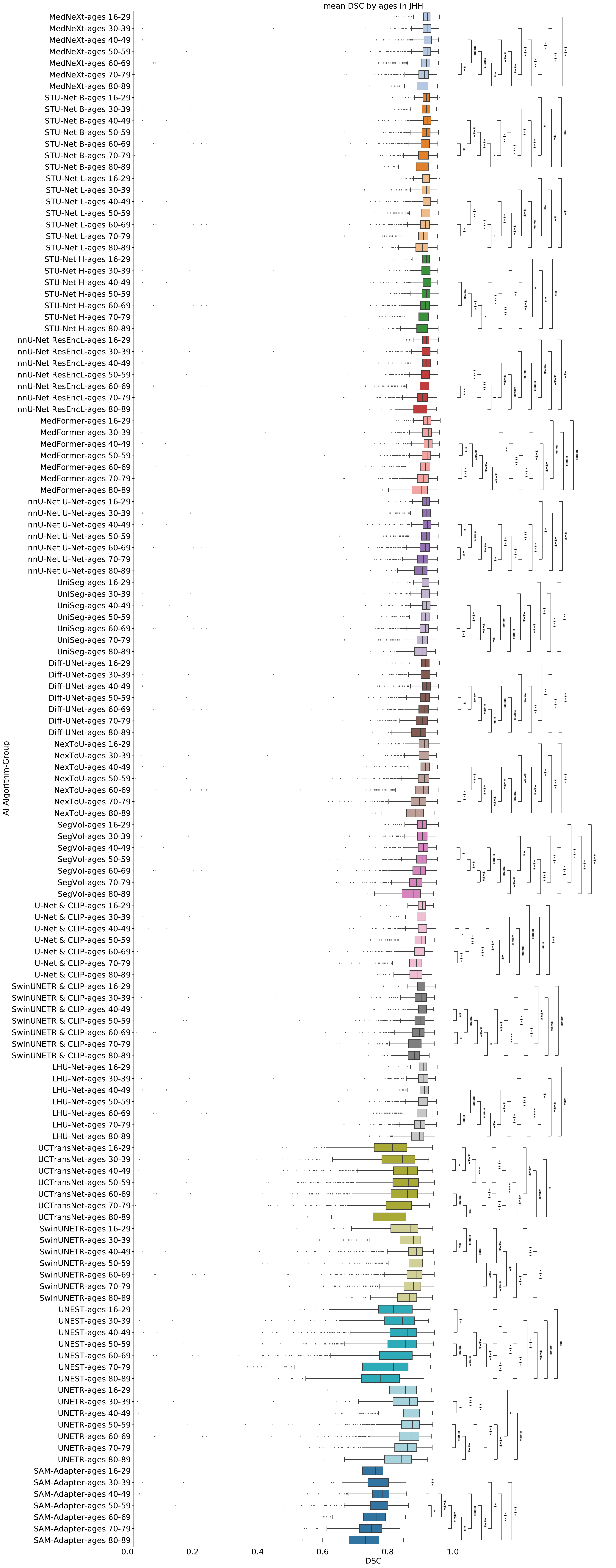}
    \caption{\textbf{Boxplot showing average DSC score by age in JHH.} Statistical significance is indicated by stars: * p < 0.05, ** p <0.01, *** p < 0.001, **** p < 0.0001. We perform Kruskal–Wallis tests followed by post-hoc Mann-Whitney U Tests with Bonferroni correction. Here, we did not perform statistical comparisons between diverse AI algorithms. Significant (at least p<0.05) reductions in DSC score for groups with advanced age are observed for all AI algorithms.}
\end{figure}

\begin{figure}[!h]
	\centering
	\includegraphics[height=0.9\textheight]{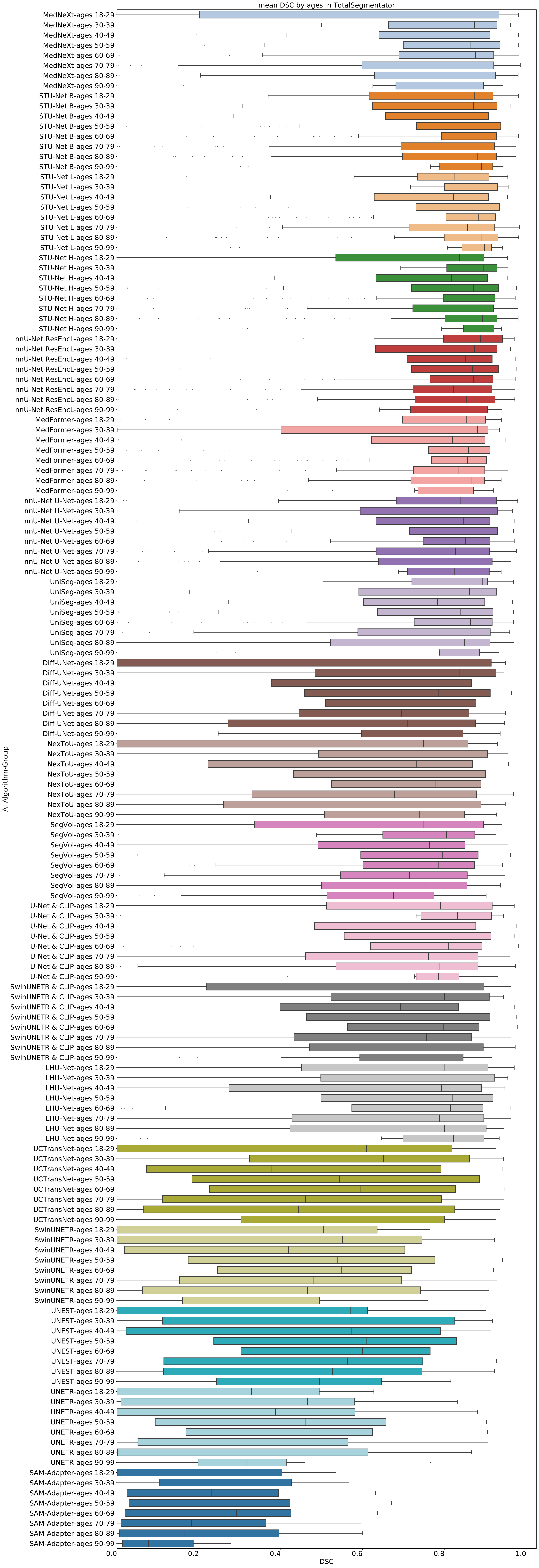}
    \caption{\textbf{Boxplot showing average DSC score by age in the whole TotalSegmentator dataset.} Statistical significance is indicated by stars: * p < 0.05, ** p <0.01, *** p < 0.001, **** p < 0.0001. We perform Kruskal–Wallis tests followed by post-hoc Mann-Whitney U Tests with Bonferroni correction. Here, we did not perform statistical comparisons between diverse AI algorithms. Significant differences are not observed, possibly due to the higher variability in the TotalSegmentator results, when compared to other datasets.}
\end{figure}

\clearpage
\subsubsection{Diagnosis}

\begin{figure}[h]
	\centering
	\includegraphics[width=0.9\columnwidth]{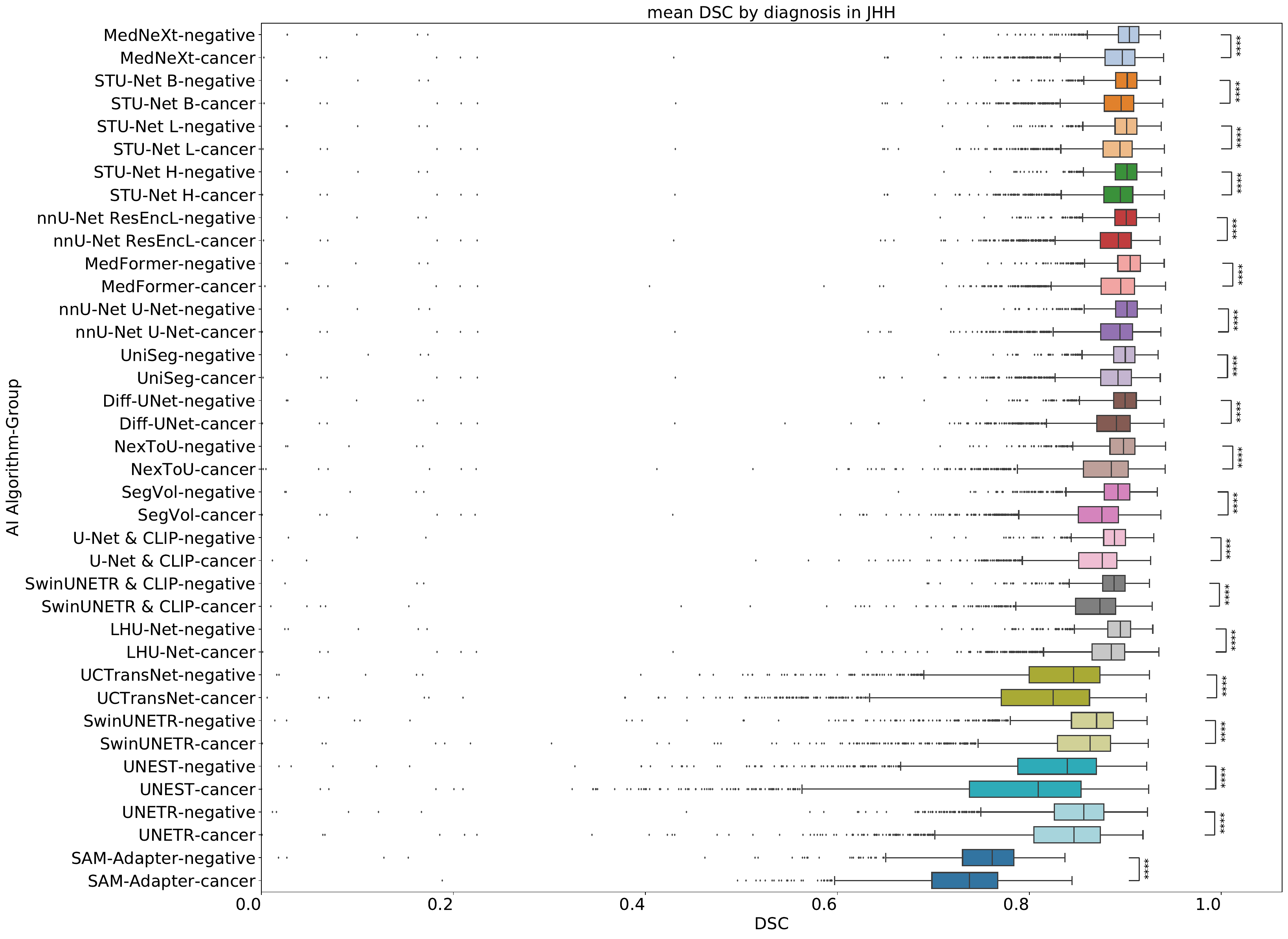}
    \caption{\textbf{Boxplot showing average DSC score by diagnosis in JHH.} Statistical significance is indicated by stars: * p < 0.05, ** p <0.01, *** p < 0.001, **** p < 0.0001. We perform Kruskal–Wallis tests followed by post-hoc Mann-Whitney U Tests with Bonferroni correction. Here, we did not perform statistical comparisons between diverse AI algorithms.}
\end{figure}

\begin{figure}[h]
	\centering
	\includegraphics[height=0.9\textheight]{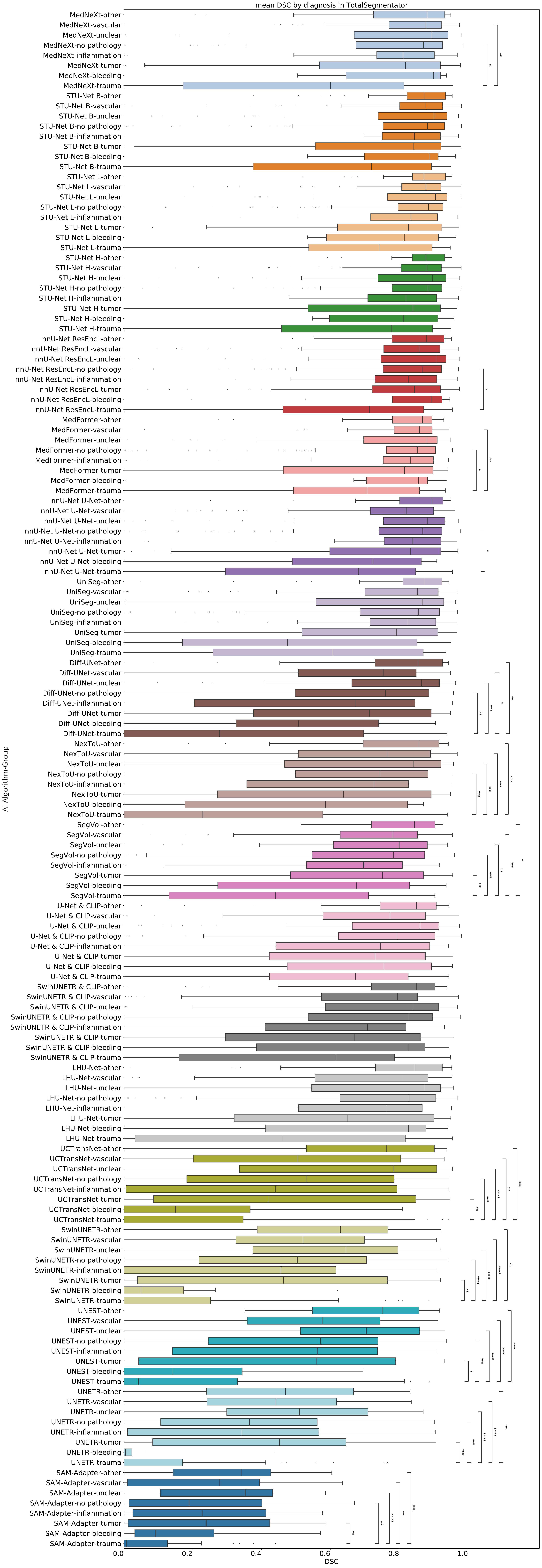}
    \caption{\textbf{Boxplot showing average DSC score by diagnosis in the whole TotalSegmentator dataset.} Statistical significance is indicated by stars: * p < 0.05, ** p <0.01, *** p < 0.001, **** p < 0.0001. We perform Kruskal–Wallis tests followed by post-hoc Mann-Whitney U Tests with Bonferroni correction. Here, we did not perform statistical comparisons between diverse AI algorithms. }
\end{figure}

\clearpage
\subsubsection{Sex}

\begin{figure}[h]
	\centering
	\includegraphics[width=0.9\columnwidth]{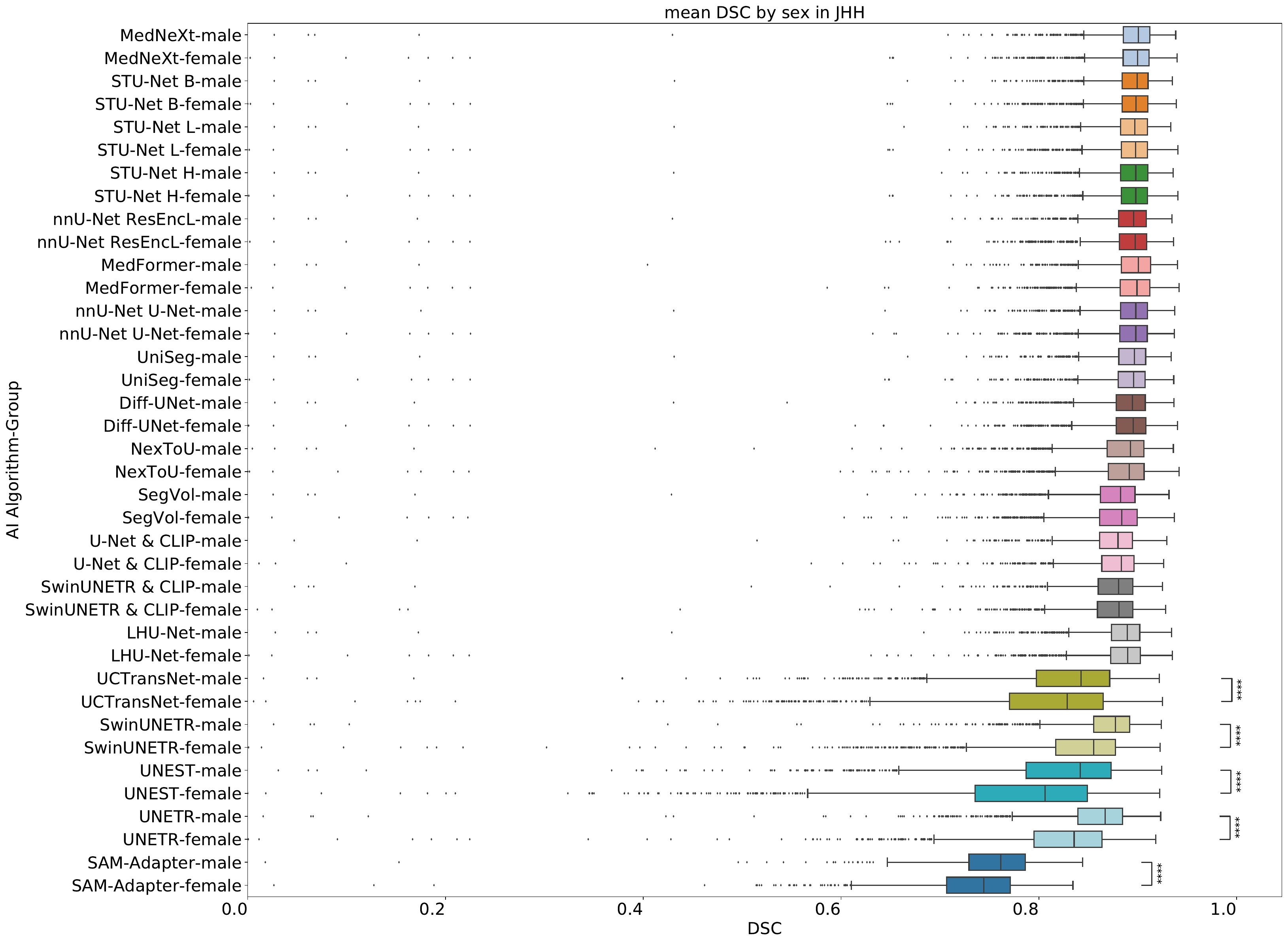}
    \caption{\textbf{Boxplot showing average DSC score by sex in JHH.} Statistical significance is indicated by stars: * p < 0.05, ** p <0.01, *** p < 0.001, **** p < 0.0001. We perform Kruskal–Wallis tests followed by post-hoc Mann-Whitney U Tests with Bonferroni correction. Here, we did not perform statistical comparisons between diverse AI algorithms. Only the worst performing algorithms show significant performance difference for the male and female groups, with better scores for male. The best performing models show no significant difference.}
\end{figure}

\begin{figure}[h]
	\centering
	\includegraphics[width=0.9\columnwidth]{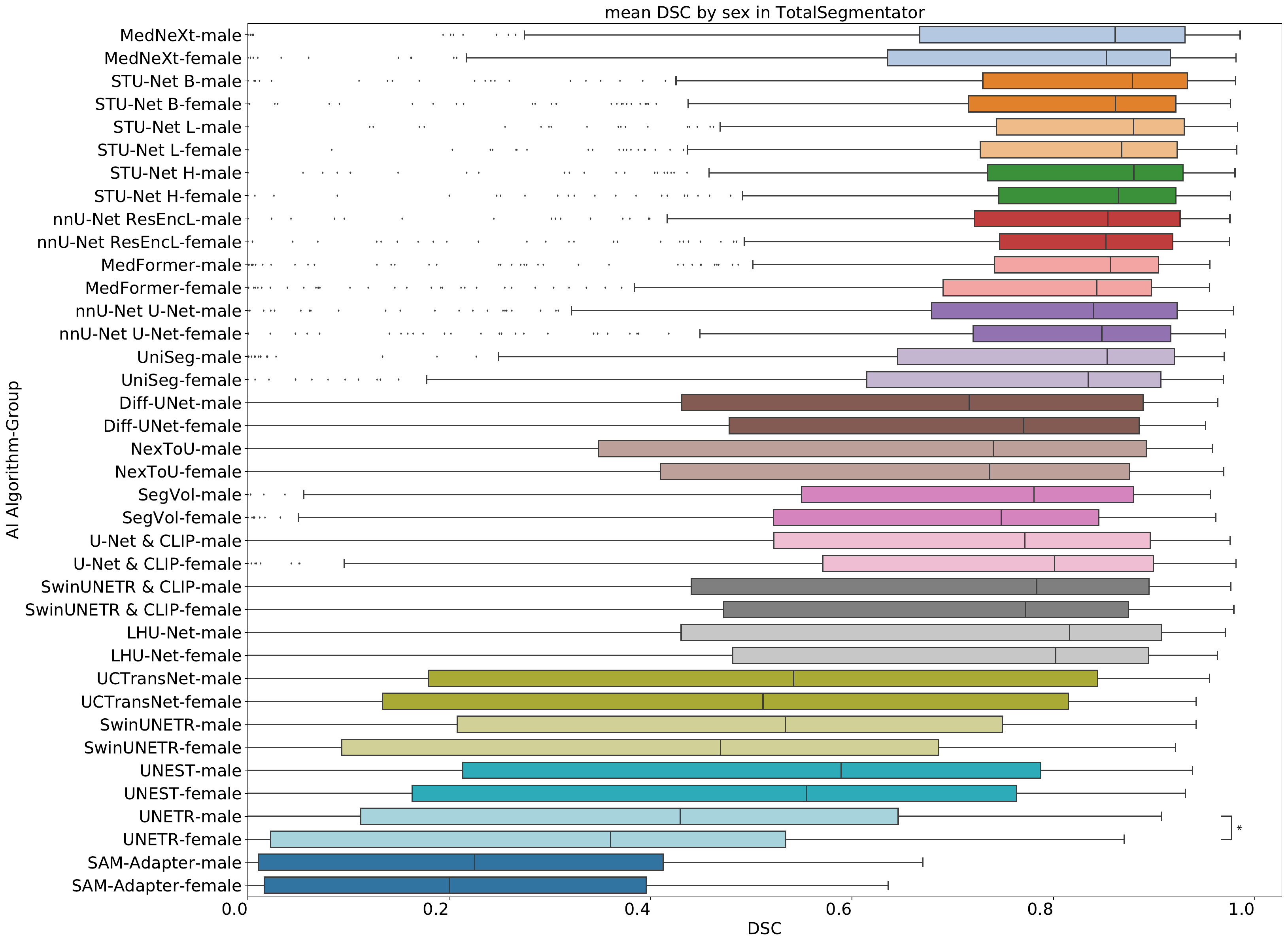}
    \caption{\textbf{Boxplot showing average DSC score by sex in the whole TotalSegmentator dataset.} Statistical significance is indicated by stars: * p < 0.05, ** p <0.01, *** p < 0.001, **** p < 0.0001. We perform Kruskal–Wallis tests followed by post-hoc Mann-Whitney U Tests with Bonferroni correction. Here, we did not perform statistical comparisons between diverse AI algorithms. Only the worst performing algorithms show significant performance difference for the male and female groups, with better scores for male. The best performing models show no significant difference.}
\end{figure}

\clearpage
\subsubsection{Race}

\begin{figure}[h]
	\centering
	\includegraphics[width=0.65\columnwidth]{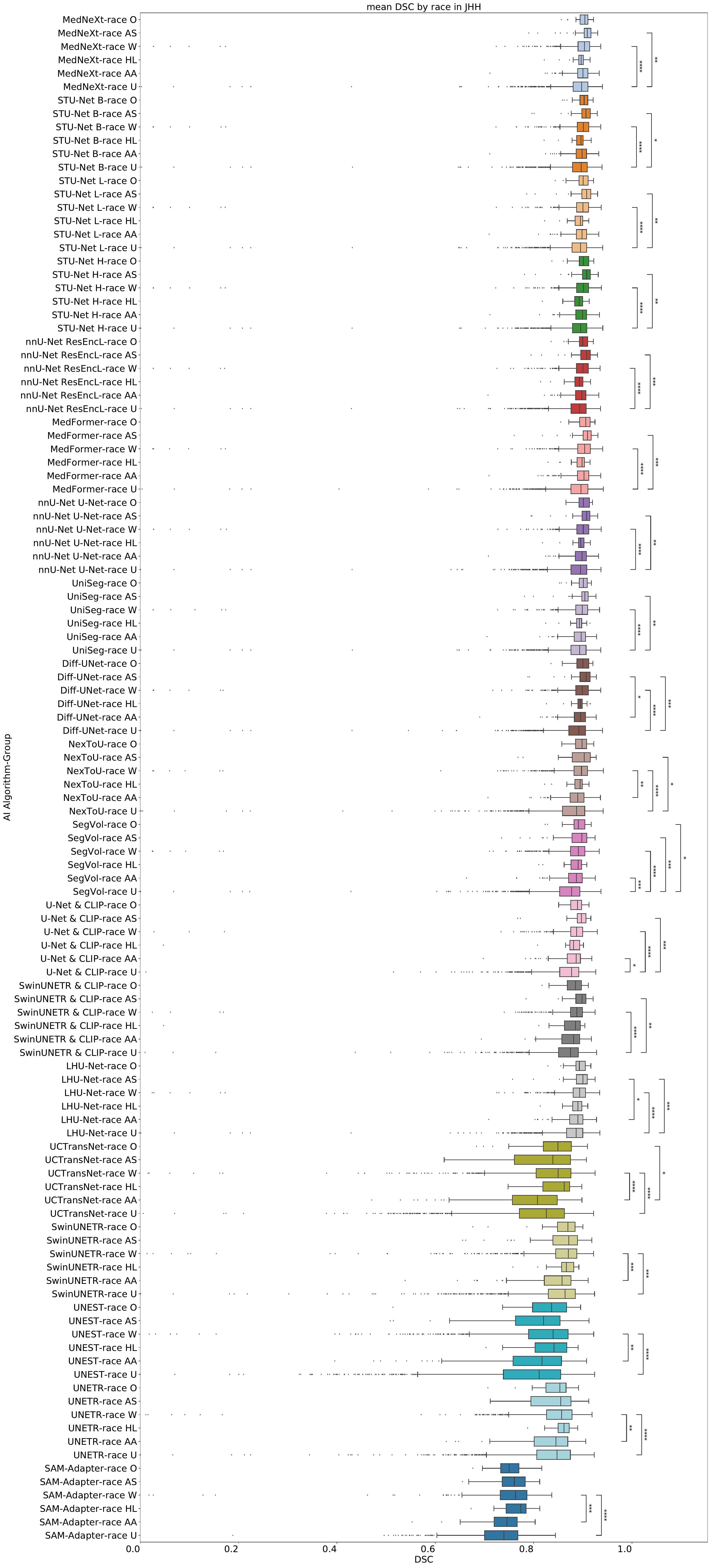}
    \caption{\textbf{Boxplot showing average DSC score by race in JHH.} Statistical significance is indicated by stars: * p < 0.05, ** p <0.01, *** p < 0.001, **** p < 0.0001. We perform Kruskal–Wallis tests followed by post-hoc Mann-Whitney U Tests with Bonferroni correction. Here, we did not perform statistical comparisons between diverse AI algorithms. Only some algorithms show significant performance differences across race groups. In these cases, the white or Asian groups have significantly better results than African American or Hispanic Latino (usually than African American). Possibly, this finding indicates a predominance of white and Asian people in the training data, and the necessity of increasing the proportion of African Americans and Hispanic Latinos in the training dataset.}
\end{figure}

\clearpage
\subsubsection{Manufacturer}

\begin{figure}[h]
	\centering
	\includegraphics[width=0.9\columnwidth]{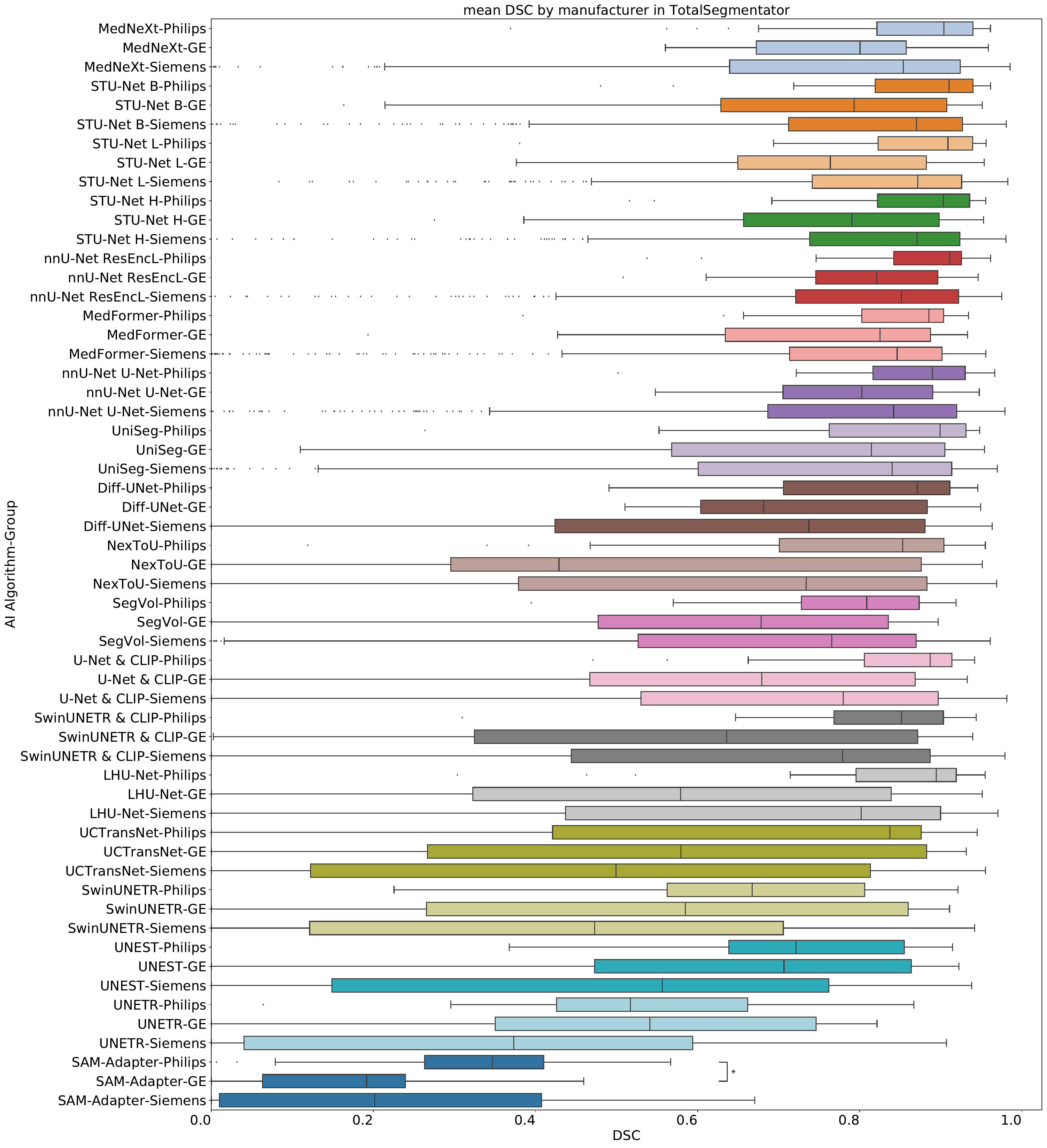}
    \caption{\textbf{Boxplot showing average DSC score by manufacturer in the whole TotalSegmentator dataset.} Statistical significance is indicated by stars: * p < 0.05, ** p <0.01, *** p < 0.001, **** p < 0.0001. We perform Kruskal–Wallis tests followed by post-hoc Mann-Whitney U Tests with Bonferroni correction. Here, we did not perform statistical comparisons between diverse AI algorithms. }
	\label{fig:plot_TS_manufacturer}
\end{figure}

\clearpage
\subsubsection{Institutes}

\begin{figure}[h]
	\centering
	\includegraphics[height=0.7\textheight]{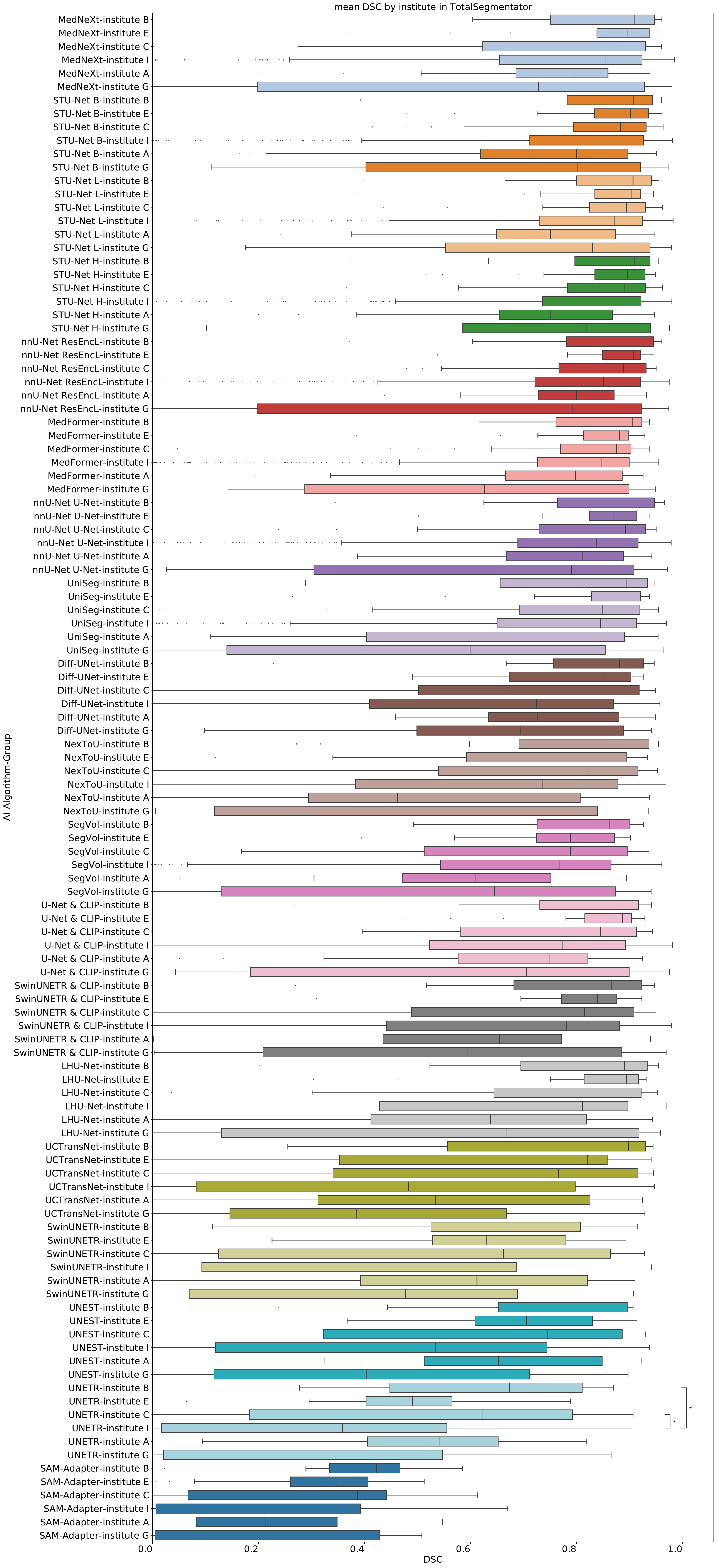}
    \caption{\textbf{Boxplot showing average DSC score by institute in the whole TotalSegmentator dataset.} Statistical significance is indicated by stars: * p < 0.05, ** p <0.01, *** p < 0.001, **** p < 0.0001. We perform Kruskal–Wallis tests followed by post-hoc Mann-Whitney U Tests with Bonferroni correction. Here, we did not perform statistical comparisons between diverse AI algorithms. Significant differences across institutes are observed for most AI algorithms, even though all institutes are located on the same country (Switzerland). This finding shows the difficulty of OOD generalization.}
	\label{fig:plot_TS_institute}
\end{figure}

\clearpage
\subsubsection{Age: per-class analysis in JHH}

\begin{figure}[!h]
	\centering
	\includegraphics[height=0.8\textheight]{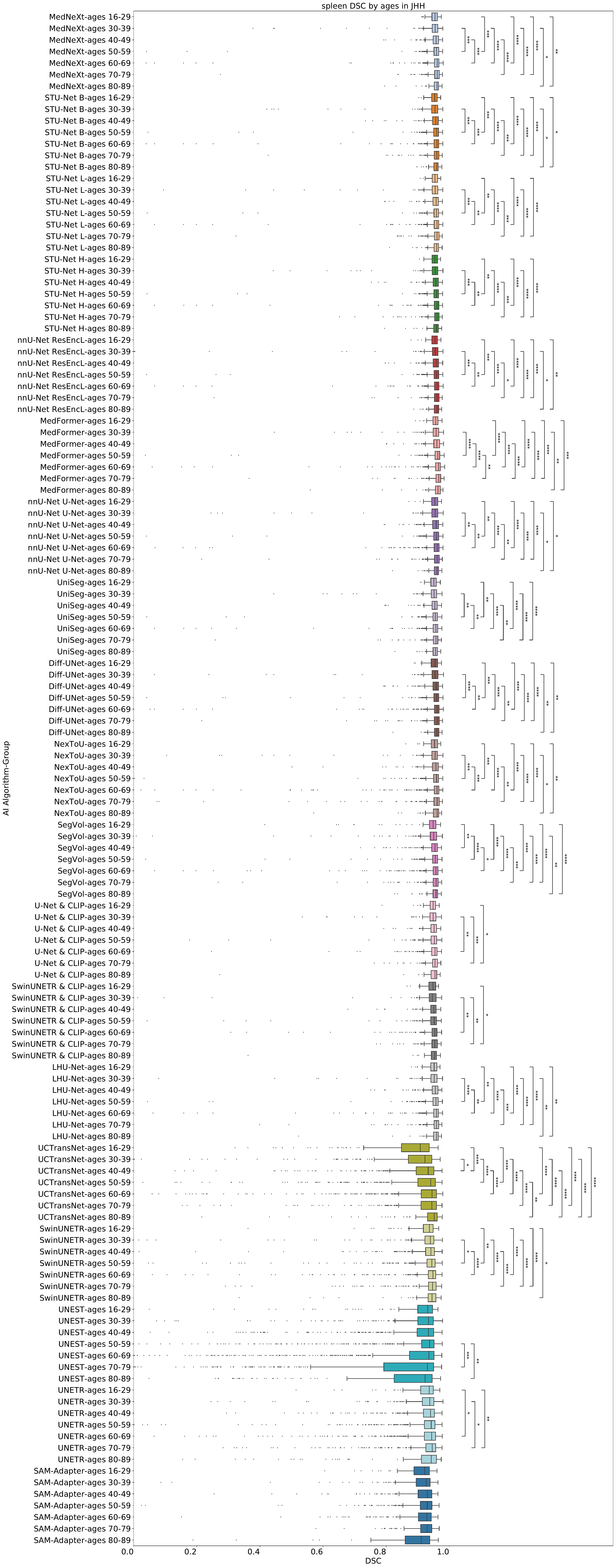}
    \caption{\textbf{Boxplot showing spleen DSC score by age in JHH.} Statistical significance is indicated by stars: * p < 0.05, ** p <0.01, *** p < 0.001, **** p < 0.0001. We perform Kruskal–Wallis tests followed by post-hoc Mann-Whitney U Tests with Bonferroni correction. Here, we did not perform statistical comparisons between diverse AI algorithms.}
\end{figure}

\begin{figure}[!h]
	\centering
	\includegraphics[height=0.8\textheight]{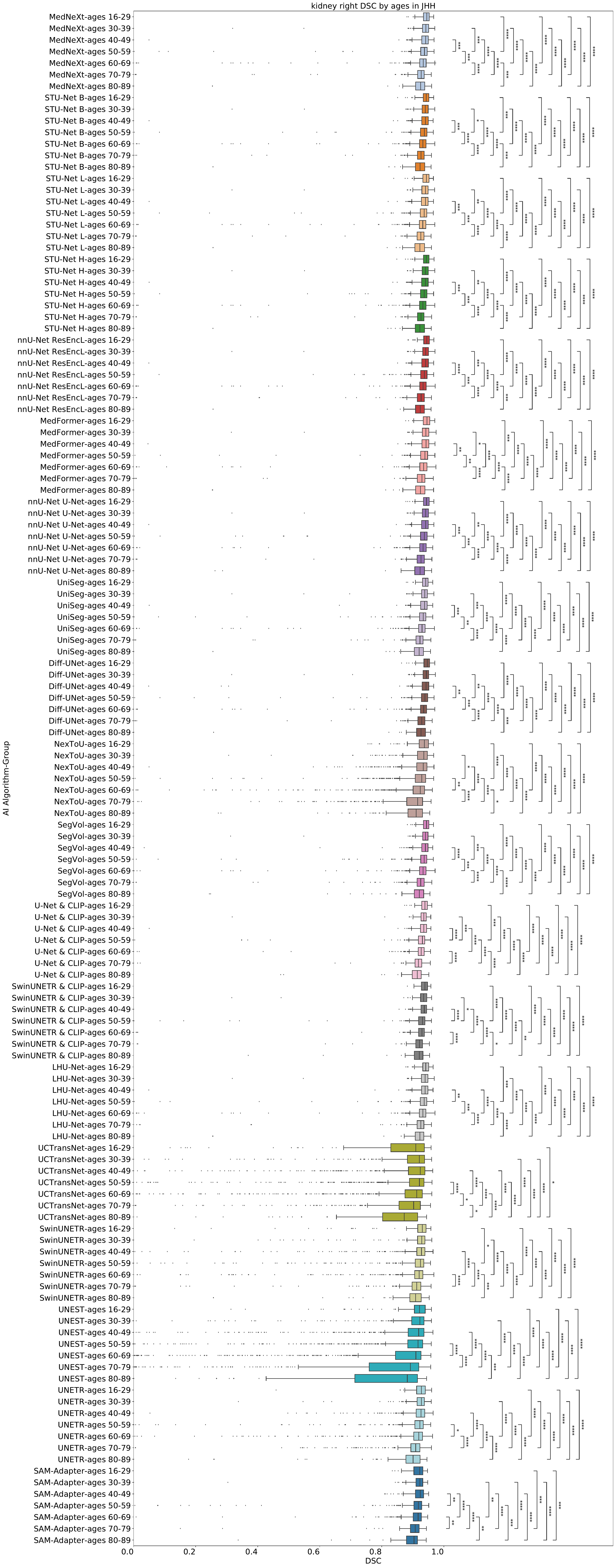}
    \caption{\textbf{Boxplot showing right kidney DSC score by age in JHH.} Statistical significance is indicated by stars: * p < 0.05, ** p <0.01, *** p < 0.001, **** p < 0.0001. We perform Kruskal–Wallis tests followed by post-hoc Mann-Whitney U Tests with Bonferroni correction. Here, we did not perform statistical comparisons between diverse AI algorithms.}
\end{figure}

\begin{figure}[!h]
	\centering
	\includegraphics[height=0.8\textheight]{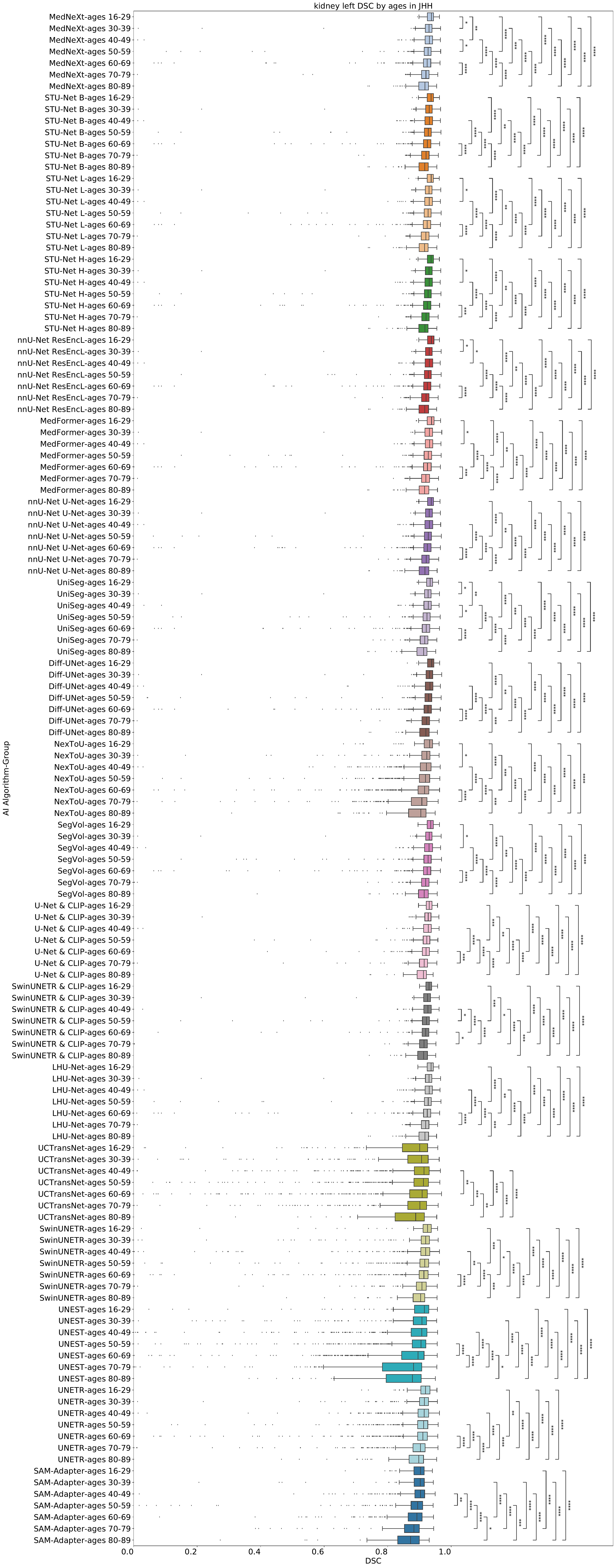}
    \caption{\textbf{Boxplot showing left kidney DSC score by age in JHH.} Statistical significance is indicated by stars: * p < 0.05, ** p <0.01, *** p < 0.001, **** p < 0.0001. We perform Kruskal–Wallis tests followed by post-hoc Mann-Whitney U Tests with Bonferroni correction. Here, we did not perform statistical comparisons between diverse AI algorithms.}
\end{figure}

\begin{figure}[!h]
	\centering
	\includegraphics[height=0.8\textheight]{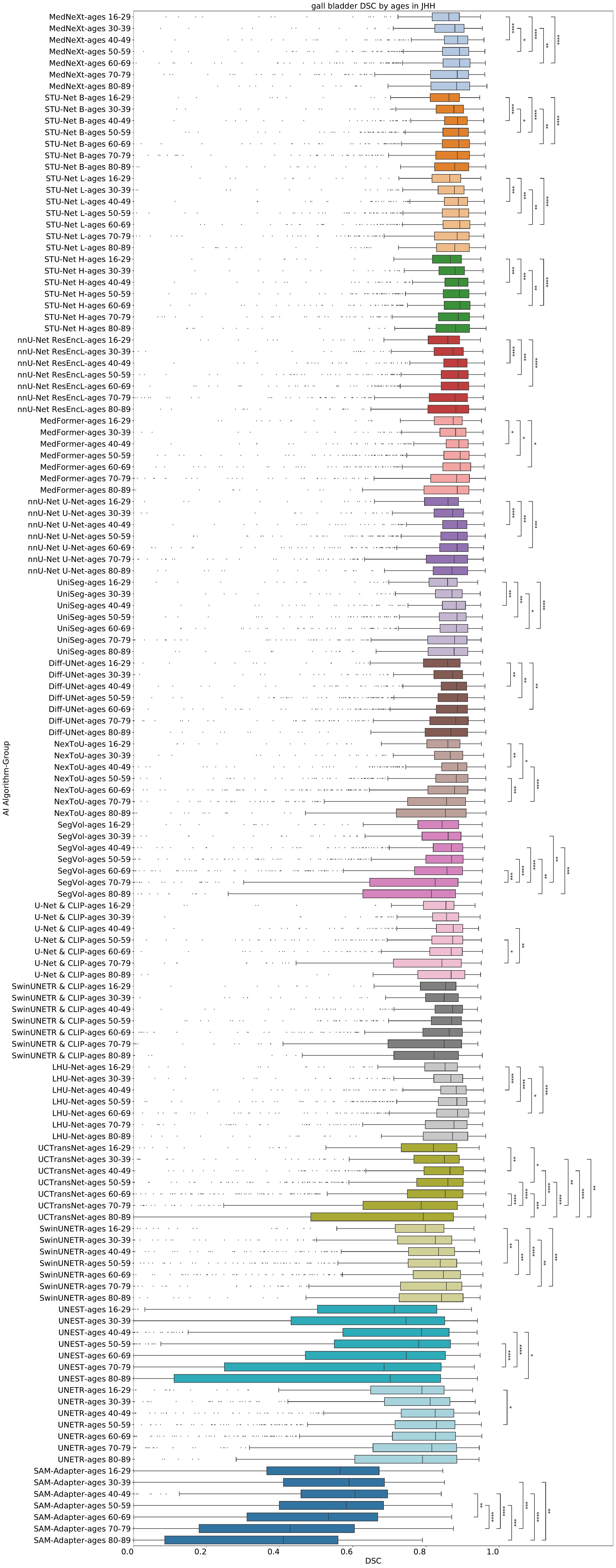}
    \caption{\textbf{Boxplot showing gall bladder DSC score by age in JHH.} Statistical significance is indicated by stars: * p < 0.05, ** p <0.01, *** p < 0.001, **** p < 0.0001. We perform Kruskal–Wallis tests followed by post-hoc Mann-Whitney U Tests with Bonferroni correction. Here, we did not perform statistical comparisons between diverse AI algorithms.}
\end{figure}

\begin{figure}[!h]
	\centering
	\includegraphics[height=0.8\textheight]{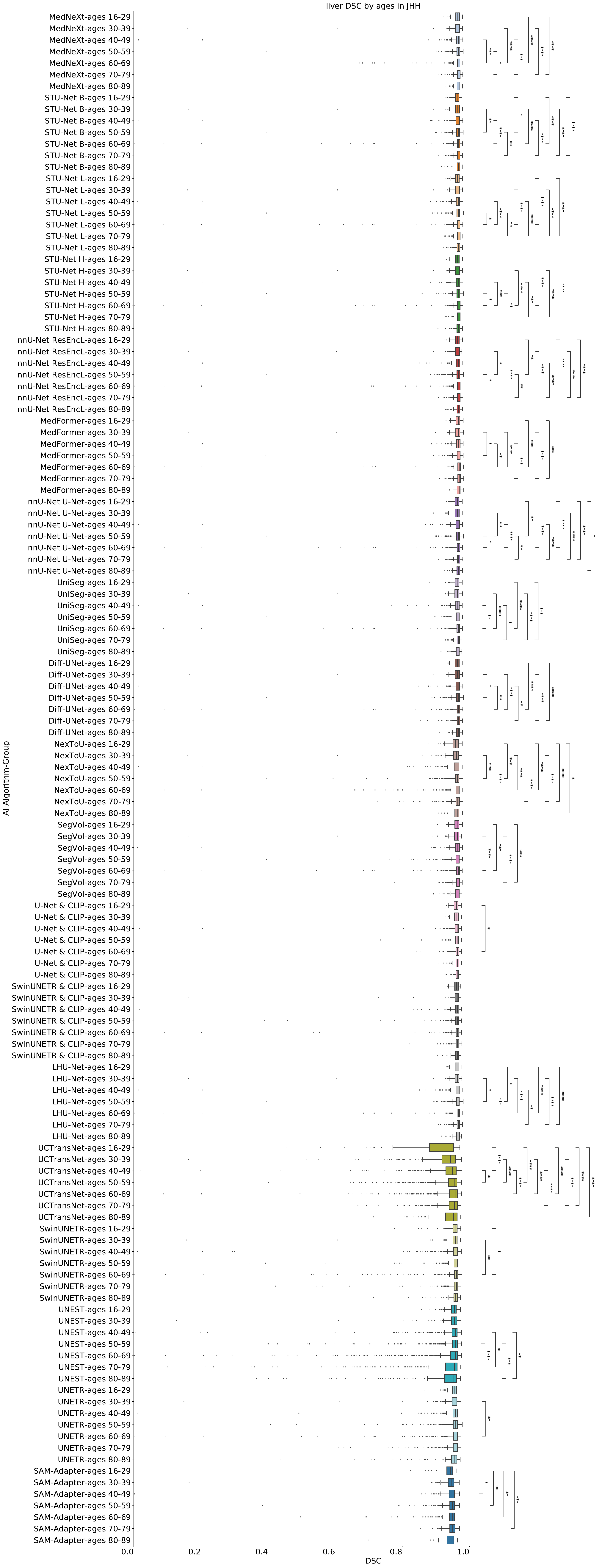}
    \caption{\textbf{Boxplot showing liver DSC score by age in JHH.} Statistical significance is indicated by stars: * p < 0.05, ** p <0.01, *** p < 0.001, **** p < 0.0001. We perform Kruskal–Wallis tests followed by post-hoc Mann-Whitney U Tests with Bonferroni correction. Here, we did not perform statistical comparisons between diverse AI algorithms.}
\end{figure}

\begin{figure}[!h]
	\centering
	\includegraphics[height=0.8\textheight]{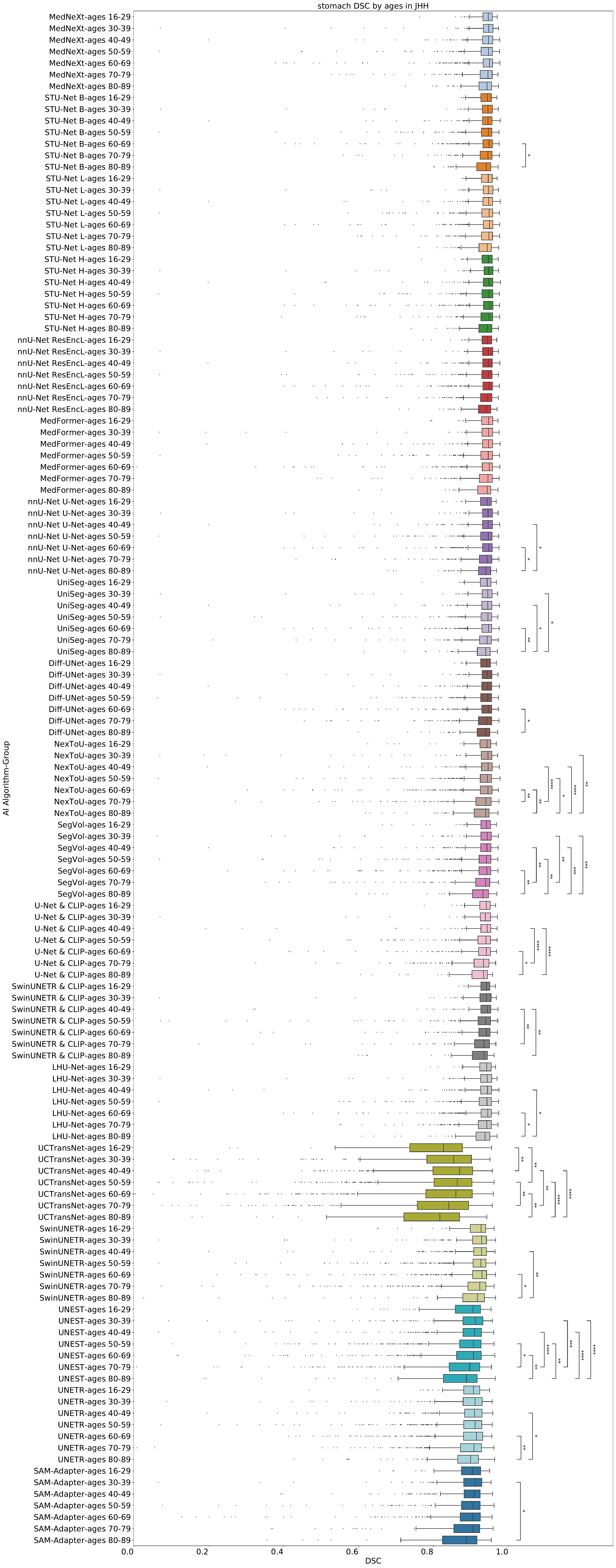}
    \caption{\textbf{Boxplot showing stomach DSC score by age in JHH.} Statistical significance is indicated by stars: * p < 0.05, ** p <0.01, *** p < 0.001, **** p < 0.0001. We perform Kruskal–Wallis tests followed by post-hoc Mann-Whitney U Tests with Bonferroni correction. Here, we did not perform statistical comparisons between diverse AI algorithms.}
\end{figure}

\begin{figure}[!h]
	\centering
	\includegraphics[height=0.8\textheight]{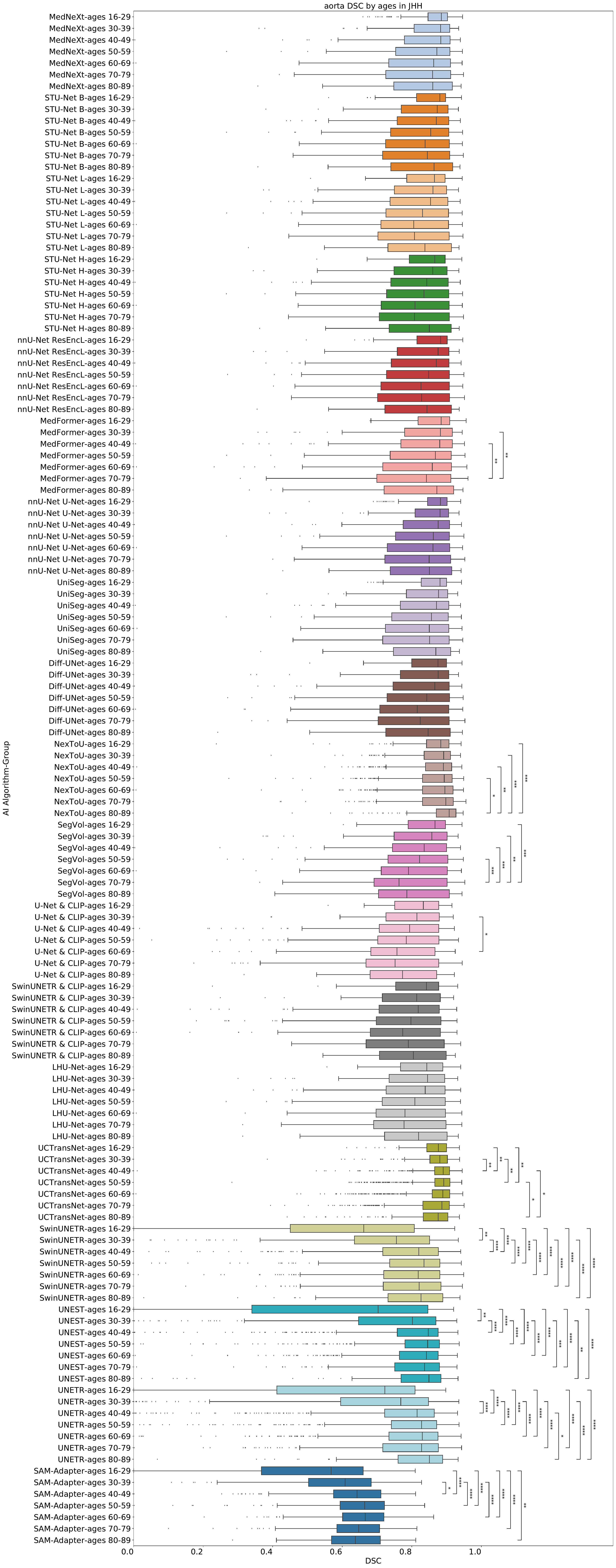}
    \caption{\textbf{Boxplot showing aorta DSC score by age in JHH.} Statistical significance is indicated by stars: * p < 0.05, ** p <0.01, *** p < 0.001, **** p < 0.0001. We observed that mean AI performance drops with advanced age, but some AI algorithm's show improving DSC score for aorta after 70. Possibly, an  explanation is that the ascending aorta and aortic arch can increase in diameter with age (due to hypertension), and the walls of the vessel will gradually show obvious calcification, possibly making the boundaries clearer. We perform Kruskal–Wallis tests followed by post-hoc Mann-Whitney U Tests with Bonferroni correction. Here, we did not perform statistical comparisons between diverse AI algorithms.}
\end{figure}

\begin{figure}[!h]
	\centering
	\includegraphics[height=0.8\textheight]{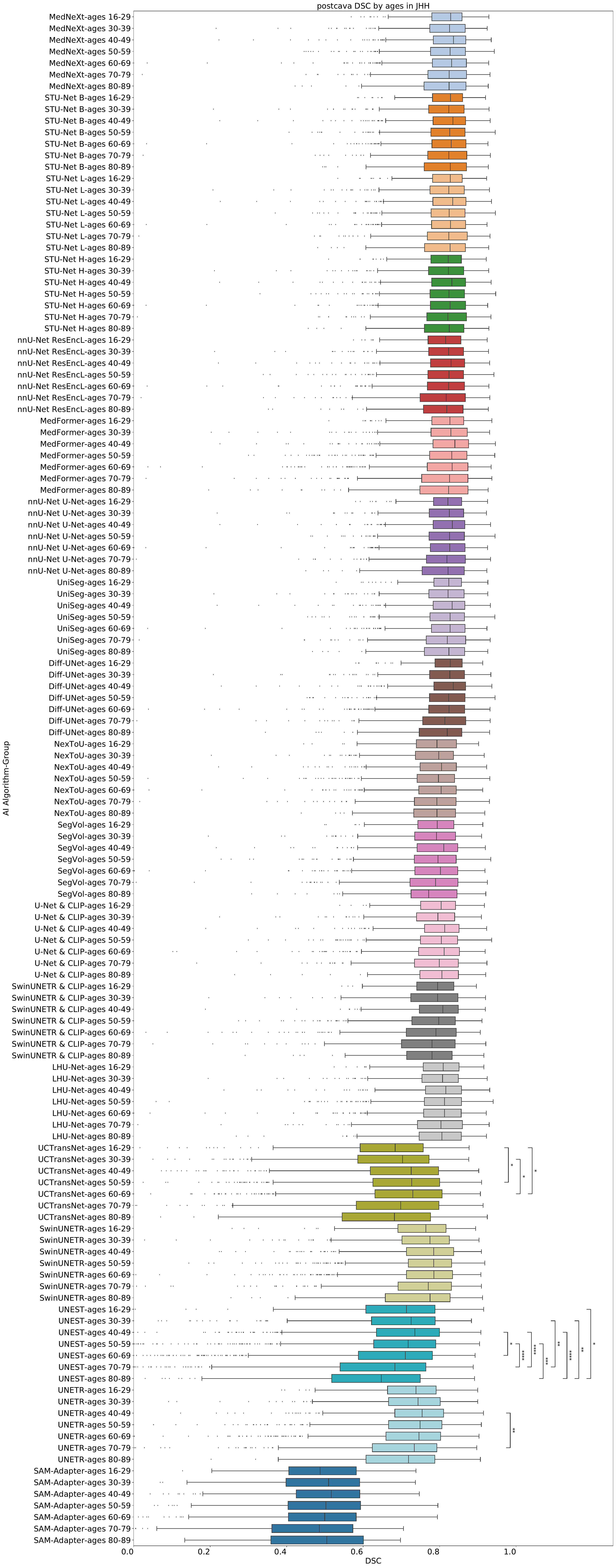}
    \caption{\textbf{Boxplot showing postcava DSC score by age in JHH.} Statistical significance is indicated by stars: * p < 0.05, ** p <0.01, *** p < 0.001, **** p < 0.0001. We perform Kruskal–Wallis tests followed by post-hoc Mann-Whitney U Tests with Bonferroni correction. Here, we did not perform statistical comparisons between diverse AI algorithms.}
\end{figure}

\begin{figure}[!h]
	\centering
	\includegraphics[height=0.8\textheight]{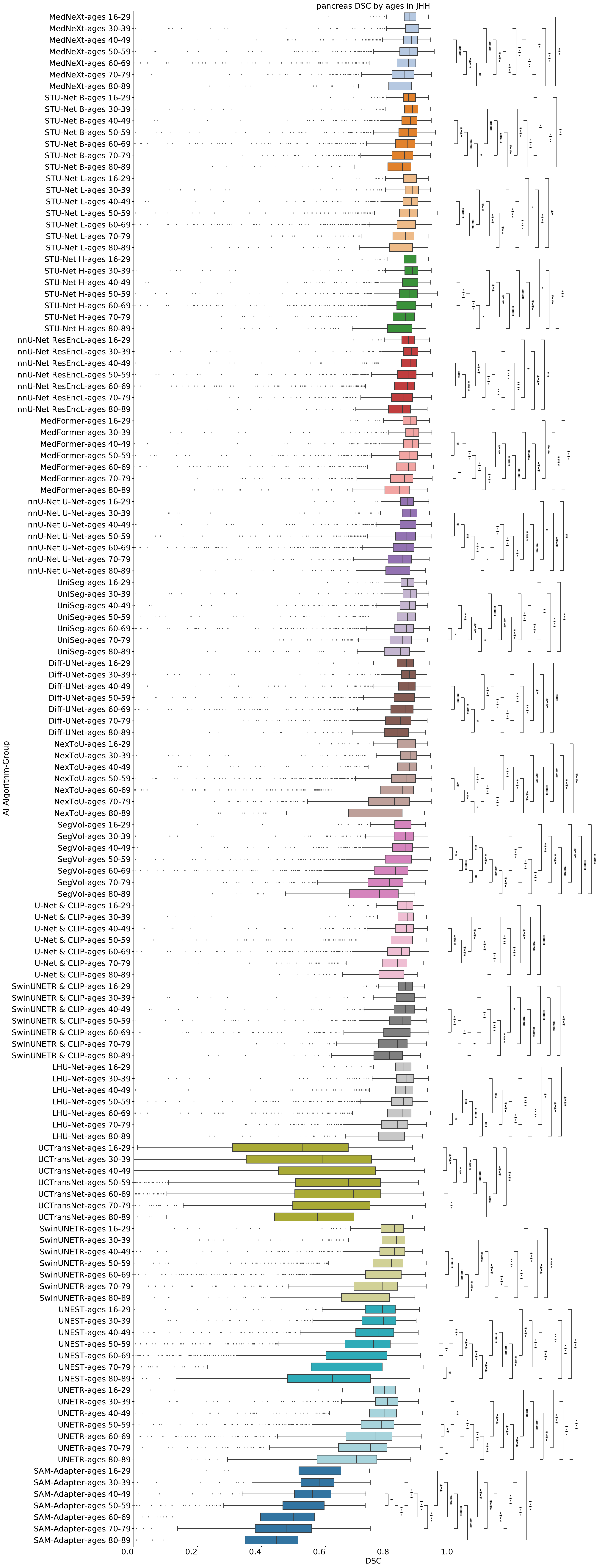}
    \caption{\textbf{Boxplot showing pancreas DSC score by age in JHH.} Statistical significance is indicated by stars: * p < 0.05, ** p <0.01, *** p < 0.001, **** p < 0.0001. We perform Kruskal–Wallis tests followed by post-hoc Mann-Whitney U Tests with Bonferroni correction. Here, we did not perform statistical comparisons between diverse AI algorithms.}
\end{figure}

\clearpage
\subsubsection{Diagnosis: per-class analysis}

\begin{figure}[h]
	\centering
	\includegraphics[width=0.9\columnwidth]{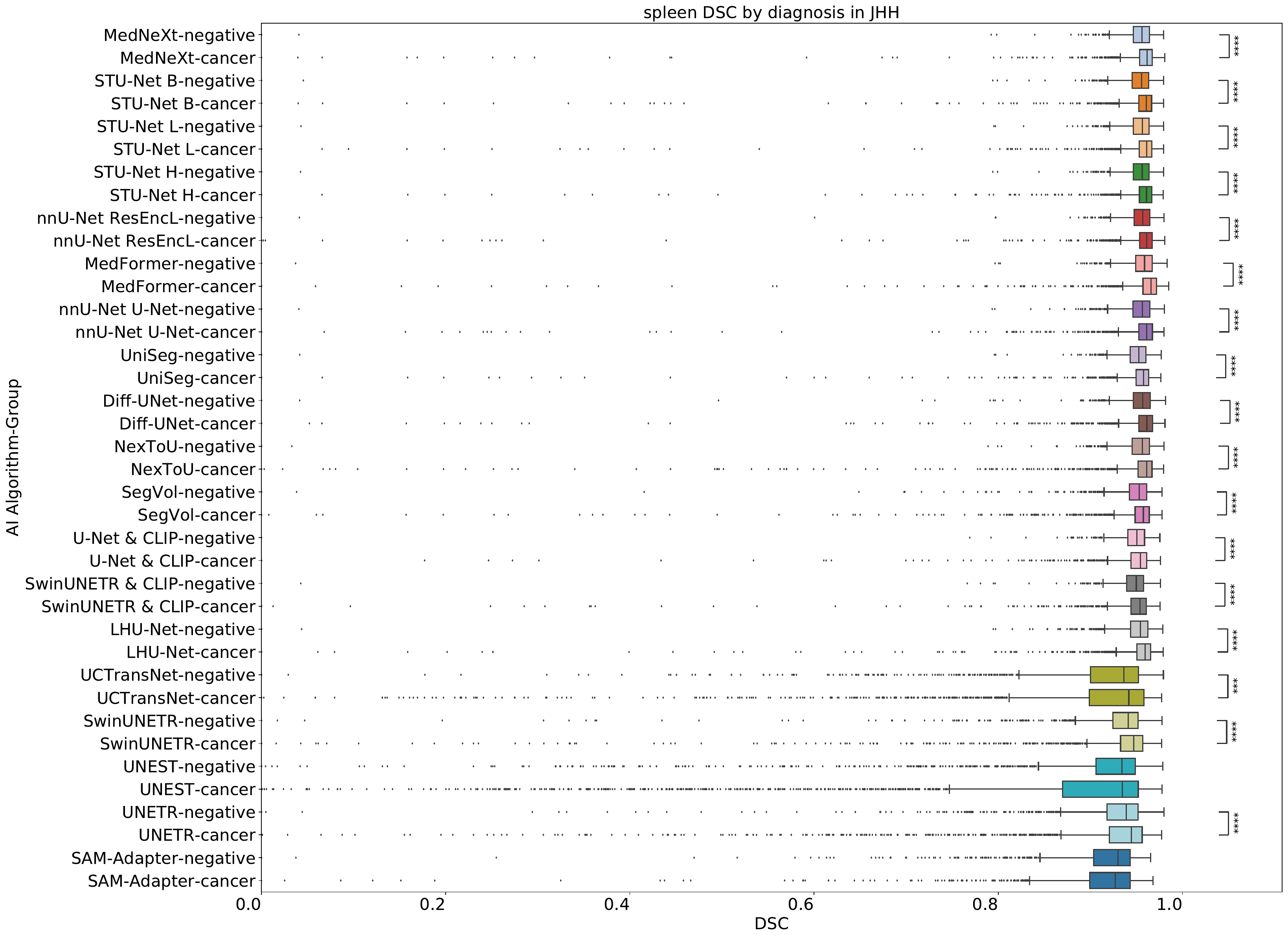}
    \caption{\textbf{Boxplot showing spleen DSC score by diagnosis in JHH.} Statistical significance is indicated by stars: * p < 0.05, ** p <0.01, *** p < 0.001, **** p < 0.0001. We perform Kruskal–Wallis tests followed by post-hoc Mann-Whitney U Tests with Bonferroni correction. Here, we did not perform statistical comparisons between diverse AI algorithms.}
\end{figure}

\begin{figure}[h]
	\centering
	\includegraphics[width=0.9\columnwidth]{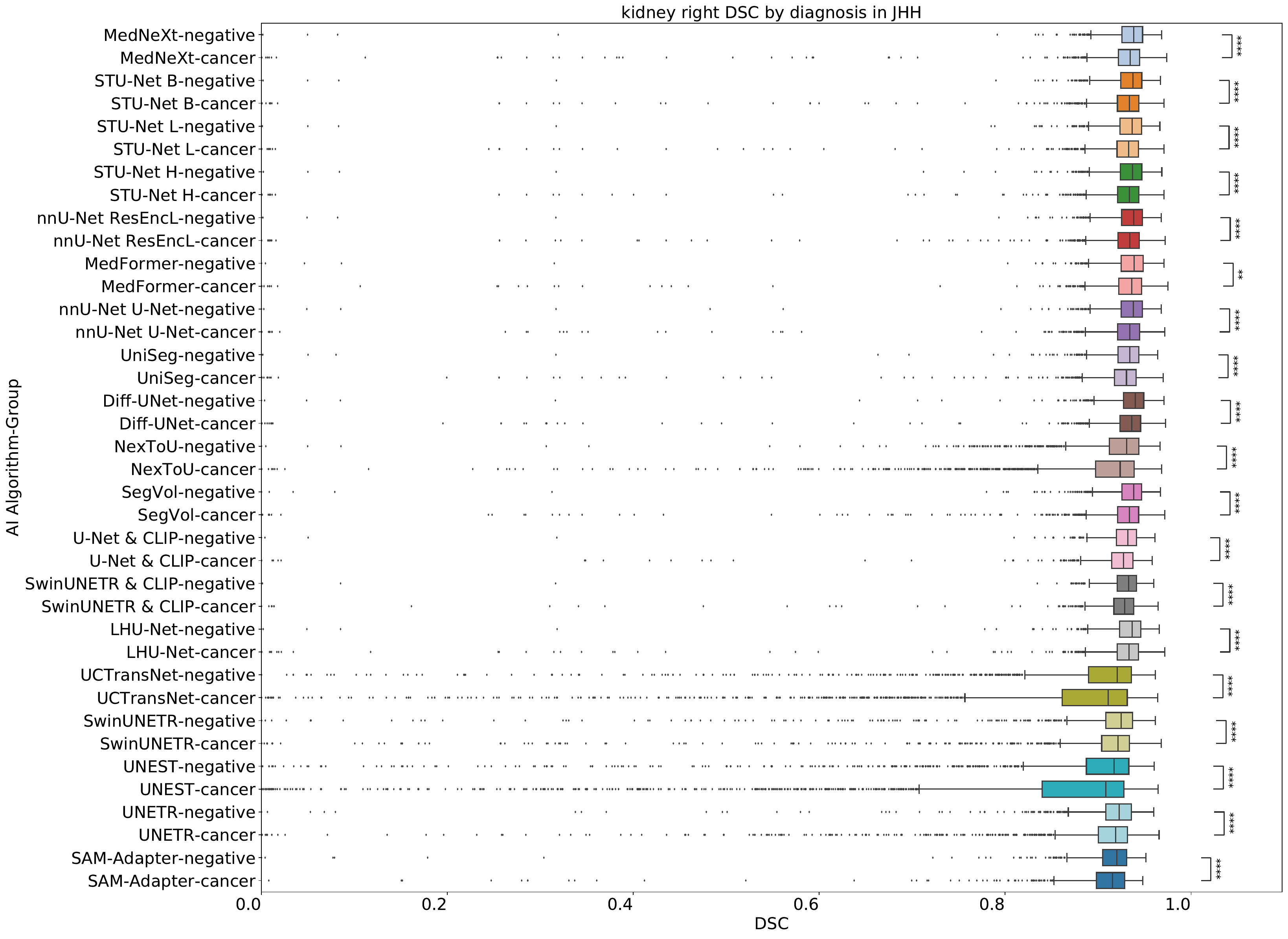}
    \caption{\textbf{Boxplot showing right kidney DSC score by diagnosis in JHH.} Statistical significance is indicated by stars: * p < 0.05, ** p <0.01, *** p < 0.001, **** p < 0.0001. We perform Kruskal–Wallis tests followed by post-hoc Mann-Whitney U Tests with Bonferroni correction. Here, we did not perform statistical comparisons between diverse AI algorithms.}
\end{figure}

\begin{figure}[h]
	\centering
	\includegraphics[width=0.9\columnwidth]{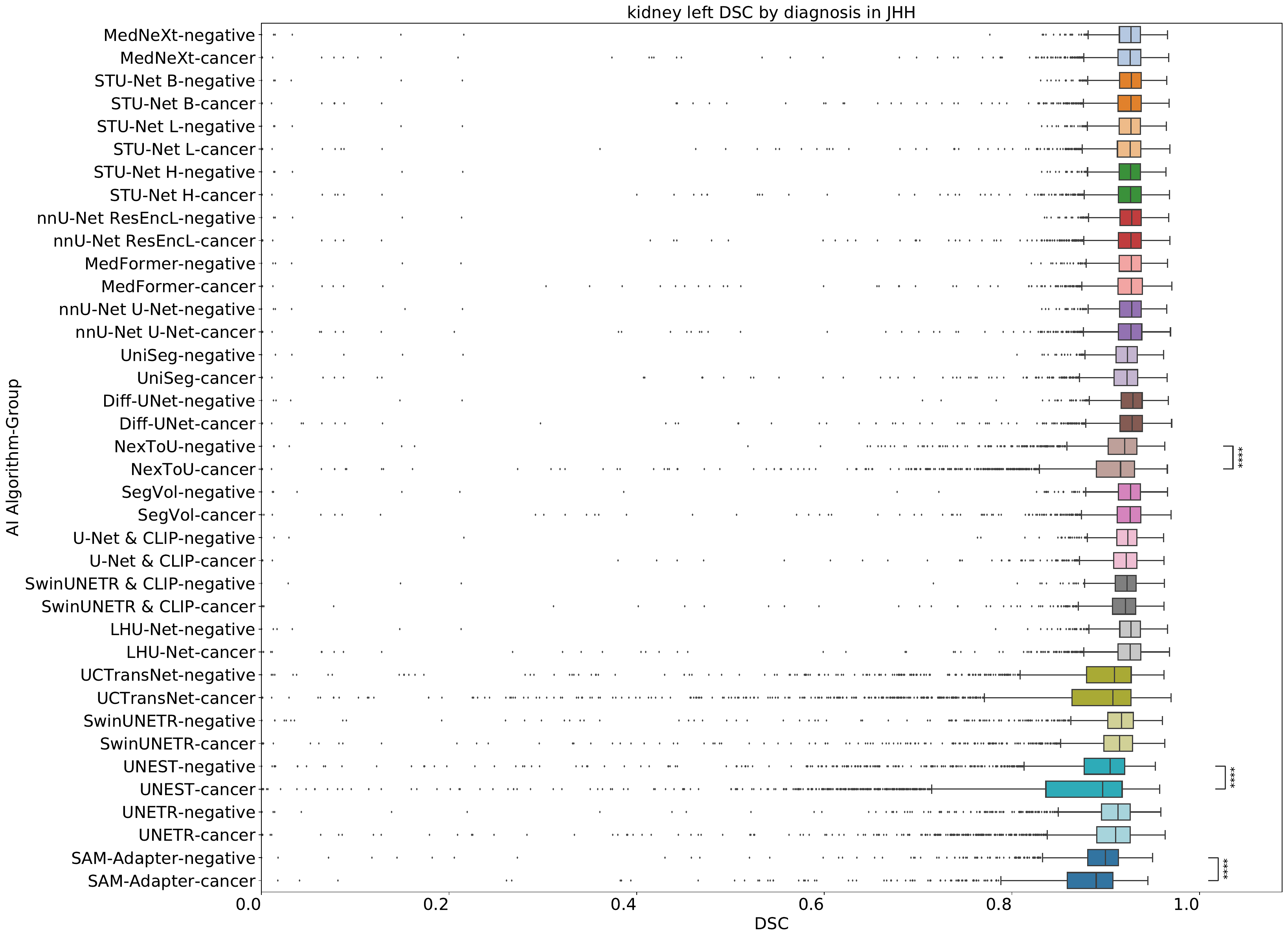}
    \caption{\textbf{Boxplot showing left kidney DSC score by diagnosis in JHH.} Statistical significance is indicated by stars: * p < 0.05, ** p <0.01, *** p < 0.001, **** p < 0.0001. We perform Kruskal–Wallis tests followed by post-hoc Mann-Whitney U Tests with Bonferroni correction. Here, we did not perform statistical comparisons between diverse AI algorithms.}
\end{figure}

\begin{figure}[h]
	\centering
	\includegraphics[width=0.9\columnwidth]{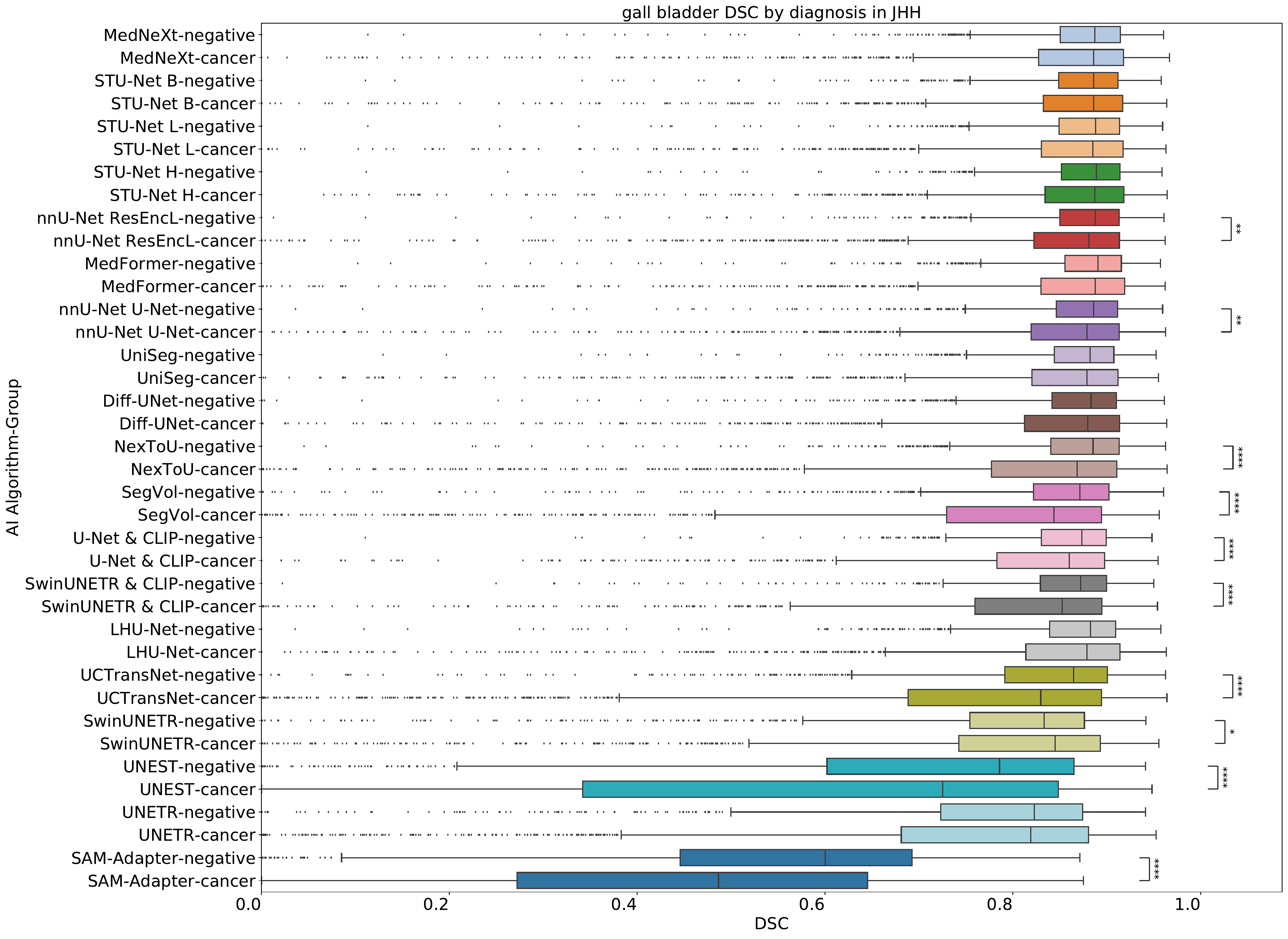}
    \caption{\textbf{Boxplot showing gallbladder DSC score by diagnosis in JHH.} Statistical significance is indicated by stars: * p < 0.05, ** p <0.01, *** p < 0.001, **** p < 0.0001. We perform Kruskal–Wallis tests followed by post-hoc Mann-Whitney U Tests with Bonferroni correction. Here, we did not perform statistical comparisons between diverse AI algorithms.}
\end{figure}

\begin{figure}[h]
	\centering
	\includegraphics[width=0.9\columnwidth]{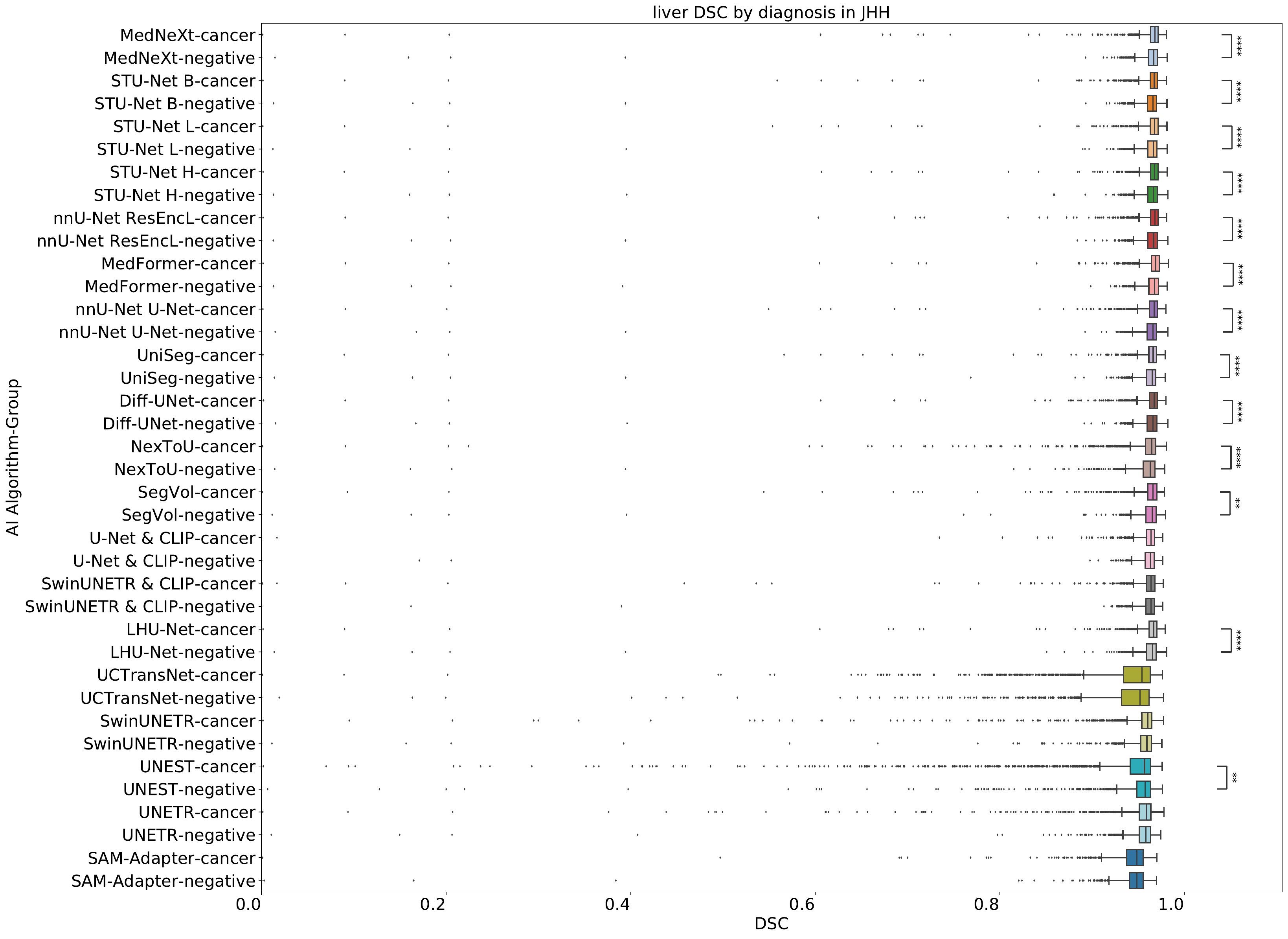}
    \caption{\textbf{Boxplot showing liver DSC score by diagnosis in JHH.} Statistical significance is indicated by stars: * p < 0.05, ** p <0.01, *** p < 0.001, **** p < 0.0001. We perform Kruskal–Wallis tests followed by post-hoc Mann-Whitney U Tests with Bonferroni correction. Here, we did not perform statistical comparisons between diverse AI algorithms.}
\end{figure}

\begin{figure}[h]
	\centering
	\includegraphics[width=0.9\columnwidth]{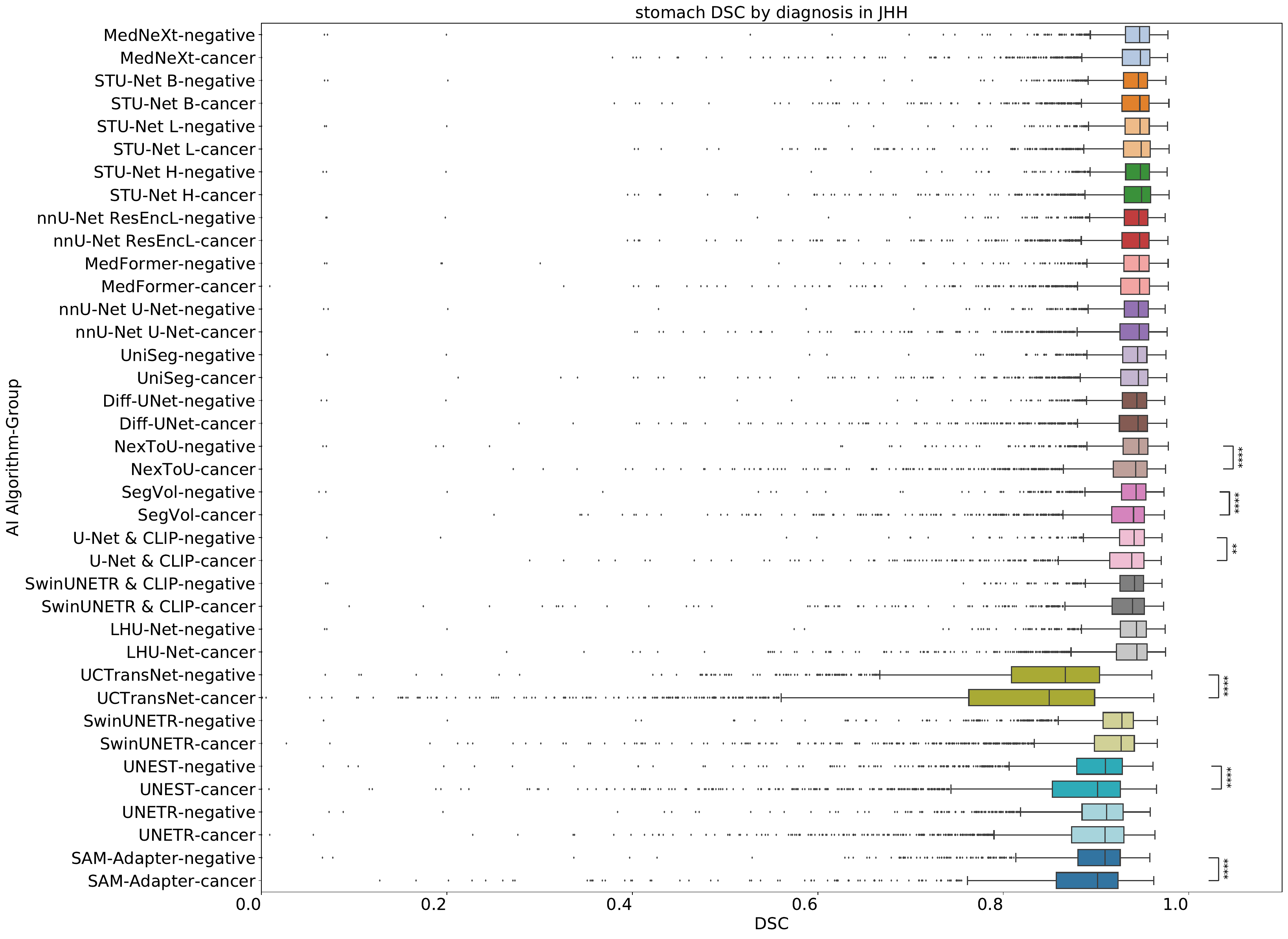}
    \caption{\textbf{Boxplot showing stomach DSC score by diagnosis in JHH.} Statistical significance is indicated by stars: * p < 0.05, ** p <0.01, *** p < 0.001, **** p < 0.0001. We perform Kruskal–Wallis tests followed by post-hoc Mann-Whitney U Tests with Bonferroni correction. Here, we did not perform statistical comparisons between diverse AI algorithms.}
\end{figure}

\begin{figure}[h]
	\centering
	\includegraphics[width=0.9\columnwidth]{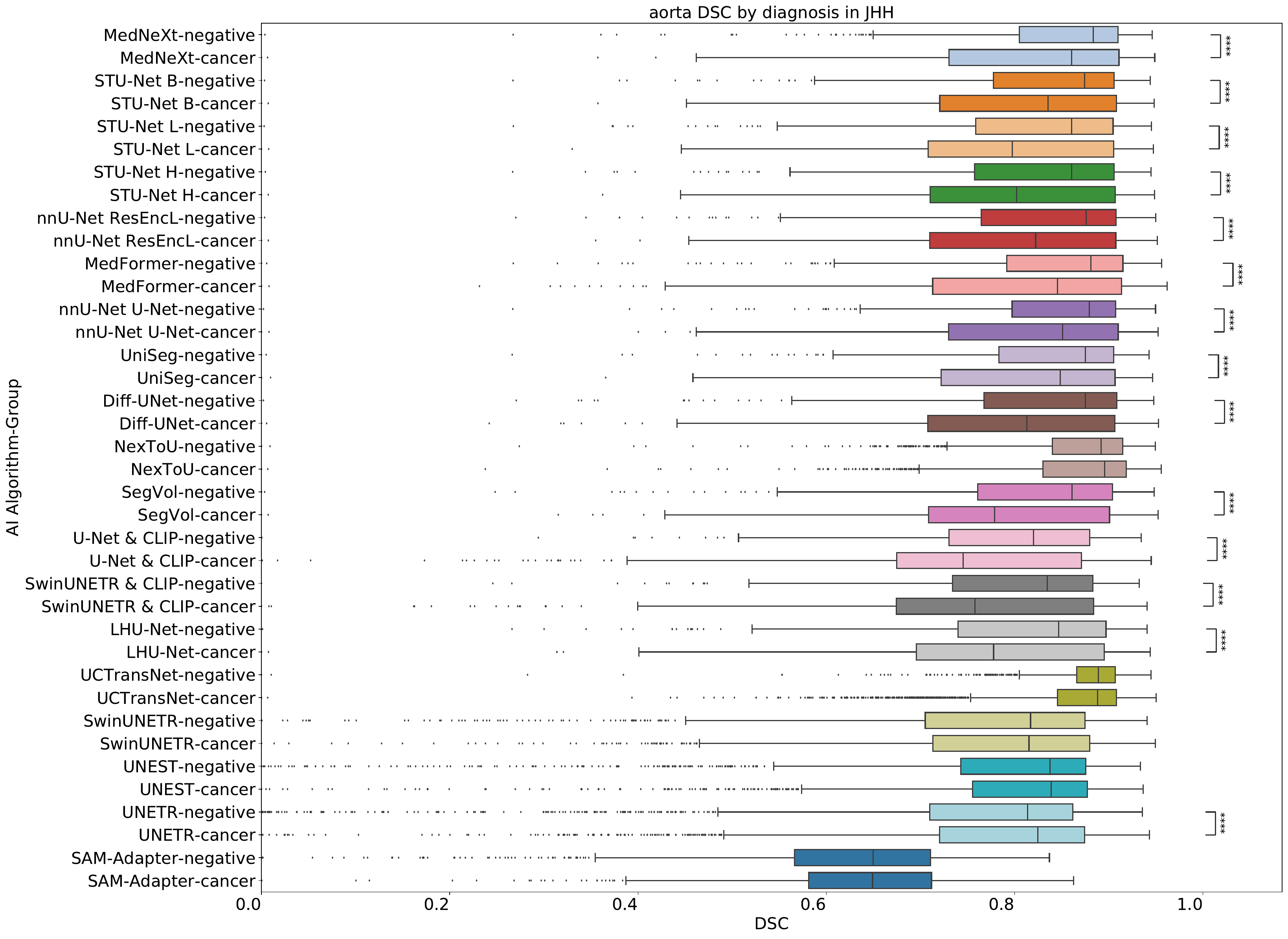}
    \caption{\textbf{Boxplot showing aorta DSC score by diagnosis in JHH.} Statistical significance is indicated by stars: * p < 0.05, ** p <0.01, *** p < 0.001, **** p < 0.0001. We perform Kruskal–Wallis tests followed by post-hoc Mann-Whitney U Tests with Bonferroni correction. Here, we did not perform statistical comparisons between diverse AI algorithms.}
\end{figure}

\begin{figure}[h]
	\centering
	\includegraphics[width=0.9\columnwidth]{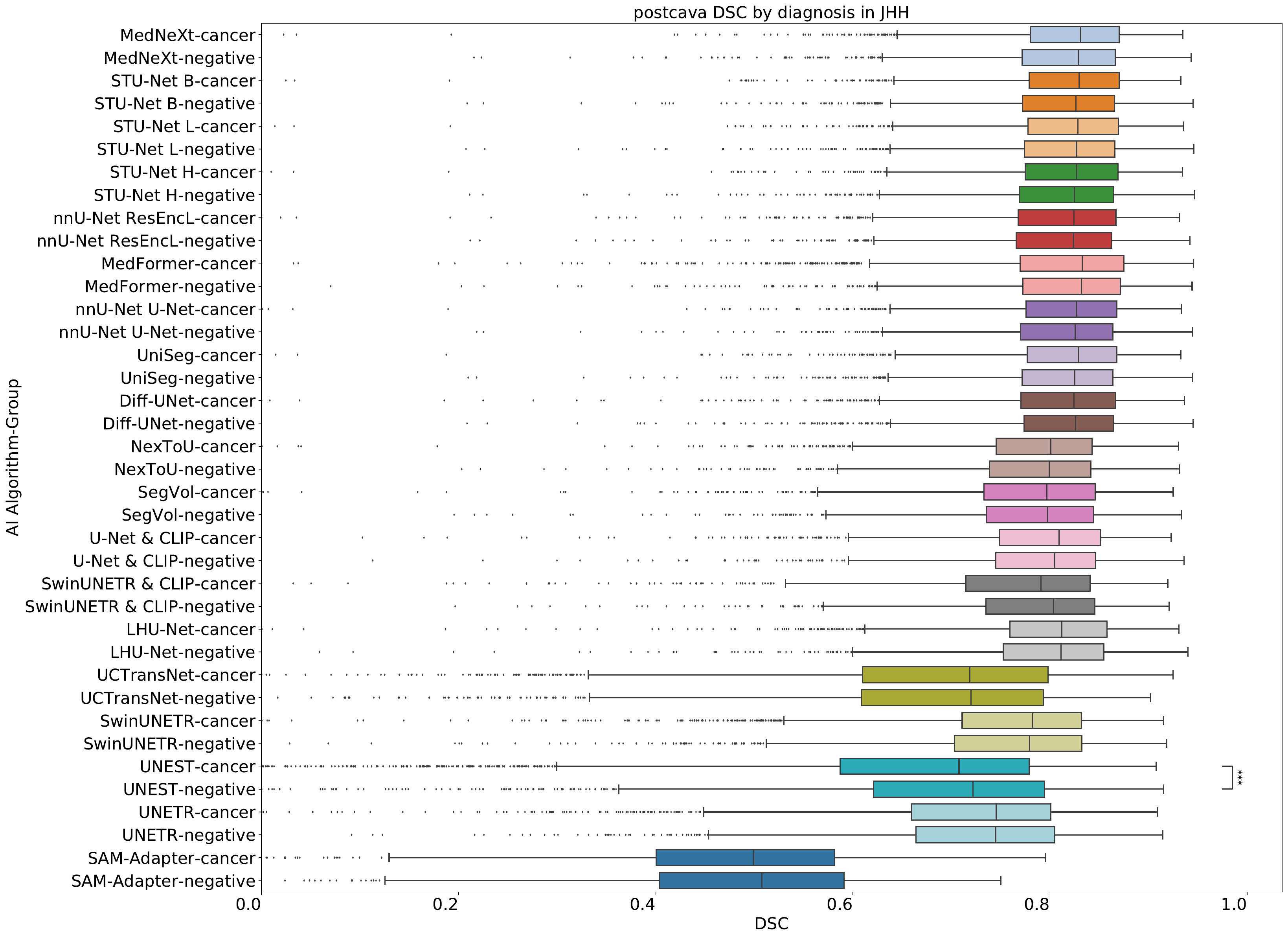}
    \caption{\textbf{Boxplot showing postcava DSC score by diagnosis in JHH.} Statistical significance is indicated by stars: * p < 0.05, ** p <0.01, *** p < 0.001, **** p < 0.0001. We perform Kruskal–Wallis tests followed by post-hoc Mann-Whitney U Tests with Bonferroni correction. Here, we did not perform statistical comparisons between diverse AI algorithms.}
\end{figure}

\begin{figure}[h]
	\centering
	\includegraphics[width=0.9\columnwidth]{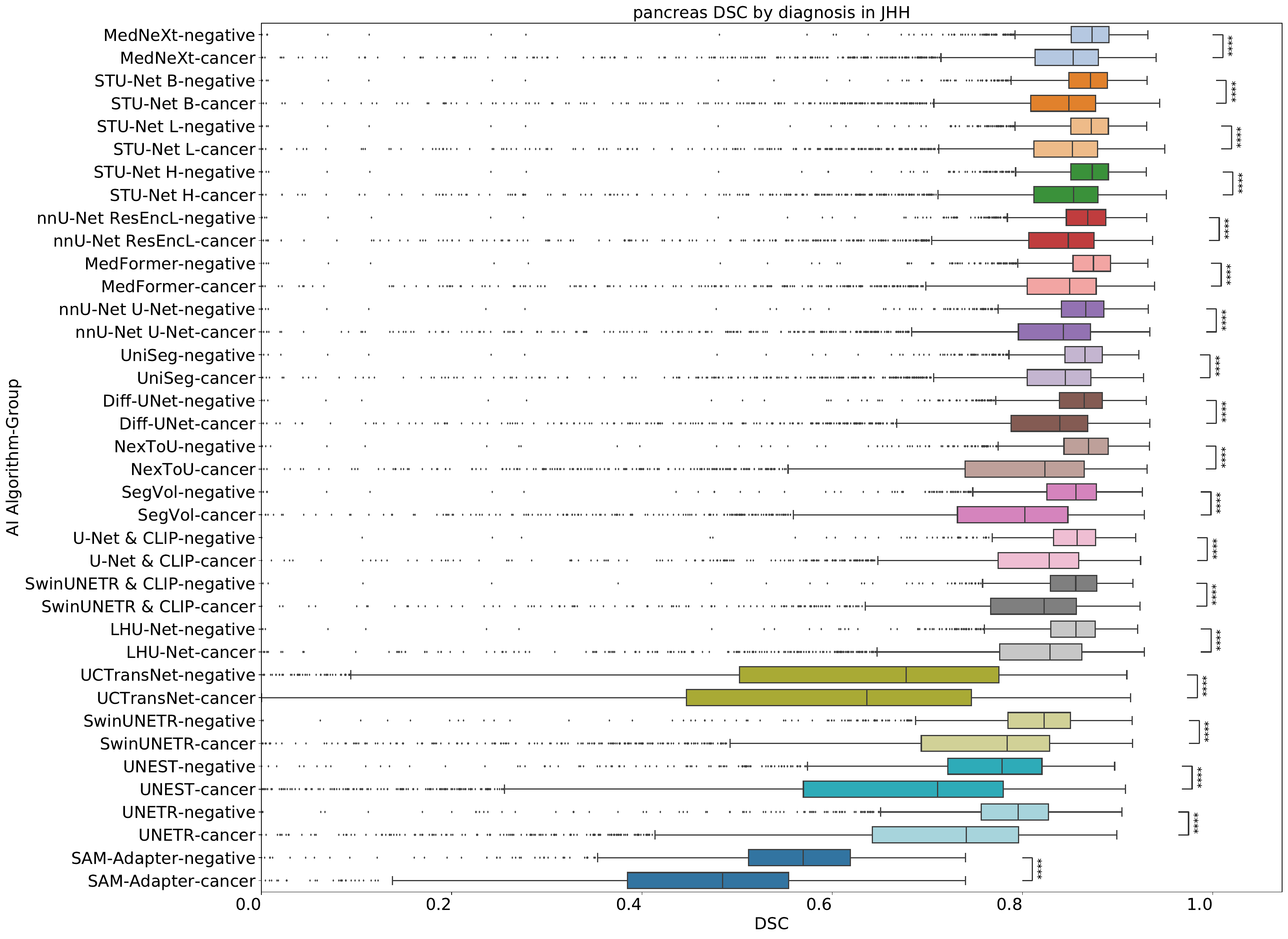}
    \caption{\textbf{Boxplot showing pancreas DSC score by diagnosis in JHH.} Statistical significance is indicated by stars: * p < 0.05, ** p <0.01, *** p < 0.001, **** p < 0.0001. We perform Kruskal–Wallis tests followed by post-hoc Mann-Whitney U Tests with Bonferroni correction. Here, we did not perform statistical comparisons between diverse AI algorithms.}
\end{figure}

\clearpage
\subsubsection{Sex: per-class analysis}

\begin{figure}[h]
	\centering
	\includegraphics[width=0.9\columnwidth]{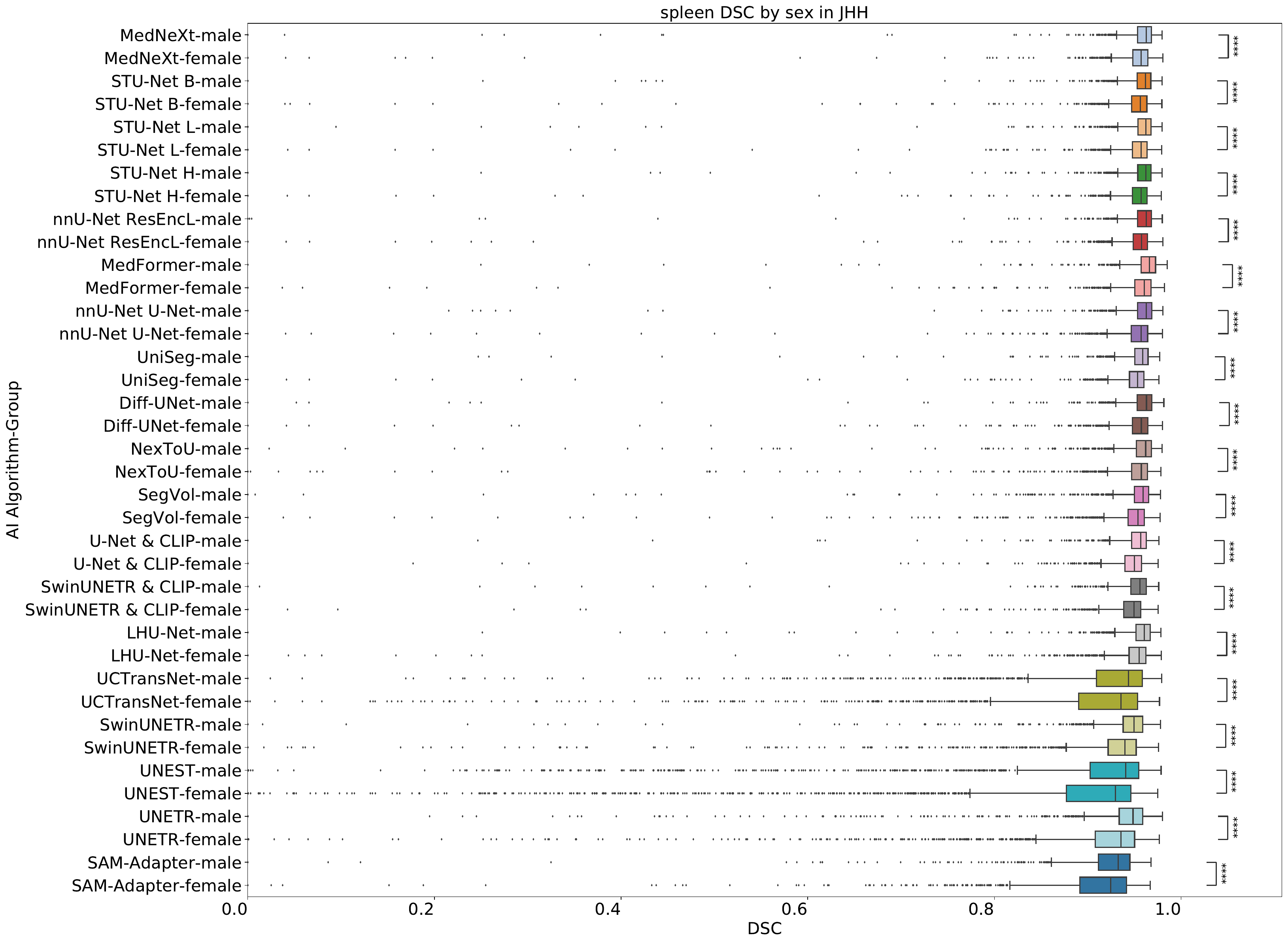}
    \caption{\textbf{Boxplot showing spleen DSC score by sex in JHH.} Statistical significance is indicated by stars: * p < 0.05, ** p <0.01, *** p < 0.001, **** p < 0.0001. We perform Kruskal–Wallis tests followed by post-hoc Mann-Whitney U Tests with Bonferroni correction. Here, we did not perform statistical comparisons between diverse AI algorithms.}
\end{figure}

\begin{figure}[h]
	\centering
	\includegraphics[width=0.9\columnwidth]{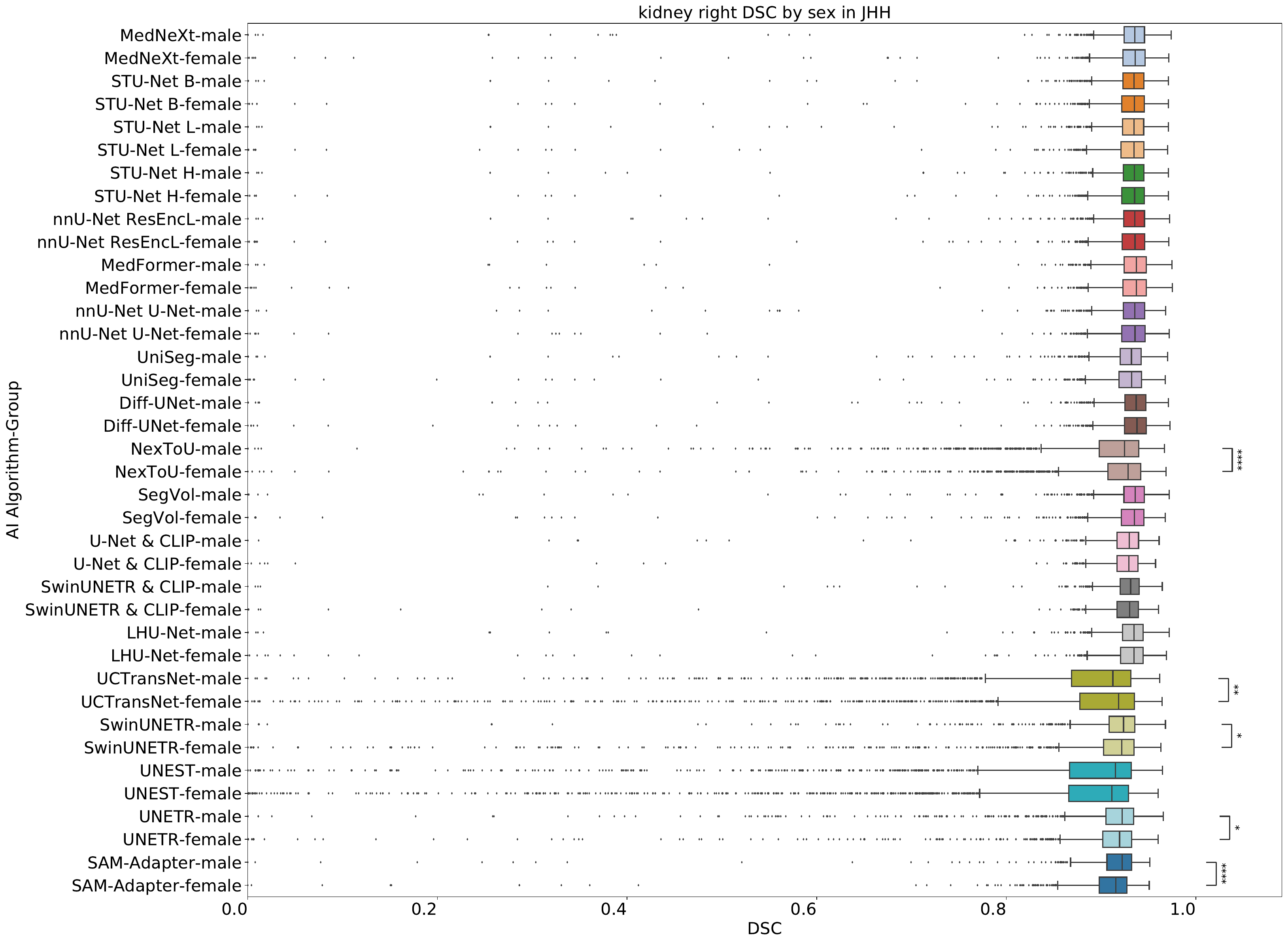}
    \caption{\textbf{Boxplot showing right kidney DSC score by sex in JHH.} Statistical significance is indicated by stars: * p < 0.05, ** p <0.01, *** p < 0.001, **** p < 0.0001. We perform Kruskal–Wallis tests followed by post-hoc Mann-Whitney U Tests with Bonferroni correction. Here, we did not perform statistical comparisons between diverse AI algorithms.}
\end{figure}

\begin{figure}[h]
	\centering
	\includegraphics[width=0.9\columnwidth]{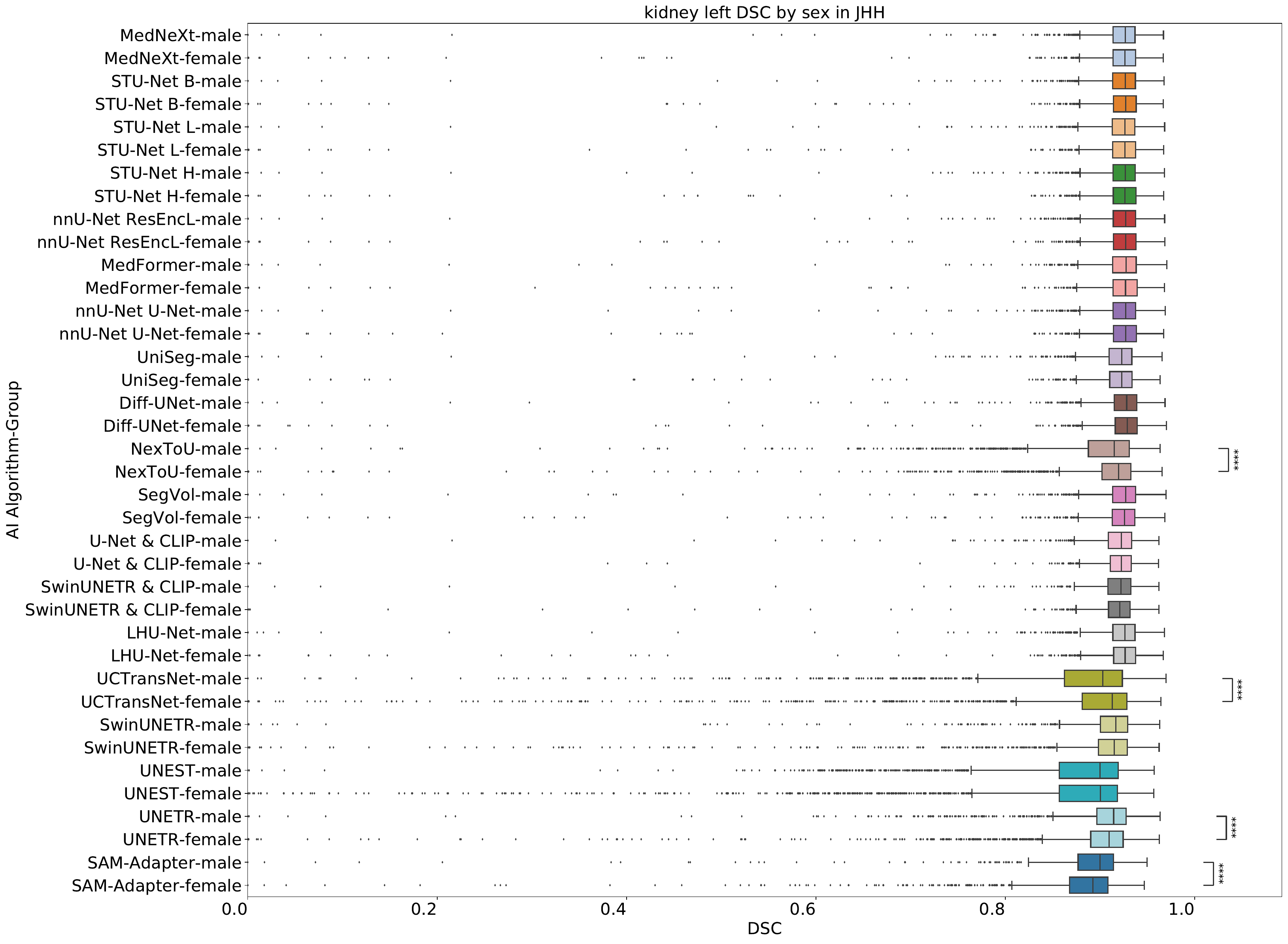}
    \caption{\textbf{Boxplot showing left kidney DSC score by sex in JHH.} Statistical significance is indicated by stars: * p < 0.05, ** p <0.01, *** p < 0.001, **** p < 0.0001. We perform Kruskal–Wallis tests followed by post-hoc Mann-Whitney U Tests with Bonferroni correction. Here, we did not perform statistical comparisons between diverse AI algorithms.}
\end{figure}

\begin{figure}[h]
	\centering
	\includegraphics[width=0.9\columnwidth]{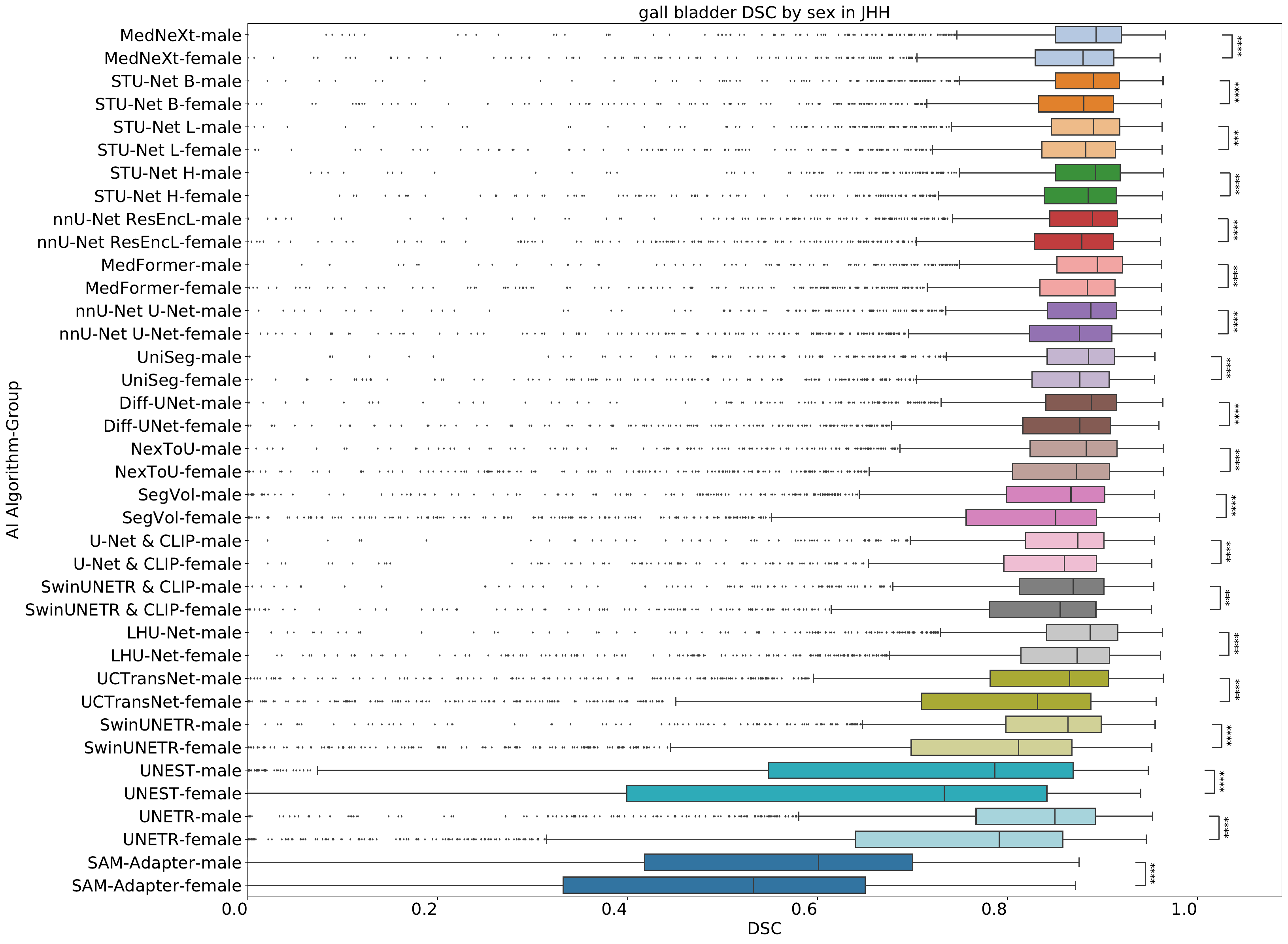}
    \caption{\textbf{Boxplot showing gallbladder DSC score by sex in JHH.} Statistical significance is indicated by stars: * p < 0.05, ** p <0.01, *** p < 0.001, **** p < 0.0001. We perform Kruskal–Wallis tests followed by post-hoc Mann-Whitney U Tests with Bonferroni correction. Here, we did not perform statistical comparisons between diverse AI algorithms.}
\end{figure}

\begin{figure}[h]
	\centering
	\includegraphics[width=0.9\columnwidth]{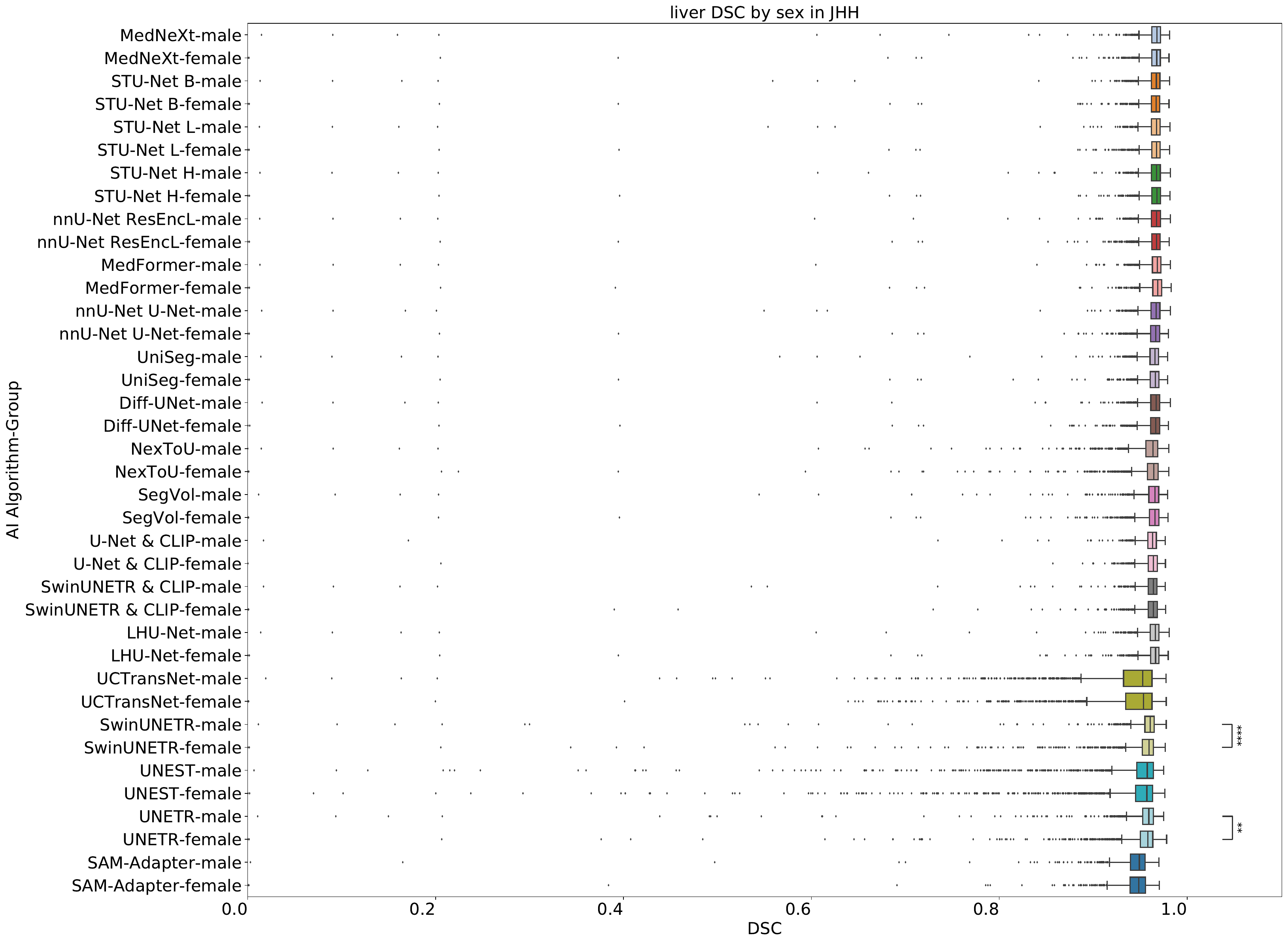}
    \caption{\textbf{Boxplot showing liver DSC score by sex in JHH.} Statistical significance is indicated by stars: * p < 0.05, ** p <0.01, *** p < 0.001, **** p < 0.0001. We perform Kruskal–Wallis tests followed by post-hoc Mann-Whitney U Tests with Bonferroni correction. Here, we did not perform statistical comparisons between diverse AI algorithms.}
\end{figure}

\begin{figure}[h]
	\centering
	\includegraphics[width=0.9\columnwidth]{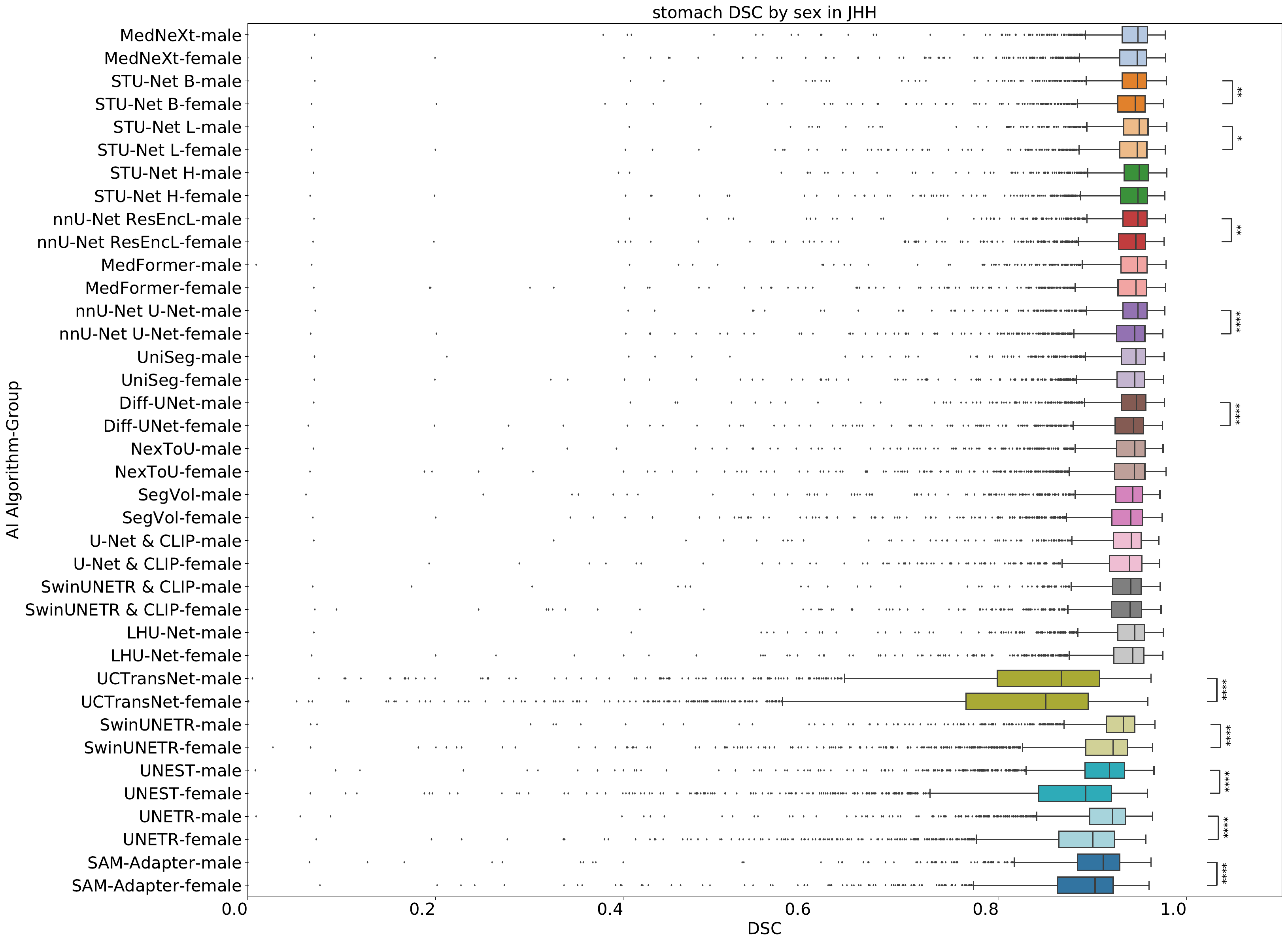}
    \caption{\textbf{Boxplot showing stomach DSC score by sex in JHH.} Statistical significance is indicated by stars: * p < 0.05, ** p <0.01, *** p < 0.001, **** p < 0.0001. We perform Kruskal–Wallis tests followed by post-hoc Mann-Whitney U Tests with Bonferroni correction. Here, we did not perform statistical comparisons between diverse AI algorithms.}
\end{figure}

\begin{figure}[h]
	\centering
	\includegraphics[width=0.9\columnwidth]{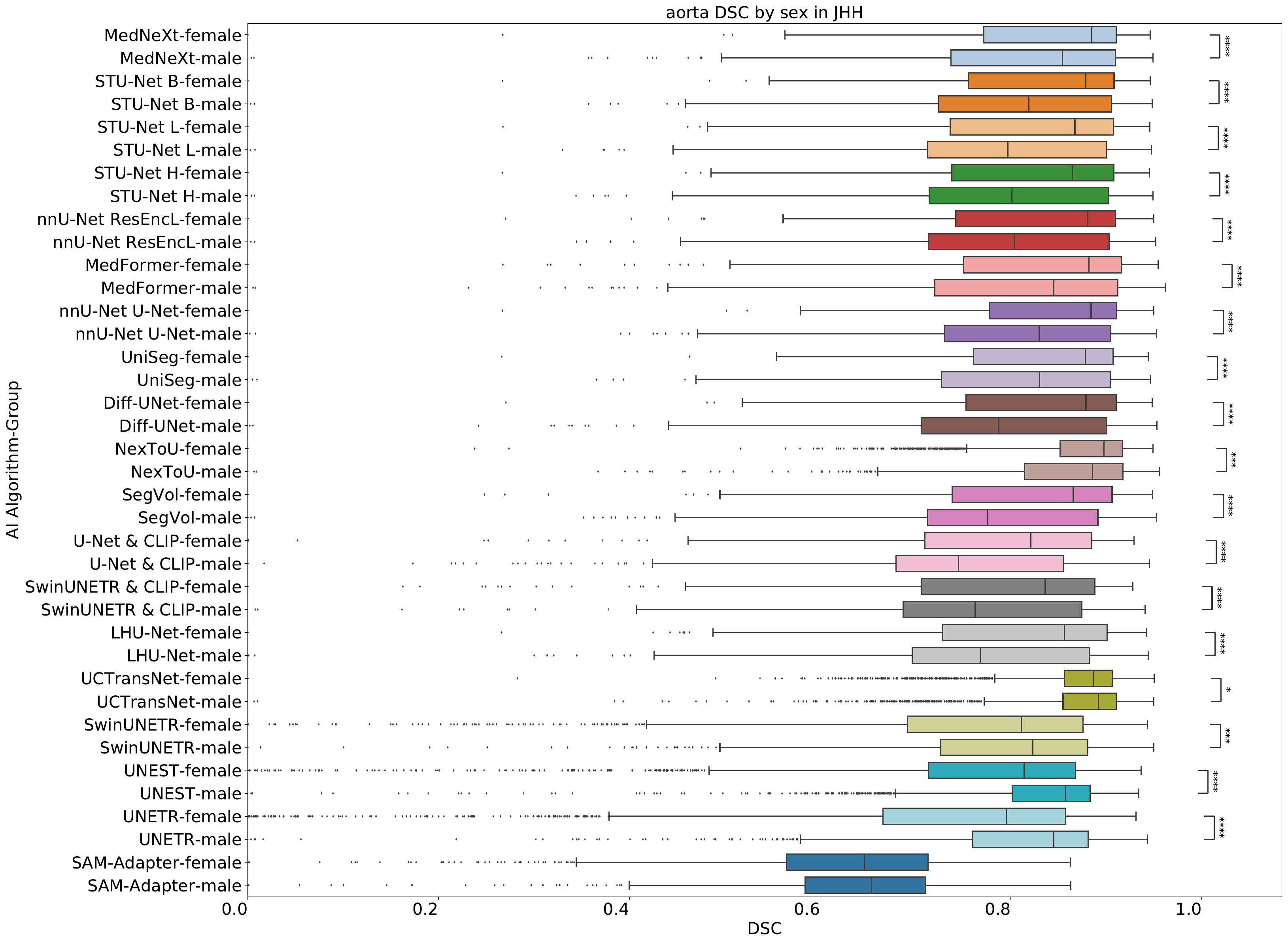}
    \caption{\textbf{Boxplot showing aorta DSC score by sex in JHH.} Statistical significance is indicated by stars: * p < 0.05, ** p <0.01, *** p < 0.001, **** p < 0.0001. We perform Kruskal–Wallis tests followed by post-hoc Mann-Whitney U Tests with Bonferroni correction. Here, we did not perform statistical comparisons between diverse AI algorithms.}
\end{figure}

\begin{figure}[h]
	\centering
	\includegraphics[width=0.9\columnwidth]{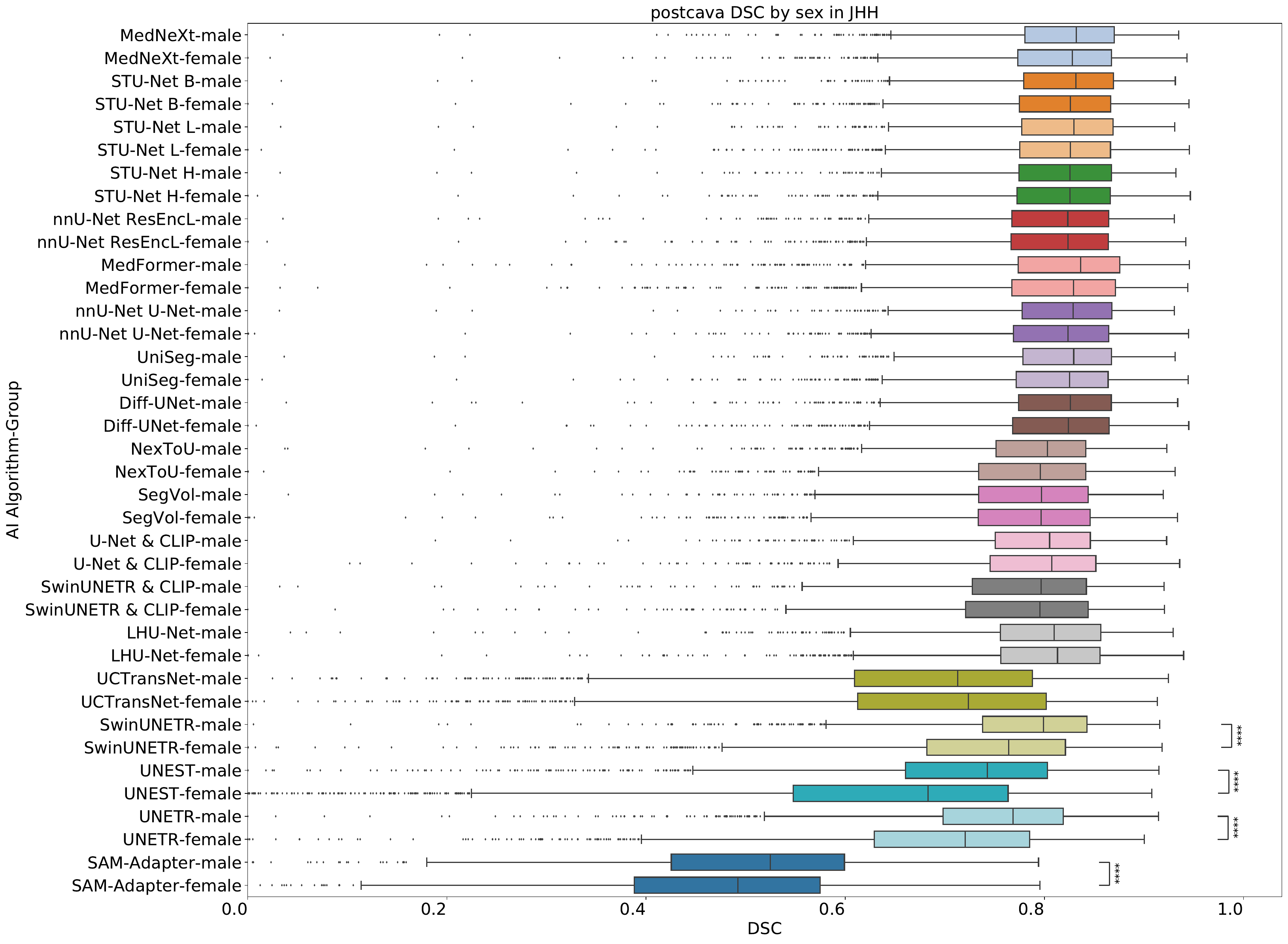}
    \caption{\textbf{Boxplot showing postcava DSC score by sex in JHH.} Statistical significance is indicated by stars: * p < 0.05, ** p <0.01, *** p < 0.001, **** p < 0.0001. We perform Kruskal–Wallis tests followed by post-hoc Mann-Whitney U Tests with Bonferroni correction. Here, we did not perform statistical comparisons between diverse AI algorithms.}
\end{figure}

\begin{figure}[h]
	\centering
	\includegraphics[width=0.9\columnwidth]{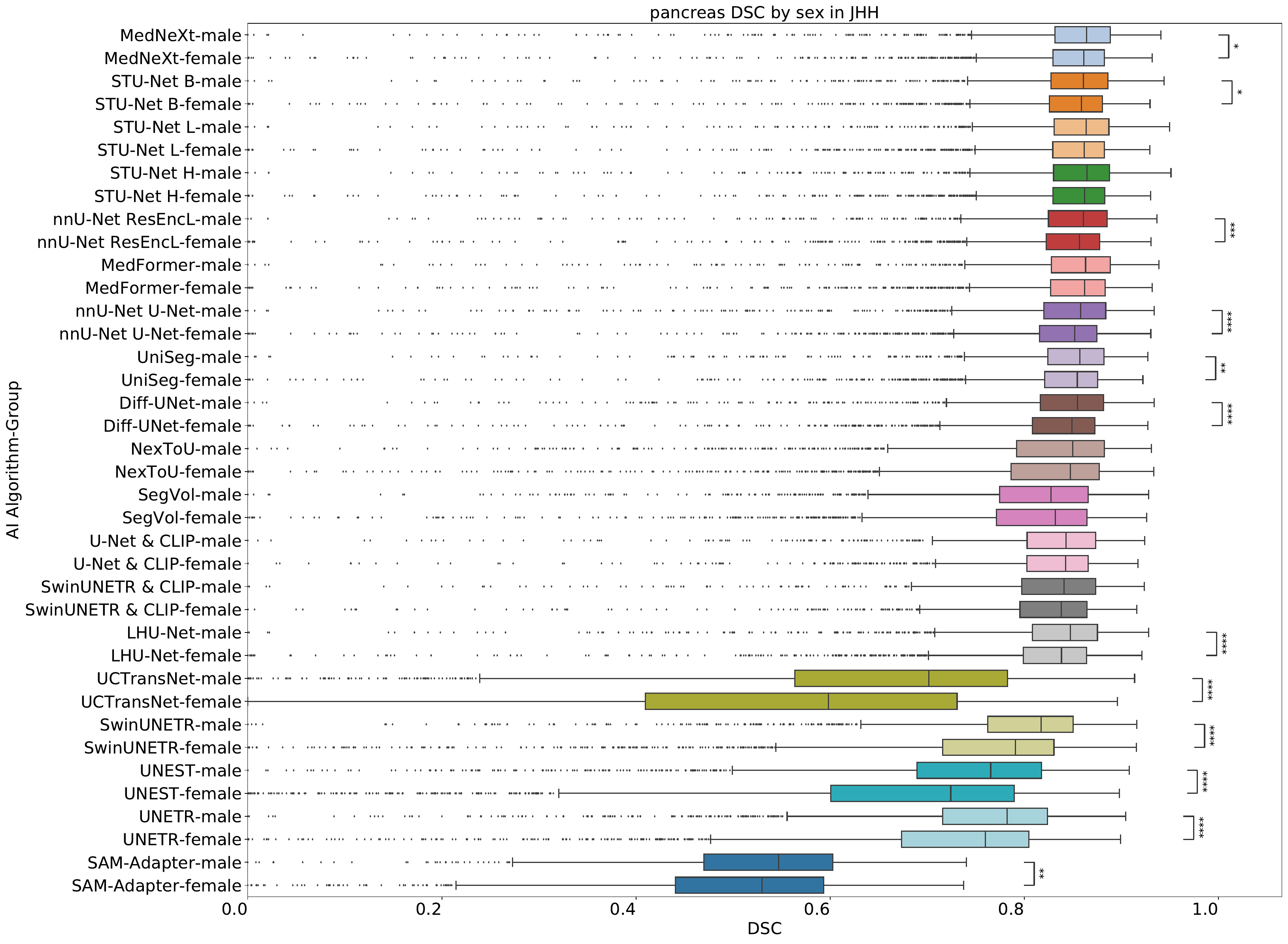}
    \caption{\textbf{Boxplot showing pancreas DSC score by sex in JHH.} Statistical significance is indicated by stars: * p < 0.05, ** p <0.01, *** p < 0.001, **** p < 0.0001. We perform Kruskal–Wallis tests followed by post-hoc Mann-Whitney U Tests with Bonferroni correction. Here, we did not perform statistical comparisons between diverse AI algorithms.}
\end{figure}

\clearpage
\subsubsection{Race: per-class analysis}

\begin{figure}[h]
	\centering
	\includegraphics[width=0.55\columnwidth]{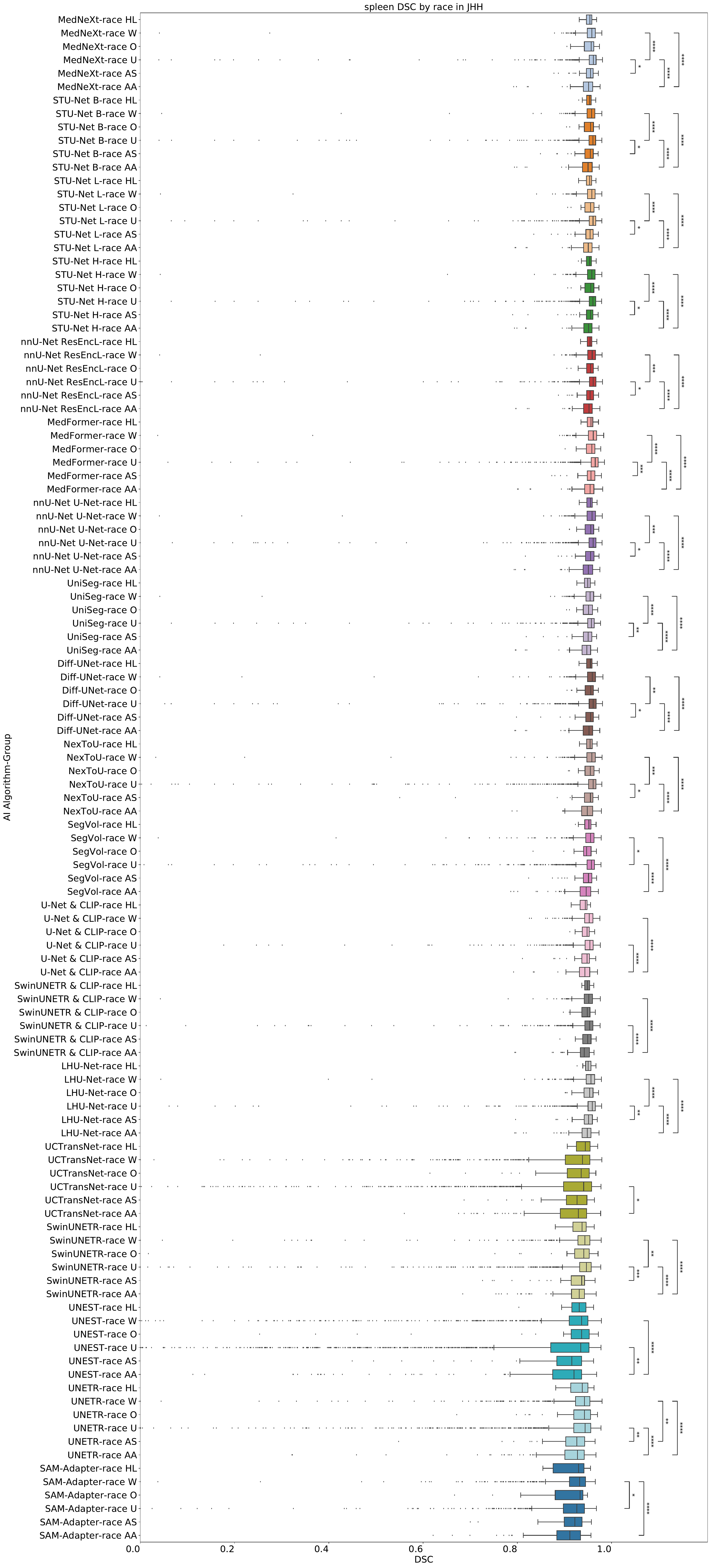}
    \caption{\textbf{Boxplot showing spleen DSC score by race in JHH.} Statistical significance is indicated by stars: * p < 0.05, ** p <0.01, *** p < 0.001, **** p < 0.0001. We perform Kruskal–Wallis tests followed by post-hoc Mann-Whitney U Tests with Bonferroni correction. Here, we did not perform statistical comparisons between diverse AI algorithms.}
\end{figure}

\begin{figure}[h]
	\centering
	\includegraphics[width=0.7\columnwidth]{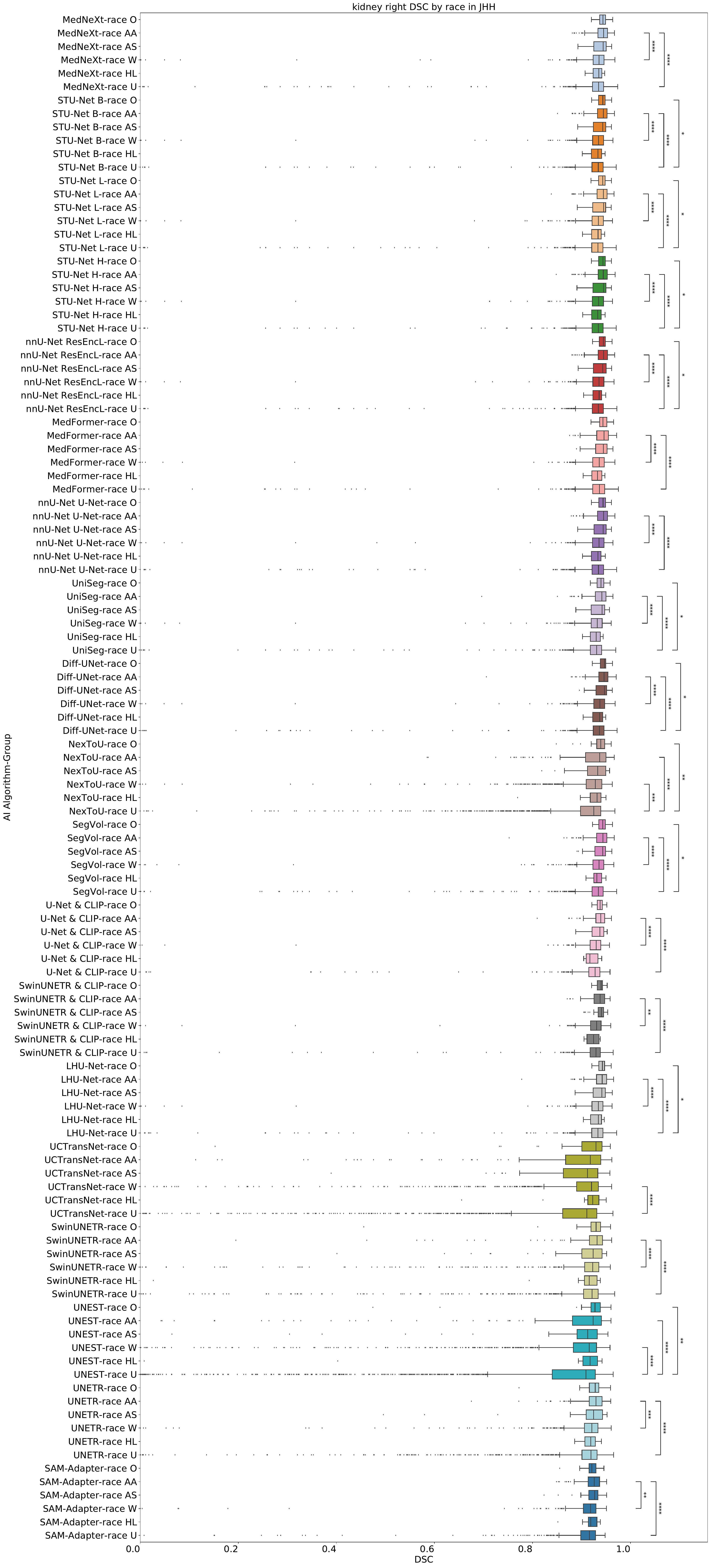}
    \caption{\textbf{Boxplot showing right kidney DSC score by race in JHH.} Statistical significance is indicated by stars: * p < 0.05, ** p <0.01, *** p < 0.001, **** p < 0.0001. We perform Kruskal–Wallis tests followed by post-hoc Mann-Whitney U Tests with Bonferroni correction. Here, we did not perform statistical comparisons between diverse AI algorithms.}
\end{figure}

\begin{figure}[h]
	\centering
	\includegraphics[width=0.7\columnwidth]{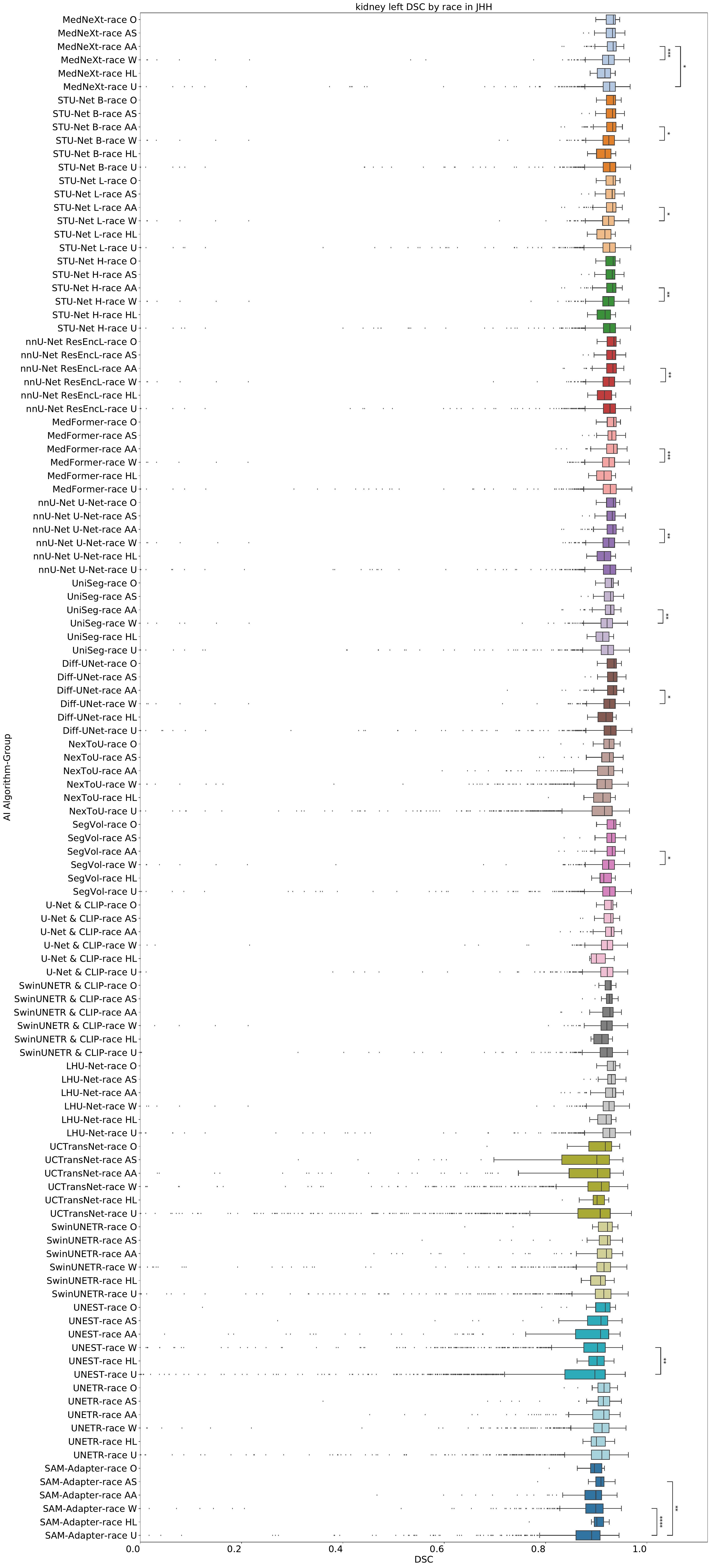}
    \caption{\textbf{Boxplot showing left kidney DSC score by race in JHH.} Statistical significance is indicated by stars: * p < 0.05, ** p <0.01, *** p < 0.001, **** p < 0.0001. We perform Kruskal–Wallis tests followed by post-hoc Mann-Whitney U Tests with Bonferroni correction. Here, we did not perform statistical comparisons between diverse AI algorithms.}
\end{figure}

\begin{figure}[h]
	\centering
	\includegraphics[width=0.7\columnwidth]{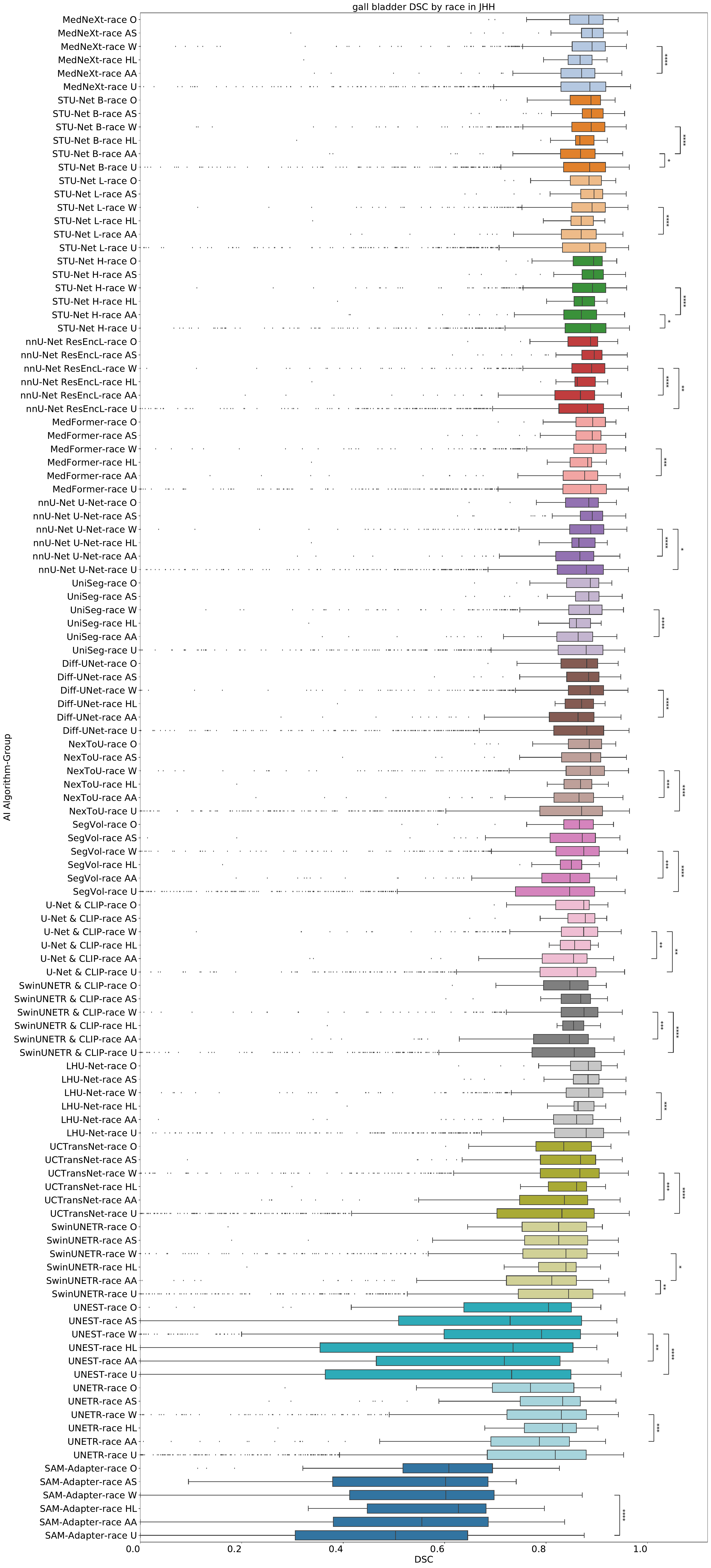}
    \caption{\textbf{Boxplot showing gallbladder DSC score by race in JHH.} Statistical significance is indicated by stars: * p < 0.05, ** p <0.01, *** p < 0.001, **** p < 0.0001. We perform Kruskal–Wallis tests followed by post-hoc Mann-Whitney U Tests with Bonferroni correction. Here, we did not perform statistical comparisons between diverse AI algorithms.}
\end{figure}

\begin{figure}[h]
	\centering
	\includegraphics[width=0.7\columnwidth]{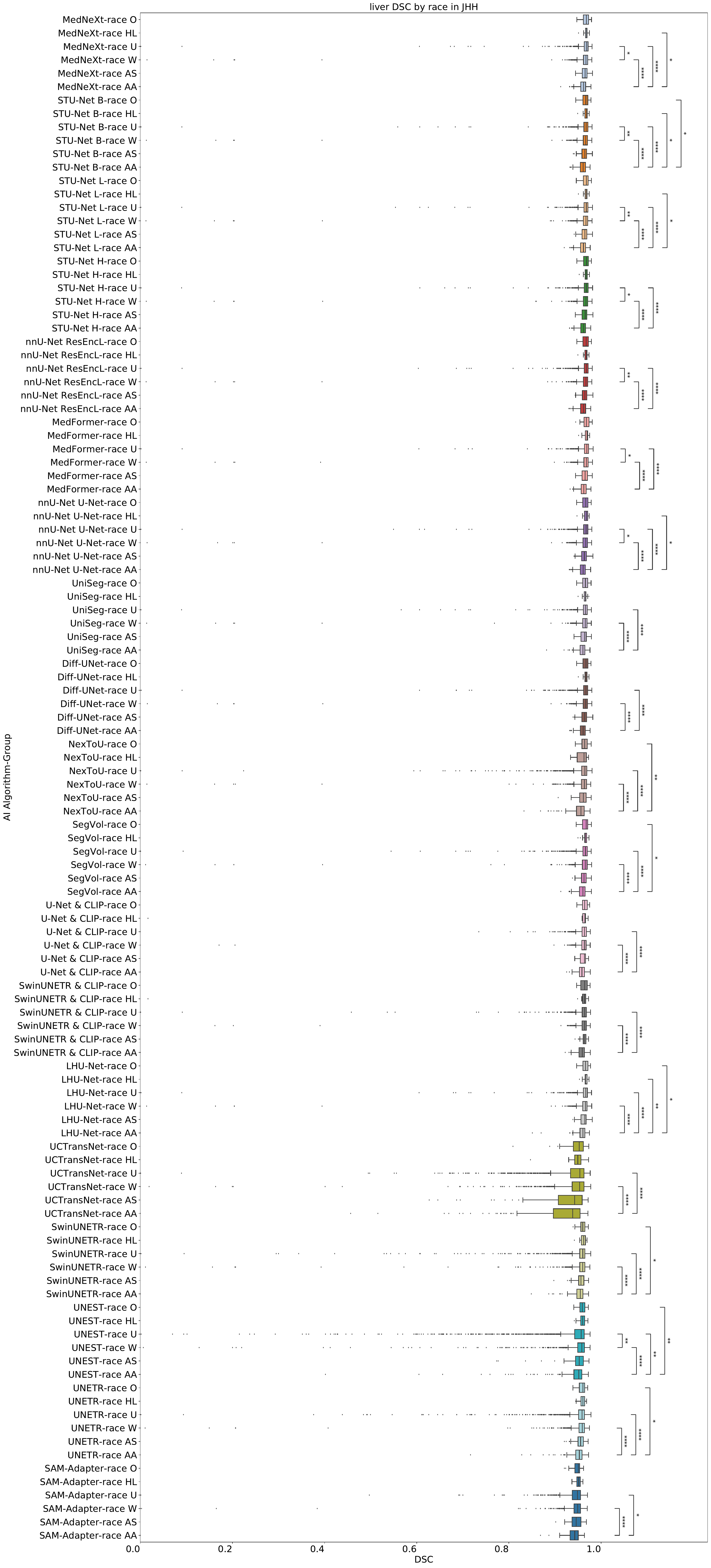}
    \caption{\textbf{Boxplot showing liver DSC score by race in JHH.} Statistical significance is indicated by stars: * p < 0.05, ** p <0.01, *** p < 0.001, **** p < 0.0001. We perform Kruskal–Wallis tests followed by post-hoc Mann-Whitney U Tests with Bonferroni correction. Here, we did not perform statistical comparisons between diverse AI algorithms.}
\end{figure}

\begin{figure}[h]
	\centering
	\includegraphics[width=0.7\columnwidth]{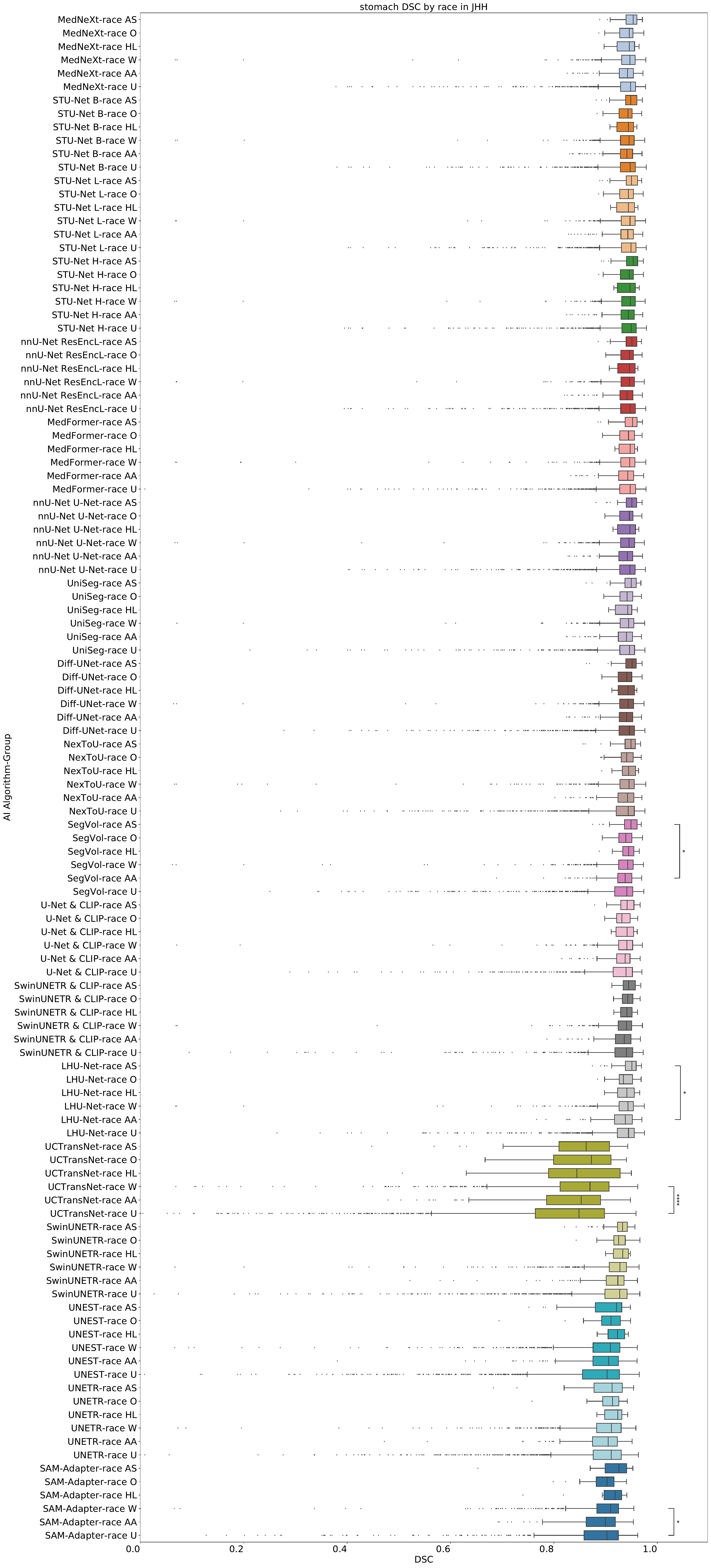}
    \caption{\textbf{Boxplot showing stomach DSC score by race in JHH.} Statistical significance is indicated by stars: * p < 0.05, ** p <0.01, *** p < 0.001, **** p < 0.0001. We perform Kruskal–Wallis tests followed by post-hoc Mann-Whitney U Tests with Bonferroni correction. Here, we did not perform statistical comparisons between diverse AI algorithms.}
\end{figure}

\begin{figure}[h]
	\centering
	\includegraphics[width=0.7\columnwidth]{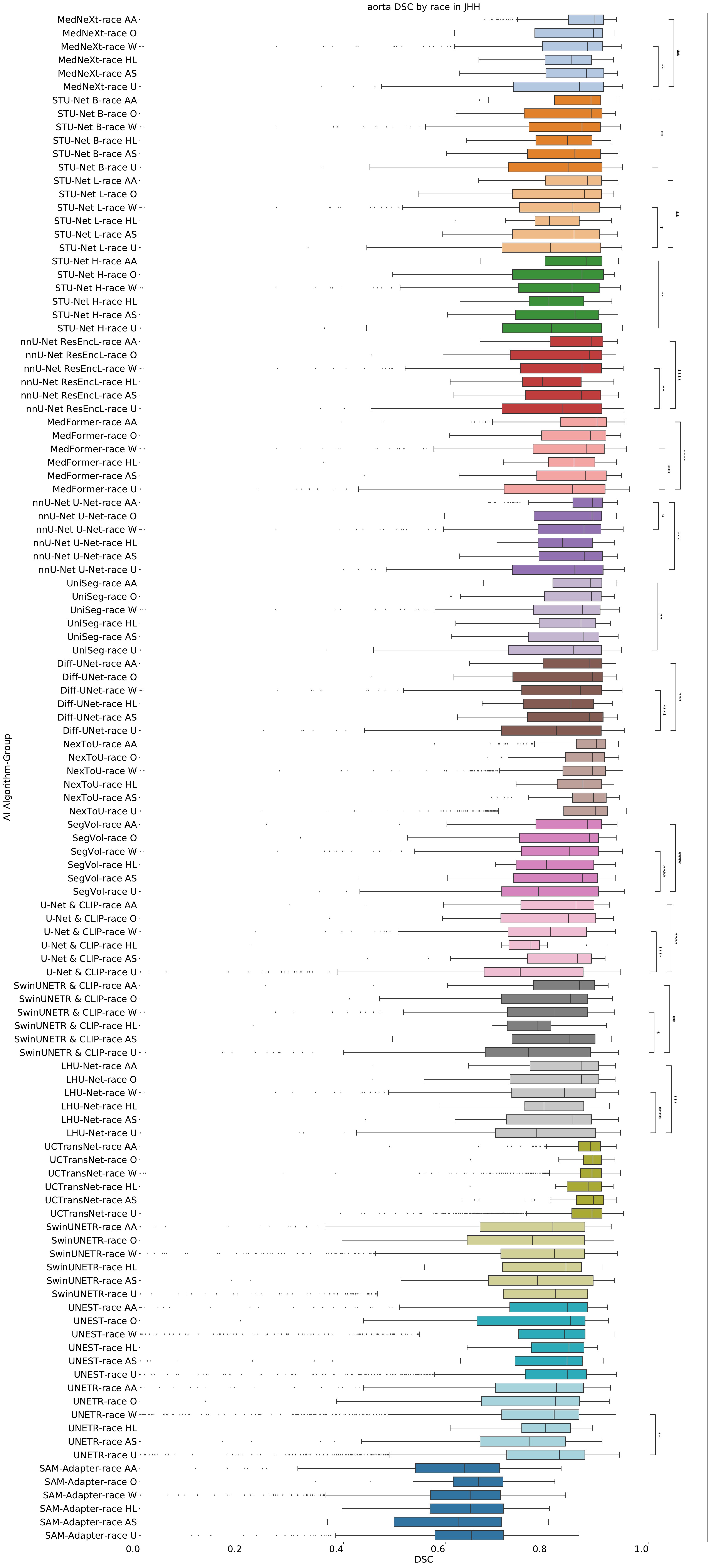}
    \caption{\textbf{Boxplot showing aorta DSC score by race in JHH.} Statistical significance is indicated by stars: * p < 0.05, ** p <0.01, *** p < 0.001, **** p < 0.0001. We perform Kruskal–Wallis tests followed by post-hoc Mann-Whitney U Tests with Bonferroni correction. Here, we did not perform statistical comparisons between diverse AI algorithms.}
\end{figure}

\begin{figure}[h]
	\centering
	\includegraphics[width=0.7\columnwidth]{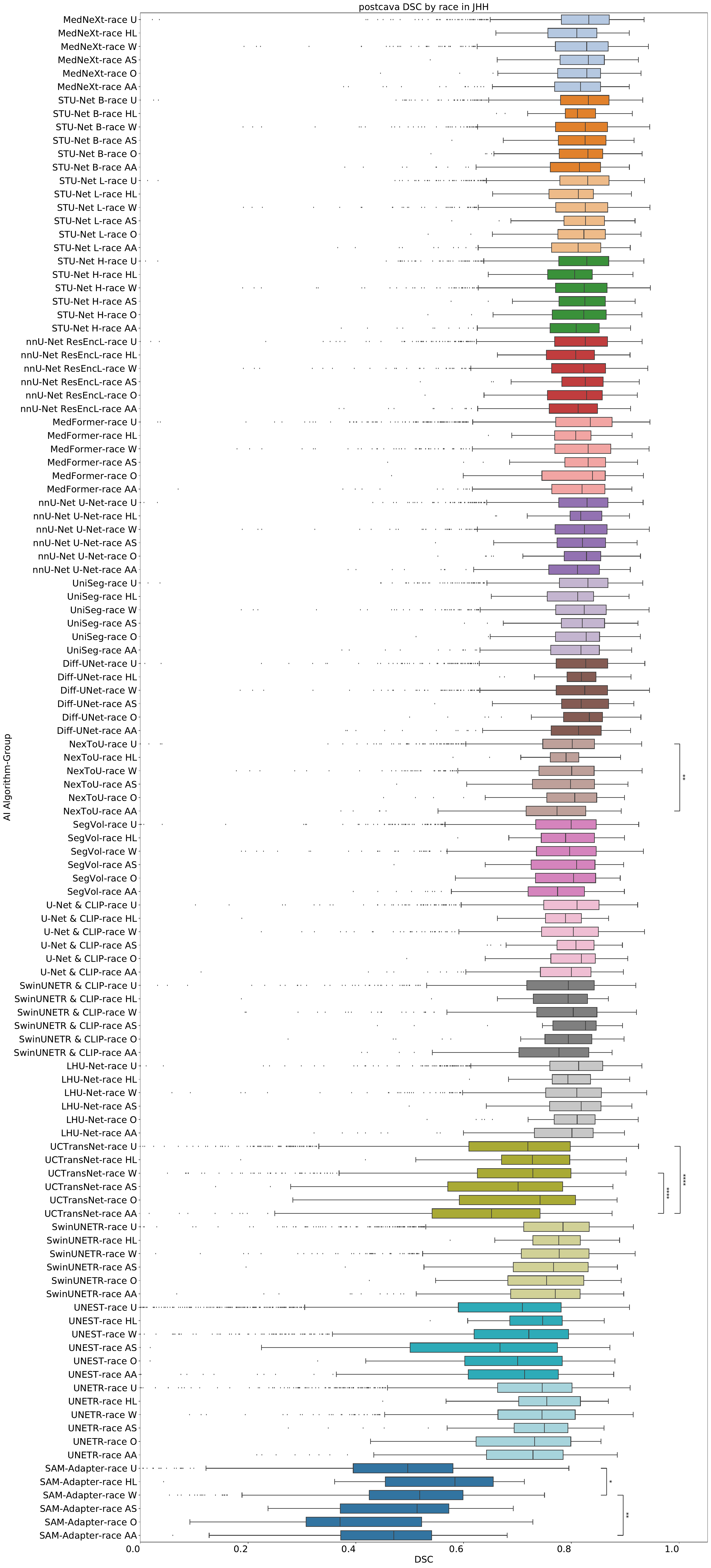}
    \caption{\textbf{Boxplot showing postcava DSC score by race in JHH.} Statistical significance is indicated by stars: * p < 0.05, ** p <0.01, *** p < 0.001, **** p < 0.0001. We perform Kruskal–Wallis tests followed by post-hoc Mann-Whitney U Tests with Bonferroni correction. Here, we did not perform statistical comparisons between diverse AI algorithms.}
\end{figure}

\begin{figure}[h]
	\centering
	\includegraphics[width=0.7\columnwidth]{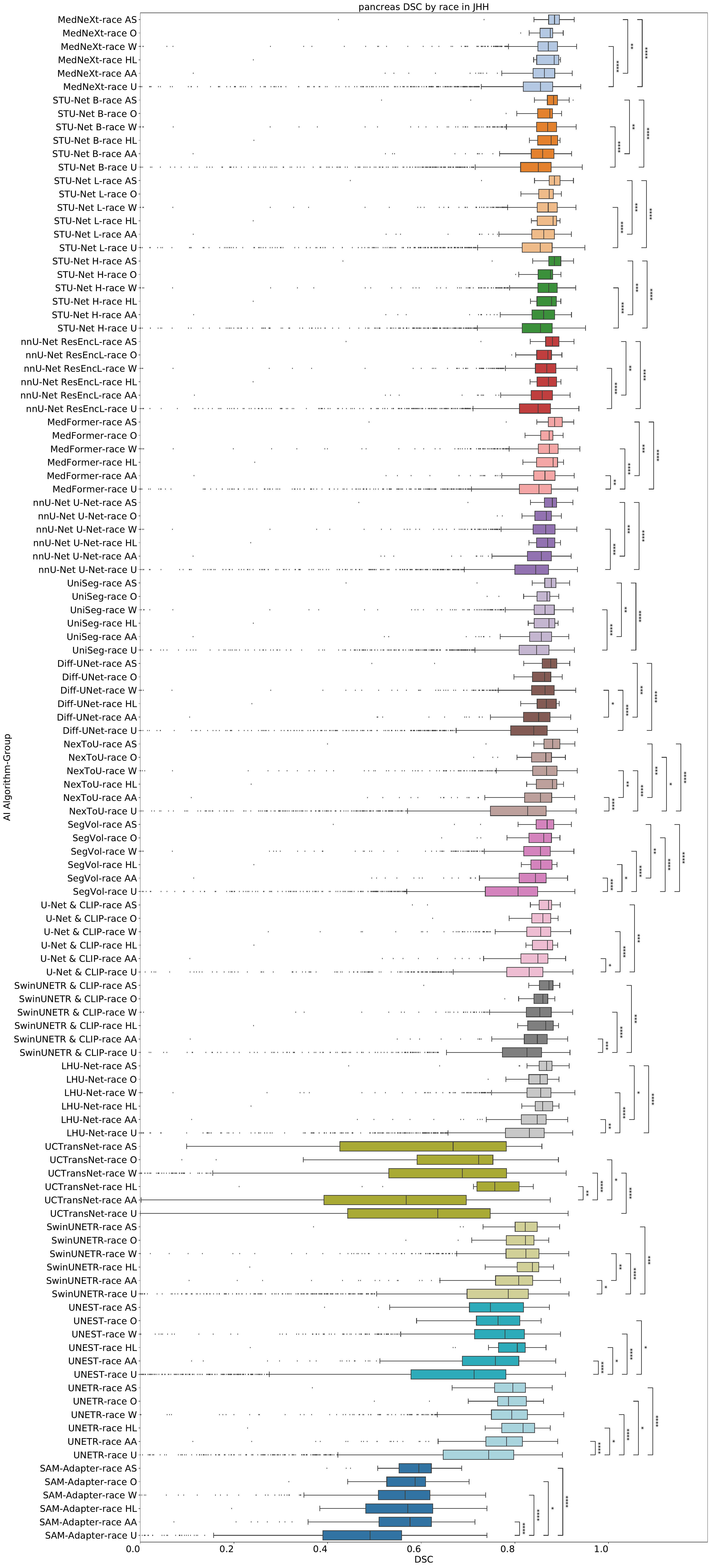}
    \caption{\textbf{Boxplot showing pancreas DSC score by race in JHH.} Statistical significance is indicated by stars: * p < 0.05, ** p <0.01, *** p < 0.001, **** p < 0.0001. We perform Kruskal–Wallis tests followed by post-hoc Mann-Whitney U Tests with Bonferroni correction. Here, we did not perform statistical comparisons between diverse AI algorithms.}
\end{figure}

\clearpage
\section{On Label Noise}\label{sec:label_noise}

\ourdataset\ is an amalgamation of \numoftrpubdataset\ public datasets (Appendix \ref{sec:abdomenatlas_construction}), which, when combined together, resulted in a partially labeled dataset. Radiologists, assisted by AI, provided all the missing labels for \numofclass\ anatomical structures, making the dataset fully-labeled \cite{qu2023annotating}. When creating \ourdataset\, we did not revise the labels that were already provided in the public datasets. However, upon future visual inspection, we found that these public datasets may share inconsistent annotation standards, also reported in Liu~\etal~\cite{liu2023clip}. For example, the aorta annotation standard is inconsistent in AbdomenCT-12organ and other datasets: part of the upper aorta region is missing in AbdomenCT-12organ, while the aorta annotation is complete in BTCV and AMOS. Moreover, since the public datasets that constitute \ourdataset\ contained both automatic and manual labels, they can also portray human and AI errors.

To address this, we developed an automatic label quality checking tool, based on anatomical priors (e.g., expected shape of organs), to detect and correct noisy labels. This tool indicated that aorta concentrated most of the label noise in \ourdataset. It has 32.4\% of noisy labels, which are mostly the aforementioned incomplete annotations. The second structure with the highest amount of detected errors was the kidneys, but its percentage of noisy labels was much lower: 2.6\%. Our tool detected less than 1\% of error in other classes. Therefore, the detected errors are mostly concentrated on one of the 9 annotated structures. Moreover, since \ourdataset\ carried the errors and annotation standard inconsistencies found in public datasets, the noise in \ourdataset\ labels represents common annotation errors and inconsistencies. Conversely, studies on AI robustness to label noise commonly rely on  artificially generated noise \cite{wang2023dealing}. Thus, we viewed the realistic and quantifiable noise in \ourdataset\ as an opportunity to perform a realistic study on AI robustness to label noise. To further increase the study's realism, we simulate the standard scenario where researchers are unaware of the noise: we did not inform the AI creators about the annotation errors in \ourdataset\ prior to model training. This approach avoided uneven label corrections by only some teams and ensured that the AI algorithms in this benchmark accurately represent the realistic scenario of AI trained on public data with common label noise, without creators actively trying to counteract the noise.

To assess AI robustness to label noise, the algorithms must be tested on datasets whose labels are less noisy than those in the training data. The JHH test set ($N$=\numoftejhhct) was entirely annotated by radiologists, manually and following a well-defined annotation standard, over 5 years \cite{park2020annotated}. Thus, it serves as a gold standard for low label noise. \ourproject\ leverages this large-scale, high-quality test dataset to verify whether AI trained noisy labels, representative of current public datasets, performs well when evaluated with high-quality manual labels. Since TotalSegmentator is not composed of multiple datasets, their annotation standards are consistent, and we detected low levels (<1\%) of label noise on them. Thus, they are also adequate for evaluating AI's robustness. Additionally, to better quantify the impact of label noise on AI accuracy, we re-trained ResEncL on \href{https://huggingface.co/datasets/MrGiovanni/AbdomenAtlas1.0C}{\ourdataset C}. This dataset, which we \href{https://huggingface.co/datasets/MrGiovanni/AbdomenAtlas1.0C}{publicly} released, is a revised version of \ourdataset, where labels were improved by radiologists assisted by AI and by our error detection tool. The aorta was the only class where the nnU-Net had large and significant performance increments (e.g., 10.35\% DSC improvement in TotalSegmentator). For other structures, improvements are mostly not significant and low, demonstrating that the AI algorithm is robust to moderate levels of label noise (e.g., less than 3\% of noisy labels according to our detection tool), but not to excessive noise. The continuous improvement of label noise detection and annotation quality, unifying annotation standards and correcting public datasets' flawed labels, is a continuous commitment of \ourproject.

\clearpage
\section{Full Affiliation List}
\label{sec:full_affiliation}
\textsuperscript{1}Department of Computer Science, Johns Hopkins University \\
\textsuperscript{2}Department of Pharmacy and Biotechnology, University of Bologna \\
\textsuperscript{3}Center for Biomolecular Nanotechnologies, Istituto Italiano di Tecnologia \\
\textsuperscript{4}NVIDIA \\
\textsuperscript{5}Division of Medical Image Computing, German Cancer Research Center (DKFZ) \\
\textsuperscript{6}Helmholtz Imaging, German Cancer Research Center (DKFZ) \\
\textsuperscript{7}ESAT-PSI, KU Leuven \\
\textsuperscript{8}Faculty of Mathematics and Computer Science, Heidelberg University \\
\textsuperscript{9}HIDSS4Health - Helmholtz Information and Data Science School for Health \\
\textsuperscript{10}Shanghai Jiao Tong University \\
\textsuperscript{11}Shanghai Artificial Intelligence Laboratory \\
\textsuperscript{12}Pattern Analysis and Learning Group, Department of Radiation Oncology, Heidelberg University Hospital\\
\textsuperscript{13}Interactive Machine Learning Group (IML), DKFZ \\
\textsuperscript{14}School of Computer Science and Engineering, Northwestern Polytechnical University \\
\textsuperscript{15}Australian Institute for Machine Learning, The University of Adelaide \\
\textsuperscript{16}College of Computer Science and Technology, Zhejiang University \\
\textsuperscript{17}Hong Kong University of Science and Technology (Guangzhou) \\
\textsuperscript{18}Hong Kong University of Science and Technology \\
\textsuperscript{19}Faculty of Informatics and Data Science, University of Regensburg \\
\textsuperscript{20}Faculty of Electrical Engineering and Information Technology, RWTH Aachen University \\
\textsuperscript{21}Fraunhofer Institute for Digital Medicine MEVIS \\
\textsuperscript{22}Electronic \& Information Engineering School, Harbin Institute of Technology (Shenzhen) \\
\textsuperscript{23}Beijing Academy of Artificial Intelligence (BAAI) \\
\textsuperscript{24}The Chinese University of Hong Kong \\
\textsuperscript{25}Peking University \\
\textsuperscript{26}Department of Electrical and Computer Engineering, Duke University \\
\textsuperscript{27}Stony Brook University \\
\textsuperscript{28}Department of Computer Science and Engineering, Department of Chemical and Biological Engineering and Division of Life Science, Hong Kong University of Science and Technology \\
\textsuperscript{29}Data Science and Computation Facility, Fondazione Istituto Italiano di Tecnologia \\
\textsuperscript{30}Ecole Polytechnique Fédérale de Lausanne \\

\clearpage
\section{Potential Negative Societal Impacts}\label{sec:negative_societal_impacts}

Potential negative societal impacts of benchmarking AI algorithms for medical image segmentation include reinforcing biases, compromising data privacy, and leading to misuse of AI systems. Standard benchmarks may suffer from in-distribution biases, small test sets, oversimplified metrics, and short-term outcome pressures, which can result in AI models that perform well on benchmarks but fail in real-world applications. These issues can undermine the reliability, fairness, and generalizability of AI systems in medical contexts, potentially causing harm and reducing trust in AI-driven healthcare solutions.

\end{document}